\documentclass[dvipsnames]{article}

\usepackage[final]{neurips_2021}

\usepackage[utf8]{inputenc} % allow utf-8 input
\usepackage[T1]{fontenc}    % use 8-bit T1 fonts
\usepackage{url}            % simple URL typesetting
\usepackage{booktabs}       % professional-quality tables
\usepackage{amsfonts}       % blackboard math symbols
\usepackage{nicefrac}       % compact symbols for 1/2, etc.
\usepackage{booktabs} 
\usepackage{multirow}
\usepackage{wrapfig}
\usepackage{amssymb}
\usepackage{epigraph}
\usepackage{makecell}

\usepackage[labelfont=bf]{caption} % For bold caption labels

\usepackage[ruled,vlined]{algorithm2e}
\usepackage{color, soul}
\usepackage{microtype}      % microtypography
\usepackage{xcolor}         % colors
\usepackage{geometry}
\usepackage{times}
\usepackage{ragged2e}
\usepackage{amsmath}
\usepackage{bm}
\usepackage{latexsym}
\usepackage{subfigure}
\usepackage{graphicx}
\usepackage{float}
\usepackage{longtable}
\usepackage{booktabs} 
\usepackage{multirow}
\usepackage{amssymb}
\usepackage{longtable}
\usepackage{tabu}
\usepackage{bbding}
\usepackage{awesomebox} % for infobox
\usepackage{bbding}
\usepackage[most]{tcolorbox}
\usepackage{etoolbox}
\usepackage{fancyhdr}

\providecommand{\keywords}[1]
{
  \small
  \textbf{\textit{Keywords---}} #1
}

\usepackage{caption} 
\usepackage{chngcntr}
\usepackage[colorlinks, linkcolor=RoyalBlue, anchorcolor=BrickRed, citecolor=RoyalBlue, urlcolor=RoyalBlue]{hyperref}

\usepackage{cleveref}
\crefname{section}{§}{§§}
\Crefname{section}{§}{§§}

\newcommand\refsec[1]{\hyperref[sec:#1]{§\ref{sec:#1}:~\textsc{#1}}}
\newcommand\refsecs[2]{\hyperref[sec:#1]{§\ref{sec:#1}:~\textsc{#1}}, \hyperref[sec:#2]{§\ref{sec:#2}:~\textsc{#2}}}

\newtheorem{thm}{Theorem}[section]

\newtcolorbox{myboxnote}[1][]{
  breakable,
  title=#1,
  colback=cyan!0,
  colbacktitle=cyan!0,
  coltitle=black,
  fonttitle=\bfseries,
  bottomrule=0pt,
  toprule=0pt,
  leftrule=1.5pt,
  rightrule=1.5pt,
  titlerule=0pt,
  arc=0pt,
  outer arc=0pt,
  colframe=lightgray,
}
\newenvironment{itemize*}%
 {\leftmargini=20pt\begin{itemize}%
  \setlength{\itemsep}{3pt}%
  \setlength{\parskip}{0pt}%
  }%
 {\end{itemize}}
\newenvironment{enumerate*}%
 {\begin{enumerate}%
  \setlength{\itemsep}{0pt}%
  \setlength{\parskip}{0pt}}%
 {\end{enumerate}}

\usepackage{fancyhdr} % to change header and footers
\usepackage{blindtext} % to quickly get a full document

\title{ Delta Tuning: A Comprehensive Study of Parameter Efficient Methods for Pre-trained Language Models}

\author{%
   Ning Ding\thanks{Equal contribution.}, Yujia Qin$^*$, Guang Yang, Fuchao Wei, Zonghan Yang, Yusheng Su, Shengding Hu, \\ \textbf{ Yulin Chen, Chi-Min Chan, Weize Chen, Jing Yi, Weilin Zhao, Xiaozhi Wang,} \\ \vspace{0.2cm} \textbf{Zhiyuan Liu, Hai-Tao Zheng, Jianfei Chen, Yang Liu, Jie Tang, Juanzi Li, Maosong Sun}
 \\ 
 {\vspace{0.2cm} Tsinghua University, BAAI} \\ 
 {\texttt{\{dingn18, qyj20\}@mails.tsinghua.edu.cn}}
}

\begin{document}

\maketitle

\vspace{-0.5cm}
\begin{abstract}
\vspace{-0.1cm}
As pre-trained language models (PLMs) have become the fundamental infrastructure for various NLP tasks and researchers have readily enjoyed themselves in the \textit{pretraining-finetuning} paradigm, evidence from emerging research has continuously proven that larger models tend to yield better performance. However, despite the welcome outcome, the process of fine-tuning large-scale PLMs brings prohibitive adaptation costs. In fact, fine-tuning all the parameters of a colossal model and retaining separate instances for different tasks are practically infeasible. This necessitates a new branch of research focusing on the parameter-efficient adaptation of PLMs.
% , dubbed as \textit{delta tuning} in this paper. 
% However, 
% % considering other potential advantages 
% to set free the imagenation of the advantages that might be
% % brought by the new adaptation method
% brought by the new adaptation method, not limited to being effective in terms of parameters, we coined a new term \textit{delta tuning} in terms of morphology.% 在改
% In order to unleash the imagination of the possible advantages of parameter-efficient adaptation methods, not limited to being parameter efficiency,, we have coined a new term \textit{delta tune} from a morphological perspective to refer to the original ``parameter efficient tuning''.
In order to unleash the imagination of the possible advantages of such methods, not limited to parameter efficiency, we coined a new term \textit{delta tuning} from a morphological point of view to refer to the original ``parameter efficient tuning''.
In contrast with the standard fine-tuning, delta tuning only fine-tunes a small portion of the model parameters while keeping the rest untouched, largely reducing both the computation and storage costs.
Recent studies have demonstrated that a series of delta tuning methods with distinct tuned parameter selection  could achieve performance on a par with full-parameter fine-tuning, suggesting a new promising way of stimulating large-scale PLMs. In this paper, we first formally describe the problem of delta tuning and then comprehensively review recent delta tuning approaches. 
We also propose a unified categorization criterion that divides existing delta tuning methods into three groups: \textit{addition-based}, \textit{specification-based}, and \textit{reparameterization-based} methods. Though initially proposed as an efficient method to steer large models, we believe that some of the fascinating evidence discovered along with delta tuning could help further reveal the mechanisms of PLMs and even deep neural networks. To this end, we discuss the theoretical principles underlying the effectiveness of delta tuning and propose frameworks to interpret delta tuning from the perspective of \textit{optimization} and \textit{optimal control}, respectively. Furthermore, we provide a holistic empirical study of representative methods, where results on over 100 NLP tasks demonstrate a \textit{comprehensive performance comparison} of different approaches. 
The experimental results also cover the analysis of \textit{combinatorial}, \textit{scaling} and \textit{transferable} properties of delta tuning. To facilitate the research of delta tuning, we are also developing an open-source toolkit, OpenDelta\footnote{\url{https://github.com/thunlp/OpenDelta} \\ \\ The contributions of the authors are listed in \hyperref[sec:contributions]{\textsc{§ contributions}}}, that enables practitioners to efficiently and flexibly implement delta tuning on PLMs. At last, we discuss a series of real-world applications of delta tuning.
%

%release the codes, dataset splits and trained delta checkpoints to facilitate future research

%At last, we discuss the open problems and the future research directions of this area. 
\end{abstract}
\vspace{-0.3cm}
\hspace{30pt}
{
}

\vspace{-0.3cm}
\hspace{180pt}\parbox[b]{0.45\textwidth}
{
\epigraph{\textit{``The lurking suspicion that something could be simplified is the world’s richest source of rewarding challenges.''}}{--- Edsger W. Dijkstra} 
}

\newpage
{
  \hypersetup{linkcolor=RoyalBlue, linktoc=page}
  \tableofcontents
}

\newpage
\section{Introduction}
\label{sec:introduction}
% earlier work: deep learning -> deep learning for language modeling -> model engineering -> PLMs 

Language lies at the heart of human intelligence. Its systematic nature allows the denotation of real objects or illustration of laws with symbolic expressions and could convey almost infinite information with a finite symbolic set; its arbitrariness shows that there are no necessary connections between the real-world and the language space, and indicates the importance of world knowledge and social convention to the effectiveness of language in human society; its richness in meaning enables the expression of extremely complex behaviors or tasks with clear and simple symbols. Understanding language is the key to understanding intelligence. The inquiry into what language is and how we acquire, store and comprehend it has never stopped among psychologists and linguists, and the charm of language will continue to impress and inspire us in the future. 
Likewise, to create the real intelligence system, researchers in the field of artificial intelligence (AI) have been dedicated to training machines to model, understand and generate language.

With the revolutionary development in computing hardware, traditional statistical methods have yielded their place to deep learning~\citep{lecun2015deep} that heavily rely on tensor computation and huge data volume. Modern natural language processing (NLP) uses deep neural networks to implicitly model probability and capture language representations~\citep{hochreiter1997long, bengio2000neural, grefenstette2014convolutional, kim-2014-convolutional, vaswani2017attention}. A standard pipeline involves encoding language into discrete tokens (tokenization) as model input, choosing a proper model architecture, training the network with the given corpora, and designing self-supervised tasks. Experimented with various model architecture, the Transformer neural network~\citep{vaswani2017attention} produced state-of-the-art performances on a series of NLP tasks and has been widely acknowledged as the standard architecture for pre-trained language models (PLMs). This ushers a new era of \textit{pre-training and fine-tuning}. PLMs typically use heavily over-parameterized Transformers as the base architecture, and model natural language in bidirectional~\citep{devlin2018bert}, auto-regressive~\citep{radford2018improving, radford2019language}, or sequence-to-sequence~\citep{raffel2019exploring} manners on large-scale unsupervised corpora. Then for downstream tasks, task-specific objectives are introduced to fine-tune the PLMs for model adaptation. 
Notably, the increasing scale of PLMs (measured by the number of parameters) seems to be an irreversible trend as constant empirical results show that larger models (along with more data) almost certainly lead to better performance. For example, the 175 billion parameters GPT-3~\citep{brown2020language} generates natural language of unprecedented quality and can conduct various desired zero-shot tasks with satisfactory results given appropriate prompts. Nevertheless, performing full parameter fine-tuning on existing computing devices becomes formidable with the growing model scale. This finally leads to a desperate yet thought-provoking question: do we really need to update all the parameters?
In this context, how to efficiently and effectively adapt large models to particular downstream tasks is an intriguing research issue. 

% need a figure
% need a stat

As a predominant way to conduct model adaptations, fine-tuning initializes the model with the pre-trained weights, updates all the parameters, and produces separate instances for different tasks. But as implied by the case of GPT-3, fine-tuning becomes impractical as the model scales.
%However, vanilla fine-tuning is impractical when it comes to larger PLMs. 
In addition to the cost of deployment and computation, storing different instances for different tasks is extremely memory-intensive. To further explore the practical application rate of large models (PLMs with over 1 billion parameters), we randomly select 1000 published research papers from the recent five NLP conferences (200 for each venue), including ACL 2021, EMNLP 2021, NAACL 2021, ACL 2020, and EMNLP 2020. Then we manually count the usage of PLMs in these peer-reviewed works, specifically, we only focus on the experiments part of the papers. According to the statistics in Table~\ref{tab:stat}, although the use of PLMs has almost become standard,
there are only 0.5\% $\sim$ 4\% research papers that practically adopt large ones in the experiments. This suggests, firstly, that there is still inertia in the academic community which has resulted in scarce usage of large models in research, and also that the cost of deploying and experimentally validating large PLMs hinders the development of NLP research.

%randomly select 1000 published research papers that utilizes PLMs after the propose of T5~\citep{raffel2019exploring} from ICLR, NeurIPS, ICML, ACL, EMNLP and NAACL. According to our statistics in Table 1, there are only 1\% research papers practically use such large models in the experiments. 
% placeholders

\begin{table}[ht]
\centering
\caption{The usage of models of different sizes in research published in NLP conferences, the statistic is based on 1000 randomly selected papers. Large PLMs are defined as PLMs with over 1 billion parameters.}
\begin{tabular}{lcccc}
\toprule
\textbf{Venue}      & \textbf{No PLMs} & \textbf{Small PLMs} & \textbf{Large PLMs} & \textbf{Per. of Large PLMs} \\ \midrule

ACL 2021   &   41      &     151       &       8     &    4.0\%                      \\ 
EMNLP 2021 & 46 & 150 & 4 & 2.0\%  \\
%EMNLP 2021 &         &            &            &       \\ 
NAACL 2021 &   37      &       158     &       5     &     2.5\%                      \\ 
ACL 2020   &   107     &     92  &      1      &        0.5\%                  \\ 
EMNLP 2020 &    62     &     137       &      1      &   0.5\%     \\ 
\bottomrule
\end{tabular}
\label{tab:stat}
\end{table}

%Indeed, standard fine-tuning is parameter-inefficient as it always produces a new model with all parameters updated. Effectively identifying a small portion of the parameters and tuning them to exploit the PLMs becomes attractive in an era with large PLMs. 
To this end, a branch of parameter-efficient methods for model tuning arises. Although each of these approaches has its own emphasis on structural design, they essentially tune a ``delta'' (i.e., adaptive parameters) in the adaptation phase, we thus coin the term \textit{delta tuning}\footnote{In \refsec{delta tuning} and \refsec{theory}, we use the consistent mathematical expressions $\Delta$ and $\delta$ to describe and analyze delta tuning.} to refer to these methods. Parametric efficiency is an external manifestation of delta tuning that further exposes the low-rank or low-dimensional nature of large model adaptation in a more fundamental way.
%\footnote{In this paper, we use delta tuning to refer to methods that could achieve PLM adaptation in a parameter-efficient way, other names for this class of methods include parameter-efficient fine-tuning, parameter-efficient transfer learning, etc.}. 
Generally, delta tuning only updates a small number of parameters (inherently in the model or additionally introduced) while freezing the remaining parameters that account for the vast majority. Adapter tuning~\citep{houlsby2019parameter} is among the earliest approaches to steer pre-trained models with a limited number of parameters. It inserts adapter modules with bottleneck architecture between layers in PLMs and only these inserted modules get updated during fine-tuning. Prefix-tuning~\citep{li2021prefix} tunes the PLMs by updating the pre-pended parameters in each transformer layer. Taken insights from GPT-3, prompt tuning~\citep{lester2021power} only prepends and updates task-specific trainable parameters in the original input embeddings. BitFit~\citep{zaken2021bitfit} updates the bias terms in PLMs while freezing the remaining modules. LoRA~\citep{hu2021lora} decomposes attention weight gradient into low-rank matrices to reduce the number of trainable parameters. With the diverse flourishing research and the promising results, efforts have been made to explain and compare the essence of some popular methods. ~\citet{he2022unified} propose a unified view of the existing delta tuning methods and illustrate the difference and connections among them formulaically.

\begin{figure}[ht]
    \centering
    \includegraphics[width=0.79\textwidth]{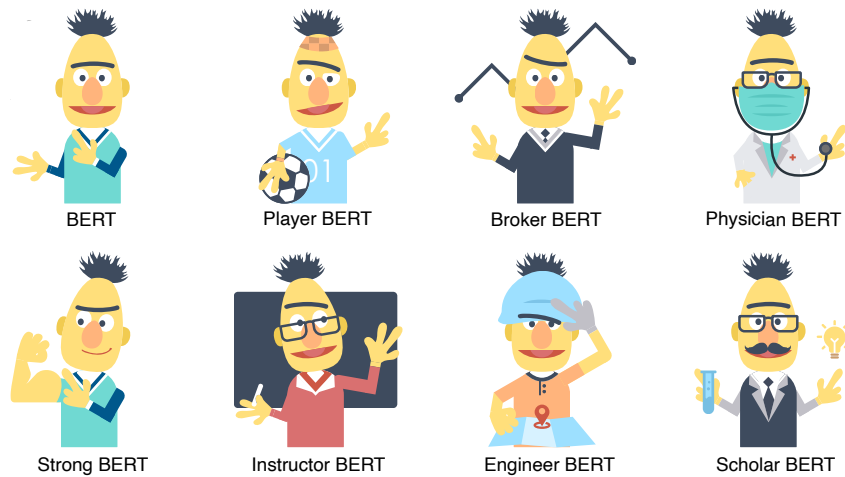}
    \caption{Delta tuning seeks to adapt and specialize PLMs with changes of a small portion of parameters.}
    \label{fig:concept}
\end{figure}

The delta tuning methods enable efficient tuning and practical usage for large pre-trained models and often achieve comparable results to the standard fine-tuning. For example, the vanilla fine-tuning of GPT-3 needs to update about 175,255 million parameters, which is almost infeasible in both industry and academia. However, if we only tune the injected low-rank decomposition matrices in each Transformer layer~\citep{hu2021lora},  only 37.7 million parameters will be involved in backpropagation. 
Delta tuning not only provides a promising way to adapt large PLMs, but also sheds light on the mechanisms behind such model adaptations. Compared to pre-training, delta tuning makes model adaptation a considerably low-cost process in terms of data volume and model optimization.
For instance, researchers find that the optimization problem of the adaptations for big models could be reparameterized into a low-dimensional ``intrinsic subspace''~\citep{aghajanyan2020intrinsic, qin2021exploring}, and various NLP tasks could be handled by only tuning very few parameters in the subspace. The empirical evidence takes us one step closer to understanding how pre-trained models work, and may even spawn new theoretical questions that are worth exploring.

This paper first attempts to survey the development and recent advances in delta tuning. 
For preliminaries, we give a description of the Transformer neural models and mainstream PLMs (\refsec{preliminaries}). Then we formally describe the delta tuning problem and propose a categorization criterion (\refsec{delta tuning}) to provide a unified view on delta tuning methods.
Categorizing delta tuning into addition-based (\refsec{addition}), specification-based (\refsec{specification}), and reparameterization-based (\refsec{reparameterization}) methods, we comprehensively introduce the technical details and empirical conclusions of the methods.

To better understand the inner connections among the delta tuning methods and the mechanisms of model adaptation, we develop theoretical analysis (\refsec{theory}) of delta tuning by proposing theoretical frameworks
from two different perspectives, optimization (\refsec{optimization}) and optimal control (\refsec{optimal control}). Our theoretical discussion is summarized as follows:

\begin{itemize*}
    \item \textbf{Optimization.} \ Based on the intrinsic low dimension in a large pre-trained language model, we show that delta tuning is essentially a subspace optimization method with respect to the solution space or functional space. The discussion justifies the designs of the existing delta tuning methods and explains some phenomena in the experiments. 
    \item \textbf{Optimal Control.} \ Inspired by the relationship between deep learning and optimal control theories, we interpret delta tuning as seeking optimal controllers for PLMs. We propose an optimal control framework that unifies different delta tuning approaches. Our analysis provides theoretical references for the novel design of delta tuning methods.
\end{itemize*}

In terms of empirical studies, we carry out extensive and systematic experiments (\refsec{experiments}) on over 100 NLP tasks to rigorously explore the performances (\refsec{performance}), combinability (\refsec{combination}), the power of scale (\refsec{scale}), transferability (\refsec{transferability}), etc.
%Then we respectively introduce delta tuning approaches categorized into adapter-based, (\cref{sec:adapter}), prompt-based (\cref{sec:prompt}), different-based (\cref{sec:delta}), intrinsic-based (\cref{sec:intrinsic}) and mixed methods (\cref{sec:uni}). delta tuning has inner connections with many previous theoretical research such as the mechanisms of  over-parameterization, thus we introduce possible theories behind delta tuning and even model adaptation in \cref{sec:theory}. 
Our main findings are summarized as follows:
\begin{itemize*}
    \item \textbf{Performance.} \ Despite the huge potential, existing delta tuning methods are still no match for the conventional fine-tuning either in performance or convergence. Among several representative delta tuning methods, no single algorithm predominantly outperforms the others. We also analyze the key properties of delta tuning such as convergence and computational efficiency.
    \item \textbf{Combinability.} \ Combining multiple delta tuning methods is more effective than a single method under most cases, despite that the optimal combination may vary for different PLM backbones, downstream tasks, and data scales.
    \item \textbf{Power of Scale.} \ The power of scale (i.e., both the performance and convergence are improved when the PLM's size is increased) is observed in all of the delta tuning methods, even in unregulated neural modules. We provide a reasonable perspective to explain this phenomenon.
    %Especially when the model scale grows to a certain degree, even optimizing minimal random parameters in the PLM could still achieve competitive performance. Furthermore, we provide a reasonable perspective and theory to explain this phenomenon. 
    \item \textbf{Transferability.} \ Existing delta tuning methods could well support knowledge transfer, showing non-trivial transferability among downstream tasks of similar categories.
\end{itemize*}

At last, we discuss the applications of delta tuning from various perspectives (\refsec{applications}), including fast training and shareable checkpoints, multi-task learning, catastrophic forgetting mitigation, and in-batch parallel computing. We also discuss the broader impacts of the delta tuning technique in terms of fairness and energy cost (\hyperref[sec:impacts]{\textsc{§ impacts}}).
Hopefully, this paper could inspire research to advance the efficient use of large models.
%Hopefully, this paper could help beginners easily understand the motivation and importance of delta tuning, the basics of current methods and tools, as well as general empirical conclusions. For researchers already working on large-scale PLMs, we hope this paper will inspire further research to advance the efficient use of large models. 
The tools, codes, data splits, trained delta checkpoints in our experiments will be publicly available to facilitate future research\footnote{This paper focuses on pre-trained language models, however, the delta tuning technique can also be seamlessly transferred to other areas where large-scale neural models come into play, such as computer vision~\citep{rebuffi2017learning, perez2018film}}.

%Adapting pre-trained language models for specific downstream tasks has become a de-facto regime in NLP~\citep{devlin2018bert, HAN2021}. And fine-tuning is the predominant way to conduct such adaptations, which updates all the parameters of the model and produce separate fine-tuned instances for different tasks. 

%PLM with an increasing number of parameters has become an irreversible trend in the field, as there are constantly empirical results showing that larger models (plus more data) almost certainly lead to better performance. In this context, traditional full parameter fine-tuning becomes very difficult

\section{Preliminaries}
\label{sec:preliminaries}

Since almost all the mainstream PLMs are developed based on the Transformer~\citep{vaswani2017attention} model, and delta tuning usually carries out operations on Transformer models, this section gives preliminaries of the Transformer~\citep{vaswani2017attention} model and mainstream PLMs with different modeling strategies. 
For more details, please refer to original papers~\citep{vaswani2017attention, devlin2018bert, brown2020language, raffel2019exploring} or related surveys~\citep{HAN2021, liu2021pre, bommasani2021opportunities}. 

%formal description of the  delta tuning problem, by first introducing the technical details of the transformer neural network. Next, we introduce mainstream transformer-based PLMs and formally introduce the delta tuning problem.

\subsection{Transformer}
\label{subsec:transformer}

The Transformer model has become a key infrastructure for existing state-of-the-art PLMs. The original Transformer is proposed as an encoder-decoder model, where both the encoder and decoder are composed of a stack of identical blocks. Each Transformer layer consists of an attention layer and a fully-connected feed-forward neural network, while the decoder block contains an extra cross-attention layer on top of the self-attention layer to capture information from the encoder. Between each layer, there are residual connection~\citep{he2016deep} and layer normalization~\citep{ba2016layer} modules.

% \bibliographystyle{my}
% \bibliography{custom}

% \newpage
% \appendix
%\section{Tasks in Experiments}

%TODO 引用一下preliminary里的表达式
%TODO 维度的记号统一

\textbf{Attention Layer.} Attention layers are the key to the success of Transformer. It involves a query matrix $\mathbf{Q}\in \mathbb{R}^{n \times d_k}$, a key matrix $\mathbf{K}\in \mathbb{R}^{m \times d_k}$, and a value matrix $\mathbf{V}\in \mathbb{R}^{m \times d_v}$, where each row in the matrices corresponds to one sample and the $i$-th row in $\mathbf{K}$ and $\mathbf{V}$ together form a \textit{key-value} pair accordingly. The attention mechanism can be formally represented as
\begin{equation}
\begin{aligned}
    & \mathbf{H} = \text{ATT} (\mathbf{Q},\mathbf{K},\mathbf{V}) = \text{Softmax}(\frac{\mathbf{Q}\mathbf{K}^\top}{\sqrt{d_k}})\mathbf{V}.
\end{aligned}
\end{equation}

Intuitively, each row in $\mathbf{H} \in \mathbb{R}^{n \times d_v}$ is a weighted sum of row vectors in $\mathbf{V}$, while the weights are decided by the dot product of the query vector and the key matrix.
The specific attention adopted in the Transformer model is termed as \textit{self-attention}, as the three matrices $\mathbf{Q}, \mathbf{K}, \mathbf{V}$ are derived from the same feature matrix $\mathbf{X} \in \mathbb{R}^{n \times d}$ from the previous layer, parameterized by three weight matrices $\mathbf{W_q} \in \mathbb{R}^{d \times d_k}, \mathbf{W_k} \in \mathbb{R}^{d \times d_k}, \mathbf{W_v} \in \mathbb{R}^{d \times d_v}$ as follows:

\begin{equation}
\begin{aligned}
    \mathbf{Q} = \mathbf{X}\mathbf{W}_q, \mathbf{K} = \mathbf{X}\mathbf{W}_k, \mathbf{V} = \mathbf{X}\mathbf{W}_v 
\end{aligned}
\end{equation}

Moreover, Transformer uses multi-head self-attention with multiple sets of $\mathbf{Q}^{(i)}, \mathbf{K}^{(i)}, \mathbf{V}^{(i)}$, each set corresponding to a distinct set of weight matrix $\mathbf{W}_q^{(i)} \in \mathbb{R}^{d \times d_k}, \mathbf{W_k} \in \mathbb{R}^{d \times d_k}, \mathbf{W_v} \in \mathbb{R}^{d \times d_h}$, where $d_h$ is usually set to $\frac{d_v}{h}$. The final output $\mathbf{H} \in \mathbb{R}^{n \times d_o}$ is obtained by projecting the concatenation of a series of $\mathbf{H}_i$ into a new feature space with a new weight matrix $\mathbf{W}_o \in \mathbb{R}^{d_v \times d_o}$.  
\begin{equation}
\begin{aligned}
     \mathbf{H} &= \text{MH-ATT}  (\mathbf{Q},\mathbf{K},\mathbf{V}) \\
    &= \text{Concat}(\mathbf{H}_1,\ldots,\mathbf{H}_h)\mathbf{W}_o,\\
     \mathbf{H}_i &= \text{ATT}(\mathbf{Q}^{(i)}, \mathbf{K}^{(i)}, \mathbf{V}^{(i)}) \\
     &= \text{ATT}(\mathbf{X}\mathbf{W}_q^{(i)}, \mathbf{X}\mathbf{W}_k^{(i)}, \mathbf{X}\mathbf{W}_v^{(i)}),
\end{aligned}
\end{equation}

For decoder blocks, however, there is an additional mask operation that prevents query vectors from attending to the future positions yet to be decoded. Besides, there is an extra \textit{cross-attention} layer following the self-attention layer, where the query matrix $\mathbf{Q}$ is derived from the output of the previous layer in the decoder, and the key and value matrices $\mathbf{K}, \mathbf{V}$ are transformed from the output of the last layer of the encoder. It is designed to avoid foreseeing the true label while considering information from the encoder when decoding.

\begin{wrapfigure}{r}{0.39\textwidth}
\centering
\includegraphics[width=1\linewidth]{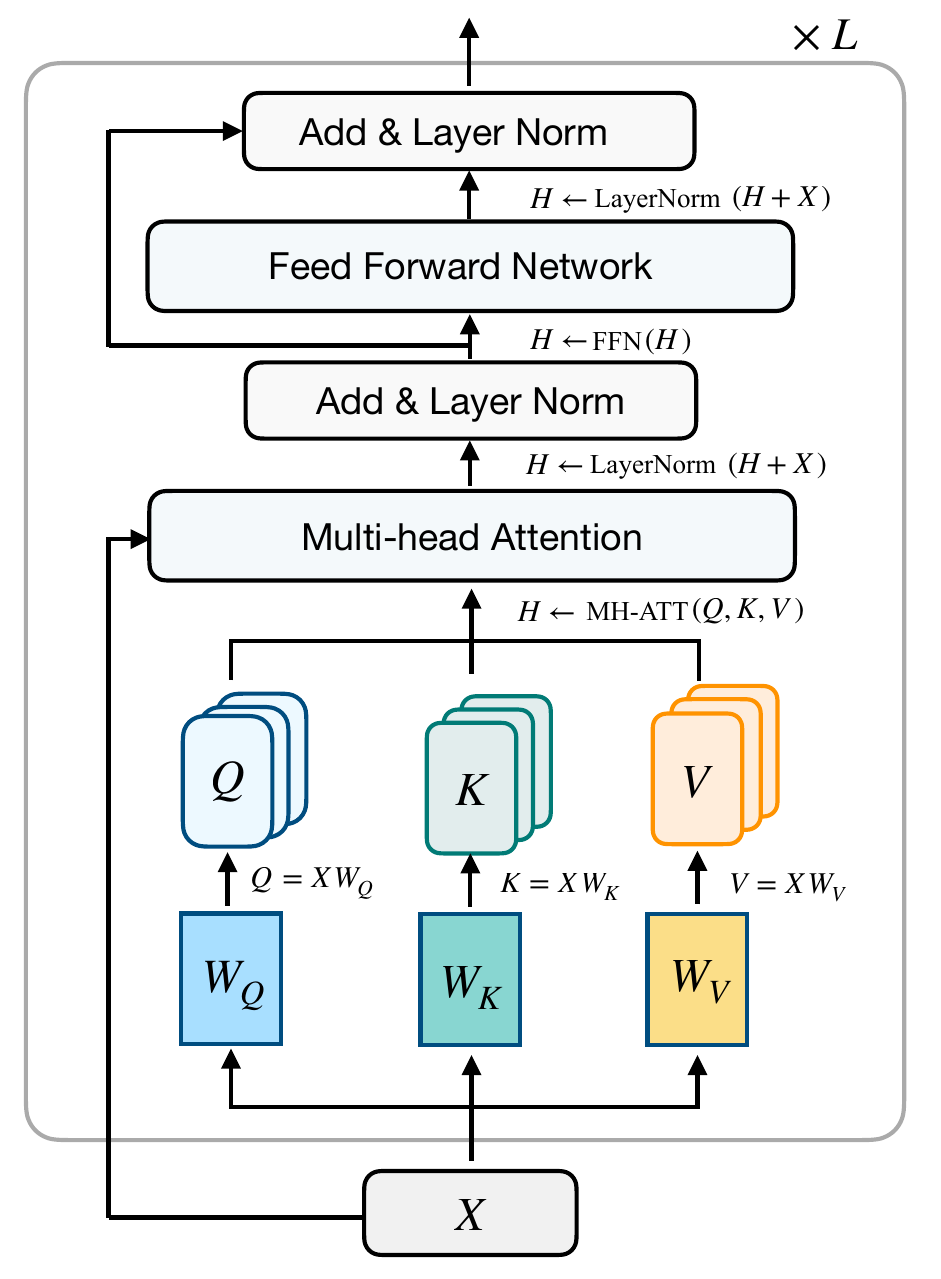}
\caption{An illustration of a Transformer block. Generally speaking, a delta tuning method could be applied to any positions in a Transformer model.}
\label{fig:geo}
\end{wrapfigure}

\textbf{Fully-connected Feed-Forward Layer.} The fully-connected feed-forward layer following the attention layer is composed of two linear transformation and a non-linear activation function. Denote the input matrix as $\mathbf{X} \in \mathbb{R}^{n \times d_i}$, the output of the feed-forward layer is
\begin{equation}
    \mathbf{F} = \text{FFN}{(\mathbf{X})} = \sigma(\mathbf{X}\mathbf{W}_1+\mathbf{b}_1)\mathbf{W}_2+\mathbf{b}_2,
\end{equation}
where $\sigma(\cdot)$ is the activation function (usually the ReLU function), and $\mathbf{W}_1 \in \mathbb{R}^{d_i \times d_m}$, $\mathbf{b}_1 \in \mathbb{R}^{d_m}$, $\mathbf{W}_2\in \mathbb{R}^{d_m \times d_o}$, $\mathbf{b}_2 \in \mathbb{R}^{d_o}$ are all learnable parameters. Empirically, $d_i$ is set equal to $d_o$, $d_m$ is set to be much larger than $d_i$ and $d_o$.

\textbf{Residual Connection and Normalization.} Following each attention layer and each feed-forward layer, residual connection and layer normalization are applied. They conduce to retaining information when the model is considerably deep and thus guarantees the model performance. Formally, given a neural layer $f(\cdot)$, the residual connection and normalization layer is defined as
\begin{equation}
    \text{A\&N}(\mathbf{X}, f) = \text{LayerNorm}(\mathbf{X}+f(\mathbf{X})),
\end{equation}
where $\text{LayerNorm}(\cdot)$ denotes the layer normalization operation, and $\text{A\&N}$ means ``add and norm''.

\textbf{Typical Transformer Layer.} As depicted in Figure \ref{fig:geo}, a typical transformer layer can be expressed as 
\begin{equation}\label{eq:transformer_arch}
\begin{aligned}
    \mathbf{M}&=\text{A\&N}(\mathbf{X},\mathbf{H})\\
    \mathbf{Y}&=\text{A\&N}(\mathbf{M},\mathbf{F}),
\end{aligned}
\end{equation}
where $\mathbf{M}$ is the intermediate representation after the attention block, and $\mathbf{Y}$ denotes the output of the layer with respect to input $\mathbf{X}$.

\subsection{Pre-trained Language Models}
\label{subsec:plms}
\begin{figure}[ht]
    \centering
    \includegraphics[width=0.95\textwidth]{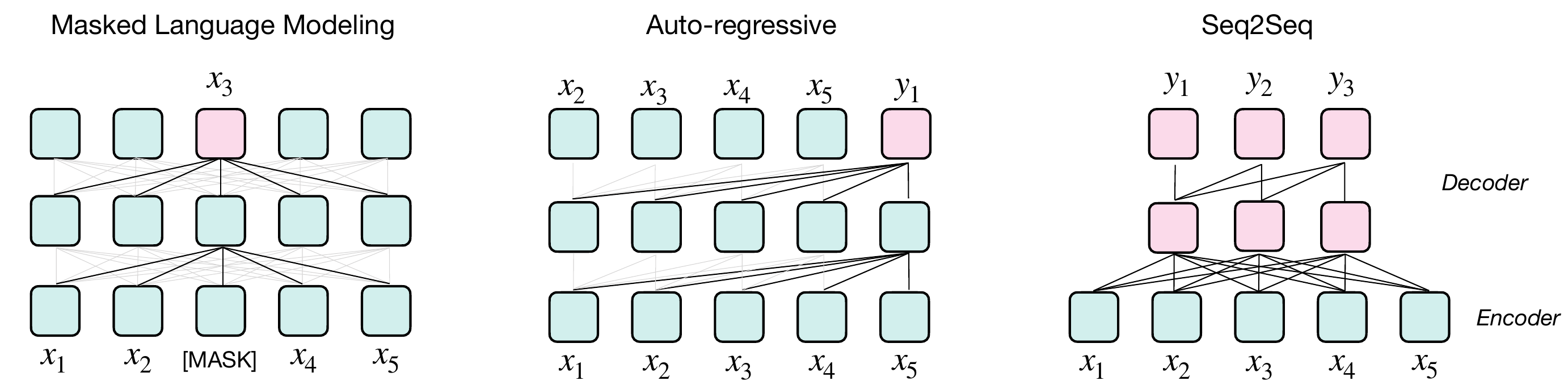}
    \caption{Different modeling strategies of PLMs.}
    \label{fig:plm}
\end{figure}

% bidirectional, transformer encoder. BERT, Roberta, XLNet...
% auto-regressive, transformer decoder. GPT, GPT-2
% seq2seq, encoder-decoder. T5, BART
Current pre-trained language models are almost consistently based on the Transformer model. However, they usually vary in the specific structure adopted (e.g. only using Transformer encoder or decoder, or both). This section briefly reviews some of the popular PLMs with respect to different modeling strategies (as shown in Figure~\ref{fig:plm}).

\paragraph{Masked Language Modeling.} The first group of PLMs are bidirectional models based on the Transformer encoder, among which BERT~\citep{devlin2018bert} is the most representative one. It is pre-trained with masked language modeling (MLM) task and next sentence prediction (NSP) task. When pre-training, the input is a pair of sequences, where special tokens [\texttt{CLS}] and [\texttt{SEP}] are added to the original input, and tokens are randomly replaced with [\texttt{MASK}] tokens. MLM loss seeks to maximize the conditional probability of label tokens at [\texttt{MASK}] position, as shown in equation~\eqref{eq:mlm}, where $M(\mathbf{x})$ contains all masked token positions. While the final representation of [\texttt{CLS}] is used to predict whether the two sentences are coherent. 
\begin{equation}
    \mathcal{L}_\text{MLM} = -\sum_{x_m \in M(\mathbf{x})}\log \mathbf{P}(x_m|\mathbf{x}_{\setminus M(x)})
    \label{eq:mlm}
\end{equation}
RoBERTa~\citep{liu2019roberta} is almost identical to BERT, except that it removes the NSP task, applies the more robust dynamic masking to the input, and is trained with larger batch sizes, the longer time, and more data. Bidirectional models are powerful in generating contextual representations of tokens and language understanding. \\
\paragraph{Auto-regressive Language Modeling.} Another set of PLMs are language models purely based on the Transformer decoder. They are also termed as \textit{auto-regressive} language models. The objective of language modeling (LM) is to model the probability of the given sequence by factorizing it into the probability of the $i$-th token given the previous tokens:
\begin{equation}
    \mathcal{L}_\text{LM} = - \log \mathbf{P}(\mathbf{x}) = - \sum_{i=1}^T \log \mathbf{P}(x_i |x_{<i}),
\end{equation}
where $x_0$ is a special token indicating the start of a sentence. It is natural to take the advantage of masking in the Transformer decoder to model the conditional probability. During pre-training, the final output at each position is further fed into a softmax layer to predict the next token. The most well-known models are GPT~\citep{radford2018improving} and GPT-2~\citep{radford2019language}, while GPT-2 is trained to be more robust to diverse tasks. The unidirectional characteristic of these models enables high-quality language generation. \\
\paragraph{Sequence to Sequence Modeling.} The last type of PLMs are sequence-to-sequence models built upon a complete Transformer architecture. Common models of this type include T5~\citep{raffel2019exploring} and BART~\citep{mike2019bart}. Both models adopt span-level corruption as the major pre-training task, i.e. to randomly replace a sequence of the text of arbitrary length with a single \textit{mask} token and ask the model to fill in the original tokens. It is also termed as \textit{Seq2Seq MLM loss}, whose objective is to maximize the probability of target sequence given a corrupted sequence:
\begin{equation}
    \mathcal{L}_\text{Seq2Seq MLM} = -\sum_{\mathbf{x}_{i:j} \in M(\mathbf{x})} \sum_{t=i}^j \log \mathbf{P}(x_t|\mathbf{x}_{\setminus M(\mathbf{x})}, \mathbf{x}_{i:t-1})
\end{equation}
where $M(\mathbf{x})$ contains all corrupted text spans and $\mathbf{x}_{i:j}$ is a single masked span. While BART requires a different fine-tuning paradigm to assist in the classification task, T5 unifies all tasks under the text-to-text generation paradigm. As a combination of both bidirectional encoder and auto-regressive decoder, sequence-to-sequence models are powerful in both language understanding and generation tasks.

%\subsection{Problem Statement}
%\label{subsec:problem}
% Given a pre-trained model $\Theta$ that is composed of a stack of Transformer layers $\{l_1, l_2, ..., l_N\}$, where $l_i=\text{A\&N}(\text{FFN}(\text{MH-ATT}(\cdot)))$, the objective is to produce a fine-tuned model $\Theta' = \{l_1', l_2', ..., l_N', \mathbf{W}\}$, where $\mathbf{W}$ represents the newly-added parameters and is optional. Empirically, $|\mathbf{W}| \ll |\Theta|$, where $|\cdot|$ denotes the number of parameters. Depending on tuning strategy, most parameters in original Transformer layers $l_i$ may stay fixed, while others may be updated or replaced.

\section{Delta Tuning}
\label{sec:delta tuning}

Given a pre-trained model $\Theta=\{w_1, w_2, ..., w_N\}$ and training data $\mathcal{D}$, the objective of PLM adaptation is to produce the adapted model $\Theta'=\{w_1', w_2', ..., w_M'\}$. Define $\Delta \Theta=\Theta'-\Theta$ as the operation on top of the original model $\Theta$. In vanilla fine-tuning, $N=M$ and $\Delta \Theta = \nabla f_\Theta(\mathcal{D})$ is the update value of all parameters in $\Theta$ with respect to training data. While in delta tuning, $\Delta \Theta$ refers to modification of a small number of parameters. Empirically, $|\Delta \Theta|=|\Theta|$ in vanilla fine-tuning, while for delta tuning, $|\Delta \Theta| \ll |\Theta|$, where $|\cdot|$ indicates the number of parameters involved.

\begin{figure}[ht]
    \centering
    \includegraphics[width=0.78\textwidth]{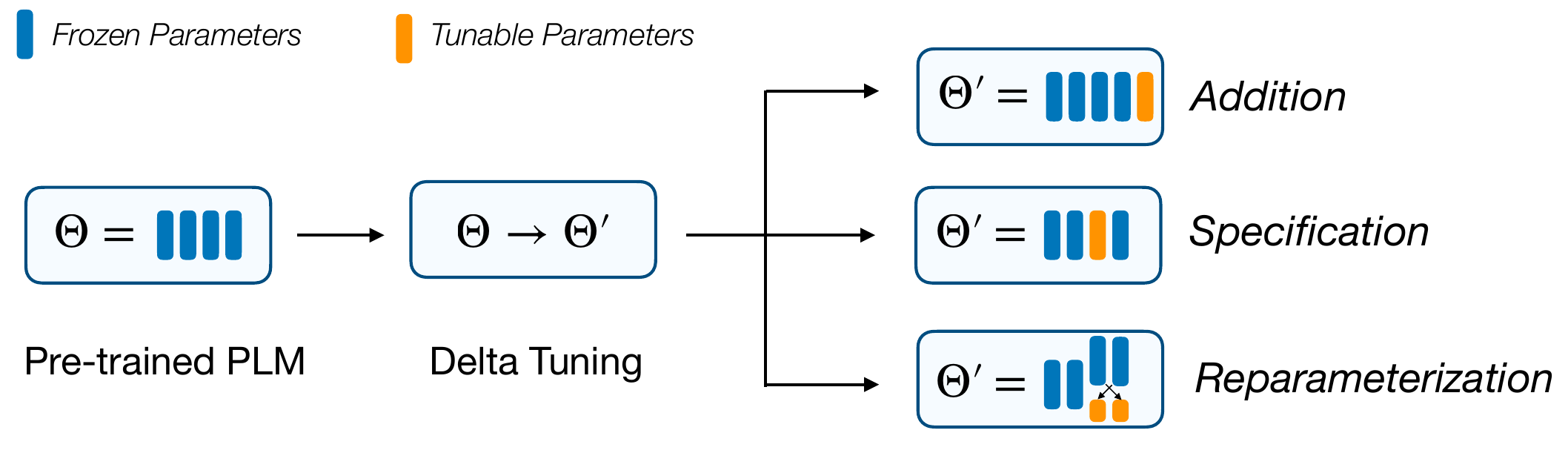}
    \caption{The categorization criterion of delta tuning, where $\Theta$ denote the pre-trained parameters, and $\Theta'$ represent the well-tuned parameters.}
    \label{fig:cate}
\end{figure}

To organize them under a unified framework, we categorize the delta tuning methods into three groups according to the operations on the delta parameters (as illustrated in Figure~\ref{fig:cate}): addition-based, specification-based, and reparameterization-based approaches. 

\begin{itemize*}
    \item \textbf{Addition-based} methods introduce extra trainable neural modules or paramters that do not exist in the original model or process. In addition-based methods, $M\geq N$ and $\Delta \Theta = \{w_{N+1}, w_{N+2}, ..., w_{M}\}$.
    \item \textbf{Specification-based}  methods specify certain parameters in the original model or process become trainable, while others are frozen. Denote the set of trainable parameters as $\mathcal{W}$, then $\Delta \Theta=\{\Delta w_1, \Delta w_2, ..., \Delta w_N\}$. When $w_i \in \mathcal{W}$, $\Delta w_i$ is the incremental value from $w_i$ to $w_i'$, else, $\Delta w_i=0$.
    %$\Delta w_i=\nabla_{w_i} f_{\Theta}(D)$ if $w_i \in \mathcal{W}$, else, $\Delta w_i=0$.
    \item \textbf{Reparameterization-based} methods reparameterize existing parameters to a parameter-efficient form by transformation. Denote the set of parameters to be reparameterized as $\mathcal{W}$, and suppose that each $w_i \in \mathcal{W}$ is reparameterized with new parameters $R(w_i) = \{u_1, u_2, ..., u_{N_i}\}$, then $\Delta \Theta=(\Theta \setminus \mathcal{W}) \cup \mathcal{U}$, where $\mathcal{U}=\{u_j|\exists w_i \in \mathcal{W}, u_j \in R(w_i)\}$.
\end{itemize*}

\subsection{Addition-based Methods}
\label{sec:addition}
With the above definition in mind, addition-based methods introduce additional parameters to the neural network. In this section, we introduce two branches of representative addition-based methods, \textit{adapter-based tuning} and \textit{prompt-based tuning}.

\paragraph{Adapters-based Tuning.} As a seminal work in delta tuning, adapter-based methods inject small-scale neural modules (adapters) to the Transformer layers and only tune these adapters for model adaptation. Although such a strategy leaves an open choice of adapter structures, a simple instantiation~\citep{houlsby2019parameter} achieves impressive performance and has become the most widely used baseline in recent research. 
Specifically, one adapter module contains a down-projection and an up-projection. For an input feature $\mathbf{h} \in \mathbb{R}^d$, a down-projection projects the input to a $r$-dimensional space with a parameter matrix $\mathbf{W}_d \in \mathbb{R}^{d \times r}$, 
after which a nonlinear function $f(\cdot)$ is applied. Then the up-projection $\mathbf{W}_u$ maps the $r$-dimensional representation back to $d$-dimensional space. Added with a residual connection, the complete computation could be written as 
 \begin{align}
    \mathbf{h} \leftarrow  f(\mathbf{h} \mathbf{W}_d) \mathbf{W}_u + \mathbf{h}. \label{adapter-computation}
 \end{align}

In each block, the adapter modules are separately inserted after the multi-head self-attention and the feed forward network sublayers, which reduces the tunable parameters per layer to $ 2\times (\ 2dr\ \text{(projection matrices)} + d\  \text{(residual connection)} + r\ \text{(bias term)})$. Practically, 0.5\%$\sim$8\% parameters of the whole model~\citep{houlsby2019parameter} could be involved in tuning process under such strategy.

% Structural improvements to the adapters have proven to have little impact on the results, so the derivative work focuses on exploring the scenarios and specific nature of the use of such a strategy.

Although adapter works with significantly fewer tunable parameters than vanilla fine-tuning, some work attempts for a more rigorous saving strategy by introducing inductive biases into the structure of the adapter layer. For example, Compacter~\citep{mahabadi2021compacter} propose to use a combination of hypercomplex multiplication and parameter sharing. The hypercomplex multiplication parameterizes the original linear layer as the sum of the Kronecker products of two small matrices. Taking the down-projection as an example, 
\begin{align}
    \mathbf{W}_d = \sum_{i=1}^{n} \mathbf{A}_{i}\otimes \mathbf{B}_i, \text{where} \ \mathbf{A}\in\mathbb{R}^{n\times n} , \mathbf{B}\in\mathbb{R}^{\frac{d}{n}\times \frac{r}{n}}
\end{align}

 Their method reduces the number of parameters in the adapter layer to $\frac1n$ without harming the performance, where $n$ is the number of divisions of the linear layer. It also shows that a simple low-rank decomposition of the linear layer leads to comparable performance with the adapter layer, i.e.,
\begin{align}
    \mathbf{W}_d = \mathbf{A}\mathbf{B}^{T}, \text{where} \  \mathbf{A}\in\mathbb{R}^{d\times n} ,\mathbf{B}\in\mathbb{R}^{r\times n} \text{and}\ n\ll \operatorname{min}(d, r).
\end{align}

As an addition-based approach, adapter-based tuning has the advantage of placing multiple adapter instances on a pre-trained model simultaneously, which can benefit many application scenarios. For example, multi-task learning~\citep{stickland2019bert, Mahabadi2021ParameterefficientMF} is an advantageous setting for adapter-based methods, inserted with adapter modules in parallel with the self-attention module, PLMs could demonstrate impressive representational capacity in the multi-task setting.
In contrast to directly conducting multi-task learning on adapters, adapterFusion~\citep{pfeiffer2020adapterfusion} first pre-train task-specific adapters and then combine the representations of the pre-trained adapters to leverage the cross-task knowledge and enhance the performance of transfer learning. 

In terms of computational efficiency, the training of adapters 
could be 60\% faster than vanilla fine-tuning while the inference is only 4\%-6\% slower. And the computational cost could be further reduced dynamically by removing adapters from lower transformer layers~\citep{ruckle-etal-2021-adapterdrop}. Research also shows that adapter-based fine-tuning demonstrates better robustness than fine-tuning. Specifically, adapter-based fine-tuning could perform better than vanilla fine-tuning on few-shot and cross-lingual scenarios~\citep{he2021effectiveness} and is more robust under adversarial attacking~\citep{han2021robust}. We provide a comparison of different adapters, as well as other delta tuning methods in Table~\ref{tab:different_adapters}.

To sum up, adapters are lightweight additional neural modules that could be trained in a task-specific style, which could be regarded as ``encapsulation'' of task information (in fact, this perspective can be applied to all the ``deltas''). Although in an ideal world, adapters could be freely shared and reused by researchers, in practice, sharing and reusing such modules face substantial obstacles. Taking the first step, AdapterHub~\citep{pfeiffer2020adapterhub} provides a feasible platform and toolkit to deploy adapters inside the transformer-based models. 
%Adapters

\label{sec:prompt}
%Prefix and prompt
\paragraph{Prompt-based Tuning.} Instead of injecting neural modules to the Transformer model, prompt-based methods wrap the original input with additional context. As a strategy to stimulate pre-training language models by mimicking pre-trained objectives in the downstream tasks, prompt-based learning has achieved promising performance in various NLP tasks~\citep{Gao2021MakingPL, Hu2021KnowledgeablePI, tan2021msp}, especially in low-data settings~\citep{Scao2021HowMD}. The introduction of the technique and implementations of prompt-based learning have already been comprehensively presented in other literature~\citep{liu2021pre, ding2021openprompt}. In this paper, we primarily focus on the parameter-efficient attribute of prompt-based learning (only prefixes or prompts are optimized) and pay less attention to the settings where the models and prompts are simultaneously optimized.

An important seminal work of this branch of research is prefix-tuning~\citep{li2021prefix}, which prepends trainable continuous tokens (prefixes) to the input and hidden states of each Transformer layer. Each prefix is drawn from a newly initialized trainable parameter matrix $\mathbf{P}$, while other parameters of the pre-trained model remain unchanged during training. During generation, if an activation $h_i$ is in a prefix position, it is the direct copy of the corresponding trainable parameter; otherwise, the activation is computed by the model as $h_i = \text{LM}(z_i, h_{<i})$. It is worth noting that the paradigm could be applied to both autoregressive and encoder-decoder models. \cite{liu2021p} demonstrate that such a strategy could be effectively applied to natural language understanding (NLU) with different scales of models.

Compared to prefix-tuning which adds tunable prefixes to every intermediate Transformer layer, prompt tuning~\citep{lester2021power} proposes a more simplified strategy that only adds soft prompts to the input layer. Similar to prefix-tuning, the newly introduced prompts are not parameterized by the pre-trained model but an additional parameter matrix. And during training, the parameters of soft prompts are updated by gradient descent while the model parameters keep frozen. As the model size increases, the performance gap between prompt-tuning and full parameter fine-tuning is narrowed. Particularly, when the model scales to T5-XXL with 11 billion parameters, prompt tuning yields comparable performance on SuperGlue with fine-tuning. This strategy also exhibits sensitivity to the length and initialization of the soft prompts. Prompts could also be injected in the pre-training stage to seek a satisfying initialization point~\citep{gu2021ppt}. Moreover, similar to other methods, prompt tuning also demonstrates transferability across tasks~\citep{vu2021spot,su2021transferability}, which suggests that appropriate initialization could be substantially beneficial for downstream tasks.

\textbf{The Training Curse of Prompt-based Methods}\quad Although prompt-based methods exhibit a promising future for the adaptation of large pre-trained models, especially that prompt tuning does not need to modify anything inside the neural network, there still exist unsolved challenges. 
In practice, prompt tuning is difficult to optimize, and generally, this phenomenon becomes more apparent as the volume of data and the size of the model decreases. Even though soft prompts can be trained successfully, they converge significantly slower than full parameter fine-tuning and other delta tuning methods during training. In our experiments, we validate the phenomenon across different datasets (\refsec{performance}), indicating that it is an interesting topic to train soft prompt to converge stably in various situations. 

\begin{table*}[ht]
\renewcommand\arraystretch{2}
  \small
  \centering
  \caption{Comparison between different delta tuning methods, we use the \textcolor{PineGreen}{green} color to denote tunable parameters and modules. 
  $[:]$ is the concatenation operation; $d_h$ means the hidden dimension of transformer model; $d_m$ is the intermediate dimension between down projection and up projection, where $d_m$ is far smaller than $d_h$. \textsc{Compacter} utilize hypercomplex matrix multiplication and low-rank decomposition to reduce the amount of parameters; \textsc{AdapterDrop} randomly dropout adapters in the first $n$ layers and also bring down back-propagation time; \textsc{Prefix-Tuning} add prefix of $n$ past key value.}
  \vspace{0.2cm}
  \scalebox{0.90}{
    \begin{tabular}{l|l|l}
    \toprule
    \textbf{Name \& Refs} & \textbf{Method} & \#\textbf{Params} \\
    \midrule
    \makecell[l]{\textsc{Sequential Adapter} \\ \cite{houlsby2019parameter}} & \multirow{3}{*}{\makecell[l]{$\text{LayerNorm}(\mathbf{X}+\mathbf{H}(\mathbf{X})) \to \text{LayerNorm}(\mathbf{X}+\textcolor{PineGreen}{\text{ADT}}(\mathbf{H}(\mathbf{X})))$
    \\ \\
    $\text{LayerNorm}(\mathbf{X}+\mathbf{F}(\mathbf{X}))\to\text{LayerNorm}(\mathbf{X}+\textcolor{PineGreen}{\text{ADT}}(\mathbf{F}(\mathbf{X})))$ 
    \\ \\ $\textcolor{PineGreen}{\text{ADT}}(\mathbf{X}) = \mathbf{X}~+~\sigma(\mathbf{X}\textcolor{PineGreen}{\mathbf{W}_{d_h\times d_m}})\textcolor{PineGreen}{\mathbf{W}_{d_m\times d_h}}, ~~\sigma=\text{activation}$}} & $L \times 2 \times (2 d_h d_m)$ \\
    \makecell[l]{\textsc{Compacter} \\ \cite{mahabadi2021compacter}} & & $L\times 2\times (2(d_h+d_m))$  \\
    \makecell[l]{\textsc{AdapterDrop} \\ \cite{ruckle-etal-2021-adapterdrop}} & & $(L-n)\times 2\times(2d_h d_m)$  \\
    \midrule
    \makecell[l]{\textsc{Parallel Adapter} \\ \cite{he2022unified}} & \makecell[l]{$\text{LayerNorm}(\mathbf{X}+\mathbf{H}(\mathbf{X})) \to \text{LayerNorm}(\mathbf{X}+\textcolor{PineGreen}{\text{ADT}}(\mathbf{X})+\mathbf{H}(\mathbf{X}))$
    \\ \\
    $\text{LayerNorm}(\mathbf{X}+\mathbf{F}(\mathbf{X}))\to\text{LayerNorm}(\mathbf{X}+\textcolor{PineGreen}{\text{ADT}}(\mathbf{X})+\mathbf{F}(\mathbf{X}))$ 
    \\ \\
    $\textcolor{PineGreen}{\text{ADT}}(\mathbf{X}) = \sigma(\mathbf{X}\textcolor{PineGreen}{\mathbf{W}_{d_h\times d_m}})\textcolor{PineGreen}{\mathbf{W}_{d_m\times d_h}}, ~~\sigma=\text{activation}$} & $L \times 2 \times (2d_h d_m)$ \\
    %\midrule
    %\makecell[l]{\textsc{PALs} \\ \cite{stickland2019bert}} & \makecell[l]{$\text{LayerNorm}(\mathbf{M}+\mathbf{F}(M))\to \text{LayerNorm}(\mathbf{X}+\mathbf{F}(\mathbf{M})+\textcolor{PineGreen}{\text{ADT}}(\mathbf{X}))$ \\ \\$\textcolor{PineGreen}{\text{ADT}}(\mathbf{X}) = \textcolor{PineGreen}{f}(\mathbf{X}\textcolor{PineGreen}{\mathbf{W}_{d_h\times d_m}})\textcolor{PineGreen}{\mathbf{W}_{d_m\times d_h}},~~\textcolor{PineGreen}{f}=\textcolor{PineGreen}{\text{MH-ATT}}$} & $2d_h d_m + L * 3d_m^2$ \\
    \midrule
    \makecell[l]{\textsc{AdapterBias} } & \makecell[l]{$\text{LayerNorm}(\mathbf{X}+\mathbf{F}(\mathbf{X}))\to\text{LayerNorm}(\textcolor{PineGreen}{\text{ADT}}(\mathbf{X})+\mathbf{F}(\mathbf{X}))$
    \\ \\
    $\textcolor{PineGreen}{\text{ADT}}(X) = \mathbf{X}\textcolor{PineGreen}{\mathbf{W}_{d_h\times1}} \textcolor{PineGreen}{\mathbf{W}_{1\times d_h}}$} & $L\times2\times d_h$ \\ \midrule
    \makecell[l]{\textsc{Prefix-Tuning} \\  \cite{li2021prefix}} & \makecell[l]{$\mathbf{H}_i = \text{ATT}(\mathbf{X}\mathbf{W}_q^{(i)}, [\textcolor{PineGreen}{\text{MLP}_k^{(i)}(\mathbf{P}'_k)}:\mathbf{X}\mathbf{W}_k^{(i)}], [\textcolor{PineGreen}{\text{MLP}_v^{(i)}(\textbf{P}'_v)}:\mathbf{X}\mathbf{W}_v^{(i)}])$ \\ \\ $\textcolor{PineGreen}{\text{MLP}^{(i)}}(\mathbf{X})=\sigma(\mathbf{X}\textcolor{PineGreen}{\mathbf{W}_{d_m\times d_m}})\textcolor{PineGreen}{\mathbf{W}^{(i)}_{d_m\times d_h}}$ \\ \\ $\textcolor{PineGreen}{P'}=\textcolor{PineGreen}{\mathbf{W}_{n\times d_m}}$} & \makecell[l]{$n\times d_m+d_m^2$ \\ + $L \times 2 \times d_h d_m$} \\
    \midrule
    \makecell[l]{\textsc{LoRA} \\ \cite{hu2021lora}} & \makecell[l]{$\mathbf{H}_i = \text{ATT}(\mathbf{X}\mathbf{W}_q^{(i)}, \textcolor{PineGreen}{\text{ADT}_k}(\mathbf{X})+\mathbf{X}\mathbf{W}_k^{(i)}, \textcolor{PineGreen}{\text{ADT}_v}(\mathbf{X})+\mathbf{X}\mathbf{W}_v^{(i)})$ \\ \\
    $\textcolor{PineGreen}{\text{ADT}}(\mathbf{X}) = \mathbf{X}\textcolor{PineGreen}{\mathbf{W}_{d_h\times d_m}\mathbf{W}_{d_m\times d_h}}$} & $L \times 2 \times (2d_h d_m)$\\
    %\midrule
    %\makecell[l]{\textsc{HYPERFORMER} \\ \cite{karimi-mahabadi-etal-2021-parameter}} & & \\
    \midrule
    \makecell[l]{\textsc{BitFit}\\\cite{zaken2021bitfit}} & $f(\mathbf{X})\to f(\mathbf{X})+\textcolor{PineGreen}{\mathbf{B}}, ~~\text{for all function }f$ & $L\times (7\times d_h + d_m)$\\
    
    %\midrule
    %\makecell[l]{\textsc{AdapterFusion} \\ \cite{pfeiffer2020adapterfusion}} & \makecell[l]{$\text{LayerNorm}(\mathbf{M}+\mathbf{F}(M))\to$}& \\
    \bottomrule
    \end{tabular}
  }
  \label{tab:different_adapters}
\end{table*}

\subsection{Specification-based Methods}
\label{sec:specification}
Specification-based methods fine-tune a few inherent parameters while leaving the majority of parameters unchanged in model adaptation. This approach does not seek to change the internal structure of a model but to optimize a small number of internal parameters to solve particular tasks. Generally, such specifications could be implemented based on heuristics or training supervision.

\paragraph{Heuristic Specification.} Specification-based methods do not introduce any new parameters in the model, but directly specify part of the parameters to be optimized. The idea is simple but surprisingly effective, \citet{lee2019would} only fine-tune one-fourth of the final layers of BERT and RoBERTa and could produce 90\% of the performance of full parameter fine-tuning.
BitFit~\citep{zaken2021bitfit} empirically proves that by only optimizing the bias terms inside the model and freezing other parameters, the model could still reproduce over 95\% performance on several benchmarks. Empirical results in BitFit also show that even if we use a small random set of parameters for delta tuning (which obviously will degrade the performance), the model could still yield passable results on the GLUE benchmark. Unfortunately, the work only applies this trick to small-scale models, and there is no guarantee that randomly choosing some parameters to be tuned would remain competitive for larger models. Another valuable observation is that different bias terms may have different functionalities during model adaptation.

%The decrease of generalization gap is also observed in BitFit.

%Such a conclusion may change as the model scaling
%Unfortunately, the work only apply this setting on small-scale models, the conclusion
%in our experiments, we find that when the amount of model parameters increases to 10b, the performance of the random setting will substantially increase and the gap with full parameter fine-tuning becomes very small (in \cref{sec:exp_scale}). This suggests that quantitative change may lead to qualitative change, when the model is extremely over-parameterized, the design of delta tuning becomes less important. 

\paragraph{Learn the Specification.} Rather than manually or heuristically specify which parameters to be updated, one alternative is to ``learn'' such specifications. Following the definition in~\refsec{delta tuning}, diff pruning~\citep{guo2021parameter} reparameterizes the fine-tuned model parameters $\Theta'$ as the summation of the pre-trained parameters $\Theta$ and the difference vector $\Delta \Theta$, i.e., $\Theta' = \Theta + \Delta \Theta$, where $\left| \Theta  \right| = \left| \Theta' \right|$. Hence, the key issue is to encourage the difference vector to be as sparse as possible, this work regularizes the vector by a differentiable approximation to the $L_0$-norm penalty to achieve the goal of sparsity. Practically, because new parameters to be optimized are introduced in the learning phase, diff pruning takes up more GPU memory than full parameter fine-tuning, which may establish barriers in the application on large PLMs. The masking method~\citep{zhao2020masking} learns selective masks for PLMs to only update the critical weights for particular tasks. 
To learn such a set of masks, a binary matrix associated with the model weights is introduced, where each value is generated by a thresholding function. During back-propagation, the matrix is updated by a noisy estimator.

%In adapter-based and prompt-based methods, the ``delta parameters'' are additionally inserted to (or before) the language model. In this section, we detail another group of delta-tuning strategies, difference-based methods, which use the ``replace'' operation to produce delta parameters rather than adding new ones. In this case, we take the parameters of a pre-trained model $\Theta$ as the start point and the parameters of the fine-tuned model $\tilde{\Theta}$ as the endpoint. The basic idea of this branch is to efficiently learn the difference $\Delta \Theta$ between $\Theta$ and $\tilde{\Theta}$. 

\subsection{Reparameterization-based Methods}
\label{sec:reparameterization}
Reparameterization-based methods transform the adaptive parameters during optimization into parameter-efficient forms. This branch of delta tuning is typically motivated by the hypothesis that PLM adaptations towards most downstream tasks are inherently low-rank, and could thus be equivalently completed in a parameter-efficient way. 

\paragraph{Intrinsic Dimensions of PLM Adaptation.} \citet{aghajanyan2020intrinsic} empirically show that the full-parameter fine-tuning process of pre-trained models can be reparameterized into optimization within a low-dimensional subspace, i.e., fine-tuning has a low \textit{intrinsic dimension}~\citep{li2018measuring}, which measures the minimum number of parameters needed to reach satisfactory performance. In experiments, they find that a relatively low-dimensional (e.g., thousands) reparameterization could achieve over $85$\% fine-tuning performance. In this sense, PLMs may serve as general compression frameworks, which compress the optimization complexity from high dimensions to low dimensions. They also demonstrate that, larger PLMs generally have smaller intrinsic dimensions, and the process of pre-training implicitly reduces PLM's intrinsic dimension. Taking inspiration from these observations, reparameterization-based delta tuning methods are proposed, which reparameterize (a part of) original model parameters with low-dimensional proxy parameters and only optimize the proxy parameters and thus reduce the computation and memory cost.

\paragraph{Intrinsic Rank of Weight Differences.}
Inspired by \citet{aghajanyan2020intrinsic}, LoRA~\citep{hu2021lora} hypothesizes that the change of weights during model tuning has a low \textit{intrinsic rank}. Based on this hypothesis, they propose to optimize the low-rank decomposition for the change of original weight matrices in the self-attention modules. In deployment, the optimized low-rank decomposition matrices are multiplied to obtain the delta of self-attention weight matrices. In this way, LoRA could match the fine-tuning performance on the GLUE benchmark. They demonstrate the effectiveness of their methods on PLMs of various scales and architectures.

\begin{figure}[ht]
    \centering
    \includegraphics[width=0.99\textwidth]{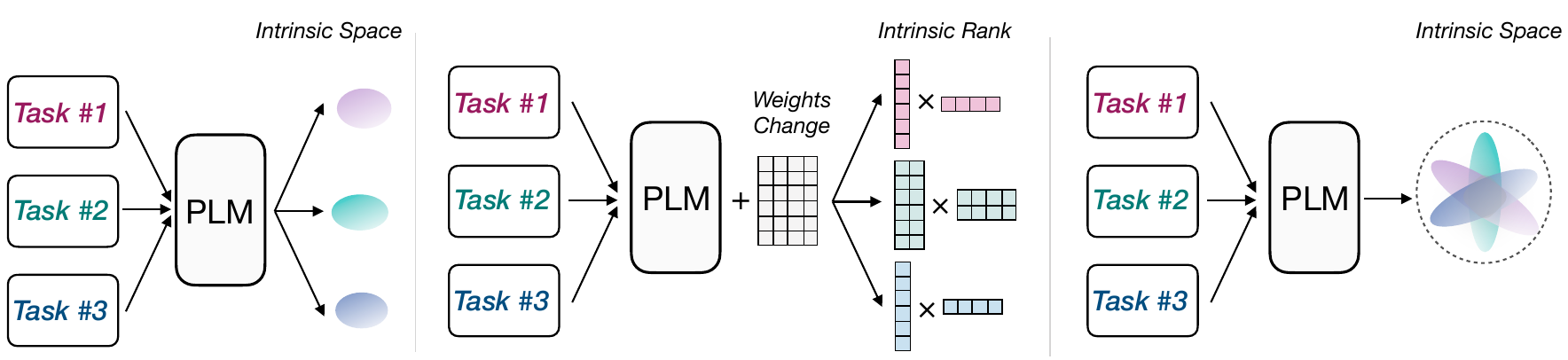}
    \caption{Conditioned on a PLM, \citet{aghajanyan2020intrinsic} hypothesize that there exist a low-dimensional intrinsic subspace that could reparameterize one specific fine-tuning process (the left part). \citet{hu2021lora} hypothesize that the change of weights during adaptation has a low intrinsic rank (the middle part). 
    And~\citet{qin2021exploring} hypothesize there may exist a common intrinsic space that could handle the fine-tuning for various NLP tasks (the right part).}
    \label{fig:intrinsic}
\end{figure}

\paragraph{Intrinsic Space of Multiple Adaptations.}
Furthermore, \citet{qin2021exploring} make a stronger hypothesis that the adaptations to multiple tasks could be reparameterized into optimizations within the same low-dimensional intrinsic subspace. Instead of resorting to a random subspace~\citep{aghajanyan2020intrinsic}, they try to find a common subspace shared by various NLP tasks, which is implemented through decomposing the trained soft prompts of multiple NLP tasks into the same low-dimensional nonlinear subspace, and then learn to adapt the PLM to unseen tasks or data by only tuning parameters in the subspace. Experiments show that in a $250$-dimensional subspace found with $100$ random tasks, by only tuning $250$ free parameters, $97$\% and $83$\% of the full prompt tuning performance can be recovered for $100$ seen tasks (using different training data) and $20$ unseen tasks, respectively. This provides strong evidence for their universal reparameterization hypothesis and may inspire future work. Moreover, \citet{qin2021exploring} also shows that the low-dimensional reparameterization can significantly improve the stability of prompt tuning. Their method could also be leveraged as a tool for analyzing the similarity and differences for various NLP tasks. The motivation differences of the above three works are visualized in Figure~\ref{fig:intrinsic}.

\section{Theoretical Perspectives of Delta Tuning}
\label{sec:theory}

Are these methods essentially doing the same thing? We are interested in the theoretical principles behind delta tuning.
 A pre-trained language model usually can be easily adapted to almost any downstream tasks with only a very small cost (compared to pre-training), and this phenomenon leads to theoretical issues that are worth exploring in depth. In this section, we propose two frameworks to introduce theoretical insights of delta tuning from the perspectives of optimization (\refsec{optimization}) and optimal control (\refsec{optimal control}).

\subsection{Optimization Perspective for Delta Tuning}
\label{sec:optimization}

 The delta tuning technique seeks to tune a small portion of parameters so as to match the performance of the fine-tuning in the original large language model while reducing the memory footprint. From the perspective of optimization, we analyze the effects of delta tuning and discuss the designs of several delta tuning methods under the low dimension assumption.
 
 Let $\mathcal{F}(\theta)$ denote the objective function of the original language model. Then the new objective function optimized by delta tuning is 
 $ \tilde{\mathcal{F}}(\theta, \delta)$. Here $\theta$ denotes the parameters of the original language model, and $\delta$ denotes the specific parameters tuned by delta tuning \footnote{The variables $\theta$ and $\delta$ correspond to the concepts of $\Theta$ and $\Delta\Theta$ in \refsec{delta tuning}. Also, $\delta$ is not necessarily independent of $\theta$.}. The starting point is $(\theta_0,\delta_0)$, where $\theta_0$ is the pre-trained language model parameters and $\delta_0$ is the initialization of $\delta$. In principle, though one may adopt some initialization of $\delta$ to facilitate the training process of delta tuning, there still exists the $\delta_0$ such that
 \begin{equation}
  \tilde{\mathcal{F}}(\theta,\delta_0) = \mathcal{F}(\theta),  \label{eq:start_points}
 \end{equation}
 which guarantees that $\tilde{\mathcal{F}}$ is identical to $\mathcal{F}$ if the delta tuning is disabled. 
 %The \cref{eq:start_points} is used to analyze the approximation of delta tuning to the full-parameter fine-tuning on the original language model. 
 Thus, the following relations hold:
 \begin{align}
  \min_{\theta, \delta}\tilde{\mathcal{F}}(\theta, \delta) \leq \min_\theta \tilde{\mathcal{F}}(\theta,\delta_0) = \min_{\theta}\mathcal{F}(\theta), \quad
  \min_{\theta, \delta}\tilde{\mathcal{F}}(\theta, \delta) \leq \min_{\delta}\tilde{\mathcal{F}}(\theta_0,\delta),
 \end{align}
 which suggests that simultaneously tuning $\theta$ and $\delta$ may be beneficial. Nonetheless, we are only interested in analyzing the case that either $\theta$ or $\delta$ is fine-tuned. 
 Let $\theta^{+} = \arg\min_{\theta}\tilde{\mathcal{F}}(\theta,\delta_0)$ and 
  $\delta^{+} = \arg\min_{\delta}\tilde{\mathcal{F}}(\theta_0,\delta)$.  %We assume that $\tilde{\mathcal{F}}$ is differentiable and denote the gradients of $\tilde{\mathcal{F}}$ with respect to $\theta$, $\delta$ as $\nabla_\theta \tilde{\mathcal{F}}$, $\nabla_\delta \tilde{\mathcal{F}}$, respectively.  
 The delta tuning essentially does no harm to the tuning of the original model under some conditions.  For instance,
  assuming that $\tilde{\mathcal{F}}$ is twice Lipschitz continuously differentiable, it can be proved that 
 \begin{align}  
  \vert\tilde{\mathcal{F}}(\theta^+,\delta_0) -\tilde{\mathcal{F}}(\theta_0,\delta^+) \vert = \mathcal{O}(\|\theta^+-\theta_0\|_2^2 + \|\delta^+-\delta_0\|_2^2),
 \end{align}
 in a local small region around $(\theta^+,\delta_0)$ and $(\theta_0,\delta^+)$. For a  sufficiently good starting point, the error bound holds. However,
 to guarantee the effectiveness of delta tuning, it is essential to exploit the problem structures to design $\tilde{\mathcal{F}}$. The intuition is to leverage the intrinsic low dimensions of the problems. Basically, there are two approaches that turn out to be useful in practice:
 \begin{itemize*}
  \item The solution is updated in  a lower dimensional subspace;
  \item The objective function (with constraints) is approximated in a certain smaller functional subspace. 
 \end{itemize*}
 For the applications in deep learning, the optimization of the objective function often has lots of local minimizers due to over-parameterization. Therefore, these approaches typically work well when the starting point is close to a local minimizer, where only some searching directions matter or the objective function can be well approximated by some simpler function in the trust region. Moreover, the small dimensional optimization can lead to a more efficient and more stable training process.
 
 \paragraph{Low dimensional representation in solution space.} As it is observed that the optimization trajectory of $\theta$ approximately follows a manifold \citep{aghajanyan2020intrinsic}, we can embed the hidden manifold to a low dimensional space of $\delta$, i.e., 
 %suppose that there exists a low dimensional representation that 
  $\theta = \psi(\delta)+\epsilon$, where $\epsilon$ is the error term depending on $\theta_0, \theta^+$. 
 Then, 
 \begin{equation}
  \tilde{\mathcal{F}}(\theta,\delta_0) = \mathcal{F}(\theta),  \quad 
  \tilde{\mathcal{F}}(\theta_0,\delta) = \mathcal{F}(\psi(\delta)). 
 \end{equation}
 If $\epsilon = 0$, the delta tuning finds the exact solution of the fine-tuning of the original language model. Otherwise, the final discrepancy depends on the approximation error, the condition number of the objective function, and the stability of the training process. Let $\delta^+ = \arg\min_{\delta}\mathcal{F}(\psi(\delta))$, and $\theta^+ = \psi(\delta')+\epsilon'$. Suppose that $\mathcal{F}$ and $\mathcal{F}\circ\psi$ are Lipschitz continuous and the Lipschitz constants are $L_1$ and $L_2$, respectively. 
 %Assume that the Lipschitz constants of $\mathcal{F}$ and $\mathcal{F}\circ\psi$ are $L_1$ and $L_2$, respectively.
 Then, we have the following bound of the approximation error of delta tuning to the full-parameter fine-tuning of the original language model:
 \begin{align}
  \vert \mathcal{F}(\theta^+) - \mathcal{F}(\psi(\delta^+)) \vert 
  &\leq \vert \mathcal{F}(\theta^+)-\mathcal{F}(\psi(\delta')) \vert
   + \vert \mathcal{F}(\psi(\delta'))-\mathcal{F}(\psi(\delta^+)) \vert  \nonumber \\
  & \leq L_1\|\epsilon'\|_2 + L_2\|\delta' -\delta^+\|_2
     \leq L_1\|\epsilon'\|_2 + L_2(\|\delta'\|_2 +\|\delta^+\|_2).  \label{ineq:diff}
 \end{align}
 The error $\epsilon'$ is controlled by the approximation of the low dimensional representation $\psi$. Since the minimization of $\mathcal{F}(\psi(\delta))$ can be viewed as minimization of a perturbed objective function of $\mathcal{F}(\theta)$, the    $\| \delta'-\delta^+\|_2$ is bounded provided that  $\mathcal{F}$ is well-conditioned and the optimization algorithm is stable. Also, if the magnitudes of $\delta'$ and $\delta^+$ are small, the bound \eqref{ineq:diff} can still lead to the good quality of $\mathcal{F}(\psi(\delta^+))$. 
 
 Some delta tuning methods benefit from such approach. In LoRA \citep{hu2021lora}, 
  the  weight matrix $W\in\mathbb{R}^{d\times n}$  is constructed as $W \approx W_0+AB$, where $W_0$ is the corresponding part of $\delta_0$, and $A\in\mathbb{R}^{d\times r}, B\in\mathbb{R}^{r\times n} $ are rank-$r$ low rank matrices  learned by the training process. The numerical results  validate the assumption of low-rank approximation \citep{hu2021lora}. 
  In BitFit \citep{zaken2021bitfit} or diff pruning \citep{guo2021parameter}, some coordinates of $\delta$ are selected to be optimized during the training process. Thus,  the approximation is $\theta = \theta_0+Vy$, where the columns of $V$ consist of columns chosen from the identity matrix and $y$ is the low dimensional vector to be learned. We can also apply some suitable transformation to $\theta$ to make the $\mathcal{F}$ easier to be optimized. For example, choose a transformation matrix $S$, and let $S\theta = S\theta_0+Vy$. The resulting delta tuning method can be viewed as a re-parameterization followed by a diff pruning. %These methods are linear transformation. To exploit a nonlinear counterpart, 
   %Besides, we can  use a nonlinear decomposition $W\approx W_0+\varphi(A)B$ to generalize LoRA, where $\varphi$ is the nonlinear transformation. In \cite{he2022unified},  this method  has been tested and has promising results. 
 
 \paragraph{Low dimensional representation in functional space.} Another approach is to directly design an approximate function that matches the final $\mathcal{F}(\theta^+)$, i.e., we seeks to find $\hat{\mathcal{F}}(\delta)$, such that
 \begin{equation}
  \vert \mathcal{F}(\theta) -\hat{\mathcal{F}}(\delta) \vert < \epsilon, \label{ineq:diff2}
 \end{equation}
 where $\epsilon$ is the approximation error. By this way, we recognize that $\tilde{\mathcal{F}}(\theta,\delta_0) = \mathcal{F}(\theta)$ and $\tilde{\mathcal{F}}(\theta_0,\delta) = \hat{\mathcal{F}}(\delta)$. The construction of $\hat{\mathcal{F}}$ can be characterized by an incremental  network \citep{houlsby2019parameter} or an augmented feature space \citep{lester2021power}. 
 Since we are more interested in the final performance of the language model, rather than the model parameters, it is promising to directly model the function $\mathcal{F}(\theta)$ which is approximately restricted in a small manifold in the functional space, and we discard the need to estimate the error \eqref{ineq:diff} from model parameters. 
 
 The construction of $\hat{\mathcal{F}}$ is an art and differs in practice. The simplest strategy is to freeze some certain parts of the networks like BitFit \citep{zaken2021bitfit}. Consider the more sophisticated construction that can improve the approximation. 
 Since the action of the function is characterized by the data flow, one natural idea is to inject the low-rank representation in the data path of the original neural networks and the resulting new language model is an incremental network like Adapter \citep{houlsby2019parameter}, as shown in \eqref{adapter-computation}. The approximation error \eqref{ineq:diff2}  is determined by the representation capacity of the incremental network. Due to the universal approximation property \citep{leshno1993multilayer} of multilayer feedforward networks, %with a nonpolynomial activation function, 
 the quality of the approximation is guaranteed. It is worth noting that a similar  architecture  is also proposed in the area of computer vision \citep{rebuffi2017learning,ye2020light}. 
 
 By exploiting the autoregressive structure of Transformer, some more dedicated functional approximation can be conceived. Note that the objective function generally is determined by the input and model parameters, namely $\mathcal{L}\bigl((X,Y);\theta\bigr)$. In principle, we can frozen the model parameters $\theta$, and make $X$ and $Y$ variable. In some cases, it may be convenient  to swap the positions of $(X,Y)$ and $\theta$ to obtain a more tractable optimization problem. Since $\theta$ is the pre-trained language model that is unwieldy to process, it makes sense to use some trainable $\tilde{X}$ to replace $X$ since the feature space of $X$, namely ${\rm range}(X)$, is generally limited in a few thousand of dimension in a language task. This idea has been exploited in the prompt tuning \citep{lester2021power}, where a series of prompt tokens $P$ are prepended to  the input $X$. The prompt $P$ is not parameterized by $\theta$; instead, it has individual trainable parameters $\theta_P$. By this way, the feature space is augmented but still feasible for training. Owing to the autoregressive property of Transformer, the approximate function serves as a good surrogate of the original function $\mathcal{F}$ to be optimized and steers the language model to focus on the specific task. The prefix tuning \citep{li2021prefix} further makes some activations in the intermediate layers trainable, thus leading to a more accurate functional approximation, or in other words,  enlarging the representation capability of the modified language model. Regarding to the performance, since the prompt tuning is a dimension reduction method that models the probability distribution with less parameters,  the effectiveness is closely related to the model size and data size. With larger model size and data size, prompt tuning is prone to achieve better performance that is consistent with the dimension reduction theory \citep{wright2021high}. Moreover, for high dimensional problems, it is possible to have more freedoms to choose the subspace for the functional approximation, \citet{su2021transferability} and our experimental results in~\refsec{scale} also verify this intuition. 
 
%  Regarding to this intuition, \citet{su2021transferability} and our experimental results in Section~\ref{sec:exp_scale} verify its existence.
 
 \paragraph{The unified view of the approximations in solution space and functional space.} Generally speaking, the representations in solution space and function space often lead to similar constructions of the approximate $\tilde{\mathcal{F}}$ due to the duality relation. In fact,  a unified view of Adapter \citep{houlsby2019parameter}, prefixing tuning \citep{li2021prefix} and LoRA  \citep{hu2021lora} is proposed in \citep{he2022unified} by analyzing the data flow in the modified language model. It is pointed out that these delta tuning methods all construct  low dimensional modifications of the original data flow, i.e., $h \leftarrow h+\Delta h$, where $\Delta h$ is parameterized by some low dimensional parameters. This view can be recognized as understanding the different delta tuning methods from the perspective of functional space. Some useful empirical results about the ways to design the functional approximation can also be found in \citep{he2022unified}. 
 
 Our discussion suggests that the performance of all these delta tuning methods rely on the low dimension assumption. In fact, it can even be found that there may exist some common subspace among various tasks \citep{qin2021exploring}, \citet{su2021transferability} and our experimental results in~\refsec{transferability} also show the transferability of delta tuning in different tasks. 
 %In summary, the existing delta tuning methods all leverage the intrinsic low dimension of the problems and
 Since the actual performance of a delta tuning method is inevitably problem-dependent, it is promising to exploit more specific structures in the tasks at hand or build some hybrid algorithm %based on the simple solution or function approximations 
 to make it more competitive with the full fine-tuning of the original language model.

%%%% 主要修改：
% ①去掉subsubsection。
% 主要改现在的4.2.2节：包括
%% ②把theta和delta的notation改掉（对齐4.1），PT改成PF。
%% ③把4.2.2改成先写统一的optimal control framework，然后底下列举不同的delta tuning方法如何统一到这个框架中。目前写了adapter，lora和prefixtuning，其中
%%% 第一个属于addition-based：adapters；
%%% 第二个属于reprameterization-based；
%%% 第三个属于addition-based：prompt；
%% ④补充一个只tune bias的方法（BitFit）纳入到框架中。

\subsection{Optimal Control Perspective for Delta Tuning}
\label{sec:optimal control}

\citet{yang2022on} propose to interpret prefix tuning from the perspective of optimal control. In this section, we generalize the optimal control view to different delta tuning scenarios.

\textbf{Relationship Between Optimal Control And Deep Learning}. We start with interpreting deep learning from the optimal control perspective. According to Section 4 in \citet{pmp-dl}, we review the theorems in the following and directly follow their notations:
%first provide a brief introduction of \cite{pmp-dl}, which reveals the relationship between optimal control and deep learning. We borrow the theorems in Section 4 of \cite{pmp-dl} and directly follow their notations:

\begin{thm} \label{dis-pmp}(discrete-time PMP)
    Consider the discrete-time control problem
    \begin{equation}
        \begin{aligned}
            & \min_{\{\theta_0, \dots, \theta_{T-1}\} \in \Theta^{T}} \Phi(x_{T}) + \delta \sum_{t=0}^{T-1} L(\theta_t), \\
        & x_{t+1} = x_t + \delta f_t(x_t, \theta_t), ~x_0 = x, ~0 \le t \le T-1
        \end{aligned}
    \end{equation}
    where $\Phi$ and $L$ are termination and running losses, respectively. There exists a co-process
    \begin{align}
        & x_{t+1}^{*} = g_t(x_t^*, \theta_t^*), \quad\quad\quad\quad x_0^*=x, \\
        & p_t^{*} = \nabla_x H_t(x_t^*, p_{t+1}^*, \theta_t), \quad p_{T+1}^*=-\nabla_x\Phi(x_{T+1}^*)
    \end{align}
    such that
    \begin{equation}
        H_t(x_t^*, p_{t+1}^*, \theta_t^*) \ge H_t(x_t^*, p_{t+1}^*, \theta), ~ \theta \in \Theta, ~ 0 \le t \le T-1. 
    \end{equation}
    Here $g_t(x_t, \theta_t) := x_t + \delta f_t(x_t,\theta_t)$ and
    \begin{equation}
        H_t(x,p,\theta) = p \cdot g_t(x,\theta) - \delta L(\theta)
    \end{equation}
    is the discrete Hamiltonian with a scaling factor $\delta > 0$.
\end{thm}

\begin{thm} \label{dis-msa} (discrete-time MSA)
    The discrete-time method of successive approximations (MSA) characterizes the co-process in Theorem \ref{dis-pmp}. For each iteration $k$,
    
    set $x_0^k = x$, and 
    \begin{equation}
        x_{t+1}^k = g_t(x_t^k, \theta_t^k)
    \end{equation}
    with $t$ enumerating from $0$ to $T-1$;
    
    then set $p_{T}^{k} = -\nabla_{x} \Phi(x_{T}^{k})$, and
    \begin{equation}
        p_{t}^k = \nabla_x H_t(x_t^k, p_{t+1}^k, \theta_t^k)
    \end{equation}
    with $t$ enumerating from $T-1$ to $0$;
    
    finally, with $t$ enumerating from $0$ to $T-1$, set
    \begin{equation}
        \theta_t^{k+1} = \theta_t^{k} + \eta \nabla_{\theta} H_t(x_t^k, p_{t+1}^k, \theta_t^k).
    \end{equation}
\end{thm}

\begin{thm} \label{msa-backprop} (equivalence between MSA and backpropagation)
    The MSA in Theorem \ref{dis-msa} is equivalent to the backpropagation process in deep networks.
\end{thm}

The proofs of Theorems \ref{dis-pmp}, \ref{dis-msa} and \ref{msa-backprop} are provided in \cite{pmp-dl}.

\textbf{Tuned Delta's As Optimal Controllers}. We consider delta tuning with pretrained autoregressive LMs (e.g., GPT-2) for text classification. By framing the content as the input sentence, the model generates the predicted label at the last step. For simplicity, we denote the position of label prediction as $o$. At position $o$, the model is inputted with a special token $[{\rm ANS}]$ and is expected to generate the prediction. 

Denote $\theta$ as the parameters of the $L$-layer PLM. We use the training set $\mathcal{D}_{tr}$ to optimize the delta parameters at each layer: $\{\delta^{(0)}, \dots, \delta^{(L-1)}\}$. The intermediate activation of the $j$-th layer at step $i$ is denoted as $h_{i}^{(j)}$. The optimization problem for delta tuning is formulated as
\begin{equation}
    \begin{aligned}
        & \min_{\{\delta^{(0)}, \dots, \delta^{(L-1)}\}} \mathbb{E}_{(x, y)\sim\mathcal{D}_{tr}} \left[S\left(h_{o}^{(L)}, y\right) + \sum_{j=0}^{L-1}R\left(\delta^{(j)}\right) \right] \\
        & h_{o}^{(j+1)} = h_{o}^{(j)} + \mathcal{G}_{\theta}^{(j)}\left(h_{o}^{(j)}, {\delta}^{(j)}\right), ~h_{o}^{(0)} = z_{o} = {\rm [ANS]}, ~0 \le j \le L-1,
    \end{aligned}
\label{unified-problem}
\end{equation}
where $S$ as the $\mathrm{softmax}$ scoring function, $R$ as the regularizer for delta parameters, $z_i$ as the $i$-th token in the input and $y$ as the label. The function $\mathcal{G}$ defines the altered forward propagation in the LM with the intervention of delta. Specifically, the learnable ${\delta}^{(j)}$ activates the fixed parameters from $\theta$ so that the representation $h_{o}^{(j)}$ at the $j$-th layer can be properly transformed as $\mathcal{G}_{\theta}^{(j)}\left(h_{o}^{(j)}, {\delta}^{(j)}\right)$. The representation transformation between two consecutive layers is thus described by function $\mathcal{G}$ and the residual connection in Transformer.

% Prefix tuning aims to steer the LM to maximize the probability of the ground-truth label. We use the training set $\mathcal{D}_{tr}$ to optimize the prefix parameters $P_{\delta}[i, :]$:
% \begin{equation}
%     \min_{\delta}\mathbb{E}_{(x,y)\sim \mathcal{D}_{tr}} \mathcal{L}(y|x; \delta) = \max_{\delta}\mathbb{E}_{(x,y)\sim \mathcal{D}_{tr}}\log \left[W\left(h_o^{(L)}\right)\right]_{y},
% \end{equation}
% where $W$ in the LM transforms the top-layer output $h_o^{(L)}$ to a probability vector over the vocabulary.

We proceed to show that the problem (\ref{unified-problem}) unifies various delta tuning scenarios with different instances of $\mathcal{G}$.

\textbf{I. Prefix-tuning (PF).} Prefix-tuning belongs to addition-based methods and exploits the idea of prompting. With $\mathrm{P}_{\rm idx}$ as the prefix indexes, the forward propagation at the output position $o$ can be formulated as
\begin{equation}
    h_{o}^{(j+1)} = \mathrm{LM}_\theta^{(j)}\left(h_{o}^{(j)} ~\middle|~ h_{<o}^{(j)} \right),
\label{prefix-propagate}
\end{equation}
where $h_{i}^{(j)} = P_{{\delta}^{(j)}}[i, :]$ for all $j=0$ to $L-1$, $i \in \mathrm{P}_{\rm idx}$ and $h_{i}^{(0)} = z_{i}$ for $i \notin \mathrm{P}_{\rm idx}$. ${\rm LM}_\theta^{(j)}$, the $j$-th layer of the LM, can be decomposed into a self-attention layer (${\rm SAN}_\theta^{(j)}$) and a FFN layer (${\rm FFN}_\theta^{(j)}$). Formally,
\begin{equation}
    h_{o}^{(j+1)} = h_{o}^{(j)} + {\rm SAN}_\theta^{(j)}\left(h_{o}^{(j)}, h_{<o}^{(j)} \right) + {\rm FFN}_\theta^{(j)}\left(h_{o}^{(j)} + {\rm SAN}_\theta^{(j)}\left(h_{o}^{(j)}, h_{<o}^{(j)}\right)\right)
\label{formally-recursion}
\end{equation}
with $h_{i}^{(j)} = P_{{\delta}^{(j)}}[i, :]$ for $i \in \mathrm{P}_{\rm idx}$. As Eq. (\ref{formally-recursion}) is recursive and according to the fact that the PLM is autoregressive, after unrolling the recursion for all $h_{<o}^{(j)}$ in Eq. (\ref{formally-recursion}), we have that for any $i < o$, $h_{i}^{(j)}$ is steered by the prefix $\delta^{(j)}$, namely $h_{i}^{(j)} = h_{i}^{(j)}\left(\delta^{(j)}\right)$.
As a result, for prefix-tuning, the $\mathcal{G}_{\theta}^{(j)}\left(h_{o}^{(j)}, {\delta}^{(j)}\right)$ in problem (\ref{unified-problem}) is instantiated as
\begin{equation}
    \mathcal{G}_{\theta}^{(j)}\left(h_{o}^{(j)}, {\delta}^{(j)}\right) = {\rm SAN}_\theta^{(j)}\left(h_{o}^{(j)}, h_{<o}^{(j)} \right) + {\rm FFN}_\theta^{(j)}\left(h_{o}^{(j)} + {\rm SAN}_\theta^{(j)}\left(h_{o}^{(j)}, h_{<o}^{(j)}\right)\right),
\end{equation}
where each item in $h_{<o}^{(j)}$ is a function of $\delta^{(j)}$.

\textbf{II. Adapter (AP).} Adapter belongs to addition-based methods as well. Instead of prompting, the Adapter method adds tunable modules between consecutive Transformer layers as the delta. As characterized by Eq. (\ref{adapter-computation}), the altered forward propagation is written as
\begin{equation}
    \begin{aligned}
        \Tilde{h}_{o}^{(j+1)} &= \mathrm{LM}_\theta^{(j)}\left(h_{o}^{(j)} ~\middle|~ h_{<o}^{(j)} \right), \\
        h_{o}^{(j+1)} &= \Tilde{h}_{o}^{(j+1)} + \sigma\left(\Tilde{h}_{o}^{(j+1)}W_d^{(j)}\right)W_u^{(j)}, \\
    \end{aligned}
\label{adapter-computation-2}
\end{equation}
where $\Tilde{h}_{o}^{(j+1)}$ is the transformed representation by the original $j$-th layer in the PLM, and $\sigma$ denotes the nonlinearity. $\delta^{(j)}$ in this case is defined as $\{ W_d^{(j)}, W_u^{(j)} \}$, the two projection matrices at the $j$-th layer. It is noted that the computation of $\mathrm{LM}$ in Eq. (\ref{adapter-computation-2}) also follows the formulation of Eq. (\ref{formally-recursion}) (with difference only in inputs). Substituting Eq. (\ref{formally-recursion}) in Eq. (\ref{adapter-computation-2}) yields
\begin{equation}
    \begin{aligned}
        \mathcal{G}_{\theta}^{(j)}\left(h_{o}^{(j)}, {\delta}^{(j)}\right) = & \,\, {\rm SAN}_\theta^{(j)}\left(h_{o}^{(j)}, h_{<o}^{(j)} \right) + {\rm FFN}_\theta^{(j)}\left(h_{o}^{(j)} + {\rm SAN}_\theta^{(j)}\left(h_{o}^{(j)}, h_{<o}^{(j)}\right)\right) \\
        & + \sigma\left(\mathrm{LM}_\theta^{(j)}\left(h_{o}^{(j)} ~\middle|~ h_{<o}^{(j)} \right)W_d^{(j)}\right)W_u^{(j)}. \\
    \end{aligned}
\end{equation}
Here each item in $h_{<o}^{(j)}$ is independent of $\delta^{(j)} = \{ W_d^{(j)}, W_u^{(j)} \}$, which is different from the prefix-tuning case.

\textbf{III. LoRA (LR).} LoRA belongs to reparameterization-based methods. The update by LoRA is
\begin{equation}
    \mathbf{h} \leftarrow s\mathbf{z}\mathbf{W}_d\mathbf{W}_u + \mathbf{h},
\label{lora-computation}
\end{equation}
where $\mathbf{z}$ is the input and $s \ge 1$ is a tunable scalar hyperparameter. Similar with the Adapter scenario, we still define $\delta^{(j)} = \{ W_d^{(j)}, W_u^{(j)} \}$. The function $\mathcal{G}$ in this case is 
\begin{equation}
    \begin{aligned}
        \mathcal{G}_{\theta}^{(j)}\left(h_{o}^{(j)}, {\delta}^{(j)}\right) = & \,\, {\rm SAN}_\theta^{(j)}\left(h_{o}^{(j)}, h_{<o}^{(j)} \right) + {\rm FFN}_\theta^{(j)}\left(h_{o}^{(j)} + {\rm SAN}_\theta^{(j)}\left(h_{o}^{(j)}, h_{<o}^{(j)}\right)\right) \\
        & + s\mathbf{z}W_d^{(j)}W_u^{(j)}. \\
    \end{aligned}
\end{equation}

\textbf{IV. BitFit.} BitFit belongs to specification-based methods. BitFit tunes only the bias parameters in the PLM. We define $\theta = \{\psi, \delta\}$, where $\delta$ represents the tuned bias parameters, and $\psi$ is the fixed ones. In this case, the formulation of $\mathrm{LM}$ in Eq. (\ref{formally-recursion}) becomes
\begin{equation}
    h_{o}^{(j+1)} = h_{o}^{(j)} + {\rm SAN}_\psi^{(j)}\left(h_{o}^{(j)}, \delta_{\rm S}^{(j)}, h_{<o}^{(j)} \right) + {\rm FFN}_\psi^{(j)}\left(h_{o}^{(j)} + {\rm SAN}_\psi^{(j)}\left(h_{o}^{(j)}, \delta_{\rm S}^{(j)}, h_{<o}^{(j)}\right), \delta_{\rm F}^{(j)}\right),
\label{formally-recursion-2}
\end{equation}
with $\delta^{(j)} = \{ \delta_{\rm S}^{(j)}, \delta_{\rm F}^{(j)} \}$ as the bias terms of the $\rm SAN$ and $\rm FFN$ layers. The function $\mathcal{G}$ is thus given by
\begin{equation}
    \mathcal{G}_{\theta}^{(j)}\left(h_{o}^{(j)}, {\delta}^{(j)}\right) = {\rm SAN}_\psi^{(j)}\left(h_{o}^{(j)}, \delta_{\rm S}^{(j)}, h_{<o}^{(j)} \right) + {\rm FFN}_\psi^{(j)}\left(h_{o}^{(j)} + {\rm SAN}_\psi^{(j)}\left(h_{o}^{(j)}, \delta_{\rm S}^{(j)}, h_{<o}^{(j)}\right), \delta_{\rm F}^{(j)}\right).
\end{equation}

We have listed the formulations of the function $\mathcal{G}$ in problem (\ref{unified-problem}) for different delta tuning methods. With Theorem \ref{dis-pmp}, $S$ and $R$ in problem (\ref{unified-problem}) can be viewed as the terminal and the running loss with the delta parameters as the control variables. This means that (\ref{unified-problem}) can be formulated as a discrete-time control problem. With Theorems \ref{dis-msa} and \ref{msa-backprop}, the forward and backward propagation in optimization delta's are equivalent to the calculation of the co-state process in Pontryagin’s Maximum Principle \citep{pmp}. To conclude, delta tuning can be viewed as seeking the optimal control of PLMs for specific downstream tasks. 

Our analysis sheds light on designing novel delta tuning methods that are inspired from control theories. One can refer to control theories when designing robust models \citep{Control_Theory_NEURIPS2019}. For example, \citet{yang2022on} propose robust prefix-tuning that tunes an additional robust prefix during inference to guide the LM towards correct predictions. The idea of test-time activation rectification can be viewed as close-loop feedback control \citep{close-loop-control-robustness}. We have also shown that the intervention of delta's with the PLMs is equivalent to the design of controllers. By applying the theories of controller design \citep{controller-book-1, controller-book-2}, we expect more delta methods be proposed with theoretical guarantees. The designed delta structures are in this way interpretable in principle while sufficiently exploiting the power of PLMs.

%\subsection{Lottery Ticket?}

%\subsection{Low-Rank Nature for Deep Neural Networks}

%\subsection{From Pruning to Parameter-efficient Model Stimulation}

%\subsection{Sparsity of Neural Nets?}

\vspace{-0.3cm}
\section{Comparisons and Experimental Discoveries}
\label{sec:experiments}
\vspace{-0.2cm}
As an effective engine to stimulate large-size PLMs, delta tuning presents an enormous practical potential for various real-world applications. In this section, we carry out systematic experiments to gain a deeper understanding of the attributes of different mainstream delta tuning methods.

Specifically, (1) we first conduct thorough comparisons among four representative delta tuning methods and fine-tuning in \refsec{performance}, covering the \textbf{performance},  \textbf{convergence} and the \textbf{efficiency} analysis; (2) secondly, we explore the \textbf{combinability} of three representative delta tuning methods in \refsec{combination} by comparing the performance under both the full-data and low-resource setting. We also explore the effects of manual templates for delta tuning methods; (3) furthermore, we investigate the \textbf{scaling law} in \refsec{scale} and (4) the \textbf{transferability} of delta tuning methods among different downstream tasks in \refsec{transferability}. The implementation details and tasks are described in \refsec{details} and \refsec{tasks}. We will release the codes, dataset splits and trained delta checkpoints to facilitate future research attempts.

\vspace{-0.2cm}
\subsection{Performance, Convergence and Efficiency}
\label{sec:performance}
\paragraph{Experimental Setting.}
We evaluate vanilla fine-tuning (\textbf{FT}) and four representative delta tuning methods, including prompt tuning (\textbf{PT}), prefix-tuning (\textbf{PF}), LoRA (\textbf{LR}) and adapter (\textbf{AP}). Other representative delta tuning methods~\citep{liu2021p,zaken2021bitfit,guo2021parameter,liu2022mathcal} are omitted.

% We omit the experiments of other delta tuning methods (e.g., BitFit) because it underperforms the above methods~\citep{he2022unified}.

To cover broad and diverse NLP tasks, we randomly select over $100$ representative tasks from Huggingface datasets\footnote{\url{https://huggingface.co/datasets}}~\citep{lhoest-etal-2021-datasets}. The selected tasks include text classification (e.g., sentiment analysis and natural language inference), question answering (e.g., machine reading comprehension and multi-choice question answering), conditional generation (e.g., summarization and dialogue), etc. We list the task details of each category in Table \ref{tab:ontology_1}. To handle different tasks with a single text-to-text PLM, following \citet{raffel2019exploring}, we process the input and output of each task into the same sequence-to-sequence format.

We choose $\text{T5}_{\texttt{BASE}}$~\citep{raffel2019exploring} as the mainly evaluated PLM backbone for different tuning methods, and we additionally report the performance of \textbf{PT} with $\text{T5}_{\texttt{LARGE}}$~\citep{raffel2019exploring}. For both models, we use the checkpoints released by \citet{lester2021power}, who conducted additional $100$k steps of LM adaption on the official checkpoints released by \citet{raffel2019exploring}. Such an LM adaptation objective has been demonstrated beneficial for better performance and faster convergence during downstream adaptation, compared with only the original ``span corruption'' pre-training objective of T5.
We follow the common practice for each delta tuning's implementation. For \textbf{PF}, we use $5$ prefix tokens; for \textbf{PT}, we prepend $100$ tunable soft tokens into the input embedding; for \textbf{LR}, we reparameterize all the query matrices and the value matrices in the multi-head attention modules as low-rank decompositions, and set the rank to $8$; for \textbf{AP}, we insert adapter modules into both the multi-head attention module and the feed-forward network in each Transformer layer, set the bottleneck dimension to $64$, and choose \texttt{SiLU}~\citep{elfwing2018sigmoid} as the activation function. More training details are left in \refsec{performance details}.

\paragraph{Performance Analysis.} The overall results are listed in Table~\ref{tab:overall}, from which we observe that: (1) in general, since different delta tuning methods significantly reduce the amounts of tunable parameters, they are no match for \textbf{FT} in performance under most cases. But after averaging the results over all datasets, the gap between the delta tuning methods and the fine-tuning method is not insurmountable, which demonstrates the potential of the large-scale applications of parameter-efficient adaptations. 
(2) Despite having different design elements, \textbf{PF}, \textbf{LR} and \textbf{AP} are comparable with each other in performance. Specifically, each of them is possible to show dominant performance (even better than \textbf{FT}) over others on certain tasks. According to the average results, the performances of all the methods are ranked as $\textbf{FT}> \textbf{LR} > \textbf{AP} > \textbf{PF}$ > \textbf{PT}. Interestingly, the performance of the delta tuning methods is not consistent with their number of tunable parameters, i.e., at least on small PLMs, more tunable parameters do not necessarily lead to approximately better performance, and the design of the structure for delta tuning may play a greater role. 
(3) As the easiest of these methods to implement (i.e. without modifying the internal structure of the model), \textbf{PT} lags far behind other delta tuning methods in most cases when experimented on $\text{T5}_{\texttt{BASE}}$, although better \textbf{PT} performance is observed when the model size is significantly enlarged to $\text{T5}_{\texttt{LARGE}}$, which is aligned with previous findings on the power of scale for prompt tuning~\citep{lester2021power}\footnote{We found empirically that \textbf{PT} tends to perform worse and converge more slowly for small-scale PLMs.}. However, as we would show later (\refsec{scale}), other delta tuning methods also exhibit far better performance when the scale of the backbone PLM grows extremely large. That is, unlike the conclusion of (3), when the model increases sharply, the design of the structure may become less important for the delta tuning methods.

\LTcapwidth=\textwidth
    \begin{longtable}{l|rrrrrr}
	\caption{Overall (test) performance of over 100 NLP tasks comparing prompt tuning (\textbf{PT}), prefix-tuning (\textbf{PF}), LoRA (\textbf{LR}), Adapter (\textbf{AP}) and fine-tuning (\textbf{FT}). We experiment all methods on $\text{T5}_{\texttt{BASE}}$, with the best performance highlighted in bold, and additionally report the performance of \textbf{PT} on $\text{T5}_{\texttt{LARGE}}$.}
	\vspace{-0.2cm}
	\label{tab:overall}
    \endfirsthead
    \endhead
	\toprule
    \multicolumn{1}{c}{\multirow{2}{*}{\textbf{Task}}} & \multicolumn{1}{c}{\textbf{PT}} & \multicolumn{1}{c}{\textbf{PT}} & \multicolumn{1}{c}{\multirow{2}{*}{\textbf{PF}}} & \multicolumn{1}{c}{\multirow{2}{*}{\textbf{LR}}} & \multicolumn{1}{c}{\multirow{2}{*}{\textbf{AP}}} & \multicolumn{1}{c}{\multirow{2}{*}{\textbf{FT}}} \\
    \multicolumn{1}{c}{}& \multicolumn{1}{c}{\textbf{(base)}}& \multicolumn{1}{c}{\textbf{(large)}}& & & & \\
	\midrule
	\textit{Ratio of tunable parameters} & 0.03\% & 0.01\%  & 7.93\%  & 0.38\% & 2.38\%  &  100\% \\

    \midrule
    \textsc{acronym\_identification} & 93.35  & 96.68  & \textbf{96.12}  & \textbf{96.12} & 95.57  & \textbf{96.12} \\
    \textsc{ade\_corpus\_v2-classification} & 41.76  & 94.42  & 93.25  & \textbf{94.47} & 93.91  & 94.27  \\
    \textsc{ade\_corpus\_v2-dosage} & 78.57  & 89.29  & 82.14  & \textbf{85.71} & 82.14  & 82.14  \\
    \textsc{ade\_corpus\_v2-effect} & 59.15  & 61.35  & \textbf{63.25}  & 62.52  & 60.91   &62.66 \\
    \textsc{adversarial\_qa} & 34.10  & 54.60  & 43.17  & 46.40  & 45.35  & \textbf{48.56} \\
    \textsc{ag\_news} & 91.37  & 93.61  & 93.42  & 94.63  & 94.60  & \textbf{95.19} \\
    \textsc{anli} & 25.85  & 44.96  & 43.88  & 45.27  & 49.19  & \textbf{50.54} \\
    \textsc{aslg\_pc12} & 15.78  & 44.07  & 47.71  & 73.72  & 80.65  & \textbf{92.92} \\
    \textsc{blimp-anaphor\_gender\_agreement} & \textbf{100.00} & 100.00 & \textbf{100.00} & \textbf{100.00} & \textbf{100.00} & 99.00  \\
    \textsc{blimp-anaphor\_number\_agreement} & 49.00  & 100.00 & \textbf{100.00} & \textbf{100.00} & \textbf{100.00} & \textbf{100.00} \\
    \textsc{blimp-determiner\_noun\_agreement} & \multirow{2}{*}{46.00}  & \multirow{2}{*}{100.00} & \multirow{2}{*}{\textbf{100.00}} & \multirow{2}{*}{\textbf{100.00}} & \multirow{2}{*}{\textbf{100.00}} & \multirow{2}{*}{\textbf{100.00}} \\
    ~~~\textsc{\_with\_adj\_irregular\_1}& & & & & & \\
    \textsc{blimp-ellipsis\_n\_bar\_1} & 49.00  & 100.00 & \textbf{100.00} & \textbf{100.00} & \textbf{100.00} & \textbf{100.00} \\
    \textsc{blimp-existential\_there} & \multirow{2}{*}{53.00}  & \multirow{2}{*}{100.00} & \multirow{2}{*}{\textbf{100.00}} & \multirow{2}{*}{\textbf{100.00}} & \multirow{2}{*}{\textbf{100.00}} & \multirow{2}{*}{\textbf{100.00}} \\
    ~~~\textsc{\_quantifiers\_1} & & & & & & \\
    \textsc{blimp-irregular\_past} & \multirow{2}{*}{\textbf{100.00}} & \multirow{2}{*}{100.00} & \multirow{2}{*}{\textbf{100.00}} & \multirow{2}{*}{\textbf{100.00}} & \multirow{2}{*}{\textbf{100.00}} & \multirow{2}{*}{\textbf{100.00}} \\
        ~~~\textsc{\_participle\_adjectives}& & & & & & \\
    \textsc{blimp-sentential\_negation} & \multirow{2}{*}{54.00}  & \multirow{2}{*}{100.00} & \multirow{2}{*}{\textbf{100.00}} & \multirow{2}{*}{\textbf{100.00}} & \multirow{2}{*}{\textbf{100.00}} & \multirow{2}{*}{\textbf{100.00}} \\
    ~~~\textsc{\_npi\_scope}& & & & & & \\
    \textsc{blimp-wh\_questions\_object\_gap} & 55.00  & 100.00 & \textbf{100.00} & \textbf{100.00} & \textbf{100.00} & \textbf{100.00} \\
    \textsc{boolq} & 61.28  & 77.43  & 77.55  & 80.00  & 78.47  & \textbf{81.77} \\
    \textsc{circa} & 13.51  & 77.39  & 80.16  & 82.38  & 82.93  & \textbf{84.69} \\
    \textsc{climate\_fever} & 15.47  & 33.42  & 38.03  & 39.35  & 37.48  & \textbf{41.57} \\
    % \textsc{*codah} & 27.68  & \textbf{52.89}  & 44.98  & 43.01  & 41.22  & 45.18 \\
    \textsc{commonsense\_qa} & 58.43  & 76.76  & 58.43  & \textbf{62.52} & 60.72  & 61.21  \\
    \textsc{cos\_e} & 12.41  & 14.82  & 13.90  & 14.05  & \textbf{14.31} & 13.46  \\
    \textsc{cosmos\_qa} & 7.30  & 10.98  & 9.91  & 10.78  & 10.85  & \textbf{11.32} \\
    \textsc{crawl\_domain} & 68.16  & 76.91  & 73.04  & 73.00  & 72.76  & \textbf{75.12} \\
    \textsc{discovery} & 0.18  & 18.83  & 16.67  & 18.98  & 18.41  & \textbf{25.88} \\
    \textsc{dream} & 49.19  & 71.83  & 58.70  & 61.00  & 59.53  & \textbf{62.42} \\
    \textsc{eli5-askh} & 11.26  & 11.70  & 12.64  & 11.99  & 11.45  & \textbf{13.00} \\
    \textsc{eli5-asks} & 14.79  & 15.54  & 15.09  & 15.25  & 15.01  & \textbf{15.28} \\
    \textsc{eli5-eli5} & 14.19  & 15.38  & \textbf{15.23}  & 14.59  & 14.43  & 14.75 \\
    \textsc{emo} & 69.91  & 71.47  & 73.31  & \textbf{76.13} & 74.88  & 75.69  \\
    \textsc{emotion} & 89.19  & 88.73  & 88.29  & 88.63  & 88.98  & \textbf{89.25} \\
    \textsc{ethos-directed\_vs\_generalized} & 76.86  & 86.64  & 94.76  & 92.29  & \textbf{94.94} & \textbf{94.94} \\
    \textsc{ethos-disability} & 46.99  & 100.00 & 93.81  & 93.81  & \textbf{100.00} & 93.81  \\
    \textsc{ethos-gender} & 63.84  & 77.08  & 77.44  & \textbf{79.91} & \textbf{79.91} & 74.48  \\
    \textsc{ethos-national\_origin} & 44.30  & 81.77  & 81.77  & \textbf{87.95} & 84.72  & 84.72  \\
    \textsc{ethos-race} & 84.36  & 97.06  & 94.54  & \textbf{97.21} & 94.27  & \textbf{97.21} \\
    \textsc{ethos-religion} & 93.02  & 93.02  & 96.35  & 93.02  & 96.35  & \textbf{96.64} \\
    % \textsc{*ethos-sexual\_orientation} & 61.20  & 94.81 & 100.00  & \textbf{94.81} & \textbf{94.81} & \textbf{94.81} \\
    \textsc{financial\_phrasebank} & 97.18  & 98.36  & \textbf{98.36}  & 97.94  & 97.95  & \textbf{98.36} \\
    \textsc{freebase\_qa} & 1.90  & 6.71  & 2.63  & 3.75  & 5.86  & \textbf{23.52} \\
    \textsc{glue-cola} & 0.00  & 55.60  & 50.95  & 49.40  & 44.66  & \textbf{51.53} \\
    \textsc{glue-mnli} & 35.43  & 86.12  & 82.21  & 83.74  & 83.90  & \textbf{86.39} \\
    \textsc{glue-mrpc} & 67.65  & 88.24  & 87.25  & 87.25  & 87.25  & \textbf{89.71} \\
    \textsc{glue-qnli} & 52.34  & 93.01  & 87.48  & 92.02  & 91.58  & \textbf{92.57} \\
    \textsc{glue-qqp} & 84.65  & 86.21  & 84.62  & 86.87  & 85.93  & \textbf{89.13} \\
    \textsc{glue-rte} & 45.32  & 79.14  & 72.66  & 79.14  & 78.42  & \textbf{80.58} \\
    \textsc{glue-sst2} & 92.20  & 94.95  & 92.66  & 94.04  & 93.35  & \textbf{94.27} \\
    % \textsc{*glue-wnli} & 47.22  & \textbf{55.56} & 58.33  & \textbf{55.56} & \textbf{55.56} & \textbf{55.56} \\
    \textsc{hate\_speech\_offensive} & 73.27  & 79.08  & \textbf{75.22}  & 75.21 & 75.06  & 75.04  \\
    \textsc{hate\_speech18} & 75.57  & 74.45  & 79.42  & 79.59  & 80.86  & \textbf{80.93} \\
    \textsc{hatexplain} & 50.98  & 67.62  & 66.06  & 68.03  & \textbf{68.11} & 68.02  \\
    \textsc{health\_fact} & 39.15  & 45.60  & 50.38  & 52.05  & 51.21  & \textbf{54.19} \\
    \textsc{hellaswag} & 23.82  & 70.28  & 24.76  & 32.82  & 27.60  & \textbf{41.90} \\
    \textsc{hotpot\_qa} & 65.95  & 76.41  & 73.76  & 76.13  & 74.65  & \textbf{78.45} \\
    % \textsc{*imdb} & 0.01  & 0.00  & 9.64  & 9.56  & 9.59  & \textbf{9.61} \\
    \textsc{lama-conceptnet} & 15.25  & 26.12  & 22.63  & 34.96  & 43.62  & \textbf{70.28} \\
    \textsc{lama-google\_re} & 11.78  & 14.08  & 12.60  & 18.82  & 23.73  & \textbf{24.88} \\
    \textsc{lama-squad} & 3.23  & 16.13  & \textbf{12.90}  & 9.68 & 3.23  & 9.68 \\
    \textsc{lama-trex} & 59.13  & 63.68  & 63.91  & 66.21  & 67.23  & \textbf{69.12} \\
    \textsc{liar} & 13.23  & 28.87 & 26.46  & \textbf{28.67} & 27.08  & 28.20  \\
    \textsc{mc\_taco} & 76.25  & 88.39  & 86.02  & \textbf{88.13} & 86.81  & 87.34  \\
    \textsc{medical\_questions\_pairs} & 46.56  & 91.80  & 85.25  & 88.52  & \textbf{90.16} & 87.21  \\
    \textsc{multi\_news} & 18.09  & 19.23  & 18.81  & 19.44  & 19.10  & \textbf{19.80} \\
    \textsc{numer\_sense} & 50.53  & 56.75  & 53.30  & 56.27  & 53.97  & \textbf{57.32} \\
    \textsc{onestop\_english} & 22.53  & 98.23  & \textbf{100.00} & \textbf{100.00} & \textbf{100.00} & \textbf{100.00} \\
    \textsc{openbookqa} & 44.80  & 54.40  & 50.20  & 52.20  & 53.80  & \textbf{57.00} \\
    \textsc{paws} & 49.60  & 91.27  & 92.07  & 93.39  & 92.91  & \textbf{93.60} \\
    \textsc{poem\_sentiment} & 54.18  & 70.31  & 85.38  & \textbf{86.80} & 82.52  & 83.26  \\
    \textsc{proto\_qa} & 21.16  & 37.66  & 24.57  & 27.87  & 26.17  & \textbf{34.47} \\
    \textsc{qasc} & 19.22  & 47.73  & 33.26  & 37.80  & 33.05  & \textbf{43.63} \\
    \textsc{quarel} & 54.89  & 54.71  & 57.25  & 59.78  & 57.61  & \textbf{62.50} \\
    \textsc{quartz-no\_knowledge} & 65.43  & 68.88  & 68.49  & 67.09  & 66.96  & \textbf{69.39} \\
    \textsc{quartz-with\_knowledge} & 64.03  & 85.97  & 71.56  & 74.23  & 73.72  & \textbf{76.28} \\
    \textsc{race-high} & 34.51  & 60.09  & 42.82  & 59.52  & 58.92  & \textbf{65.95} \\
    \textsc{race-middle} & 47.21  & 74.65  & 62.67  & 68.31  & 65.46  & \textbf{70.61} \\
    \textsc{rotten\_tomatoes} & 88.36  & 91.84  & \textbf{89.96}  & 89.30  & 89.20  & 89.77 \\
    \textsc{samsum} & 39.35  & 45.12  & 43.38  & 45.00  & 44.68  & \textbf{45.73} \\
    \textsc{sciq} & 96.95  & 98.53  & 98.08  & \textbf{98.42} & 98.19  & 98.30  \\
    \textsc{scitail} & 91.02  & 95.47  & 93.04  & 93.80  & 94.04  & \textbf{94.77} \\
    \textsc{search\_qa} & 7.14  & 19.17  & 8.70  & 10.17  & 9.72  & \textbf{19.26} \\
    \textsc{sick} & 40.10  & 88.82  & 87.91  & 88.69  & 88.88  & \textbf{89.15} \\
    \textsc{sms\_spam} & 95.80  & 97.46 & 97.14  & 97.14  & \textbf{97.46} & 97.11  \\
    \textsc{spider} & 3.29  & 6.38  & 7.74  & \textbf{9.67} & 8.70  & 6.77  \\
    \textsc{superglue-cb} & 75.00  & 78.57  & \textbf{100.00} & \textbf{100.00} & 96.43  & 96.43  \\
    \textsc{superglue-copa} & 53.60  & 56.00  & 58.40  & 56.40  & \textbf{60.40} & 59.20  \\
    % \textsc{*superglue-multirc} & 0.00  & 0.00  & 0.00  & 0.05  & 0.05  & \textbf{0.16} \\
    \textsc{superglue-record} & 44.67  & 73.82  & 61.62  & 64.66  & 62.08  & \textbf{67.20} \\
    \textsc{superglue-rte} & 50.36  & 84.89  & 73.38  & 79.14  & \textbf{82.01} & 78.42  \\
    \textsc{superglue-wic} & 50.16  & 68.34  & 64.89  & 68.65  & 70.53  & \textbf{71.79} \\
    % \textsc{*superglue-wsc} & 53.85  & 61.54  & 67.31  & 63.46  & 65.38  & \textbf{67.31} \\
    \textsc{tab\_fact} & 46.65  & 50.16  & 52.53  & 56.86  & 53.42  & \textbf{57.34} \\
    \textsc{trec} & 90.80  & 91.51  & 91.38  & 93.38  & 93.36  & \textbf{94.81} \\
    \textsc{trec-finegrained} & 80.63  & 88.18  & 90.04  & \textbf{91.44} & 90.00  & 91.27  \\
    \textsc{tweet\_eval-hate} & 53.00  & 42.23  & 44.67  & 48.16  & 47.88  & \textbf{51.33} \\
    \textsc{tweet\_eval-irony} & 58.02  & 69.73  & 76.00  & 76.75  & 73.88  & \textbf{77.43} \\
    \textsc{tweet\_eval-offensive} & 75.94  & 78.87  & 80.94  & 80.97  & 80.59  & \textbf{82.05} \\
    \textsc{tweet\_eval-sentiment} & 28.90  & 72.79  & 71.78  & 71.31  & 71.90  & \textbf{71.98} \\
    \textsc{tweet\_eval-stance\_abortion} & 32.59  & 61.42  & 61.47  & \textbf{63.20} & 62.61  & 61.72  \\
    \textsc{tweet\_eval-stance\_atheism} & 56.28  & 67.58  & 71.54  & 71.77  & 71.27  & \textbf{74.41} \\
    \textsc{tweet\_eval-stance\_climate} & 47.61  & 52.43  & 52.86  & 55.92  & \textbf{59.06} & 57.38  \\
    \textsc{tweet\_eval-stance\_feminist} & 29.65  & 51.63  & 56.27  & 57.41  & \textbf{58.57} & 58.51  \\
    \textsc{tweet\_eval-stance\_hillary} & 41.34  & 63.18  & 62.15  & 65.40  & 61.74  & \textbf{66.41} \\
    \textsc{web\_questions} & 11.90  & 19.58  & 15.87  & 18.78  & 20.63  & \textbf{25.40} \\
    \textsc{wiki\_bio} & 42.39  & 44.03  & 44.84  & 45.36  & 46.19  & \textbf{47.09} \\
    \textsc{wiki\_qa} & 48.78  & 73.97  & 64.10  & 72.15  & 70.75  & \textbf{74.41} \\
    \textsc{wiki\_split} & 79.80  & 80.10  & 79.91  & 80.09  & 80.05  & \textbf{80.34} \\
    \textsc{wino\_grande} & 48.42  & 58.20  & 50.79  & 61.20  & 50.47  & \textbf{67.19} \\
    \textsc{wiqa} & 36.10  & 65.27  & 63.67  & 77.99  & 64.44  & \textbf{79.82} \\
    \textsc{xsum} & 21.35  & 26.56  & 23.84  & 25.87  & 26.07  & \textbf{29.90} \\
    \textsc{yelp\_polarity} & 95.47  & 98.18  & 97.78  & 97.37  & 97.30  & \textbf{97.92} \\
    \midrule
    \textbf{Average} & 48.81 & 65.92 & 64.07 & 66.06 & 65.58 & \textbf{67.96} \\
	\bottomrule
    \end{longtable}%

\paragraph{Convergence Analysis.} In Figure \ref{fig:convergence}, Figure \ref{fig:convergence_2} and Figure \ref{fig:convergence_3}, we visualize the performance of different delta tuning methods (\textbf{LR}, \textbf{AP}, \textbf{PF}) and fine-tuning (\textbf{FT}) at different training steps to compare their convergence rate. It could be derived that, the convergence rate of these tuning methods are ranked as: $\textbf{FT} > \textbf{AP} \approx \textbf{LR} > \textbf{PF}$. Overall, \textbf{FR} is the most stable method for convergence, and despite the fact that \textbf{PF} has the highest number of tunable parameters of all delta tuning methods, it still faces some convergence difficulties (the original paper also mentions that the convergence of \textbf{PF} is very dependent on the reparameterization).

\begin{figure*}[!th]
\subcapraggedrighttrue
\subcaphangtrue
    \centering
    \subfigure{
	    \includegraphics[width=0.3\textwidth]{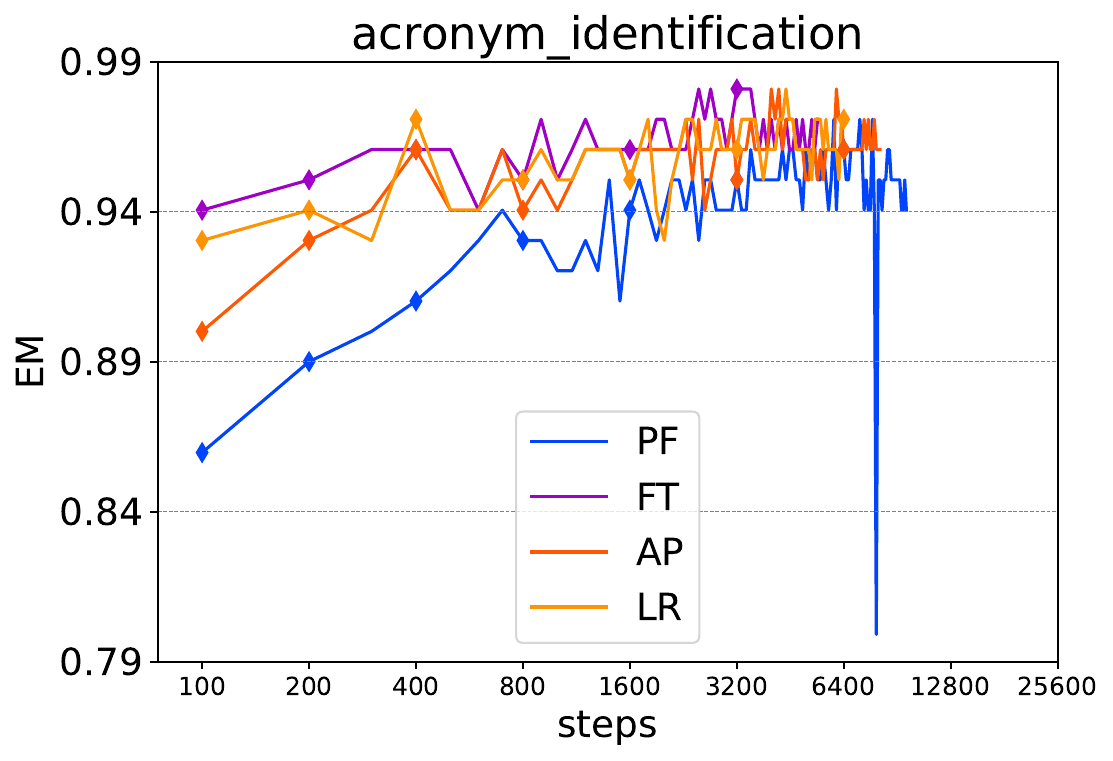}
    }
    \subfigure{
	    \includegraphics[width=0.3\textwidth]{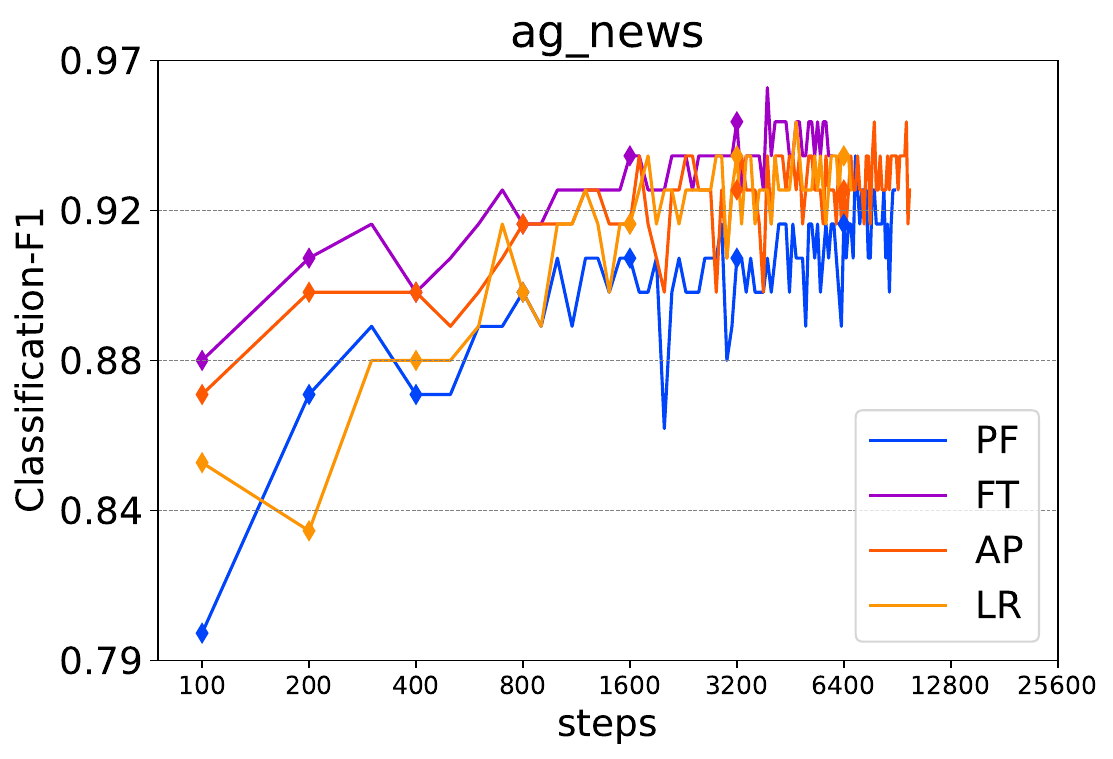}
    }
    \subfigure{
	    \includegraphics[width=0.3\textwidth]{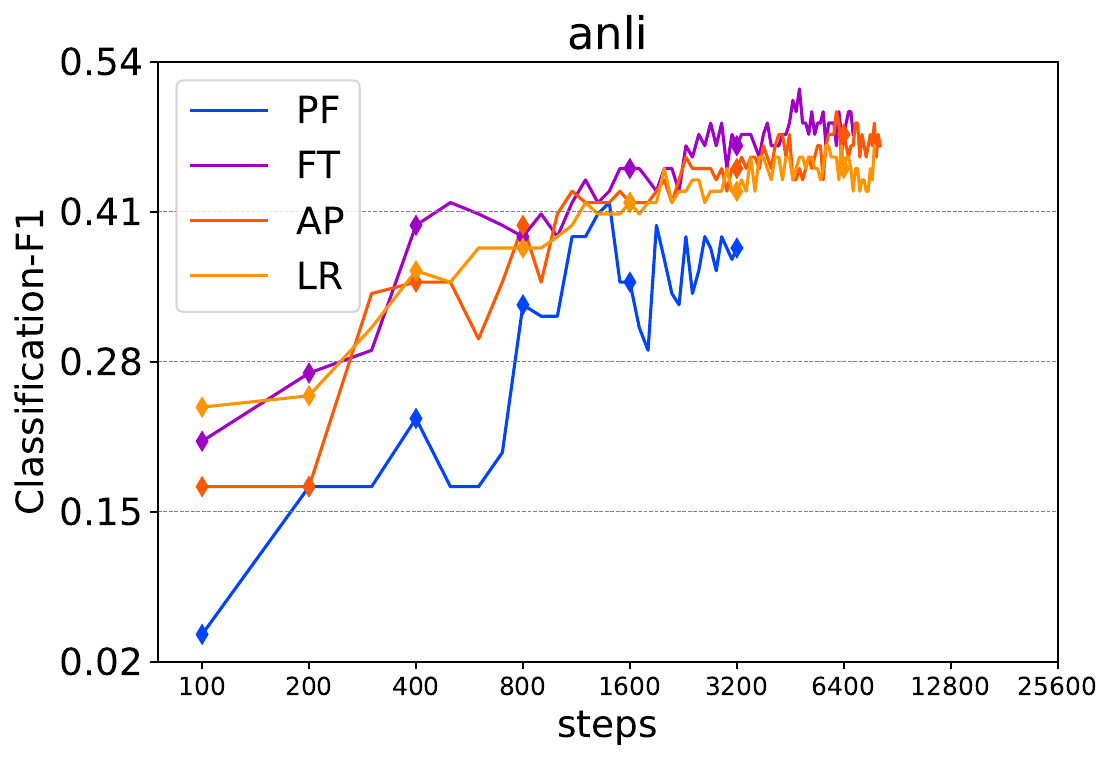}
    }
    
    \subfigure{
	    \includegraphics[width=0.3\textwidth]{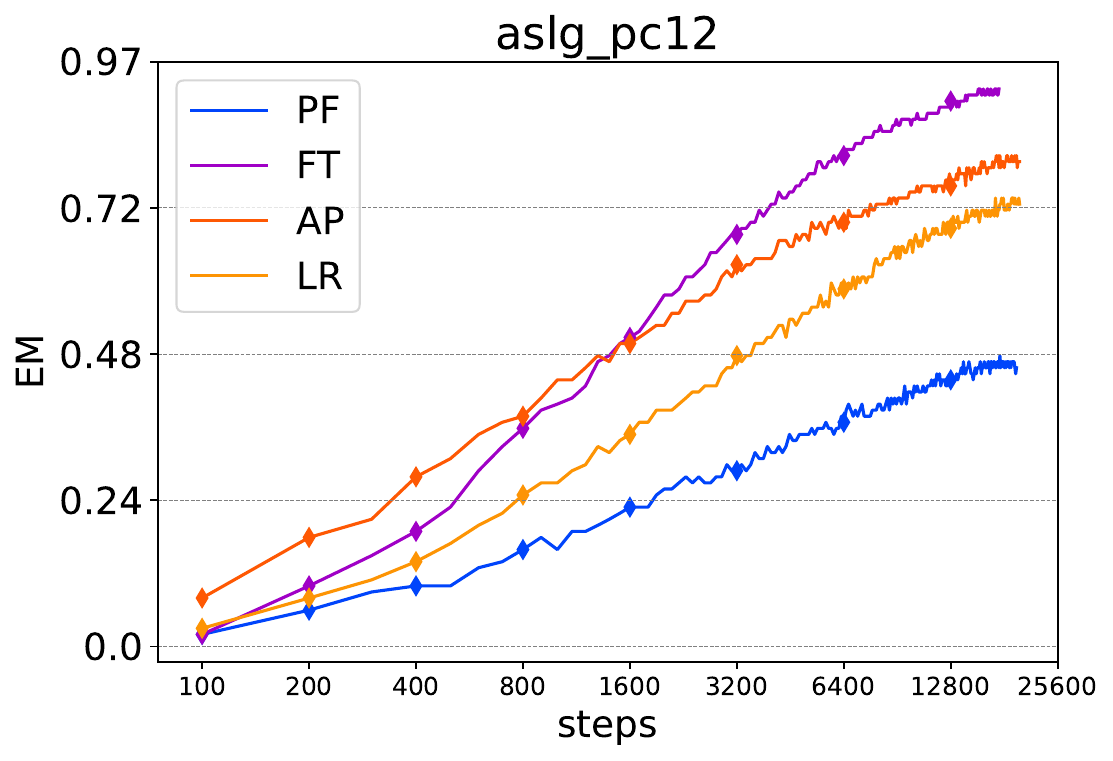}
    }
    \subfigure{
	    \includegraphics[width=0.3\textwidth]{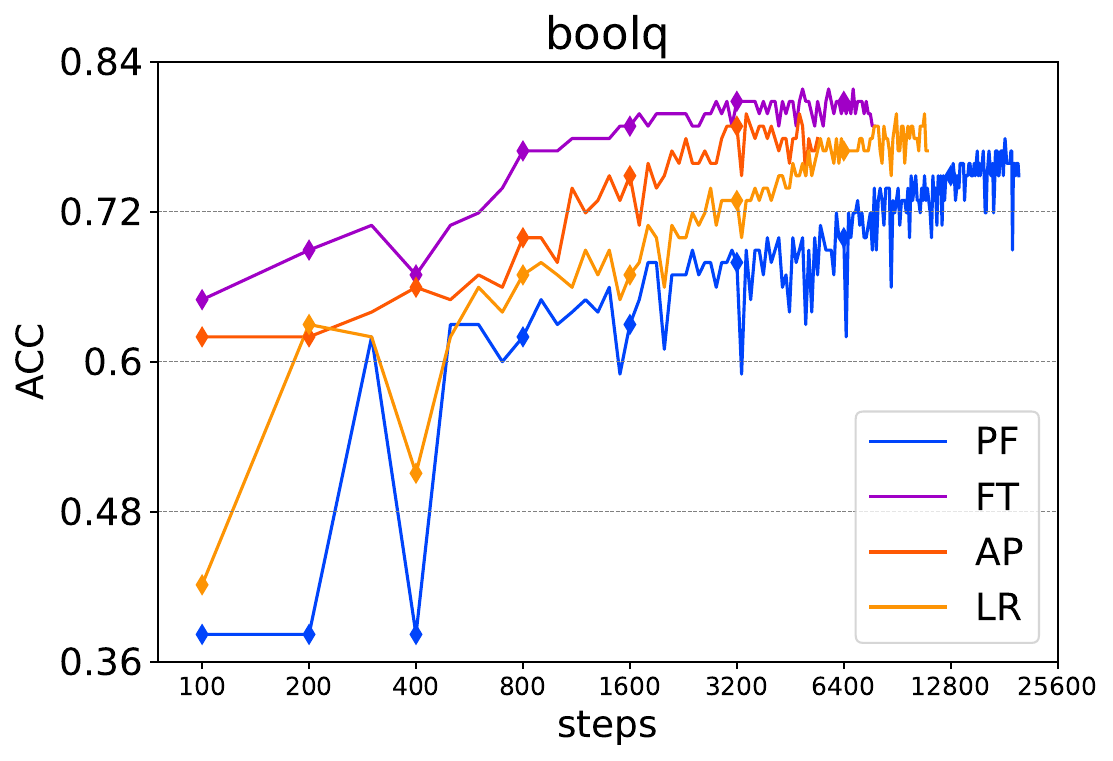}
    }
    \subfigure{
	    \includegraphics[width=0.3\textwidth]{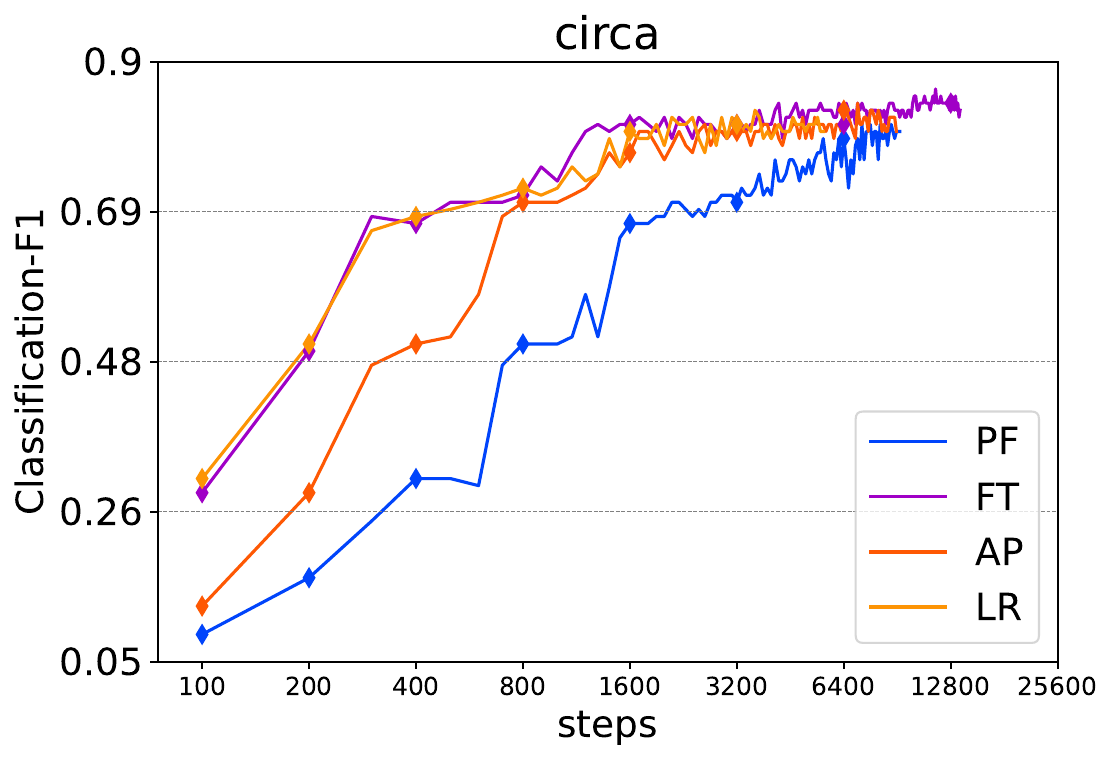}
    }
    
    \subfigure{
	    \includegraphics[width=0.3\textwidth]{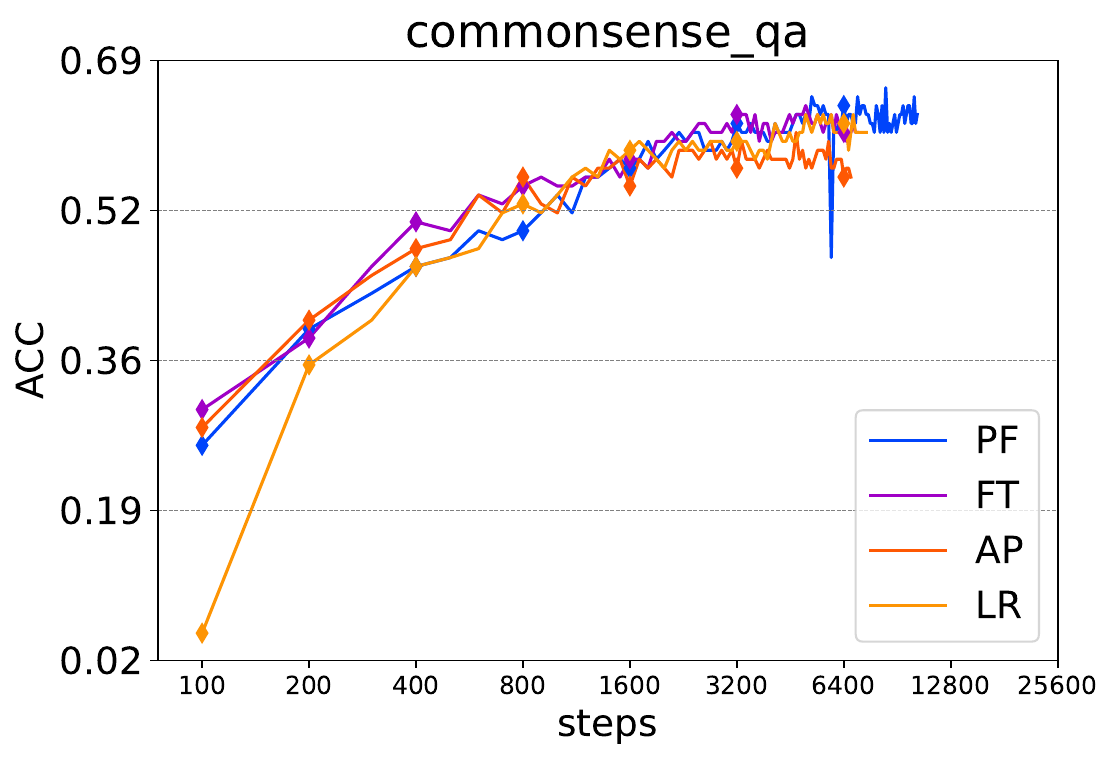}
    }
    \subfigure{
	    \includegraphics[width=0.3\textwidth]{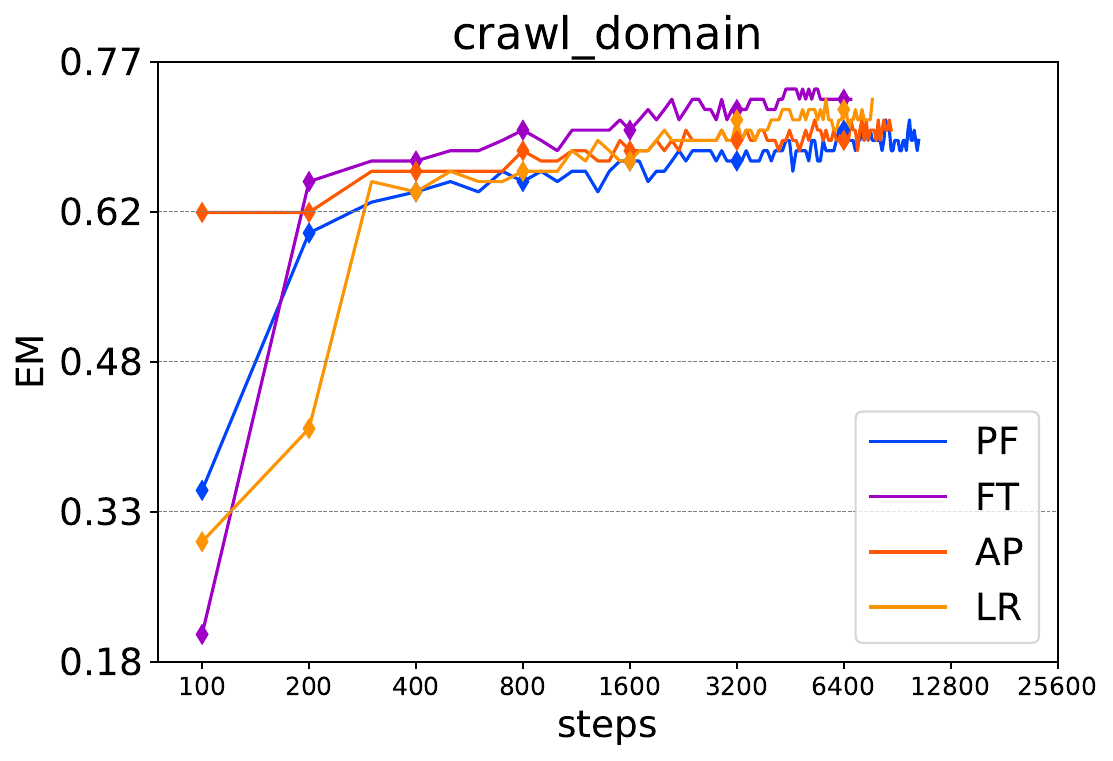}
    }
    \subfigure{
	    \includegraphics[width=0.3\textwidth]{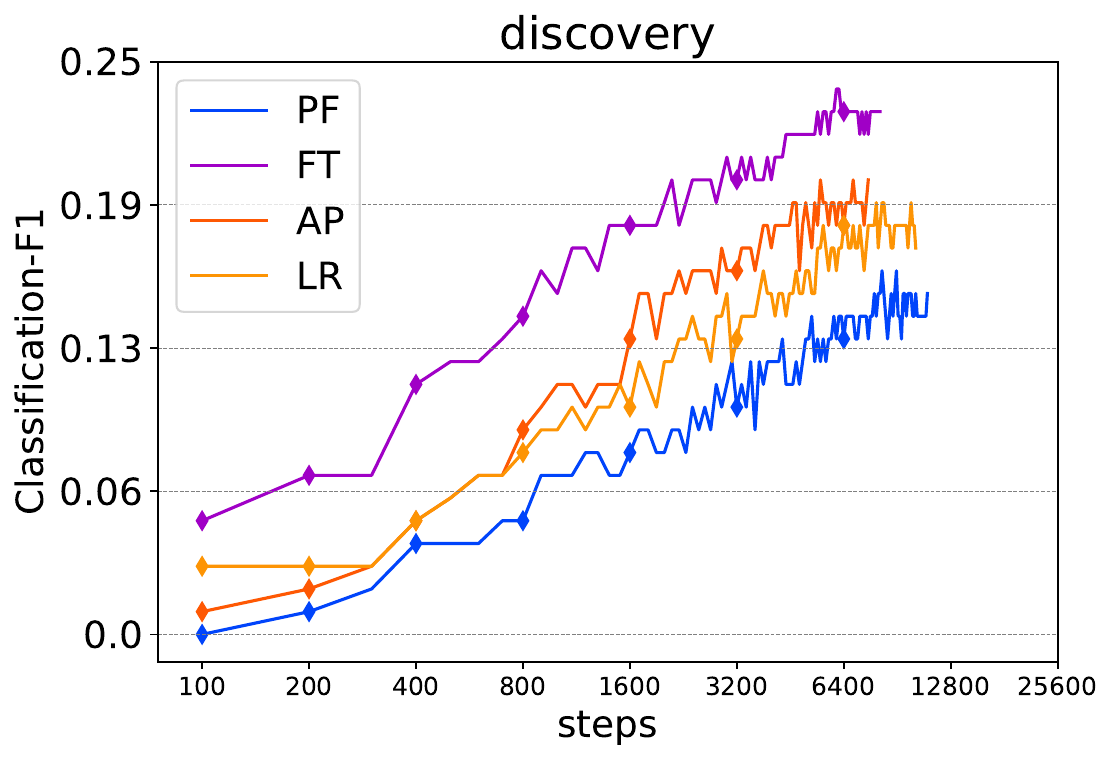}
    }
    
    \subfigure{
	    \includegraphics[width=0.3\textwidth]{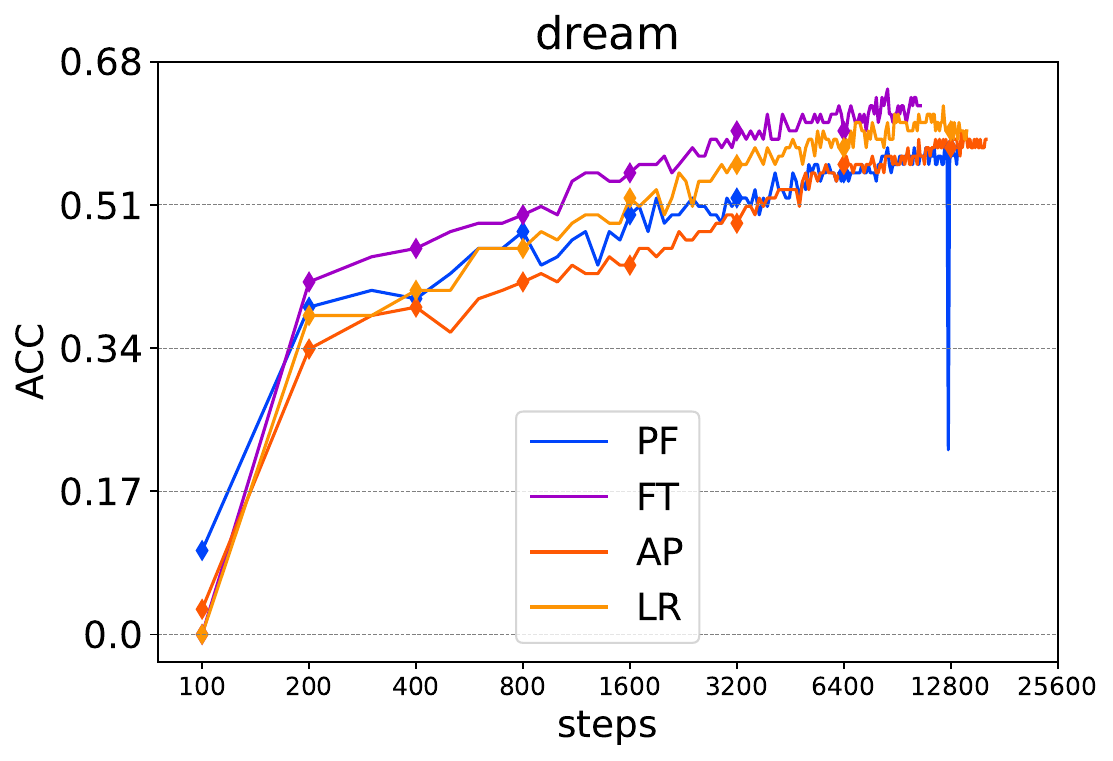}
    }
    \subfigure{
	    \includegraphics[width=0.3\textwidth]{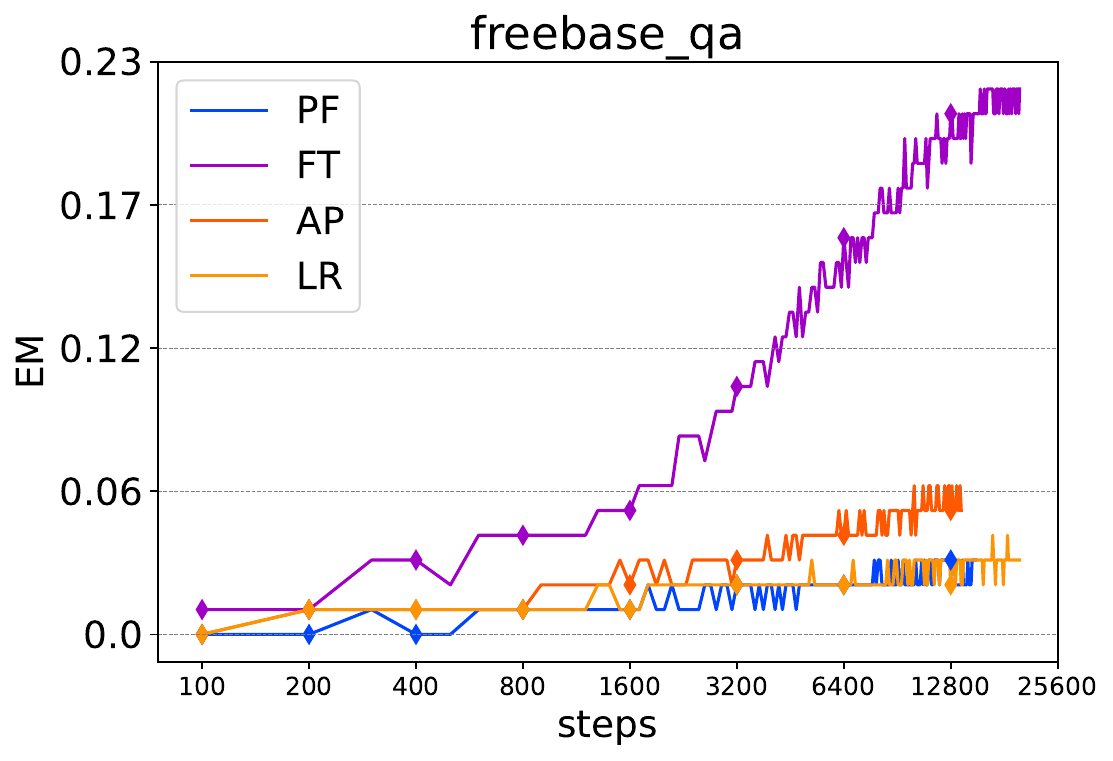}
    }
    \subfigure{
	    \includegraphics[width=0.3\textwidth]{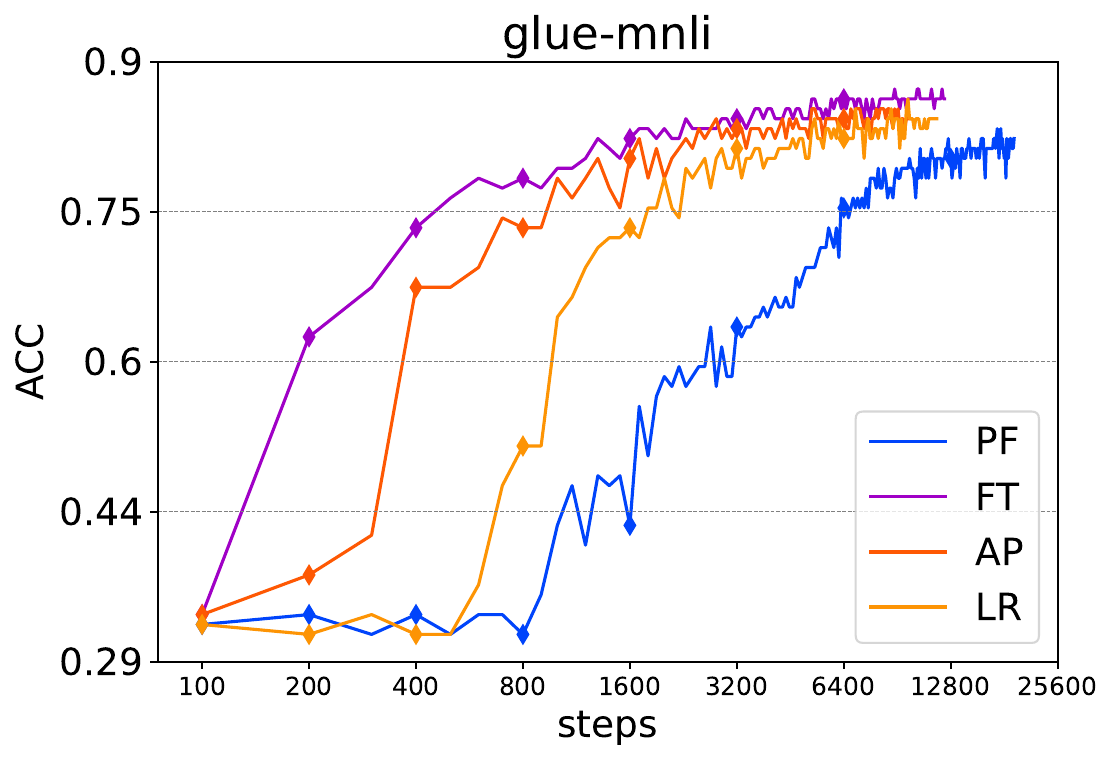}
    }
    
    \subfigure{
	    \includegraphics[width=0.3\textwidth]{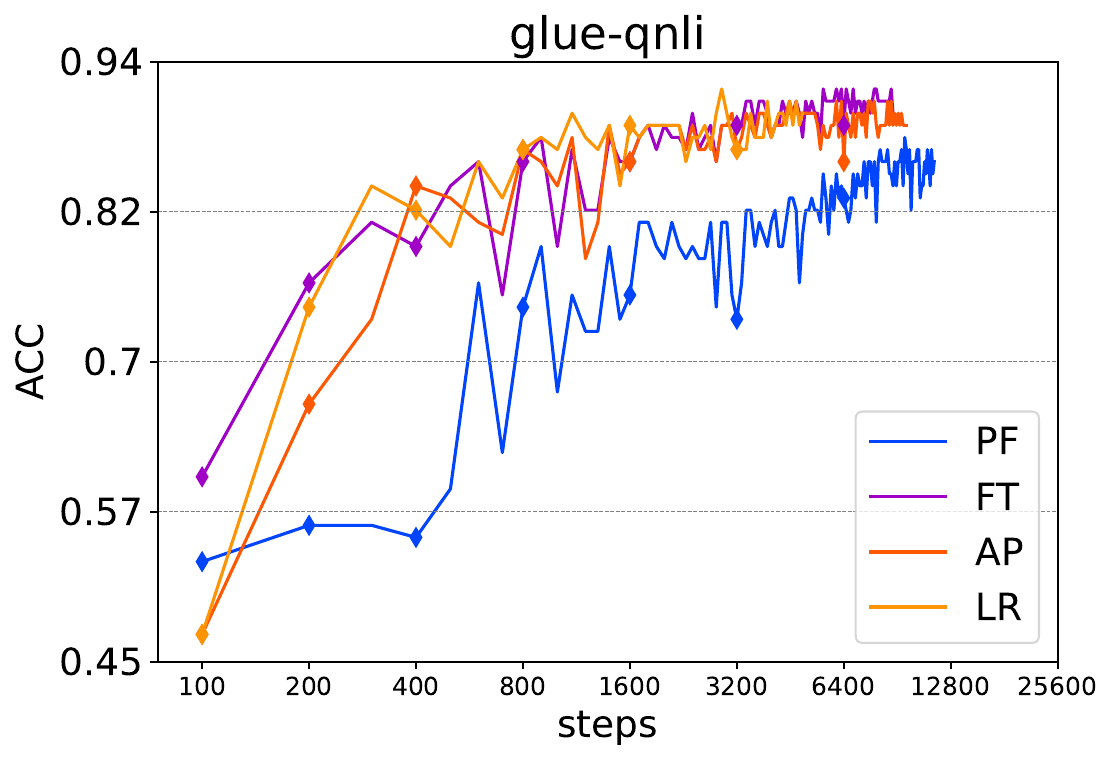}
    }
    \subfigure{
	    \includegraphics[width=0.3\textwidth]{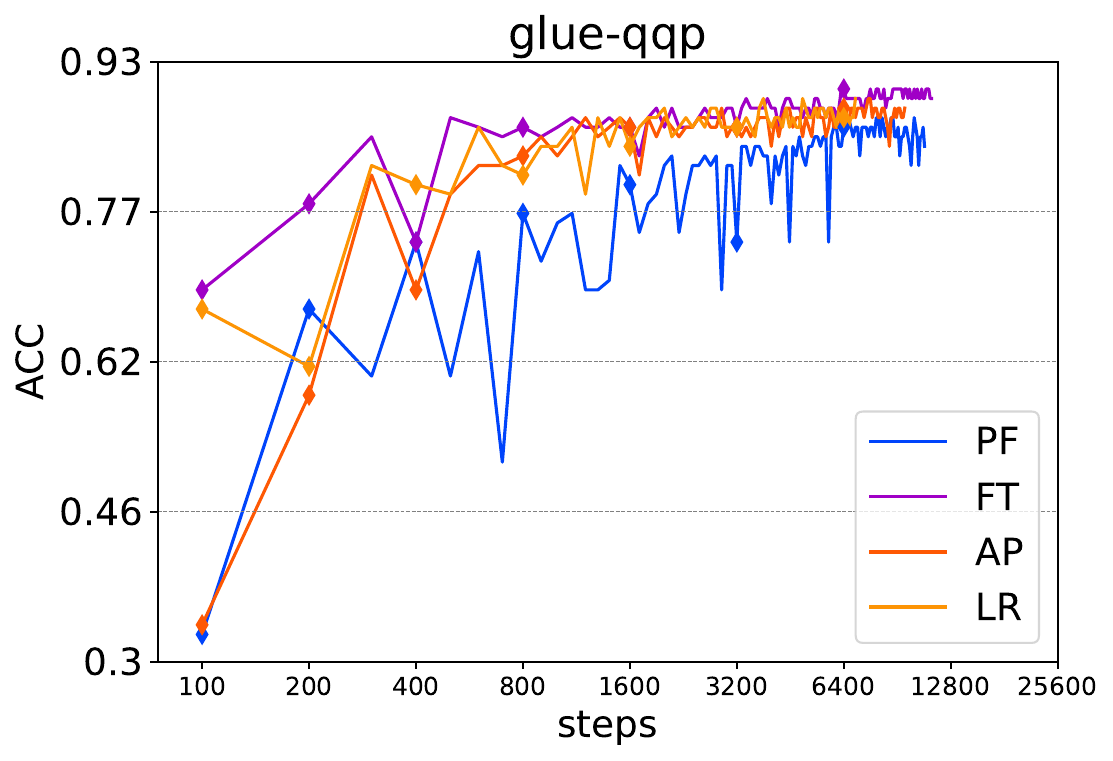}
    }
    \subfigure{
      \includegraphics[width=0.3\textwidth]{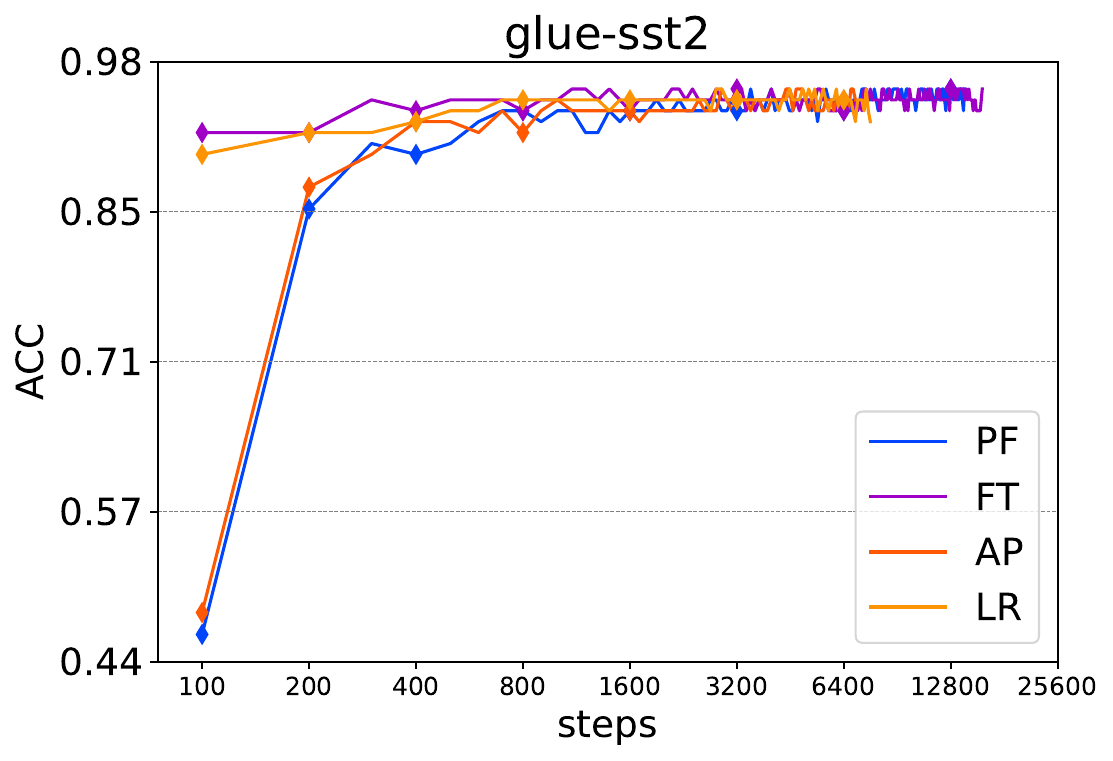}
    }
    
    \caption{The performance of $\text{T5}_{\texttt{BASE}}$ with different delta tuning methods (\textbf{LR}, \textbf{AP}, \textbf{PF}) and fine-tuning (\textbf{FT}) at different training steps. Note we apply early stop in all the experiments. The performance of \textbf{PT} is omitted since it lags far behind other tuning methods in both convergence and performance.}
    \label{fig:convergence}
\end{figure*}

Since \textbf{PT} lags far behind other tuning methods in both convergence rate and performance, we do not visualize it in the above figures. But as mentioned in \refsec{addition}, \textbf{PT} is the easiest method to implement and it is desirable to theoretically and empirically further study the convergence issue across different sizes of PLMs.  
We also found empirically that, (1) for each delta tuning method, within a reasonably broad range, both performance and convergence are not sensitive to the number of tunable parameters, but more sensitive to the structures of the methods,
and (2) with the scale of PLM growing larger, the convergence of delta tuning is also accelerated (\refsec{scale}). To summarize, our experiments yield very similar conclusions in terms of convergence and overall performance, and these conclusions are well supported by the fact that we used the same experimental and implementation setup, same model selection strategy, and plenty of datasets.

\begin{figure*}[!th]
\subcapraggedrighttrue
\subcaphangtrue
    \centering
    \subfigure{
      \includegraphics[width=0.3\textwidth]{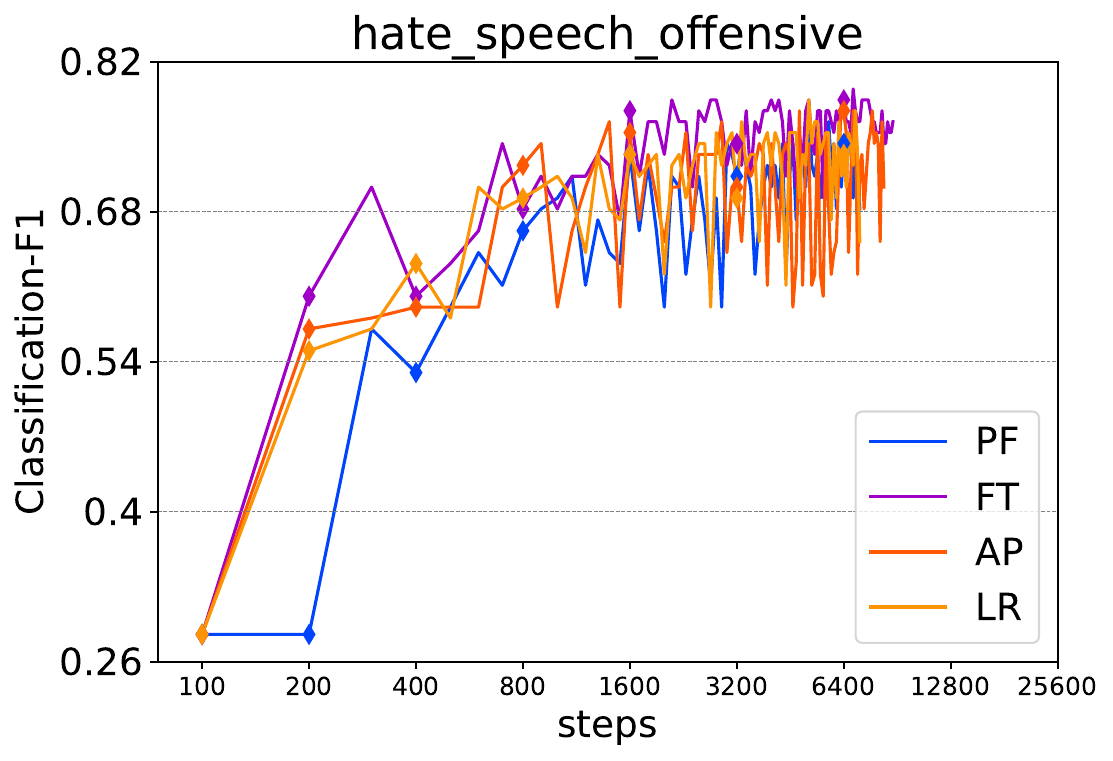}
    }
    \subfigure{
      \includegraphics[width=0.3\textwidth]{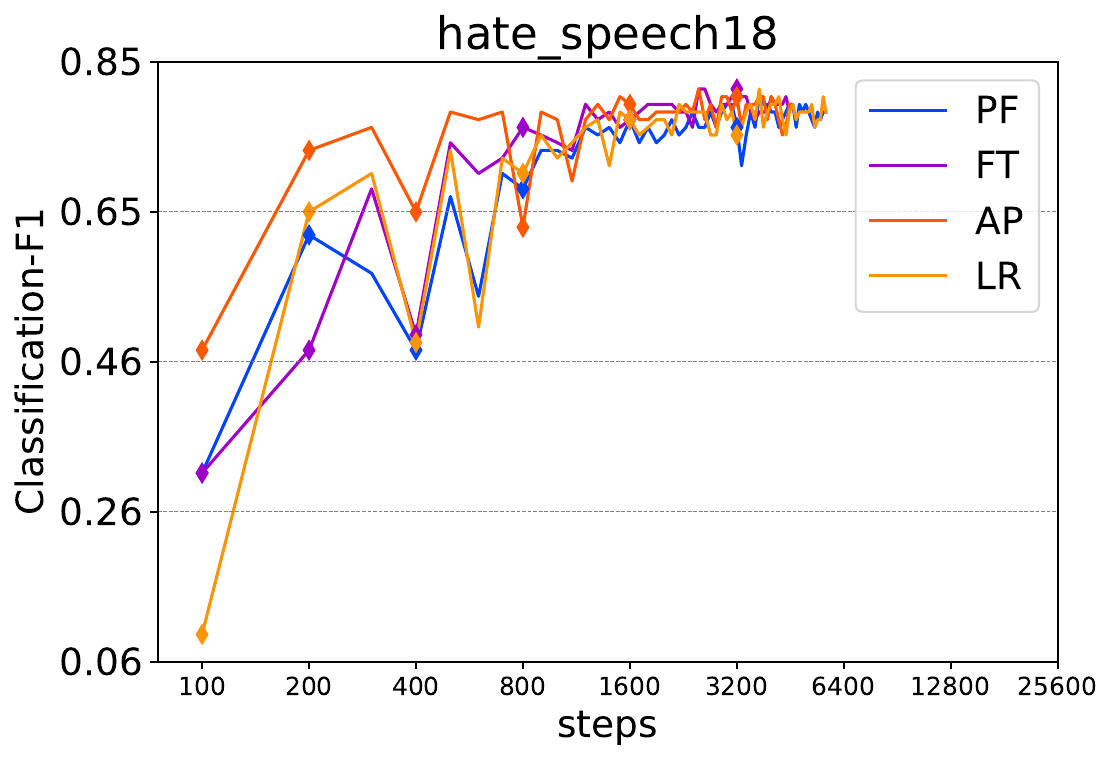}
    }
    \subfigure{
      \includegraphics[width=0.3\textwidth]{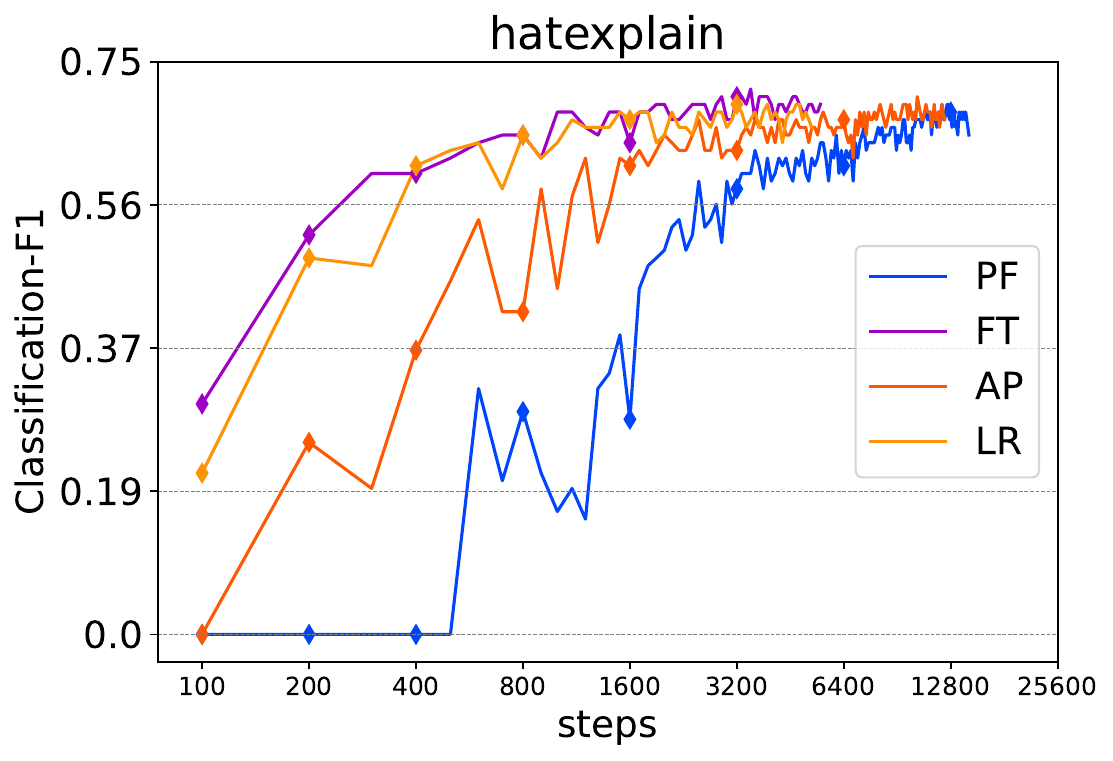}
    }
    
    \subfigure{
      \includegraphics[width=0.3\textwidth]{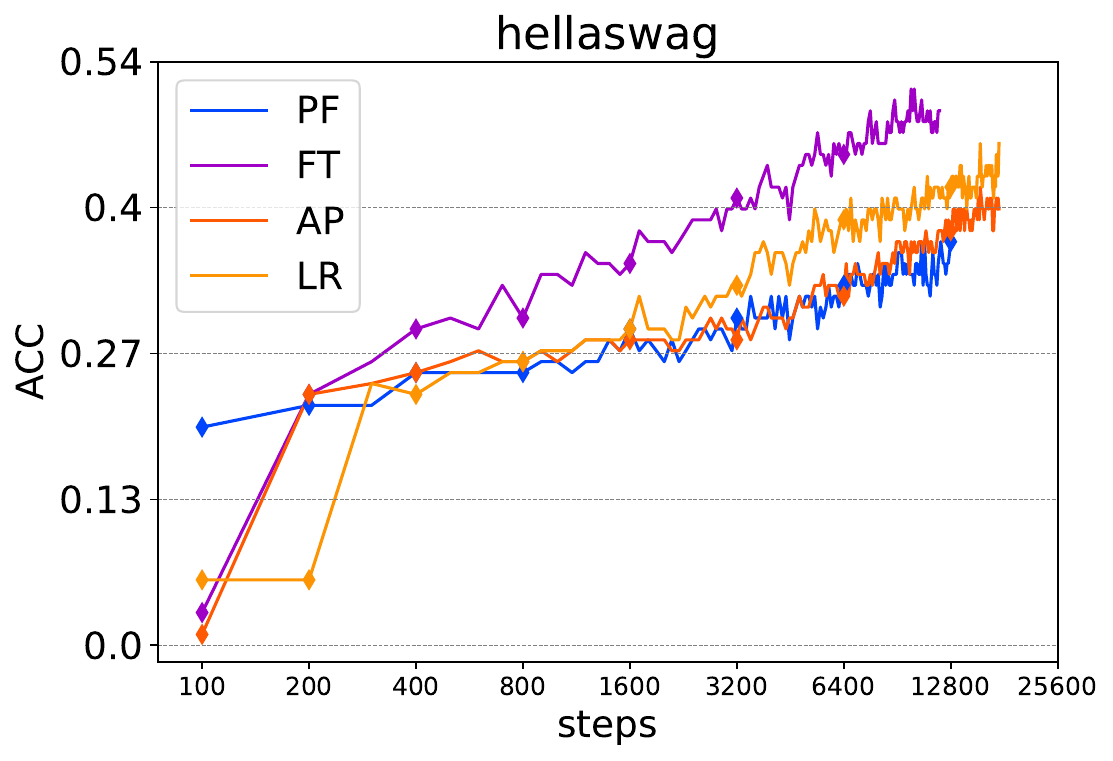}
    }
    \subfigure{
      \includegraphics[width=0.3\textwidth]{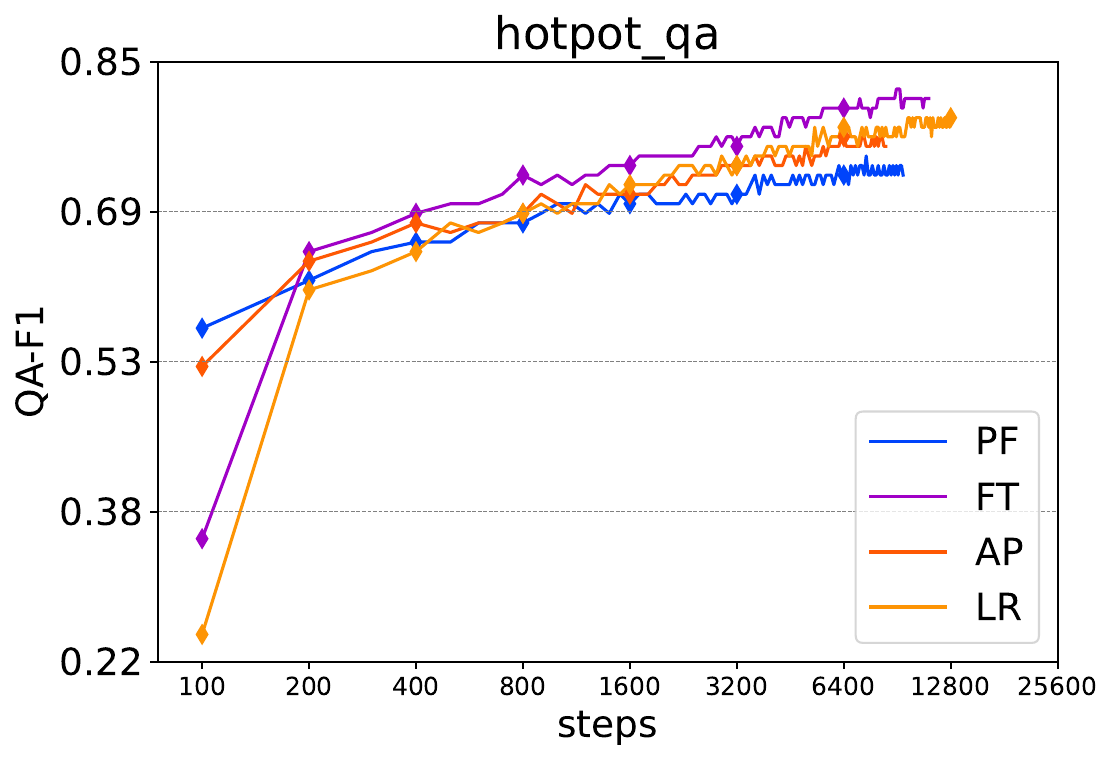}
    }
    \subfigure{
      \includegraphics[width=0.3\textwidth]{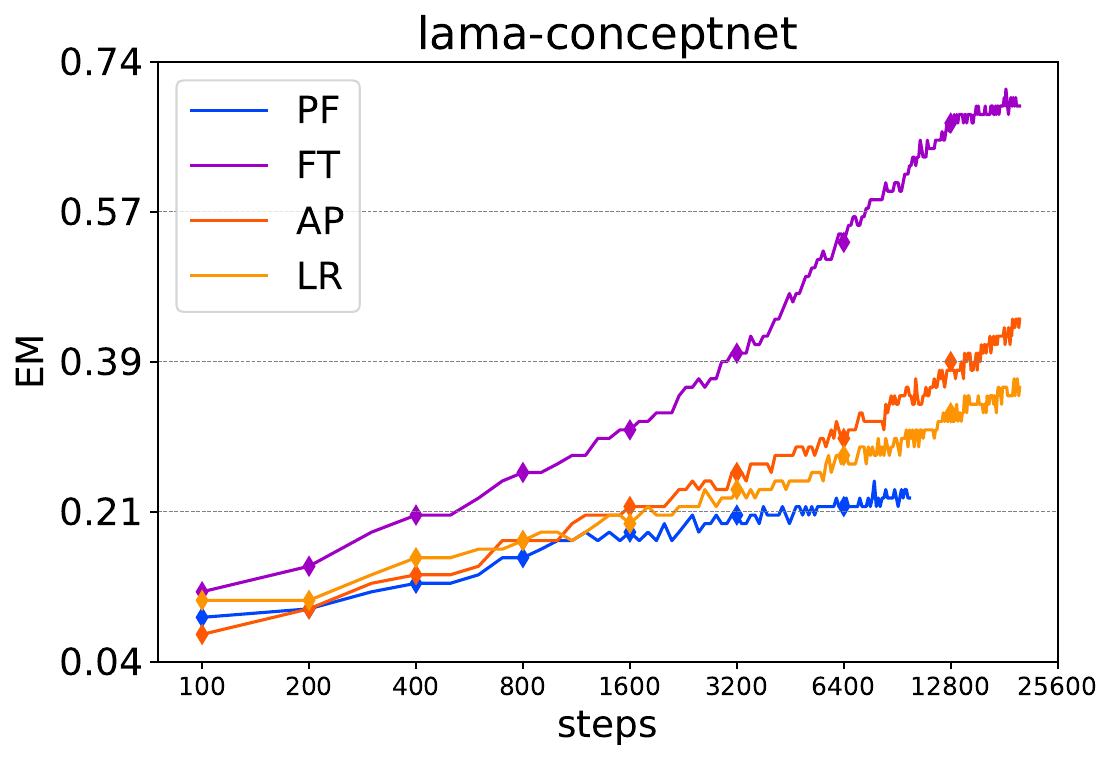}
    }
    
    \subfigure{
      \includegraphics[width=0.3\textwidth]{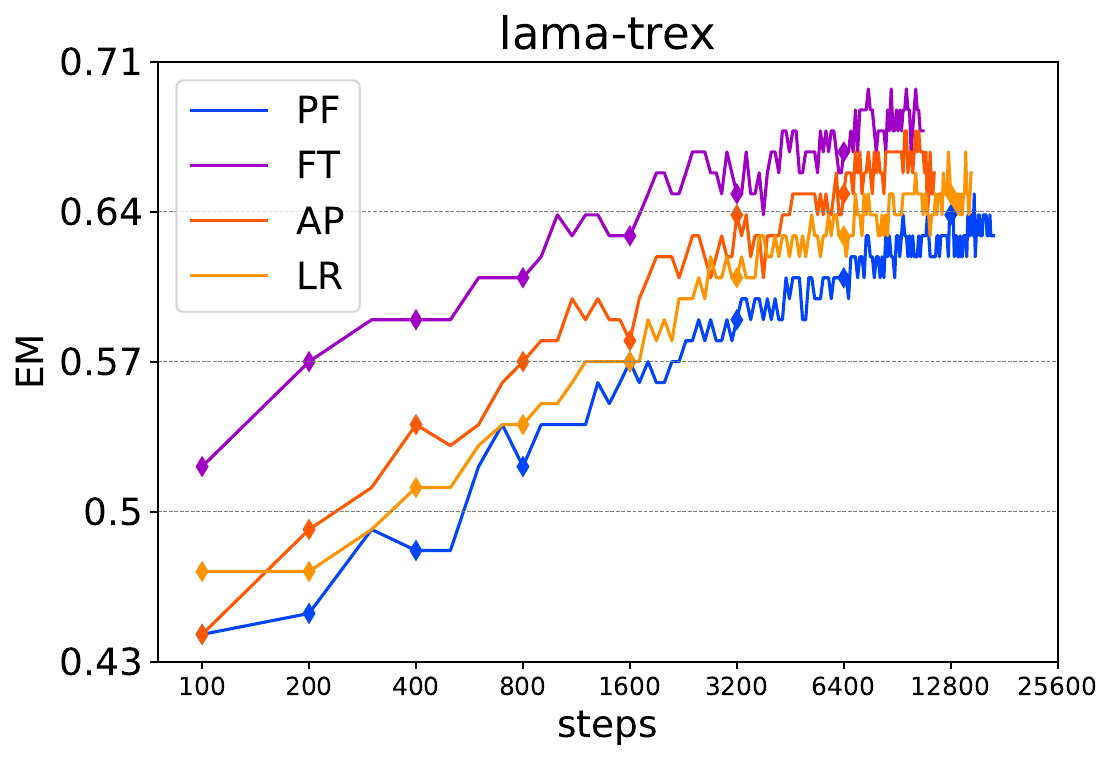}
    }
    \subfigure{
      \includegraphics[width=0.3\textwidth]{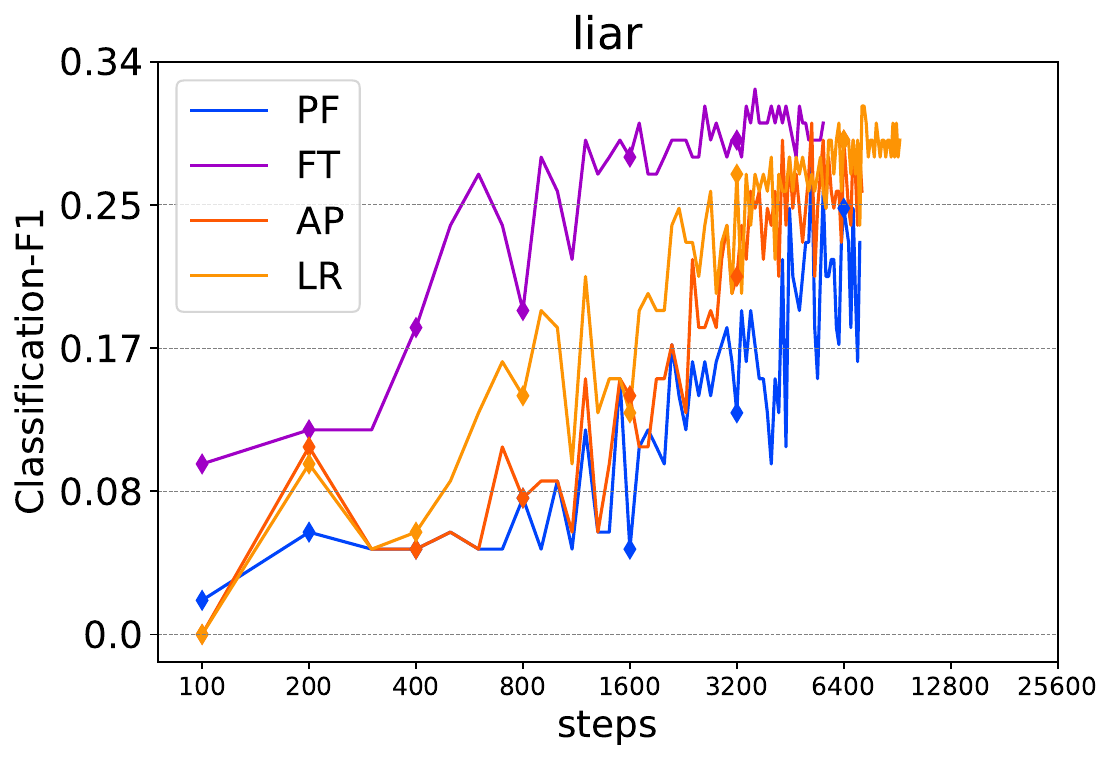}
    }
    \subfigure{
      \includegraphics[width=0.3\textwidth]{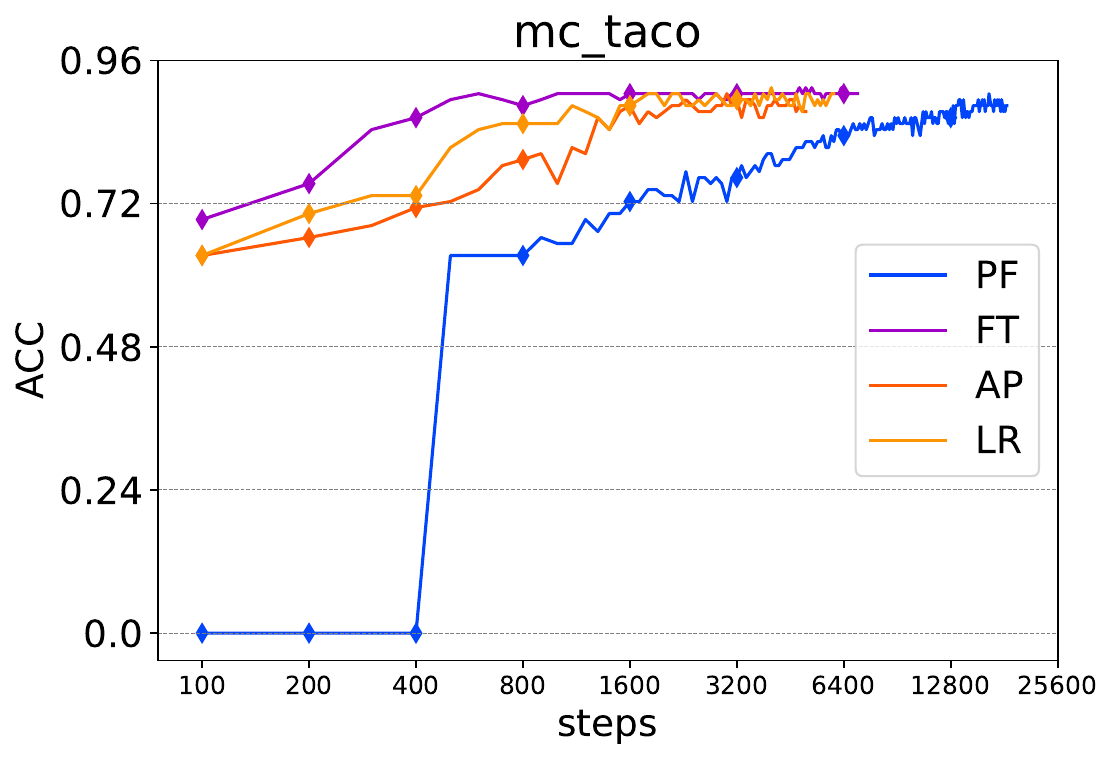}
    }
    
    \subfigure{
      \includegraphics[width=0.3\textwidth]{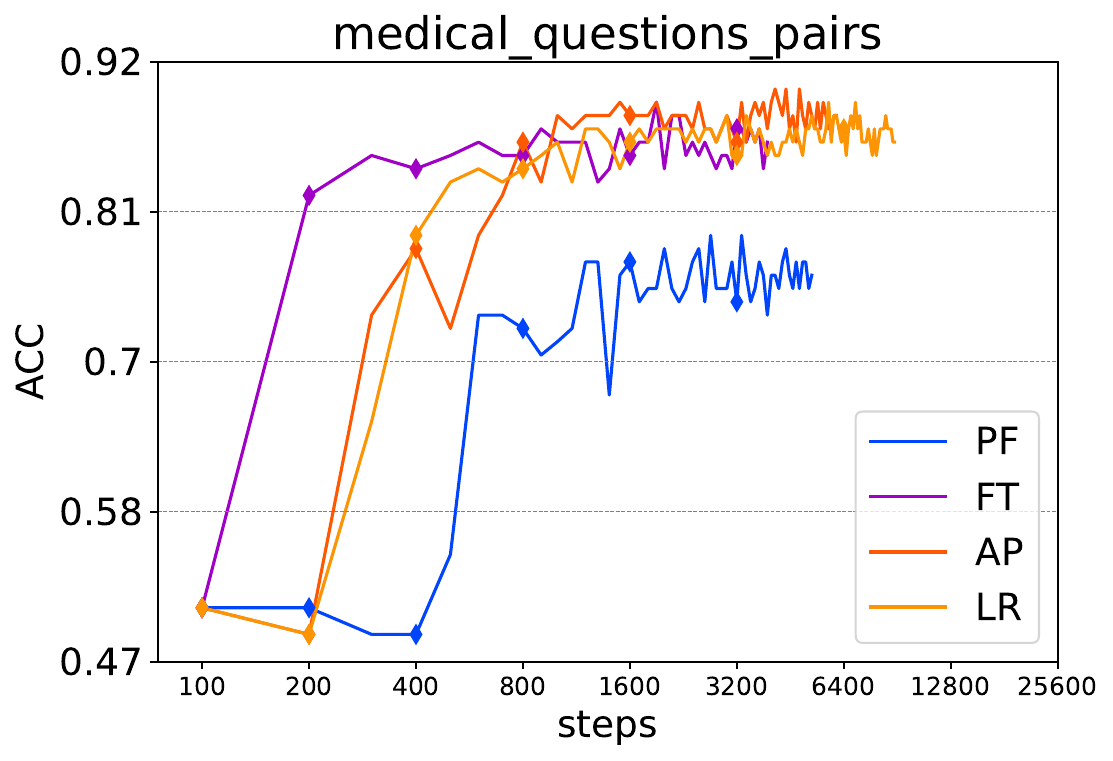}
    }
    \subfigure{
      \includegraphics[width=0.3\textwidth]{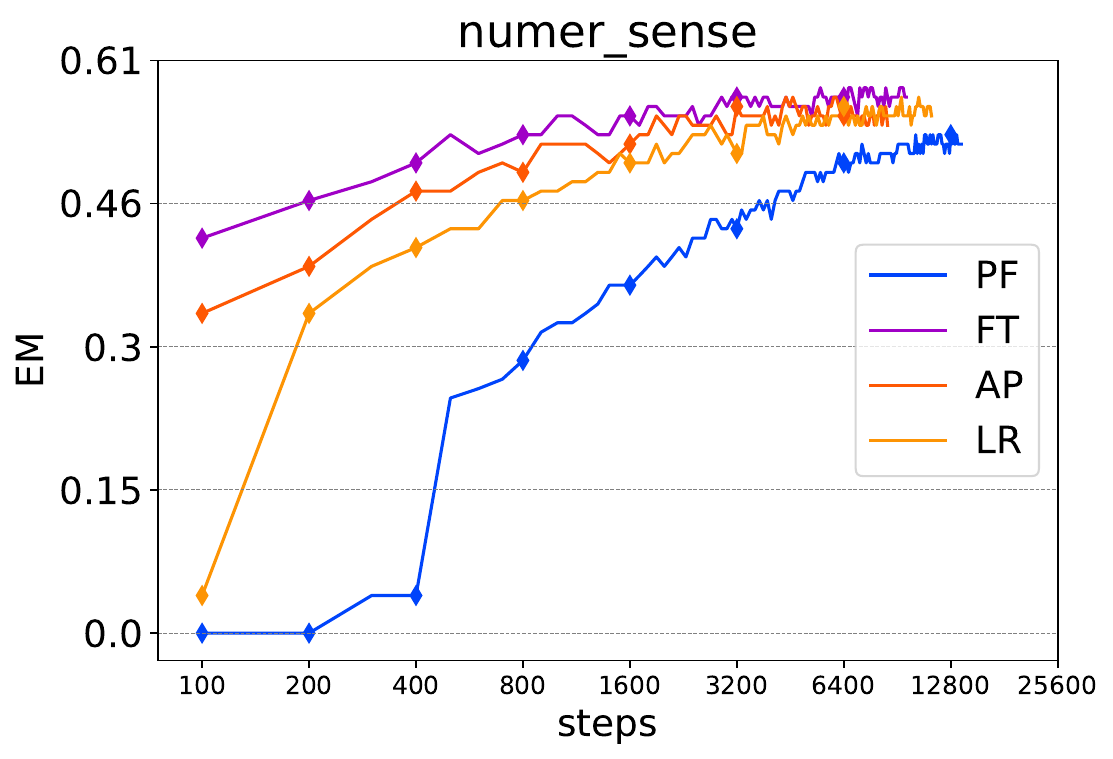}
    }
    \subfigure{
      \includegraphics[width=0.3\textwidth]{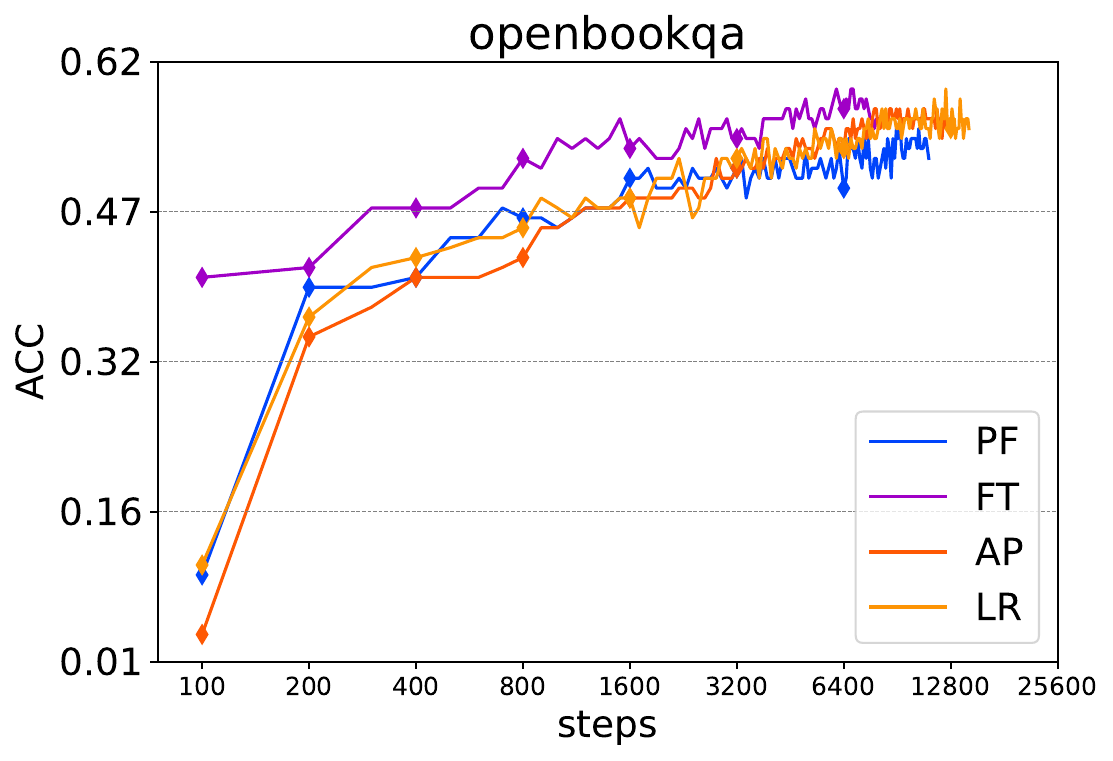}
    }
    
    \subfigure{
      \includegraphics[width=0.3\textwidth]{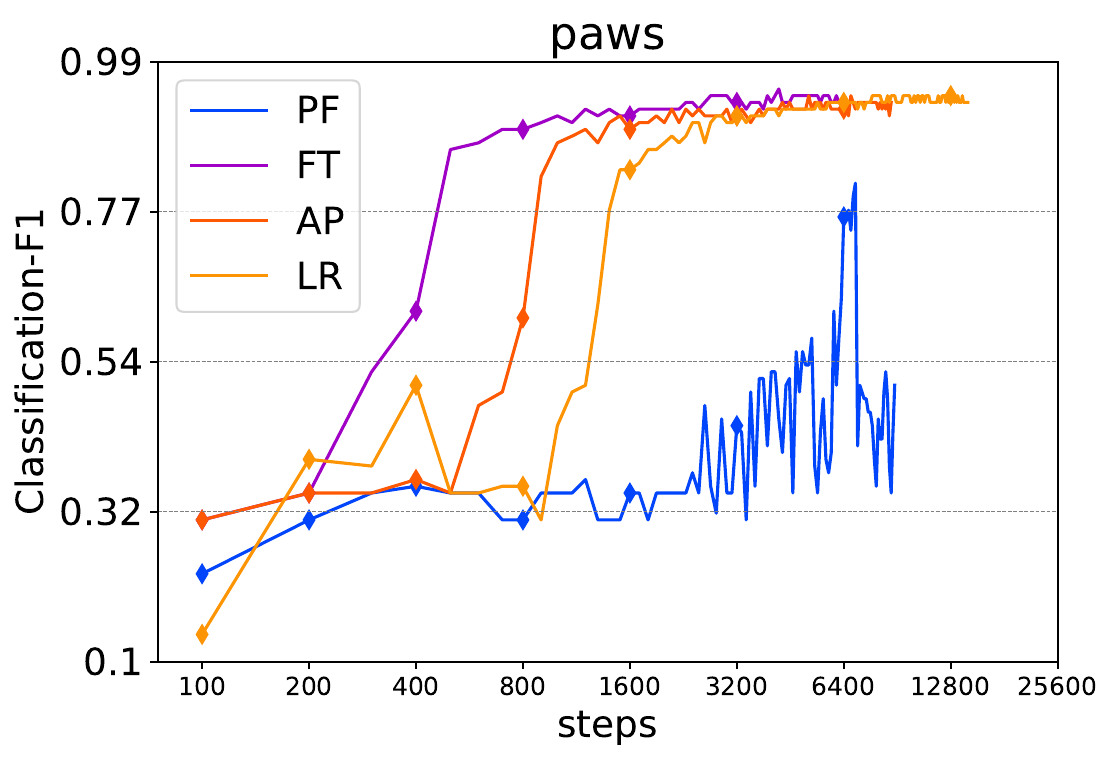}
    }
    \subfigure{
      \includegraphics[width=0.3\textwidth]{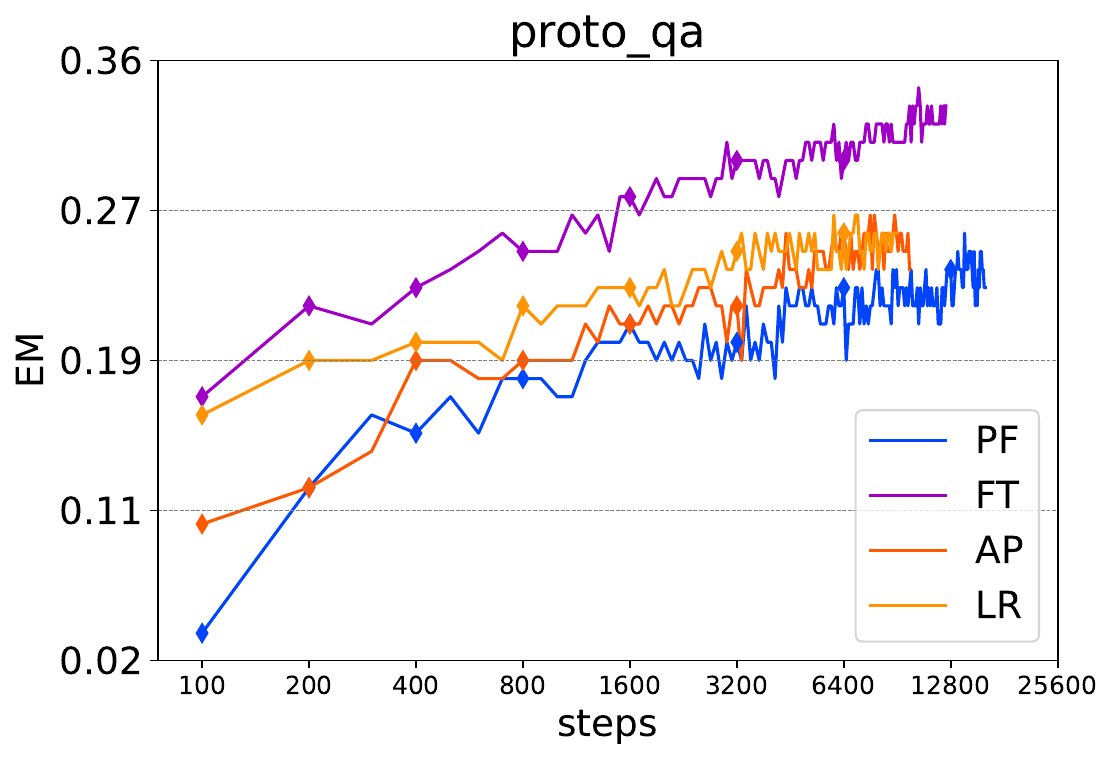}
    }
    \subfigure{
      \includegraphics[width=0.3\textwidth]{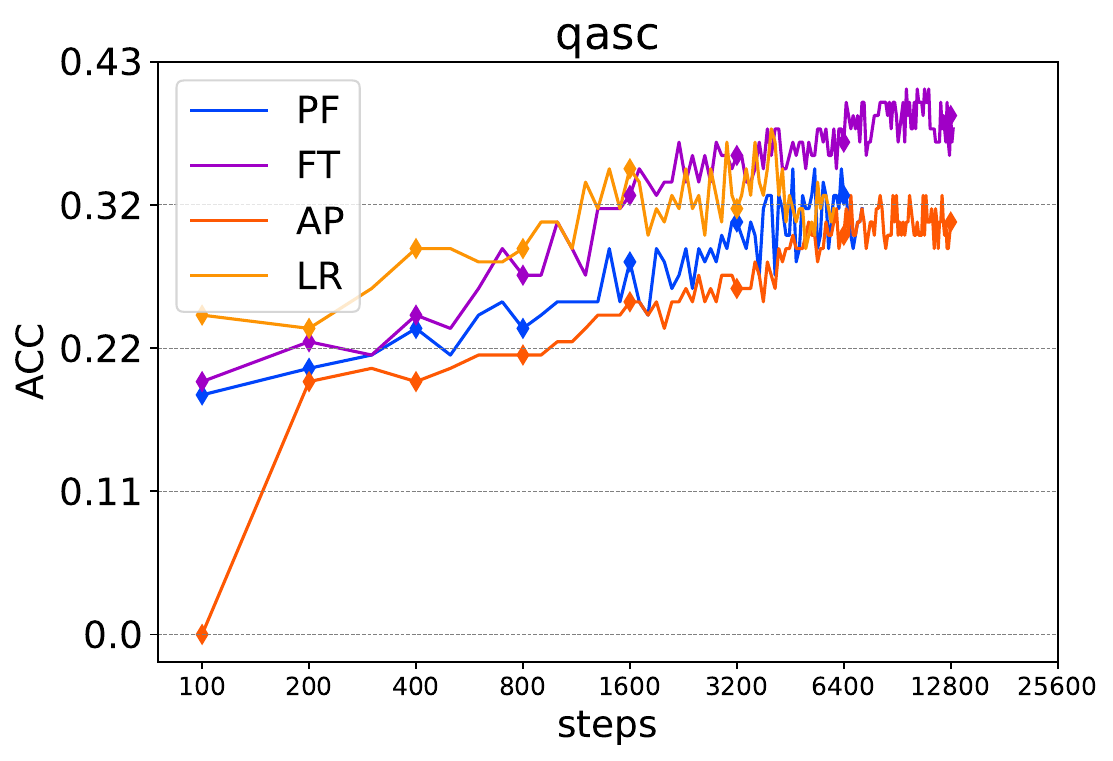}
    }
    \caption{Continued with Figure \ref{fig:convergence}. The performance of $\text{T5}_{\texttt{BASE}}$ with different delta tuning methods (\textbf{LR}, \textbf{AP}, \textbf{PF}) and fine-tuning (\textbf{FT}) at different training steps. Note we apply early stop in all the experiments.}
    \label{fig:convergence_2}
\end{figure*}

\begin{figure*}[!th]
\subcapraggedrighttrue
\subcaphangtrue
    \centering
    \subfigure{
      \includegraphics[width=0.3\textwidth]{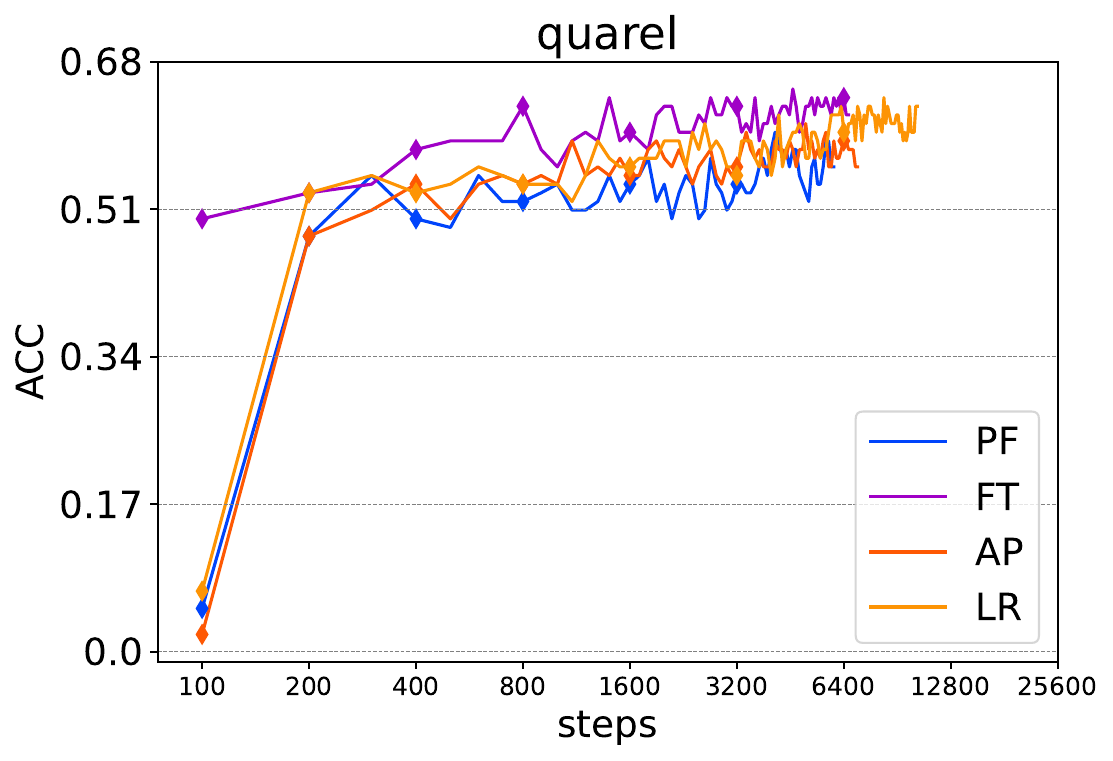}
    }
    \subfigure{
      \includegraphics[width=0.3\textwidth]{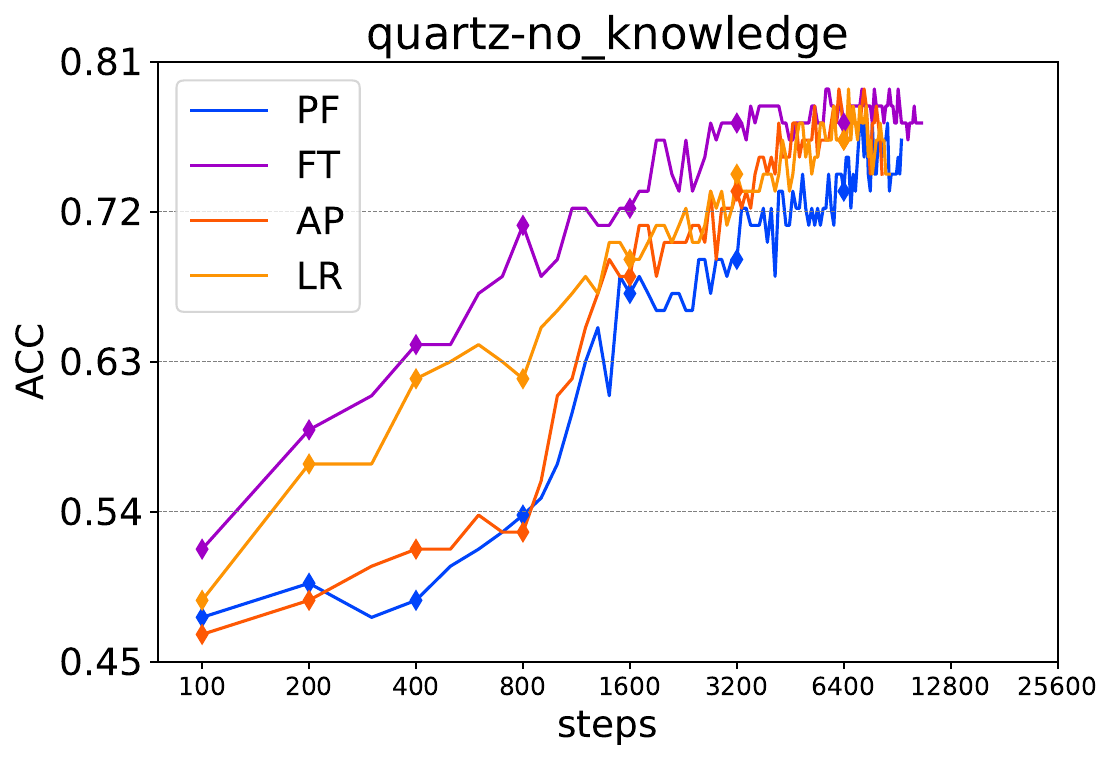}
    }
    \subfigure{
      \includegraphics[width=0.3\textwidth]{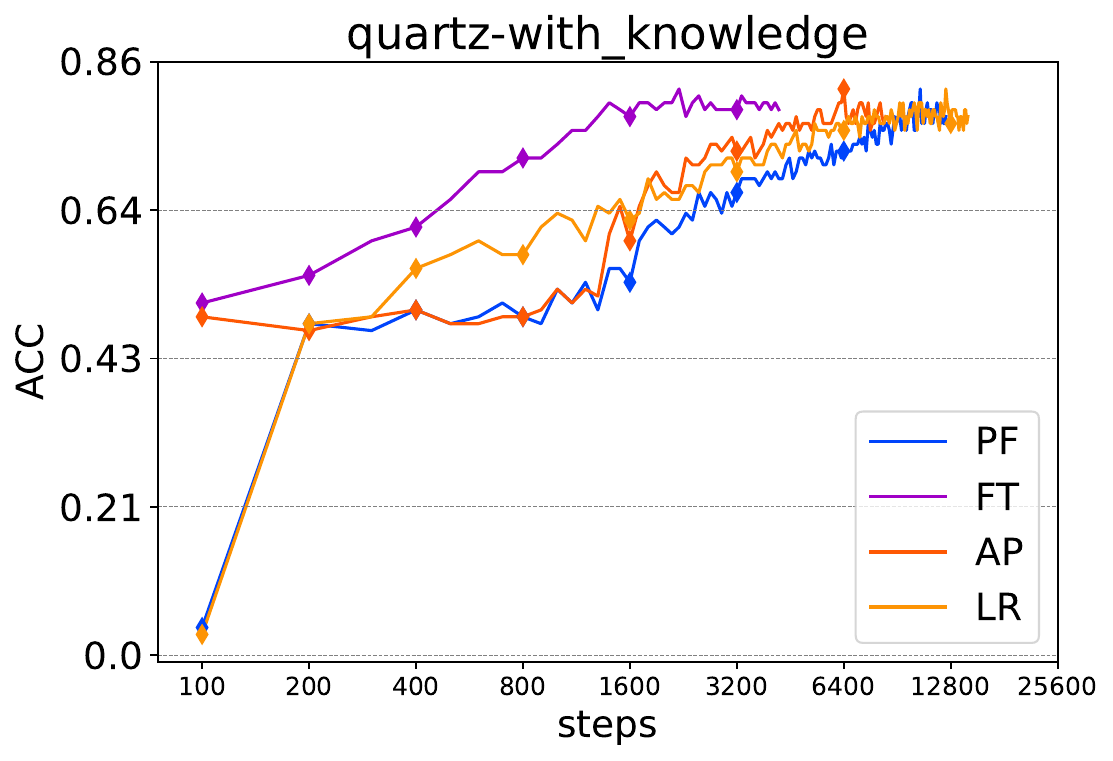}
    }
    
    \subfigure{
      \includegraphics[width=0.3\textwidth]{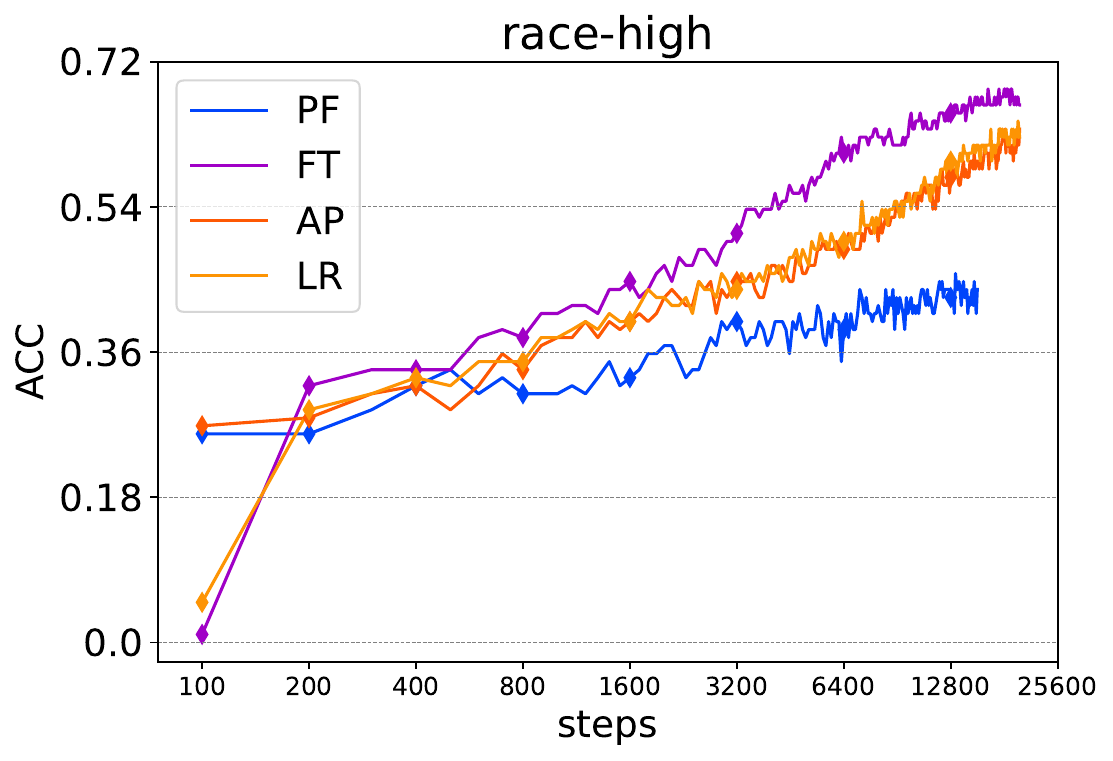}
    }
    \subfigure{
      \includegraphics[width=0.3\textwidth]{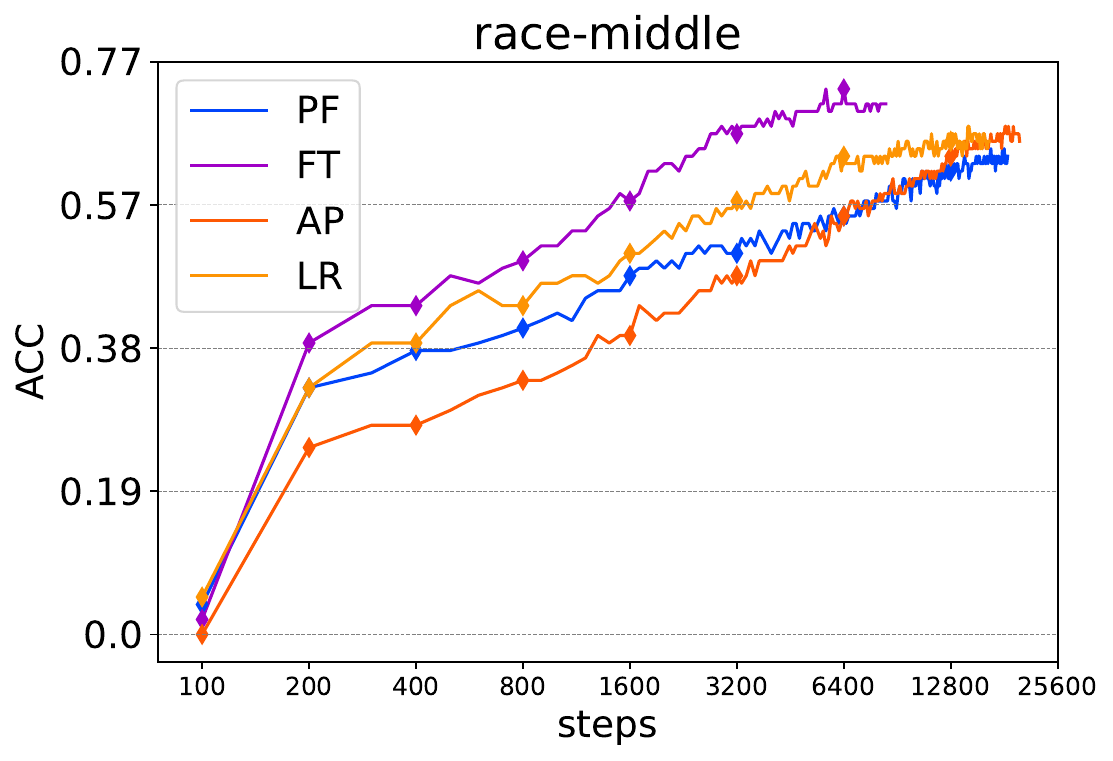}
    }
    \subfigure{
      \includegraphics[width=0.3\textwidth]{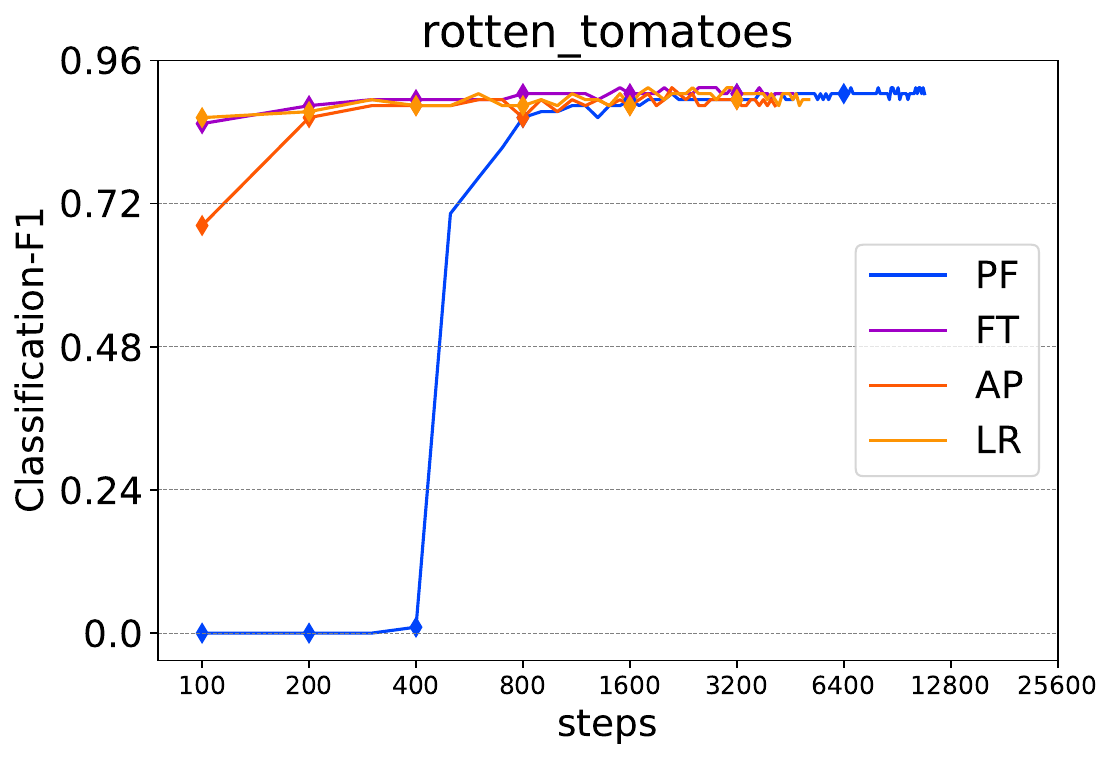}
    }
    
    \subfigure{
      \includegraphics[width=0.3\textwidth]{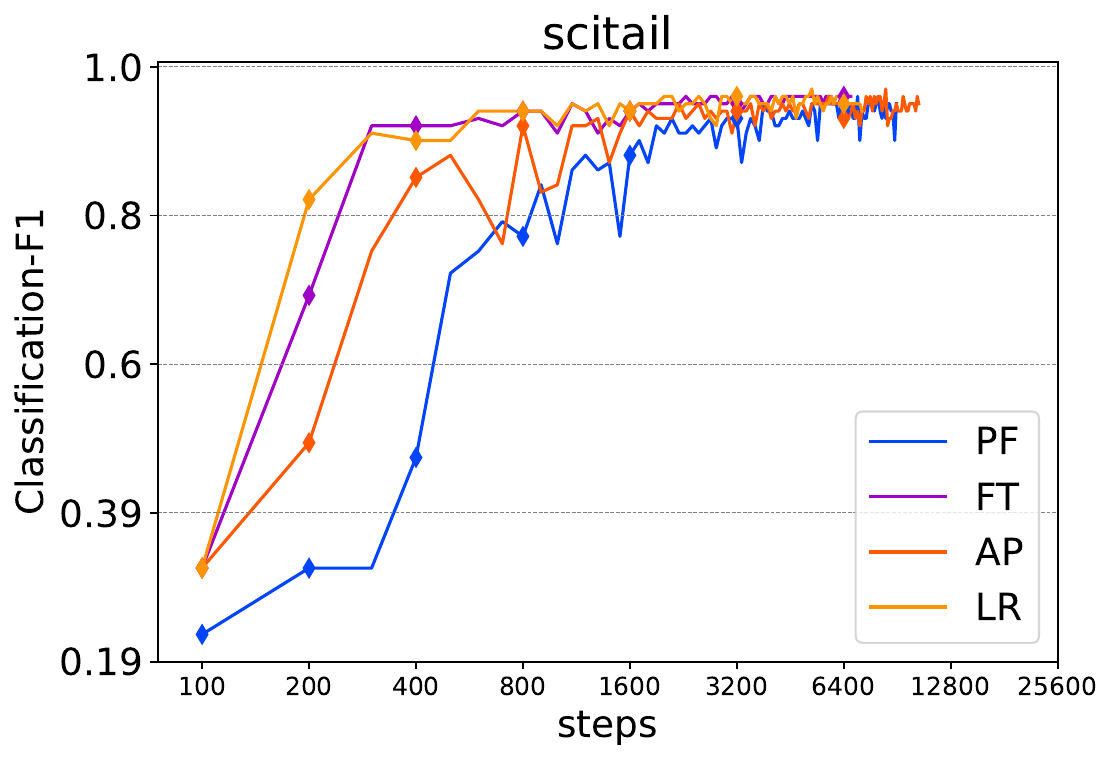}
    }
    \subfigure{
      \includegraphics[width=0.3\textwidth]{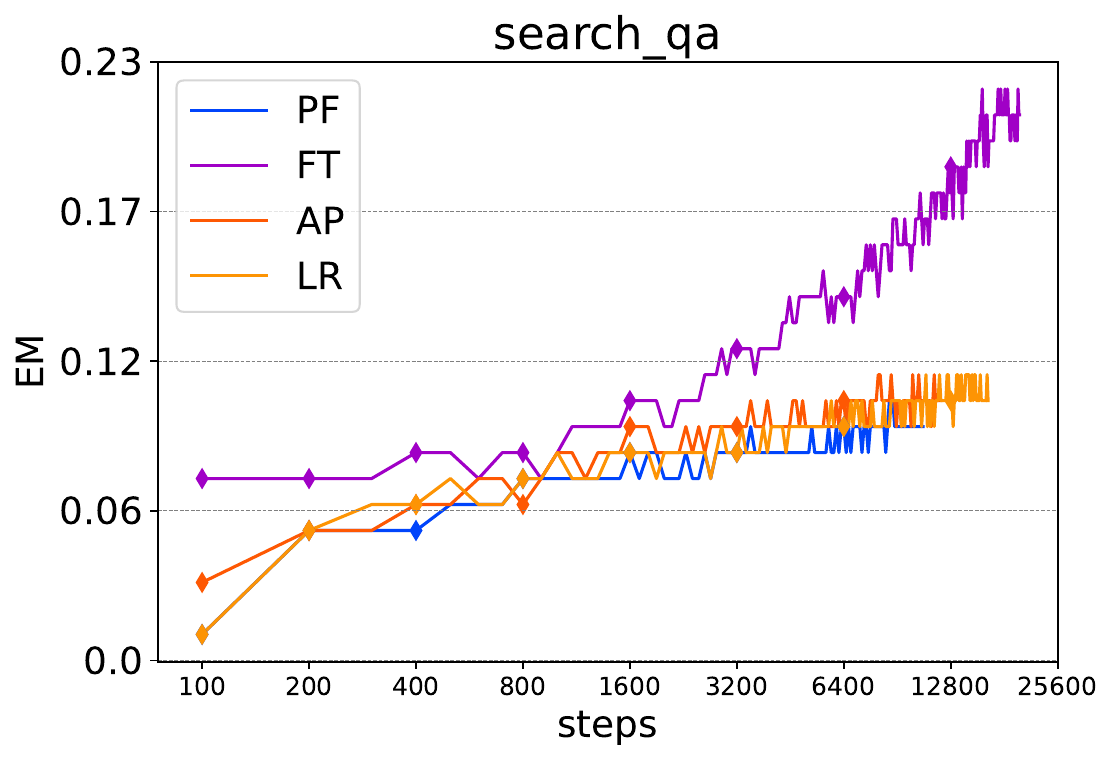}
    }
    \subfigure{
      \includegraphics[width=0.3\textwidth]{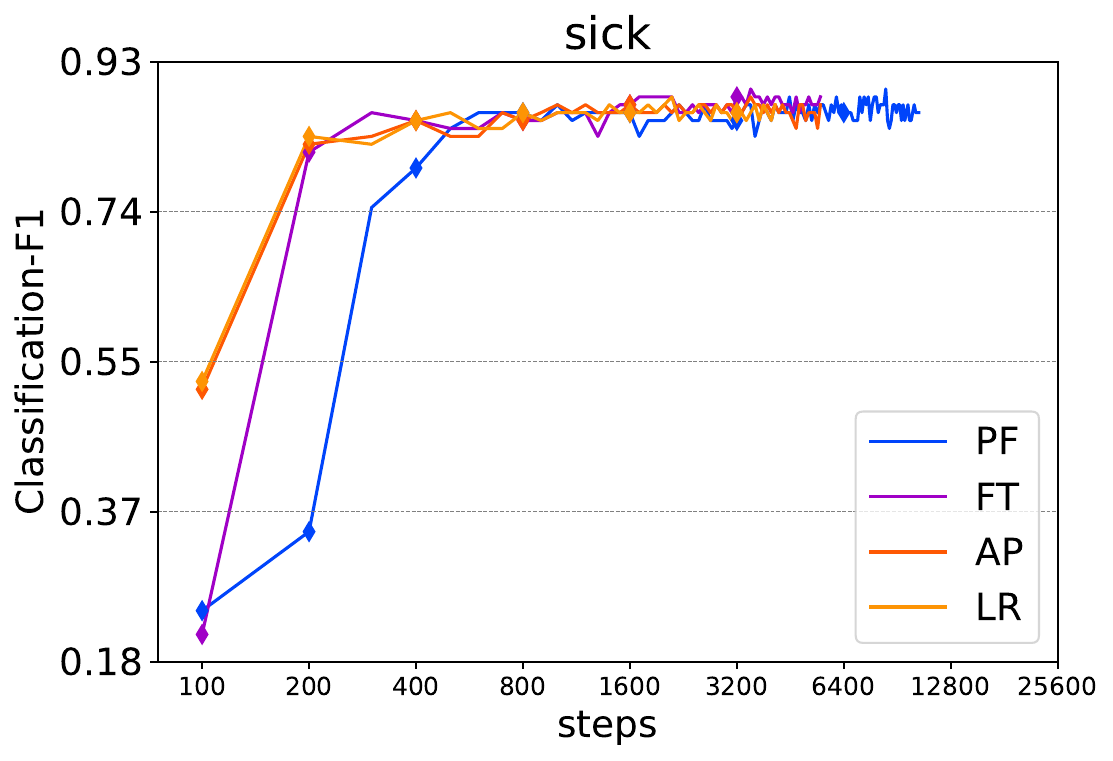}
    }
    
    \subfigure{
      \includegraphics[width=0.3\textwidth]{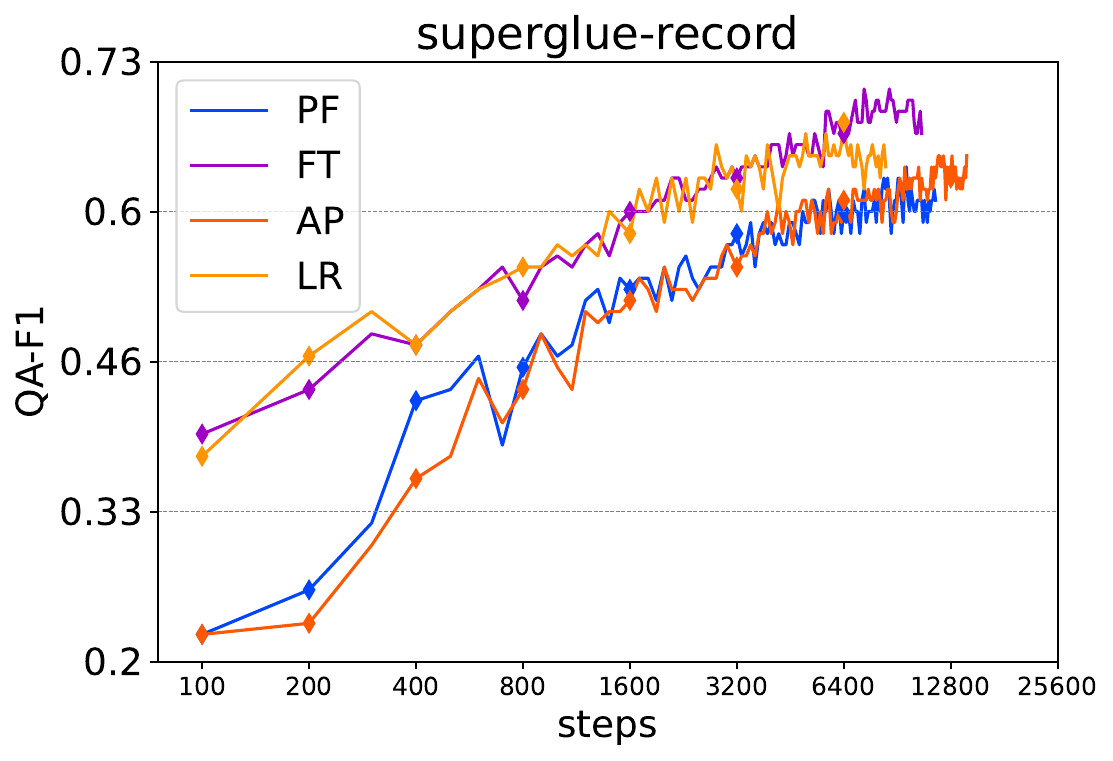}
    }
    \subfigure{
      \includegraphics[width=0.3\textwidth]{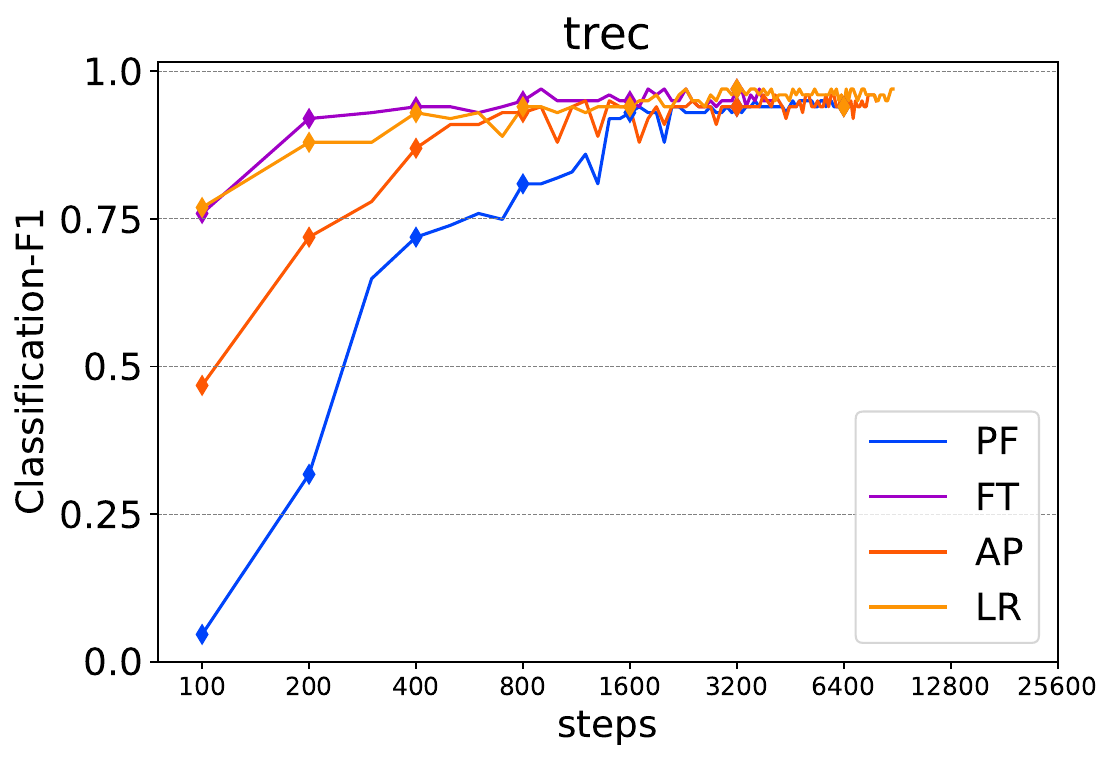}
    }
    \subfigure{
      \includegraphics[width=0.3\textwidth]{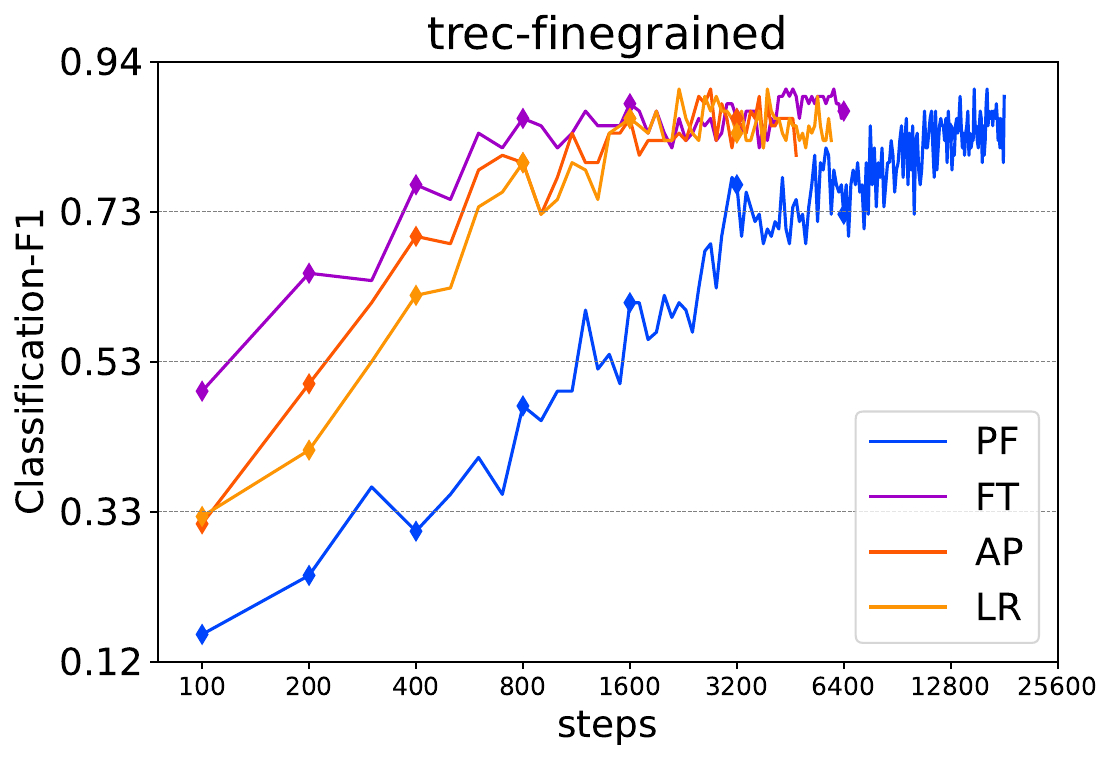}
    }
    
    \subfigure{
      \includegraphics[width=0.3\textwidth]{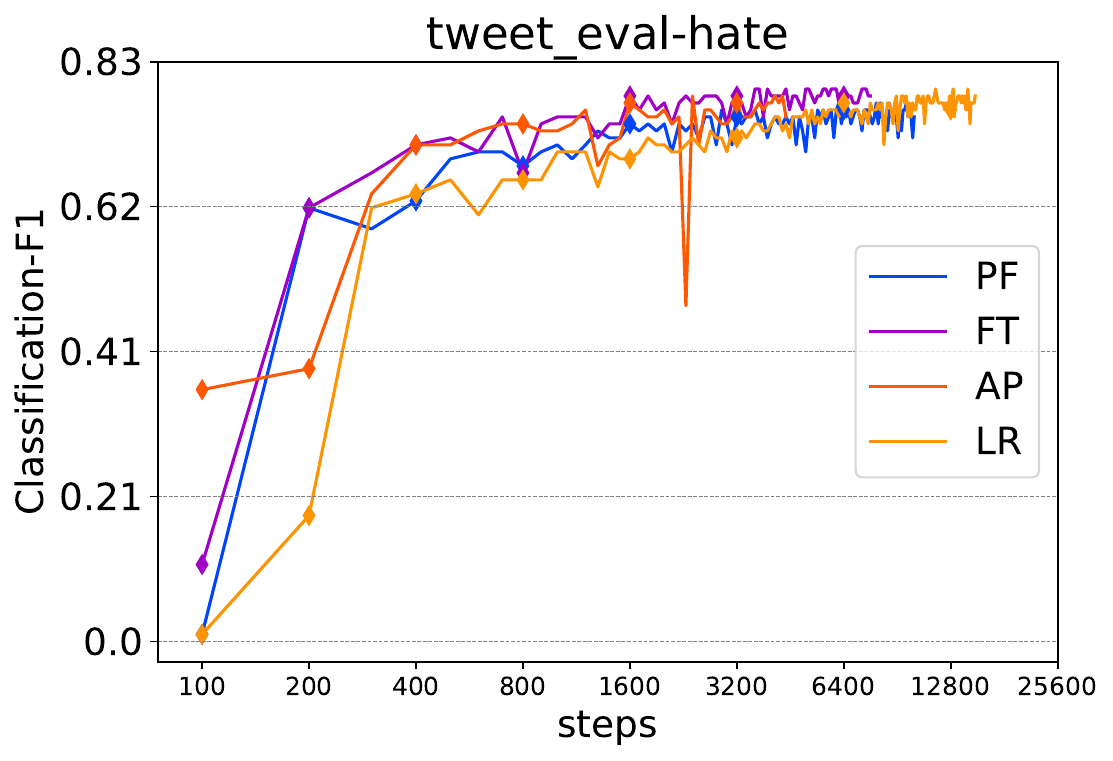}
    }
    \subfigure{
      \includegraphics[width=0.3\textwidth]{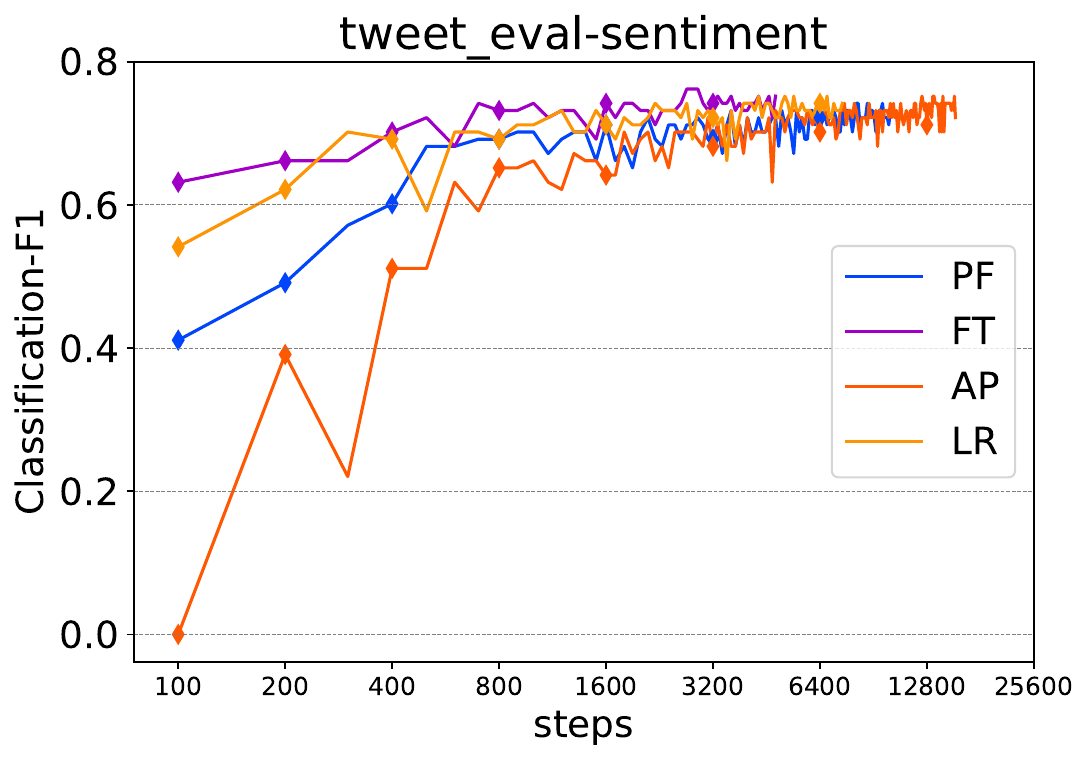}
    }
    \subfigure{
      \includegraphics[width=0.3\textwidth]{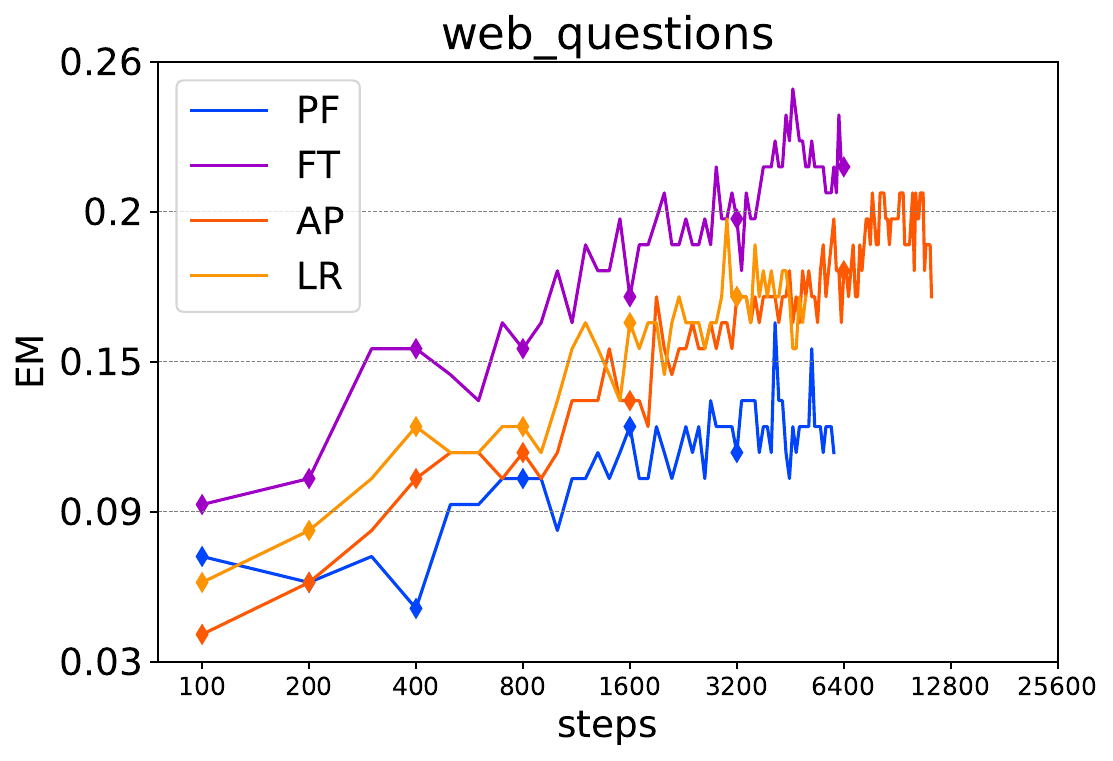}
    }
    
    %\subfigure{
    %  \includegraphics[width=0.3\textwidth]{Figures/tasks/wino_grande_acc.pdf}
    %} 
    %\subfigure{
    %  \includegraphics[width=0.3\textwidth]{Figures/tasks/wiqa_acc.pdf}
    %}
    %\subfigure{
    %  \includegraphics[width=0.3\textwidth]{Figures/tasks/xsum_acc.pdf}
    %}
    
    \caption{Continued with Figure \ref{fig:convergence_2}. The performance of $\text{T5}_{\texttt{BASE}}$ with different delta tuning methods (\textbf{LR}, \textbf{AP}, \textbf{PF}) and fine-tuning (\textbf{FT}) at different training steps. Note we apply early stop in all the experiments.}
    \label{fig:convergence_3}
\end{figure*}

\begin{figure}[ht]
    \centering
    \includegraphics[width=1.0\textwidth]{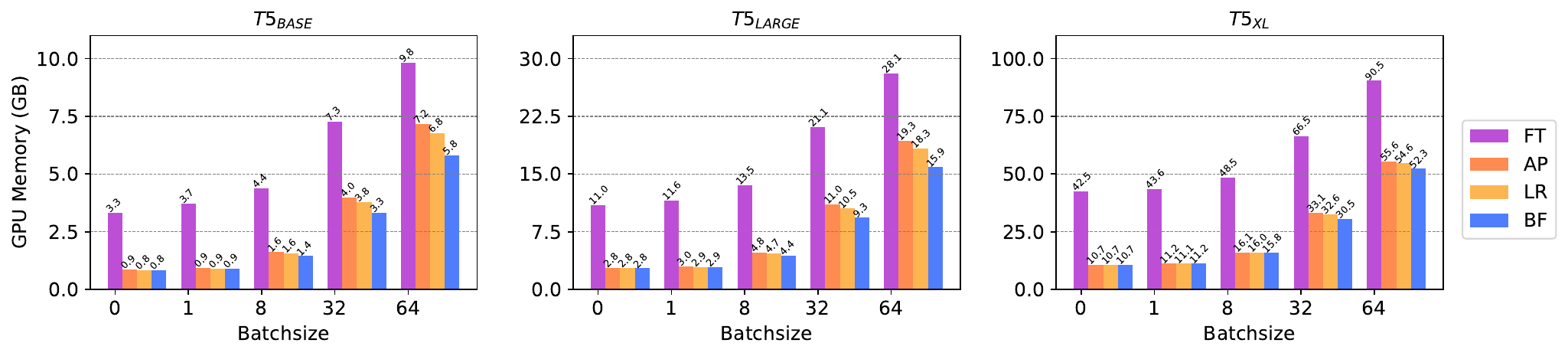}
    \caption{GPU memory consumed by each delta tuning methods compared with fine-tuning.}
    \label{fig:gpu}
\end{figure}

\paragraph{Efficiency Analysis.} Delta tuning saves GPU memory by alleviating the need for gradient computations for most parameters. To specifically verify the efficiency of GPU memory, in Figure~\ref{fig:gpu}, we conduct experiments to compare the GPU memory consumed by different delta tuning methods and fine-tuning across different PLM scales. We choose three scales of T5 model, i.e., $\text{T5}_{\texttt{BASE}}$,$\text{T5}_{\texttt{LARGE}}$, $\text{T5}_{\texttt{XL}}$, , and test the peak GPU memories achieved under different batchsizes. The static GPU memories, which leave out the intermediate tensors such as hidden states, are draw on Batchsize=0.  We use NVIDIA A100 (maximum GPU memory=39.58GB) and library OpenDelta~\footnote{\url{https://github.com/thunlp/OpenDelta}} for these experiments. For the cases which consume large GPU memory than a single A100, we parallelize the model across several GPUs using model parallelization, which doesn't introduce additional memory consumption. We can see from the figure that under small batchsizes (e.g., 1, 8), delta tuning saves up to 3/4 GPU memory, which under big batchsizes (e.g., 64), delta tuning saves at least 1/3 GPU memory. 
Given the fact that small batchsize is more preferred when applying big models to save GPU memory, delta tuning can further reduce the gpu memory dramatically.

\subsection{Combinations of Delta Tuning Methods}
\label{sec:combination}
Considering that different delta tuning methods are compatible with each other, which means they could be applied on the same PLM together, we thus investigate whether such a combination would bring additional benefits. Specifically, we evaluate both simultaneous combination and sequential combination. We choose three representative delta tuning methods, including prompt tuning, BitFit, and adapter, to explore the effects of their combinations. The training details are described in \refsec{combination details}.

\paragraph{Simultaneous Combination.} We first explore the effects of directly applying all the three delta tuning methods simultaneously. The experiments are conducted using both $\text{RoBERTa}_{\texttt{LARGE}}$~\citep{liu2019roberta} and $\text{T5}_{\texttt{BASE}}$ on eight GLUE tasks~\citep{wang2018glue}, and we report the performance on development sets. We also test the performance of $\text{RoBERTa}_{\texttt{LARGE}}$ under the few-shot setting, where we randomly sample $16$ training examples per label to construct the new training set and development set, respectively.
  
Similar to prompt-based fine-tuning~\citep{schick-schutze-2021-exploiting}, we insert a natural language prompt template into the input text for each task. Take the sentiment classification task as an example, an input sentence $x=$ (\textit{I like this movie.}) could be re-formulated as: $x_{\text{prompt}} = \texttt{[CLS]}$ $x$ $\text{It was}$ $\texttt{[MASK]}$. $\texttt{[SEP]}$, where ``$\text{It was}$ $\texttt{[MASK]}$.'' is a manual template. The PLM is trained to fill in the $\texttt{[MASK]}$ token with either ``\textit{great}'' (positive) or ``\textit{terrible}'' (negative) for classification. The manual templates are designed to bridge the gap between pre-training and downstream tuning, we also test the performance of delta tuning combinations without templates, i.e., $x_{\text{prompt}}' = \texttt{[CLS]}$ $x$ $\texttt{[SEP]}$ $\texttt{[MASK]}$ $\texttt{[SEP]}$, to evaluate manual templates' functionalities. The manual templates and label words for different GLUE tasks are listed in Table \ref{tab:template}.
  
  \begin{table}[thbp]
  \small
      \centering
      \caption{Manual templates and the corresponding label words for different tasks.}
      \begin{tabu}{l l l}
           \toprule
           \textbf{Task} & \textbf{Template} & \textbf{Label words} \\
           \midrule
           \textbf{CoLA}~\citep{warstadt-etal-2019-neural} & $\langle S_1\rangle$ This is \texttt{[MASK]} . & grammatical: correct, not\_grammatical: incorrect \\
           \textbf{SST-2}~\citep{socher-etal-2013-recursive} & $\langle S_1\rangle$ It was \texttt{[MASK]} . & positive: great, negative: terrible\\
           \textbf{MRPC}~\citep{dolan-brockett-2005-automatically} & $\langle S_1\rangle$ \texttt{[MASK]} , $\langle S_2 \rangle$ & equivalent: Yes, not\_equivalent: No\\
           \textbf{STS-B}~\citep{cer-etal-2017-semeval} & $\langle S_1\rangle$ \texttt{[MASK]} , $\langle S_2 \rangle$ & $y_u$: Yes, $y_l$: No\\
           \textbf{QQP}~\href{https://quoradata.quora.com/First-Quora-Dataset-Release-Question-Pairs}{(link)} & $\langle S_1\rangle$ \texttt{[MASK]} , $\langle S_2 \rangle$ & equivalent: Yes, not\_equivalent: No\\
           \textbf{MNLI}~\citep{williams-etal-2018-broad} & $\langle S_1\rangle$? \texttt{[MASK]} , $\langle S_2 \rangle$ & entailment: Yes, neutral: Maybe, contradiction: No \\
           \textbf{QNLI}~\citep{rajpurkar-etal-2016-squad} & $\langle S_1\rangle$? \texttt{[MASK]} , $\langle S_2 \rangle$ & entailment: Yes, no\_entailment: No \\
           \textbf{RTE}~\citep{dagan2005pascal} & $\langle S_1\rangle$? \texttt{[MASK]} , $\langle S_2 \rangle$ & entailment: Yes, no\_entailment: No \\
           \bottomrule
      \end{tabu}
      \label{tab:template}
  \end{table}
  
 % For all configurations, we report mean (and standard deviation) performance over 3 different random seeds.

% This reflects that, being confined in merely modifying the input text, prompt tuning could be seen as ``virtual templates'', which may have similar functionalities to those manually written ones. Such a phenomenon is more evident under full-data setting because the soft prompts could be better tuned through more supervisions;

We list the results of $\text{RoBERTa}_{\texttt{LARGE}}$ in Table \ref{tab:combine_roberta}, from which we could conclude that: for $\text{RoBERTa}_{\texttt{LARGE}}$, (1) under both full-data setting and the few-shot setting, introducing adapter into the combination almost always conduces to the average GLUE performance no matter whether there exist manual templates; (2) introducing prompt tuning into the combination generally harms the average performance, showing that prompt tuning may not be compatible with other two delta tuning methods; (3) introducing BitFit into the combination generally improves the average performance; (4) manual templates could significantly improve the zero-shot performance (from $23.7$ to $43.4$) by narrowing the gap between downstream tuning and pre-training. Under the few-shot setting, manual templates could also help boost the average performance evidently. However, when the training supervision is abundant (full-data setting), manual templates only exhibit marginal improvements or even harm the performance.

We list the results of $\text{T5}_{\texttt{BASE}}$ in Table \ref{tab:combine_t5}, from which we observe slightly different phenomena than $\text{RoBERTa}_{\texttt{LARGE}}$ as follows: (1) still, introducing prompt tuning into the combination would always harm the performance no matter whether there exist manual templates, showing that prompt tuning may not be compatible with other two delta tuning methods for $\text{T5}_{\texttt{BASE}}$, either; (2) introducing BitFit into the combination, however, could always conduce to the average performance; (3) adapter does not always improve the performance when there exist manual templates but could still bring benefits when there do not exist manual templates; (4) inserting manual templates into the input text would always improve the average performance. The improvements tend to be more evident than $\text{RoBERTa}_{\texttt{LARGE}}$.

\begin{table}[thbp]
  \centering
  \caption{Performance of $\text{RoBERTa}_{\texttt{LARGE}}$ on GLUE datasets. We report the average result of multiple random seeds on the validation set.}
    \begin{tabu}{l|r r r r r r r r}
    \toprule
    \textbf{Prompt} & \multicolumn{1}{c|}{\color{BrickRed}{\XSolidBrush}} & \multicolumn{1}{c|}{\color{BrickRed}{\XSolidBrush}} & \multicolumn{1}{c|}{\color{BrickRed}{\XSolidBrush}} & \multicolumn{1}{c|}{\color{BrickRed}{\XSolidBrush}} & \multicolumn{1}{c|}{\color{ForestGreen}{\Checkmark}} & \multicolumn{1}{c|}{\color{ForestGreen}{\Checkmark}} & \multicolumn{1}{c|}{\color{ForestGreen}{\Checkmark}} & \multicolumn{1}{c}{\color{ForestGreen}{\Checkmark}} \\
    
    \textbf{BitFit} & \multicolumn{1}{c|}{\color{BrickRed}{\XSolidBrush}} & \multicolumn{1}{c|}{\color{BrickRed}{\XSolidBrush}} &
    \multicolumn{1}{c|}{\color{ForestGreen}{\Checkmark}} & \multicolumn{1}{c|}{\color{ForestGreen}{\Checkmark}} &
    \multicolumn{1}{c|}{\color{BrickRed}{\XSolidBrush}} & \multicolumn{1}{c|}{\color{BrickRed}{\XSolidBrush}} &  \multicolumn{1}{c|}{\color{ForestGreen}{\Checkmark}} & \multicolumn{1}{c}{\color{ForestGreen}{\Checkmark}} \\
    
    \textbf{Adapter} & \multicolumn{1}{c|}{\color{BrickRed}{\XSolidBrush}} & \multicolumn{1}{c|}{\color{ForestGreen}{\Checkmark}} & \multicolumn{1}{c|}{\color{BrickRed}{\XSolidBrush}} & \multicolumn{1}{c|}{\color{ForestGreen}{\Checkmark}} & \multicolumn{1}{c|}{\color{BrickRed}{\XSolidBrush}} & \multicolumn{1}{c|}{\color{ForestGreen}{\Checkmark}} & \multicolumn{1}{c|}{\color{BrickRed}{\XSolidBrush}} & \multicolumn{1}{c}{\color{ForestGreen}{\Checkmark}} \\ \midrule
    \textit{Tunable parameters} & 0\% & 1.75\% & 0.09\% & 1.84\% & 0.003\% & 1.76\% & 0.09\% & 1.85\% \\
    \midrule
    \multicolumn{9}{l}{$\text{RoBERTa}_{\texttt{LARGE}}$ \textit{, full-data, without manual templates}} \\ \midrule
   \textbf{CoLA}\scriptsize{(Matt.)} & 4.6   & \textbf{66.6}$_{1.6}$ & 63.5$_{0.6}$ & 65.9$_{0.5}$ & 42.7$_{2.3}$ & 63.1$_{1.5}$ & 63.7$_{0.9}$ & 64.4$_{0.9}$ \\
    \textbf{SST-2}\scriptsize{(acc)} & 50.9  & \textbf{95.8}$_{0.1}$ & 95.6$_{0.1}$ & 95.7$_{0.2}$ & 95.3$_{0.2}$ & 95.7$_{0.1}$ & 95.3$_{0.2}$ & 95.5$_{0.1}$ \\
    \textbf{MRPC}\scriptsize{(F1)} & 1.4   & 92.7$_{0.2}$ & 91.9$_{0.4}$ & \textbf{93.0}$_{0.4}$ & 85.4$_{0.5}$ & 92.0$_{0.5}$ & 92.2$_{0.5}$ & 92.9$_{0.3}$ \\
    \textbf{STS-B}\scriptsize{(Pear.)} & -6.2  & \textbf{91.4}$_{0.1}$ & 90.7$_{0.2}$ & 90.5$_{0.1}$ & 83.0$_{2.8}$ & 90.5$_{0.4}$ & 90.3$_{0.7}$ & 90.9$_{0.1}$ \\
    \textbf{QQP}\scriptsize{(F1.)} & 6.4   & 83.5$_{0.1}$ & 83.5$_{0.0}$ & 84.4$_{0.0}$ & 77.2$_{0.4}$ & 84.3$_{0.0}$ & 83.6$_{0.1}$ & \textbf{84.4}$_{0.0}$ \\
    \textbf{MNLI}\scriptsize{(acc)} & 34.2  & 88.6$_{0.2}$ & 88.0$_{0.2}$ & \textbf{89.0}$_{0.1}$ & 77.9$_{2.5}$ & 88.9$_{0.1}$ & 88.0$_{0.2}$ & 88.9$_{0.1}$ \\
    \textbf{QNLI}\scriptsize{(acc)} & 50.6  & 93.7$_{0.3}$ & 93.4$_{0.3}$ & 94.2$_{0.1}$ & 86.2$_{0.5}$ & 94.2$_{0.1}$ & 93.2$_{0.3}$ & \textbf{94.4}$_{0.1}$ \\
    \textbf{RTE}\scriptsize{(acc)} & 47.7  & \textbf{86.8}$_{0.5}$ & 86.2$_{1.0}$ & 84.5$_{0.5}$ & 74.4$_{0.5}$ & 84.1$_{0.8}$ & 85.7$_{1.5}$ & 84.7$_{1.1}$ \\
    \textbf{Average} & 23.7  & \textbf{87.4}$_{0.4}$ & 86.6$_{0.4}$ & 87.1$_{0.2}$ & 77.7$_{1.2}$ & 86.6$_{0.4}$ & 86.5$_{0.6}$ & 87.0$_{0.3}$ \\

    % \textbf{CoLA}\scriptsize{(Matt.)} & 
    % 4.6 & \textbf{64.1}$_{4.0}$ & 58.0$_{11.3}$ & 42.0$_{29.8}$ & 3.8$_{1.4}$ & 63.3$_{4.5}$ & 58.5$_{8.4}$ & 42.3$_{30.2}$ \\
    % \textbf{SST-2}\scriptsize{(acc)} &  50.9 & 95.6$_{0.1}$ & \textbf{95.9}$_{0.1}$ & 95.8$_{0.2}$ & 92.4$_{3.5}$ & 95.8$_{0.3}$ & 95.8$_{0.4}$ & 95.6$_{0.3}$ \\
    % \textbf{MRPC}\scriptsize{(F1)} & 1.4 & 88.4$_{5.1}$ & 90.3$_{3.7}$ & 88.6$_{5.2}$ & 82.7$_{0.7}$ & \textbf{92.3}$_{1.1}$ & 89.1$_{3.8}$ & 88.1$_{4.9}$ \\
    % \textbf{STS-B}\scriptsize{(Pear.)} & -6.2 & 90.3$_{2.2}$ & 85.5$_{7.1}$ & \textbf{90.6}$_{1.6}$ & 54.3$_{23.7}$ & 90.0$_{2.9}$ & 85.0$_{7.8}$ & 63.3$_{37.3}$ \\
    % \textbf{QQP}\scriptsize{(F1)} & 6.4 & 55.2$_{39.1}$ & 81.3$_{3.6}$ & 77.2$_{9.7}$ & 75.0$_{6.9}$ & \textbf{81.9}$_{3.2}$ & 81.2$_{3.7}$ & 55.9$_{39.6}$ \\
    % \textbf{MNLI}\scriptsize{(acc)} & 34.2 & 88.0$_{1.8}$ & 84.6$_{4.9}$ & 70.5$_{24.9}$ & 63.7$_{17.4}$ & \textbf{88.6}$_{2.0}$ & 84.7$_{5.0}$ & 70.5$_{24.8}$ \\
    % \textbf{QNLI}\scriptsize{(acc)} & 
    % 50.6 & \textbf{92.2}$_{2.3}$ & 89.1$_{5.1}$ & 78.4$_{19.8}$ & 61.0$_{7.3}$ & 92.0$_{2.2}$ & 90.2$_{4.0}$ & 78.7$_{20.0}$ \\
    % \textbf{RTE}\scriptsize{(acc)} & 47.7 & 74.2$_{15.4}$ & 83.3$_{3.5}$ & 74.7$_{15.7}$ & 69.6$_{6.4}$ & \textbf{85.8}$_{2.2}$ & 83.2$_{5.3}$ & 74.8$_{15.8}$ \\
    % \textbf{Average} & 23.7 & 81.0$_{8.8}$ & 83.5$_{4.9}$ & 77.2$_{13.4}$ & 62.8$_{8.4}$ & \textbf{86.2}$_{2.3}$ & 83.5$_{4.8}$ & 71.2$_{21.6}$ \\
    \midrule
    \multicolumn{9}{l}{$\text{RoBERTa}_{\texttt{LARGE}}$ \textit{, full-data, with manual templates}} \\ \midrule
    % \textbf{CoLA}\scriptsize{(Matt.)} & 2.2 & \textbf{63.6}$_{4.0}$ & 57.8$_{9.3}$ & 41.2$_{29.3}$ & 15.2$_{14.0}$ & 41.4$_{29.5}$ & 58.7$_{8.0}$ & 41.8$_{29.7}$ \\
    % \textbf{SST-2}\scriptsize{(acc)} & 83.6 & \textbf{96.1}$_{0.2}$ & 95.9$_{0.2}$ & 96.0$_{0.2}$ & 95.5$_{0.5}$ & 95.9$_{0.3}$ & 95.7$_{0.4}$ & 95.5$_{0.8}$ \\
    % \textbf{MRPC}\scriptsize{(F1)} & 61.9 & 88.8$_{5.3}$ & \textbf{89.7}$_{3.3}$ & 88.9$_{5.5}$ & 81.9$_{1.1}$ & 88.2$_{5.0}$ & \textbf{89.7}$_{3.6}$ & 88.4$_{5.1}$ \\
    % \textbf{STS-B}\scriptsize{(Pear.)} & -3.3 & 90.1$_{2.1}$ & 86.1$_{6.6}$ & 62.0$_{39.5}$ & 45.3$_{25.3}$ & 89.9$_{2.5}$ & 85.4$_{7.1}$ & \textbf{90.5}$_{1.8}$ \\
    % \textbf{QQP}\scriptsize{(F1)} & 49.7 & \textbf{83.0}$_{1.7}$ & 81.1$_{3.1}$ & 84.7$_{1.9}$ & 52.3$_{3.1}$ & 82.4$_{2.0}$ & 80.9$_{3.5}$ & 55.8$_{39.5}$ \\
    % \textbf{MNLI}\scriptsize{(acc)} & 50.9 & 70.2$_{24.7}$ & 85.1$_{3.9}$ & \textbf{88.4}$_{1.7}$ & 68.7$_{11.0}$ & 87.2$_{1.8}$ & 84.8$_{4.5}$ & 87.0$_{1.9}$ \\
    % \textbf{QNLI}\scriptsize{(acc)} & 50.8 & 93.1$_{1.7}$ & 90.2$_{3.4}$ & \textbf{93.2}$_{1.5}$ & 68.2$_{12.8}$ & 92.1$_{1.7}$ & 89.8$_{4.1}$ & 93.1$_{1.7}$ \\
    % \textbf{RTE}\scriptsize{(acc)} & 51.3 & 85.1$_{2.5}$ & 85.3$_{3.2}$ & 86.6$_{1.1}$ & 69.3$_{8.3}$ & 85.9$_{0.3}$ & 84.6$_{4.2}$ & \textbf{86.8}$_{0.3}$ \\
    % \textbf{Average} & 43.4 & 83.8$_{5.3}$ & \textbf{83.9}$_{4.1}$ & 80.1$_{10.1}$ & 62.0$_{9.5}$ & 82.9$_{5.4}$ & 83.7$_{4.4}$ & 79.9$_{10.1}$ \\
    \textbf{CoLA}\scriptsize{(Matt.)} & 2.2   & \textbf{66.9}$_{1.1}$ & 64.2$_{0.5}$ & 65.5$_{1.0}$ & 37.8$_{20.8}$ & 64.7$_{1.3}$ & 64.8$_{0.7}$ & 64.9$_{1.0}$ \\
    \textbf{SST-2}\scriptsize{(acc)} & 83.6  & \textbf{96.3}$_{0.2}$ & 96.1$_{0.1}$ & 96.2$_{0.2}$ & 95.7$_{0.2}$ & 95.8$_{0.1}$ & 95.9$_{0.1}$ & 95.8$_{0.2}$ \\
    \textbf{MRPC}\scriptsize{(F1)} & 61.9  & 92.2$_{0.4}$ & \textbf{92.7}$_{0.6}$ & 92.7$_{0.2}$ & 84.2$_{0.5}$ & 91.8$_{0.2}$ & 92.2$_{0.4}$ & 92.0$_{0.4}$ \\
    \textbf{STS-B}\scriptsize{(Pear.)} & -3.3  & 91.3$_{0.5}$ & 90.9$_{0.1}$ & 90.7$_{0.2}$ & 79.6$_{1.3}$ & \textbf{91.9}$_{0.3}$ & 90.8$_{0.4}$ & 90.1$_{0.6}$ \\
    \textbf{QQP}\scriptsize{(F1)} & 49.7  & 83.6$_{0.1}$ & 83.6$_{0.0}$ & \textbf{84.6}$_{0.1}$ & 77.0$_{0.7}$ & 84.3$_{0.0}$ & 83.7$_{0.0}$ & 84.4$_{0.2}$ \\
    \textbf{MNLI}\scriptsize{(acc)} & 50.9  & 88.6$_{0.1}$ & 87.7$_{0.1}$ & 88.7$_{0.1}$ & 80.2$_{0.2}$ & 88.7$_{0.1}$ & 88.0$_{0.1}$ & \textbf{88.9}$_{0.1}$ \\
    \textbf{QNLI}\scriptsize{(acc)} & 50.8  & 93.6$_{0.1}$ & 93.1$_{0.2}$ & \textbf{93.8}$_{0.1}$ & 86.6$_{0.4}$ & 93.8$_{0.1}$ & 93.0$_{0.1}$ & 93.8$_{0.1}$ \\
    \textbf{RTE}\scriptsize{(acc)} & 51.3  & \textbf{86.9}$_{0.2}$ & 86.2$_{1.0}$ & 86.0$_{0.7}$ & 78.3$_{0.3}$ & 84.6$_{0.5}$ & 86.4$_{1.5}$ & 84.7$_{0.9}$ \\
    \textbf{Average} & 43.4  & \textbf{87.4}$_{0.3}$ & 86.8$_{0.3}$ & 87.3$_{0.3}$ & 77.4$_{3.0}$ & 86.9$_{0.3}$ & 86.9$_{0.4}$ & 86.8$_{0.4}$ \\
    \midrule
    \multicolumn{9}{l}{$\text{RoBERTa}_{\texttt{LARGE}}$ \textit{, $16$-shot, without manual templates}} \\ \midrule
    \textbf{CoLA}\scriptsize{(Matt.)} & 4.6  & $19.6_{9.6}$ & $15.1_{17.0}$ & $17.7_{11.4}$ &
$3.5_{0.6}$ & $21.4_{11.5}$ & $20.8_{19.6}$ & \textbf{21.5}$_{13.4}$ \\
    \textbf{SST-2}\scriptsize{(acc)} & 50.9 & $92.7_{0.4}$ & $92.7_{0.6}$ &
	\textbf{93.1}$_{0.6}$ & $74.9_{0.6}$& $91.7_{0.8}$ & $92.2_{0.5}$ &
	$91.6_{0.7}$ \\
    \textbf{MRPC}\scriptsize{(F1)} & 1.4  & $78.2_{4.4}$ & $69.8_{1.6}$ & \textbf{81.2}$_{0.0}$ &
	$6.2_{4.1}$ & $74.6_{7.1}$ & $69.3_{6.5}$ & $77.4_{5.4}$ \\
    \textbf{STS-B}\scriptsize{(Pear.)} & -0.6 & $66.5_{2.5}$ & $67.5_{8.0}$ &
	\textbf{71.0}$_{2.5}$ & $10.7_{3.5}$ & $63.3_{1.6}$ & $64.7_{5.6}$ &
	$69.6_{8.6}$ \\
    \textbf{QQP}\scriptsize{(F1)} & 6.4  & $55.9_{5.8}$ & $55.1_{6.8}$ & $54.6_{4.2}$ &
	$52.4_{1.4}$ & $58.3_{7.2}$ & $55.1_{4.8}$ & \textbf{58.5}$_{6.1}$ \\
    \textbf{MNLI}\scriptsize{(acc)} & 34.2 & $58.1_{4.5}$ & \textbf{64.6}$_{3.4}$ & $62.7_{4.1}$
	& $35.3_{0.6}$ & $61.4_{3.9}$ & $61.4_{5.1}$ & $61.0_{3.8}$ \\
    \textbf{QNLI}\scriptsize{(acc)} & 50.6 & $60.2_{3.0}$ & \textbf{69.7}$_{1.9}$ & $59.8_{1.7}$
	& $52.8_{1.0}$ & $60.2_{4.9}$ & $60.9_{4.0}$ & $61.6_{7.0}$ \\
    \textbf{RTE}\scriptsize{(acc)} & 47.7 & $55.0_{1.6}$ & $54.5_{0.8}$ & $54.9_{2.9}$ &
	$50.1_{0.7}$ & $58.2_{2.5}$ & $54.6_{2.4}$ & \textbf{58.7}$_{3.4}$ \\
	\textbf{Average} & 24.4 & $60.8_{4.0}$ & $61.1_{5.0}$	& $61.9_{3.4}$ & $35.7_{1.6}$ & $61.2_{4.9}$ & $59.9_{6.1}$ & \textbf{62.5}$_{6.0}$ \\
	\midrule
    \multicolumn{9}{l}{$\text{RoBERTa}_{\texttt{LARGE}}$ \textit{, $16$-shot, with manual templates}} \\ \midrule
    \textbf{CoLA}\scriptsize{(Matt.)} & 2.2  & \textbf{10.5}$_{15.0}$ & $4.6_{5.0}$ & $9.2_{10.2}$ & $1.4_{1.7}$ & $10.2_{4.2}$ & $5.9_{2.5}$ & $5.9_{5.5}$ \\
    \textbf{SST-2}\scriptsize{(acc)} & 83.6  & \textbf{93.1}$_{0.3}$ & $92.9_{0.1}$ & $92.1_{0.1}$ & $90.9_{0.6}$ & $91.9_{0.4}$ & $92.0_{0.4}$ & $92.2_{0.6}$ \\
    \textbf{MRPC}\scriptsize{(F1)} & 61.9 & $77.2_{1.4}$ & $74.5_{4.9}$ & \textbf{81.2}$_{0.0}$ & $72.1_{4.4}$ & $76.8_{1.3}$ & $76.1_{2.4}$ & \textbf{81.2}$_{0.0}$ \\
    \textbf{STS-B}\scriptsize{(Pear.)} & -3.3 & $65.8_{4.7}$ & $69.3_{6.0}$ & $71.0_{4.1}$ & $12.0_{8.0}$ & $61.7_{5.7}$ & \textbf{71.3}$_{6.4}$ & $67.1_{2.8}$ \\
    \textbf{QQP}\scriptsize{(F1)} & 49.7 & $66.6_{0.5}$ & $67.8_{0.5}$ & $66.3_{4.1}$ & $53.4_{1.0}$ & $66.9_{1.9}$ & \textbf{68.6}$_{1.2}$ & $67.1_{2.9}$ \\
    \textbf{MNLI}\scriptsize{(acc)} & 50.9 & $68.0_{1.4}$ & \textbf{69.4}$_{3.3}$ & $68.9_{0.4}$ & $53.2_{2.5}$ & $67.1_{1.8}$ & $67.1_{2.0}$ & $68.1_{0.3}$ \\
    \textbf{QNLI}\scriptsize{(acc)} & 50.8 & $69.5_{1.1}$ & $70.2_{3.4}$ & $68.1_{2.4}$ &
$59.4_{0.5}$ & $69.9_{2.5}$ & \textbf{72.5}$_{3.9}$ & $70.4_{2.3}$ \\
    \textbf{RTE}\scriptsize{(acc)} & 51.3 & $70.6_{3.6}$ & $67.3_{5.1}$ & \textbf{73.0}$_{2.0}$
	& $56.3_{4.6}$ & $70.4_{2.3}$ & $69.2_{3.5}$ & $72.4_{2.8}$ \\
	\textbf{Average} & 43.4 & $65.2_{3.5}$ & $64.5_{3.5}$ & \textbf{66.2}$_{2.9}$ & $49.8_{2.9}$ & $64.4_{2.5}$ & $65.3_{2.8}$ & $65.6_{2.2}$ \\
    \bottomrule
    \end{tabu}%
  \label{tab:combine_roberta}%
\end{table}%

\begin{table}[thbp]
  \centering
  \caption{Performance of $\text{T5}_{\texttt{BASE}}$ on GLUE datasets. We report the average result of multiple
random seeds on the validation set.}
    \begin{tabu}{l|r r r r r r r r}
    \toprule
    \textbf{Prompt} & \multicolumn{1}{c|}{\color{BrickRed}{\XSolidBrush}} & \multicolumn{1}{c|}{\color{BrickRed}{\XSolidBrush}} & \multicolumn{1}{c|}{\color{BrickRed}{\XSolidBrush}} & \multicolumn{1}{c|}{\color{BrickRed}{\XSolidBrush}} & \multicolumn{1}{c|}{\color{ForestGreen}{\Checkmark}} & \multicolumn{1}{c|}{\color{ForestGreen}{\Checkmark}} & \multicolumn{1}{c|}{\color{ForestGreen}{\Checkmark}} & \multicolumn{1}{c}{\color{ForestGreen}{\Checkmark}} \\
    
    \textbf{BitFit} & \multicolumn{1}{c|}{\color{BrickRed}{\XSolidBrush}} & \multicolumn{1}{c|}{\color{BrickRed}{\XSolidBrush}} &
    \multicolumn{1}{c|}{\color{ForestGreen}{\Checkmark}} & \multicolumn{1}{c|}{\color{ForestGreen}{\Checkmark}} &
    \multicolumn{1}{c|}{\color{BrickRed}{\XSolidBrush}} & \multicolumn{1}{c|}{\color{BrickRed}{\XSolidBrush}} &  \multicolumn{1}{c|}{\color{ForestGreen}{\Checkmark}} & \multicolumn{1}{c}{\color{ForestGreen}{\Checkmark}} \\
    
    \textbf{Adapter} & \multicolumn{1}{c|}{\color{BrickRed}{\XSolidBrush}} & \multicolumn{1}{c|}{\color{ForestGreen}{\Checkmark}} & \multicolumn{1}{c|}{\color{BrickRed}{\XSolidBrush}} & \multicolumn{1}{c|}{\color{ForestGreen}{\Checkmark}} & \multicolumn{1}{c|}{\color{BrickRed}{\XSolidBrush}} & \multicolumn{1}{c|}{\color{ForestGreen}{\Checkmark}} & \multicolumn{1}{c|}{\color{BrickRed}{\XSolidBrush}} & \multicolumn{1}{c}{\color{ForestGreen}{\Checkmark}} \\
    \midrule
    \textit{Tunable parameters} & 0\% & 0.89\% & 0.09\% & 0.98\% & 0.003\% & 0.893\% & 0.093\% & 0.983\% \\
    \midrule
            \multicolumn{9}{l}{$\text{T5}_{\texttt{BASE}}$ \textit{, full-data, without manual templates}} \\ 
    \midrule
        \textbf{CoLA}\scriptsize{(Matt.)} & \multicolumn{1}{c}{-}  & \textbf{59.2}$_{0.2}$ & $58.7_{1.7}$ & $58.4_{0.8}$ & $34.0_{18.6}$ & $51.2_{4.6}$ & $36.9_{21.7}$ & $57.8_{0.3}$ \\
    \textbf{SST-2}\scriptsize{(acc)} & \multicolumn{1}{c}{-}   & $94.6_{0.1}$ & $94.4_{0.1}$ & $95.0_{0.2}$ & $94.0_{0.3}$ & $94.1_{1.4}$ & $95.0_{0.1}$ & \textbf{95.1}$_{0.2}$ \\
    \textbf{MRPC}\scriptsize{(F1)} & \multicolumn{1}{c}{-}  & $89.1_{0.6}$ & $90.1_{0.4}$ & \textbf{90.8}$_{0.3}$ & $84.8_{1.2}$ & $89.2_{0.7}$ & $88.5_{0.7}$ & $88.8_{0.6}$ \\
    \textbf{STS-B}\scriptsize{(Pear.)} & \multicolumn{1}{c}{-}  & $86.7_{0.3}$ & $86.6_{0.1}$ & \textbf{86.9}$_{0.3}$ & $83.1_{1.8}$ & $86.1_{0.4}$ & $85.8_{1.4}$ & $85.8_{0.4}$ \\
    \textbf{QQP}\scriptsize{(F1)} & \multicolumn{1}{c}{-}  & $86.7_{0.2}$ & \textbf{88.3}$_{0.1}$ & $87.7_{0.3}$ & $83.4_{1.0}$ & $86.9_{0.4}$ & $88.0_{0.4}$ & $88.1_{0.2}$ \\
    \textbf{MNLI}\scriptsize{(acc)} & \multicolumn{1}{c}{-}  & $84.5_{0.4}$ & $87.1_{0.3}$ & $87.1_{0.4}$ & $81.6_{0.2}$ & $86.0_{0.2}$ & \textbf{87.3}$_{0.2}$ & $87.1_{0.4}$ \\
    \textbf{QNLI}\scriptsize{(acc)} & \multicolumn{1}{c}{-}  & $89.8_{0.1}$ & $91.6_{0.1}$ & $91.3_{0.1}$ &
$87.8_{0.3}$ & $89.1_{0.3}$ & \textbf{91.8}$_{0.3}$ & $91.7_{0.2}$ \\
    \textbf{RTE}\scriptsize{(acc)} & \multicolumn{1}{c}{-}  & $75.3_{1.0}$ & $77.5_{1.3}$ & \textbf{80.0}$_{0.8}$
	& $64.3_{0.9}$ & $71.5_{4.7}$ & $72.9_{6.6}$ & $71.5_{2.0}$ \\
	\textbf{Average} & \multicolumn{1}{c}{-}  & $83.2_{0.3}$ & $84.3_{0.5}$ & \textbf{84.7}$_{0.4}$ & $76.6_{3.1}$  & $81.8_{1.6}$ & $80.8_{3.9}$ & $83.2_{0.5}$ \\ 
	\midrule
    \multicolumn{9}{l}{$\text{T5}_{\texttt{BASE}}$ \textit{, full-data, with manual templates}} \\ \midrule
    \textbf{CoLA}\scriptsize{(Matt.)} & \multicolumn{1}{c}{-}      & $57.4_{1.8}$ & \textbf{59.5}$_{1.2}$ & $59.1_{0.4}$ & $36.3_{18.2}$ & $39.1_{22.8}$ & $58.1_{0.6}$ & $48.2_{7.4}$ \\
    \textbf{SST-2}\scriptsize{(acc)} & \multicolumn{1}{c}{-}      & $95.0_{0.1}$ & $94.8_{0.3}$ & $95.0_{0.2}$ & $93.9_{0.1}$ & $95.2_{0.1}$ & $95.0_{0.2}$ & \textbf{95.3}$_{0.3}$ \\
    \textbf{MRPC}\scriptsize{(F1)} & \multicolumn{1}{c}{-}      & $90.9_{0.3}$ & $90.7_{0.6}$ & \textbf{91.4}$_{0.4}$ & $81.3_{3.5}$ & $87.8_{0.5}$ & $89.4_{0.3}$ & $89.2_{0.4}$ \\
    \textbf{STS-B}\scriptsize{(Pear.)} & \multicolumn{1}{c}{-}      & $87.1_{0.2}$ & $87.4_{0.4}$ & \textbf{87.7}$_{0.2}$ & $83.4_{1.0}$ & $86.6_{0.1}$ & $84.7_{3.4}$ & $86.8_{0.4}$ \\
    \textbf{QQP}\scriptsize{(F1)} & \multicolumn{1}{c}{-}      & $87.4_{0.1}$ & \textbf{88.3}$_{0.1}$ & $88.2_{0.1}$ & $83.7_{1.1}$ & $87.4_{0.1}$ & $88.3_{0.1}$ & $88.3_{0.2}$ \\
    \textbf{MNLI}\scriptsize{(acc)} & \multicolumn{1}{c}{-}      & $86.1_{0.2}$ & $87.1_{0.2}$ & $86.7_{0.5}$ & $83.1_{0.5}$ & $86.3_{0.4}$ & \textbf{87.2}$_{0.4}$ & $86.9_{0.3}$ \\
    \textbf{QNLI}\scriptsize{(acc)} & \multicolumn{1}{c}{-}      & $91.9_{0.3}$ & $92.8_{0.4}$ & $92.6_{0.2}$ & $89.3_{0.7}$ & $92.2_{0.1}$ & \textbf{92.9}$_{0.1}$ & $92.7_{0.3}$ \\
    \textbf{RTE}\scriptsize{(acc)} & \multicolumn{1}{c}{-}      & $81.9_{0.7}$ & \textbf{84.0}$_{0.8}$ & $83.2_{1.5}$ & $66.8_{2.2}$ & $80.0_{0.2}$ & $81.8_{1.3}$ & $80.1_{0.6}$ \\
    \textbf{Average} & \multicolumn{1}{c}{-}  & $84.7_{0.5}$  & \textbf{85.6}$_{0.5}$ & $85.5_{0.4}$ & $77.2_{3.4}$ & $81.8_{3.0}$ & $84.7_{0.8}$ & $83.4_{1.2}$ \\ 
    \bottomrule
    \end{tabu}
  \label{tab:combine_t5}%
\end{table}%

\begin{table}[thbp]
  \centering
  \caption{The experiments of generalization gap for $\text{RoBERTa}_{\texttt{LARGE}}$ on GLUE datasets. We report the average result (train performance - dev performance) of multiple
random seeds.}
    \begin{tabu}{l|r r r r r r r r}
    \toprule
    \textbf{Prompt} &  \multicolumn{1}{c|}{\color{BrickRed}{\XSolidBrush}} & \multicolumn{1}{c|}{\color{BrickRed}{\XSolidBrush}} & \multicolumn{1}{c|}{\color{BrickRed}{\XSolidBrush}} & \multicolumn{1}{c|}{\color{ForestGreen}{\Checkmark}} & \multicolumn{1}{c|}{\color{ForestGreen}{\Checkmark}} & \multicolumn{1}{c|}{\color{ForestGreen}{\Checkmark}} & \multicolumn{1}{c|}{\color{ForestGreen}{\Checkmark}}  & \multicolumn{1}{c}{\multirow{3}{*}{\textbf{FT}}} \\
    
    \textbf{BitFit} &  \multicolumn{1}{c|}{\color{BrickRed}{\XSolidBrush}} &
    \multicolumn{1}{c|}{\color{ForestGreen}{\Checkmark}} & \multicolumn{1}{c|}{\color{ForestGreen}{\Checkmark}} &
    \multicolumn{1}{c|}{\color{BrickRed}{\XSolidBrush}} & \multicolumn{1}{c|}{\color{BrickRed}{\XSolidBrush}} &  \multicolumn{1}{c|}{\color{ForestGreen}{\Checkmark}} & \multicolumn{1}{c|}{\color{ForestGreen}{\Checkmark}} & \\
    \textbf{Adapter} &  \multicolumn{1}{c|}{\color{ForestGreen}{\Checkmark}} & \multicolumn{1}{c|}{\color{BrickRed}{\XSolidBrush}} & \multicolumn{1}{c|}{\color{ForestGreen}{\Checkmark}} & \multicolumn{1}{c|}{\color{BrickRed}{\XSolidBrush}} & \multicolumn{1}{c|}{\color{ForestGreen}{\Checkmark}} & \multicolumn{1}{c|}{\color{BrickRed}{\XSolidBrush}} & \multicolumn{1}{c|}{\color{ForestGreen}{\Checkmark}} & \\ \midrule
    \textit{Tunable parameters} & 1.75\% & 0.09\% & 1.84\% & 0.003\% & 1.76\% & 0.09\% & 1.85\% & 100\% \\
    \midrule 
            \multicolumn{9}{l}{$\text{RoBERTa}_{\texttt{BASE}}$ \textit{, full-data, without manual templates}} \\ 
    \midrule
    \textbf{CoLA} & 25.4$_{1.5}$ & 13.0$_{2.8}$ & 28.4$_{2.4}$ & 12.1$_{3.8}$ & \textbf{29.5}$_{6.8}$ & 16.2$_{7.6}$ & 18.0$_{1.8}$ & 28.2$_{2.4}$ \\
    \textbf{SST-2} & 3.0$_{1.3}$ & 1.7$_{0.5}$ & 0.9$_{0.3}$ & 1.1$_{0.5}$ & \textbf{3.6}$_{0.5}$ & 1.9$_{0.6}$ & 3.5$_{1.1}$ & 3.3$_{0.9}$ \\
    \textbf{MRPC} & 7.1$_{0.3}$ & 5.7$_{2.2}$ & 7.0$_{0.4}$ & 1.0$_{1.1}$ & \textbf{8.0}$_{0.5}$ & 4.5$_{0.5}$ & 7.1$_{0.2}$ & 6.3$_{0.7}$ \\
    \textbf{STS-B} & 5.1$_{0.0}$ & 4.9$_{0.6}$ & 7.0$_{0.8}$ & 6.7$_{1.6}$ & 6.5$_{0.3}$ & 5.6$_{0.6}$ & 6.5$_{0.4}$ & \textbf{7.5}$_{0.2}$ \\
    \textbf{QQP} & 0.6$_{0.1}$ & 0.7$_{0.1}$ & 0.8$_{0.0}$ & 0.1$_{0.0}$ & 0.8$_{0.0}$ & 0.7$_{0.1}$ & 0.8$_{0.1}$ & \textbf{1.9}$_{0.2}$ \\
    \textbf{MNLI} & \textbf{0.6}$_{0.1}$ & 0.5$_{0.1}$ & 0.6$_{0.2}$ & 0.6$_{0.4}$ & 0.5$_{0.1}$ & 0.5$_{0.2}$ & 0.5$_{0.1}$ & 0.6$_{0.0}$ \\
    \textbf{QNLI} & 0.9$_{0.1}$ & 0.7$_{0.1}$ & 0.5$_{0.2}$ & 1.6$_{0.1}$ & 0.5$_{0.2}$ & 0.8$_{0.3}$ & 0.5$_{0.2}$ & \textbf{1.6}$_{0.0}$ \\
    \textbf{RTE} & 13.1$_{0.6}$ & 13.2$_{0.7}$ & 14.9$_{0.3}$ & 9.8$_{1.6}$ & \textbf{15.9}$_{0.8}$ & 12.6$_{2.3}$ & 15.1$_{1.3}$ & 12.9$_{1.3}$ \\
    \textbf{Average} & 7.0$_{0.5}$ & 5.1$_{0.9}$ & 7.5$_{0.6}$ & 4.1$_{1.1}$ & \textbf{8.2}$_{1.2}$ & 5.3$_{1.5}$ & 6.5$_{0.7}$ & 7.8$_{0.7}$ \\
    \midrule
     \multicolumn{9}{l}{$\text{RoBERTa}_{\texttt{BASE}}$ \textit{, full-data, with manual templates}} \\ 
    \midrule
    \textbf{CoLA} & 20.9$_{5.0}$ & 25.4$_{4.4}$ & 24.3$_{7.5}$ & 11.4$_{1.2}$ & 29.1$_{3.2}$ & 29.6$_{6.4}$ & 24.6$_{10.3}$ & \textbf{30.4}$_{2.3}$ \\
    \textbf{SST-2} & 3.3$_{0.1}$ & 1.4$_{0.6}$ & 1.3$_{0.7}$ & 1.0$_{0.3}$ & 2.6$_{0.7}$ & 2.5$_{0.8}$ & 3.8$_{0.4}$ & \textbf{4.0}$_{0.1}$ \\
    \textbf{MRPC} & 6.2$_{2.5}$ & 6.5$_{0.6}$ & 6.4$_{0.3}$ & 3.8$_{2.5}$ & \textbf{8.2}$_{0.2}$ & 7.2$_{0.3}$ & 6.7$_{1.4}$ & 7.2$_{0.5}$ \\
    \textbf{STS-B} & 5.8$_{1.4}$ & 4.9$_{0.4}$ & 6.7$_{1.2}$ & \textbf{10.2}$_{0.6}$ & 6.9$_{0.5}$ & 5.5$_{0.7}$ & 6.1$_{1.5}$ & 7.5$_{0.2}$ \\
    \textbf{QQP} & 0.7$_{0.1}$ & 0.6$_{0.1}$ & 0.8$_{0.0}$ & 0.2$_{0.1}$ & 0.8$_{0.2}$ & 0.7$_{0.1}$ & 0.8$_{0.0}$ & \textbf{2.0}$_{0.1}$ \\
    \textbf{MNLI} & \textbf{0.8}$_{0.1}$ & 0.3$_{0.1}$ & 0.4$_{0.1}$ & 0.7$_{0.2}$ & 0.6$_{0.1}$ & 0.7$_{0.1}$ & 0.6$_{0.1}$ & 0.8$_{0.2}$ \\
    \textbf{QNLI} & 0.8$_{0.1}$ & 0.4$_{0.2}$ & 0.1$_{0.0}$ & 1.4$_{0.1}$ & 0.1$_{0.0}$ & 0.5$_{0.2}$ & 0.0$_{0.0}$ & \textbf{2.0}$_{0.1}$ \\
    \textbf{RTE} & 13.1$_{0.2}$ & 11.7$_{0.8}$ & 13.9$_{0.7}$ & 10.1$_{5.2}$ & 15.0$_{0.4}$ & 13.3$_{1.3}$ & \textbf{15.1}$_{0.9}$ & 12.7$_{1.1}$ \\
    \textbf{Average} & 6.4$_{1.2}$ & 6.4$_{0.9}$ & 6.7$_{1.3}$ & 4.8$_{1.3}$ & 7.9$_{0.7}$ & 7.5$_{1.2}$ & 7.2$_{1.8}$ & \textbf{8.3}$_{0.6}$ \\
    
    \bottomrule
    \end{tabu}
  \label{tab:gen_gap}%
\end{table}%

\begin{figure}[thbp]
    \centering
    \subfigure{
        \includegraphics[width=0.3\textwidth]{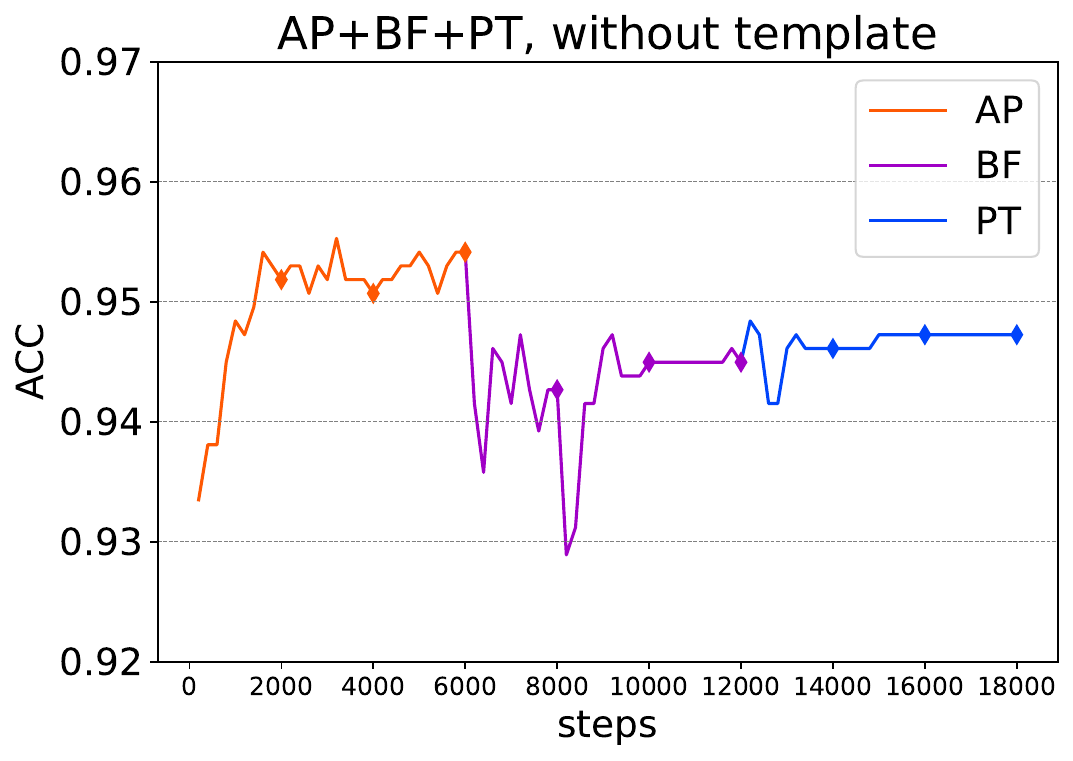}
    }
    \subfigure{
        \includegraphics[width=0.3\textwidth]{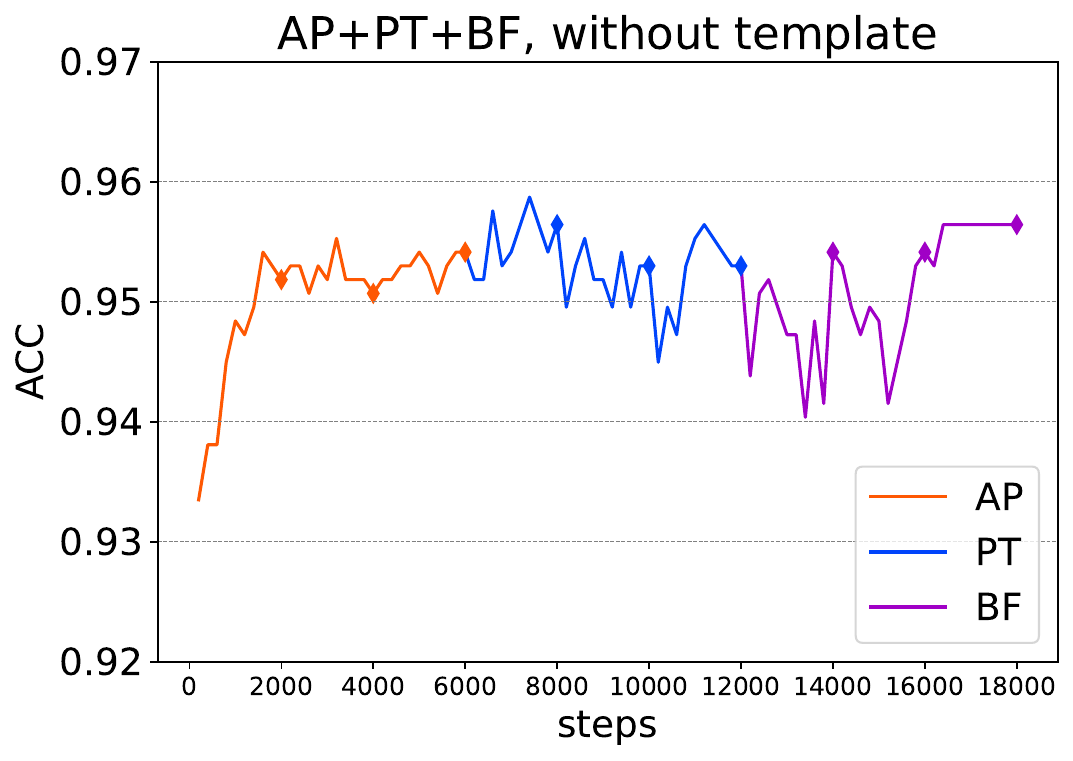}
    }
    \subfigure{
        \includegraphics[width=0.3\textwidth]{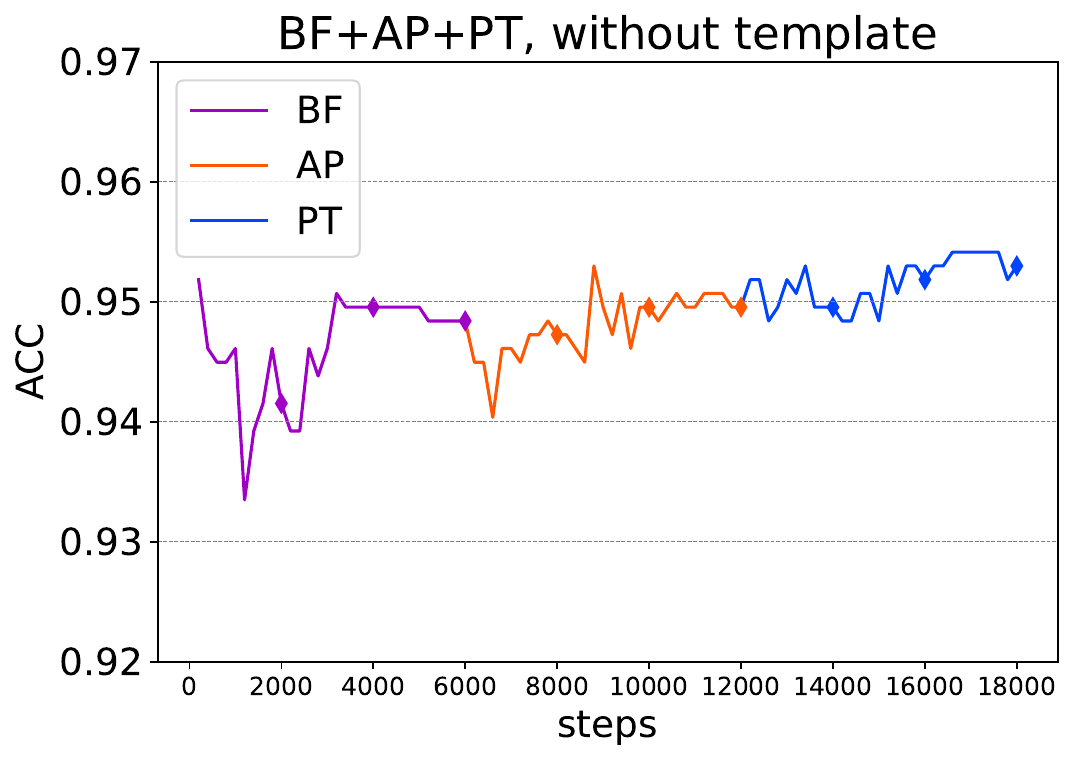}
    }
    \subfigure{
        \includegraphics[width=0.3\textwidth]{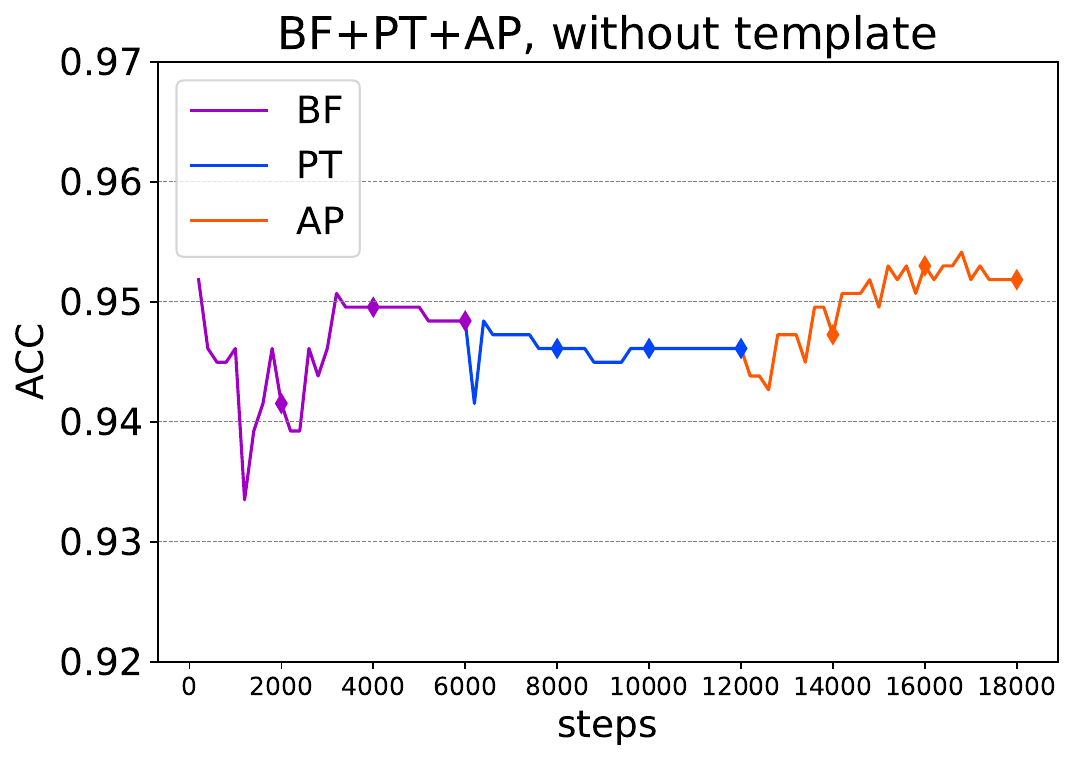}
    }
    \subfigure{
        \includegraphics[width=0.3\textwidth]{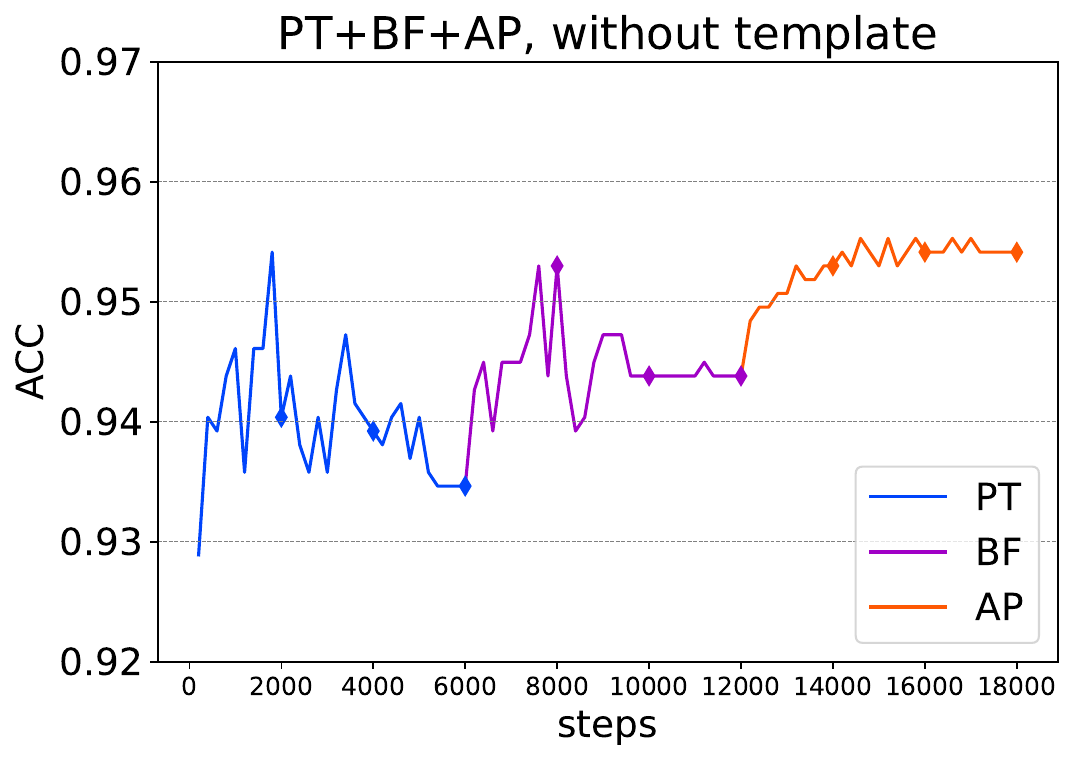}
    }
    \subfigure{
        \includegraphics[width=0.3\textwidth]{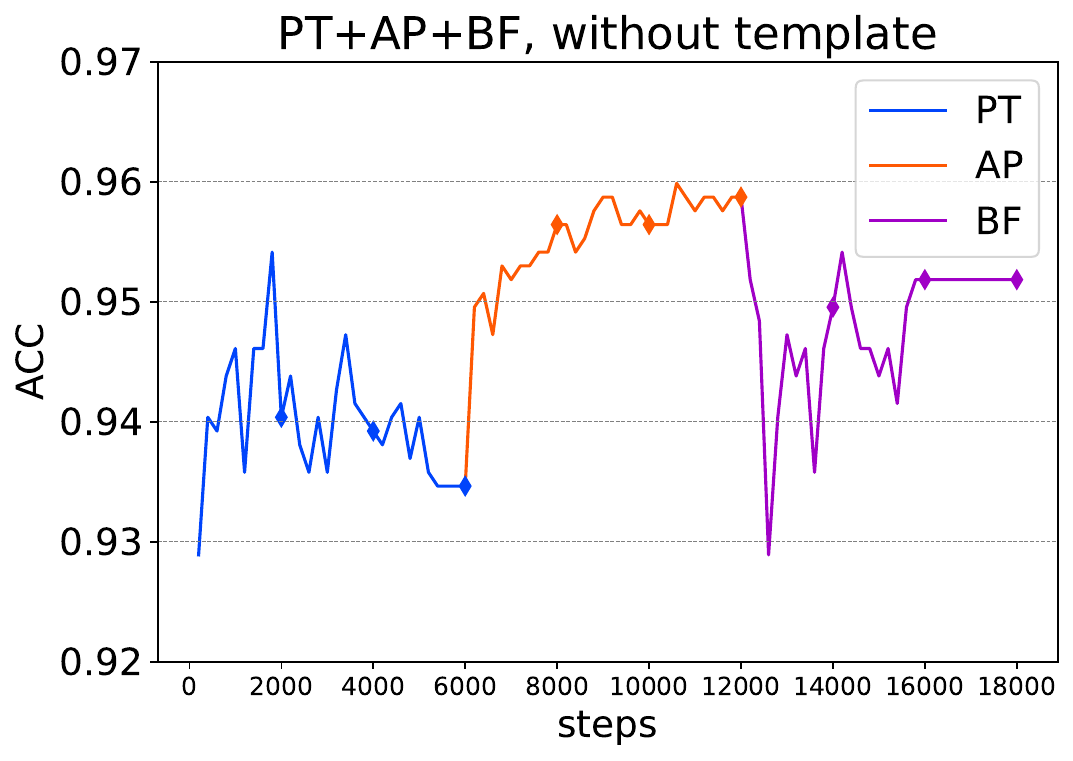}
    }
        \subfigure{
        \includegraphics[width=0.3\textwidth]{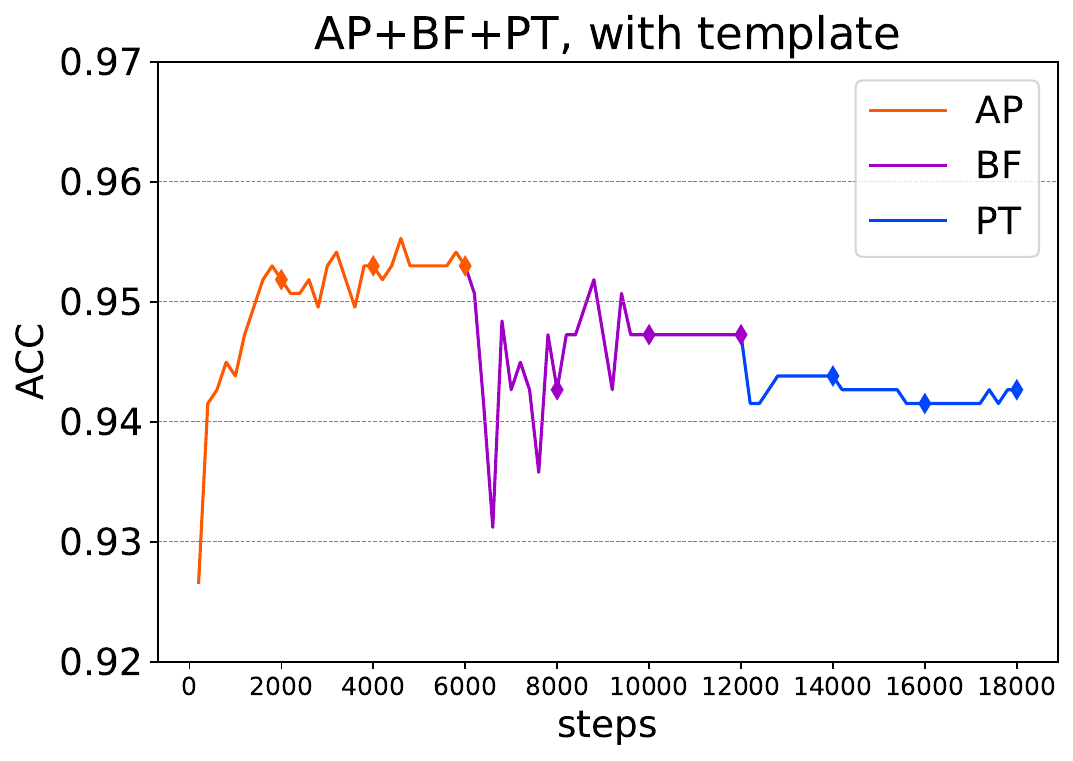}
    }
    \subfigure{
        \includegraphics[width=0.3\textwidth]{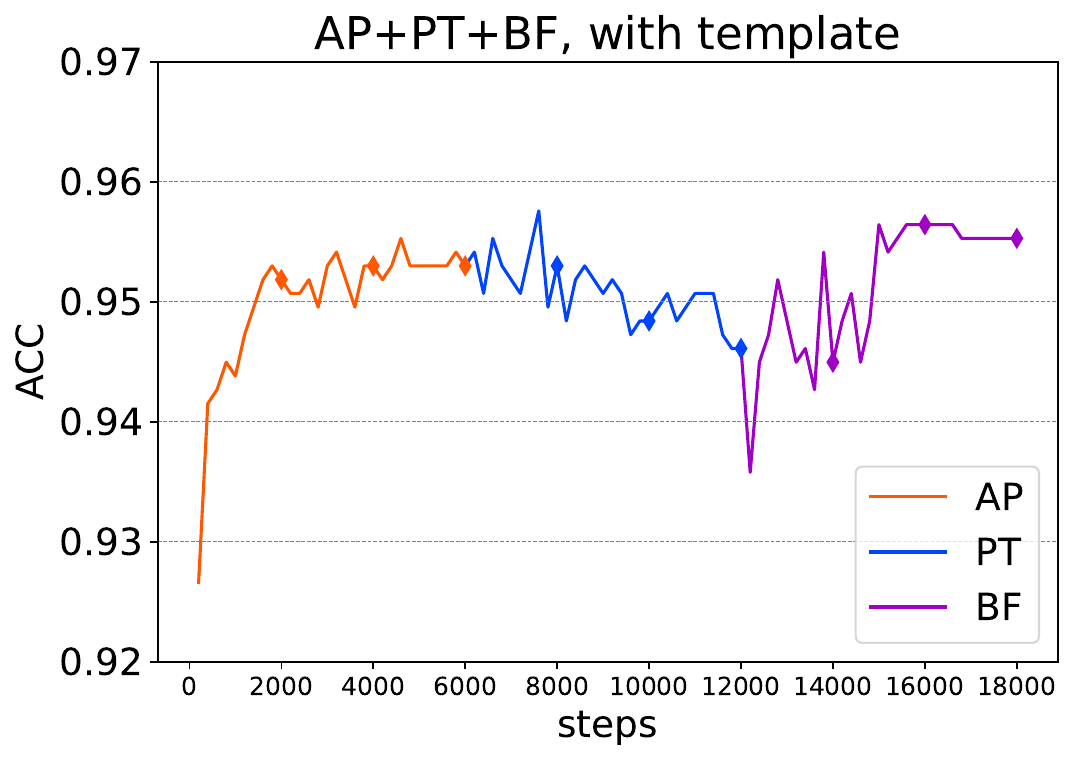}
    }
    \subfigure{
        \includegraphics[width=0.3\textwidth]{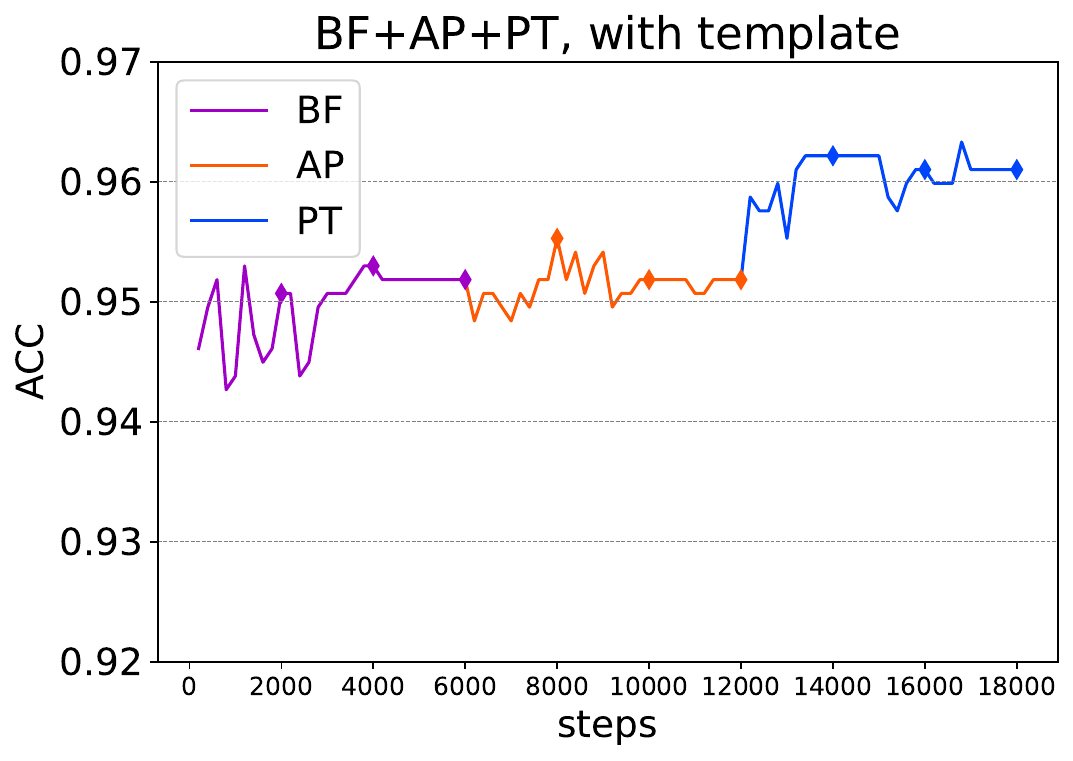}
    }
    
    \subfigure{
        \includegraphics[width=0.3\textwidth]{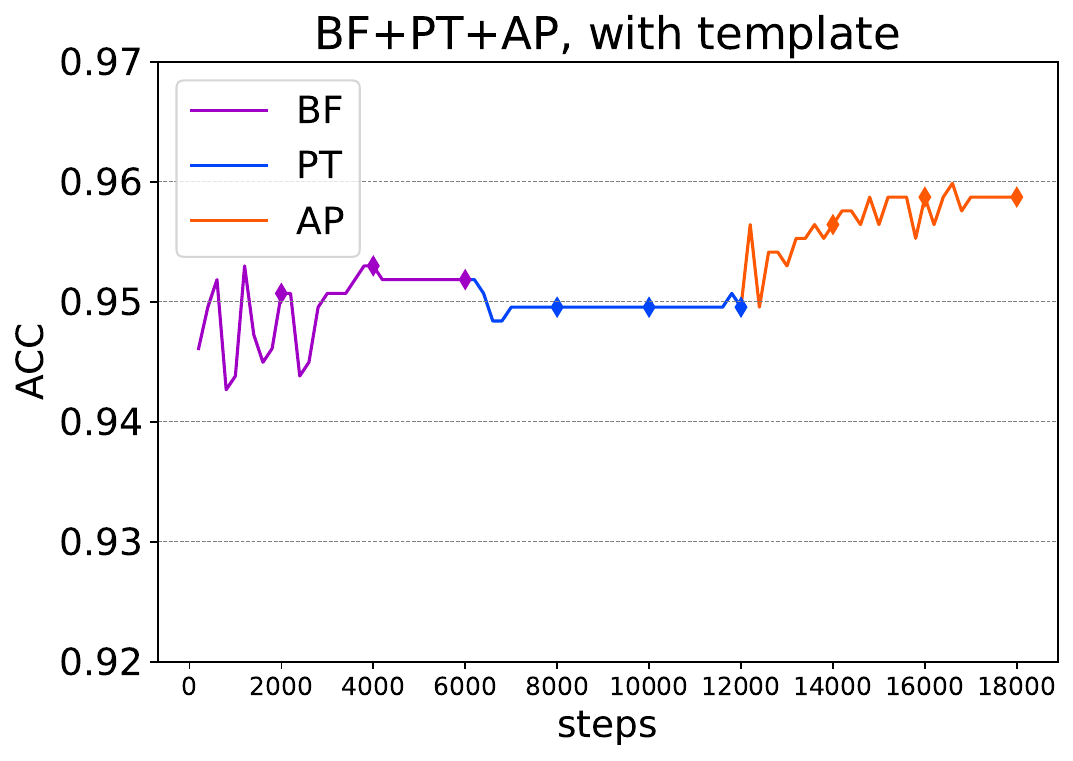}
    }
    \subfigure{
        \includegraphics[width=0.3\textwidth]{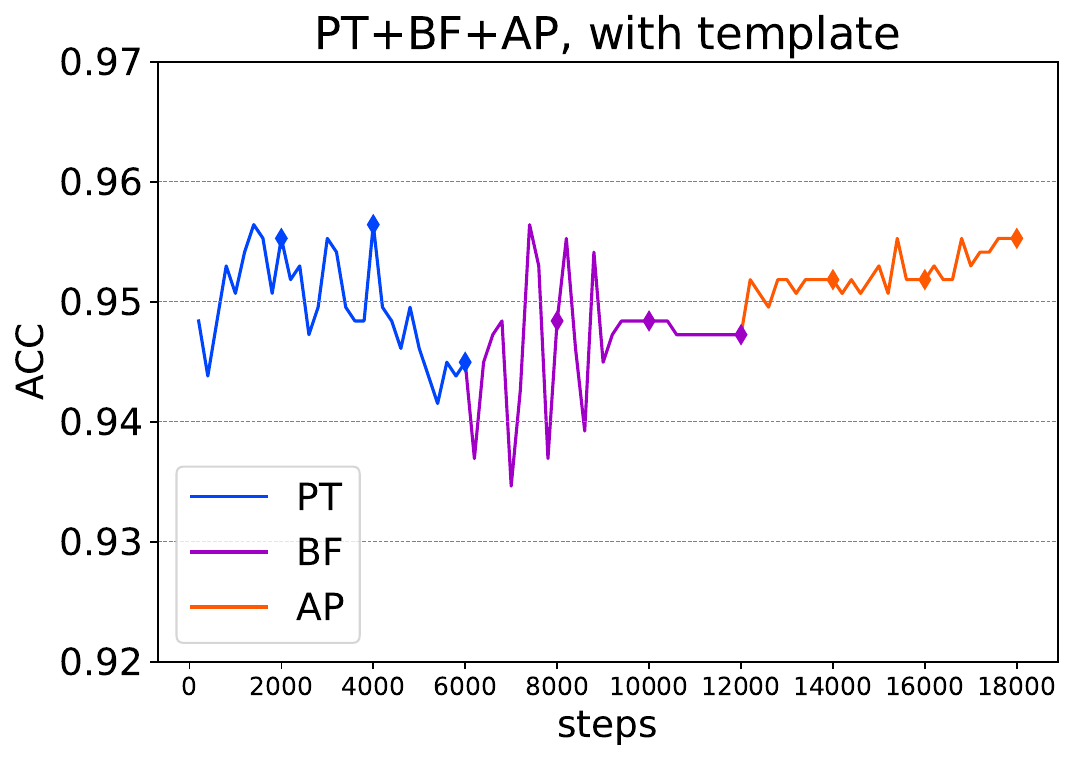}
    }
    \subfigure{
        \includegraphics[width=0.3\textwidth]{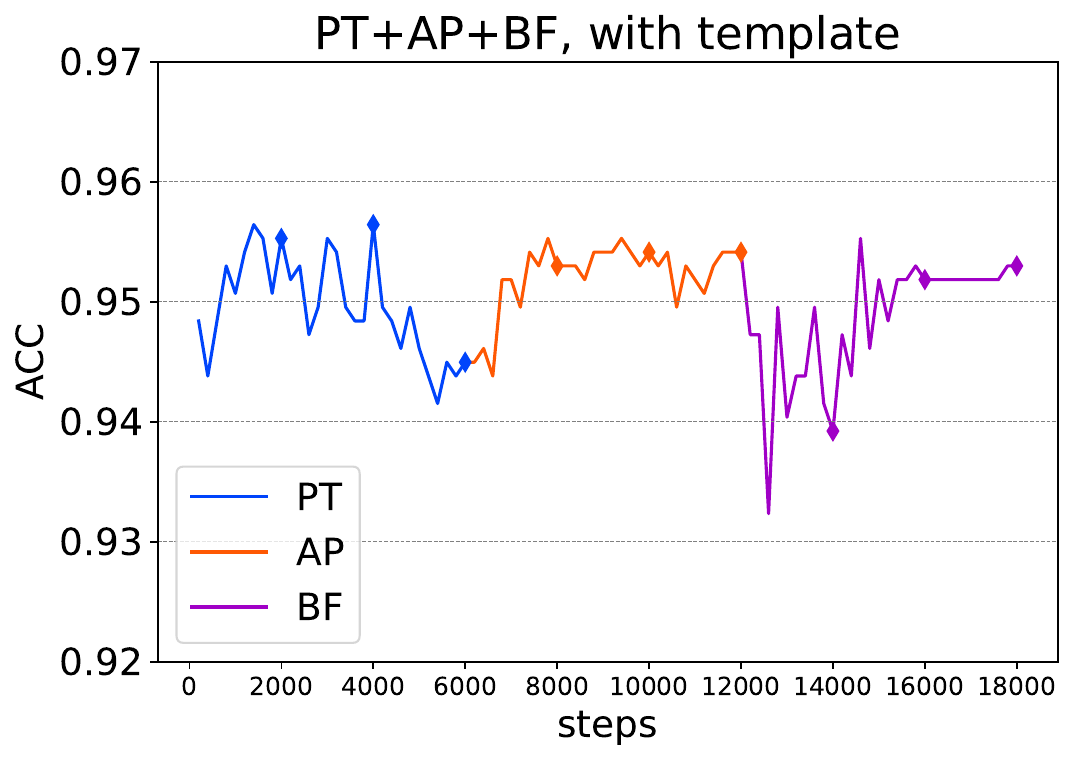}
    }
    \caption{The performance of $\text{RoBERTa}_{\texttt{LARGE}}$ when different delta tuning methods (adapter (\textbf{AP}), BitFit (\textbf{BF}) and prompt tuning (\textbf{PT})) are applied sequentially. The experiments are conducted on SST-2~\citep{socher-etal-2013-recursive}.}
    \label{fig:mulstage}
\end{figure}

\paragraph{Sequential Combination.}
In addition to the simultaneous combination, we further investigate the compatibility when the above three delta tuning methods (prompt tuning, BitFit, and adapter) are sequentially introduced. Specifically, we split the whole tuning process into $3$ stages. During each stage, we train an individual delta tuning method for $6,000$ steps; in the stages to follow, we freeze the tuned parameters in previous stages and only optimize the newly introduced delta parameters. We experiment $\text{RoBERTa}_{\texttt{LARGE}}$ on SST-2~\citep{socher-etal-2013-recursive} with  / without manual templates. The results are visualized in Figure~\ref{fig:mulstage}, from which we could derive that, under certain cases, the performance could be improved with the involvements of subsequent delta tuning methods. However, there does not exist an optimal sequential combination under different settings.

\paragraph{Generalization Gap.}
Additionally, we report the generalization gap (train performance - dev performance) for $\text{RoBERTa}_{\texttt{LARGE}}$ under the full-data setting, with the results shown in Table~\ref{tab:gen_gap}. It could be derived that, (1) the gap of a single delta tuning method is always smaller than fine-tuning, which means over-parameterization may help better memorize (overfit) training samples. Among all the delta tuning methods, prompt tuning tends to have the smallest generalization gap. Considering that each delta tuning method could already generalize well and achieve non-trivial performance on the dev set, hence overfitting the training set may not be the prerequisite for good generalization; (2) in general, combining delta tuning methods would enlarge the generalization gap, even to the extent that is comparable with fine-tuning, despite tuning far fewer parameters. This suggests that, for the investigated tasks, memorizing the training set may not require employing all of the parameters; in other words, a small model capacity during downstream adaptation may be enough for good memorization; (3) utilizing manual templates generally would not influence the generalization gap.

\paragraph{Conclusion.}
To sum up, the above experiments indicate that, different delta tuning methods have distinct functionalities for PLMs' optimization, thus combining them is generally conducive to the downstream performance. However, as shown in the above results, different PLMs may favor distinct delta tuning combinations, and the optimal combination of delta tuning methods may vary a lot under different settings. That being said, it would be interesting to explore the mechanisms behind the inductive biases brought by different delta tuning methods under different cases in the future. Besides, we also encourage future research explorations to systematically report the performance of their proposed delta methods on various PLM backbones under different settings thoroughly.

%0.48
\begin{figure*}[thbp]
\subcapraggedrighttrue
\subcaphangtrue
    \centering
    \subfigure[Adapter (MNLI).]{
	    \includegraphics[width=0.31\textwidth]{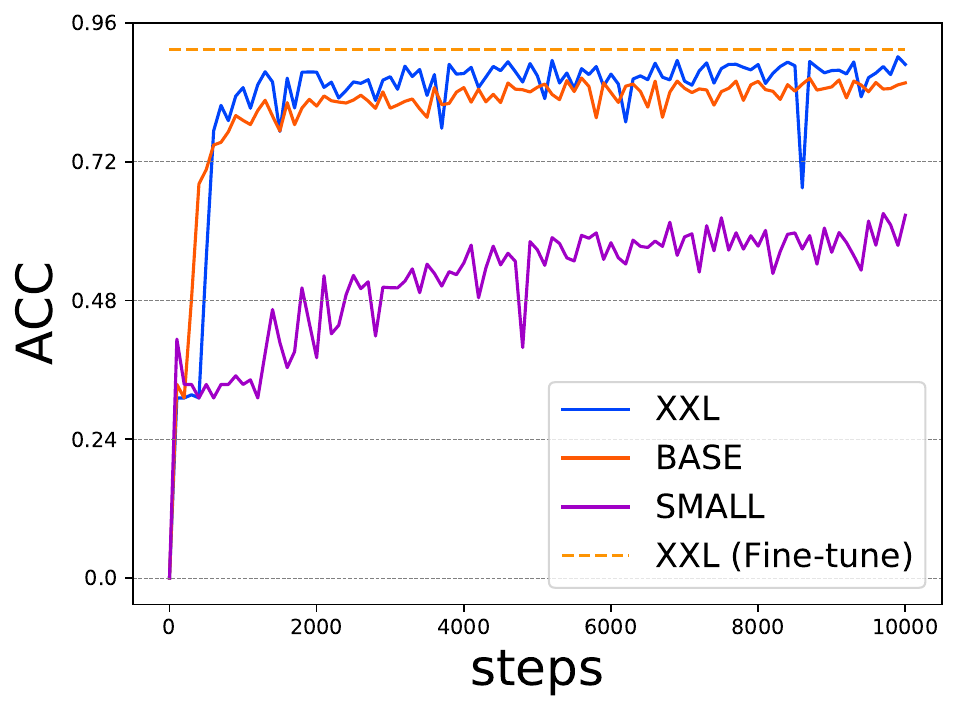}
    }
    \subfigure[Adapter (QNLI).]{
	    \includegraphics[width=0.31\textwidth]{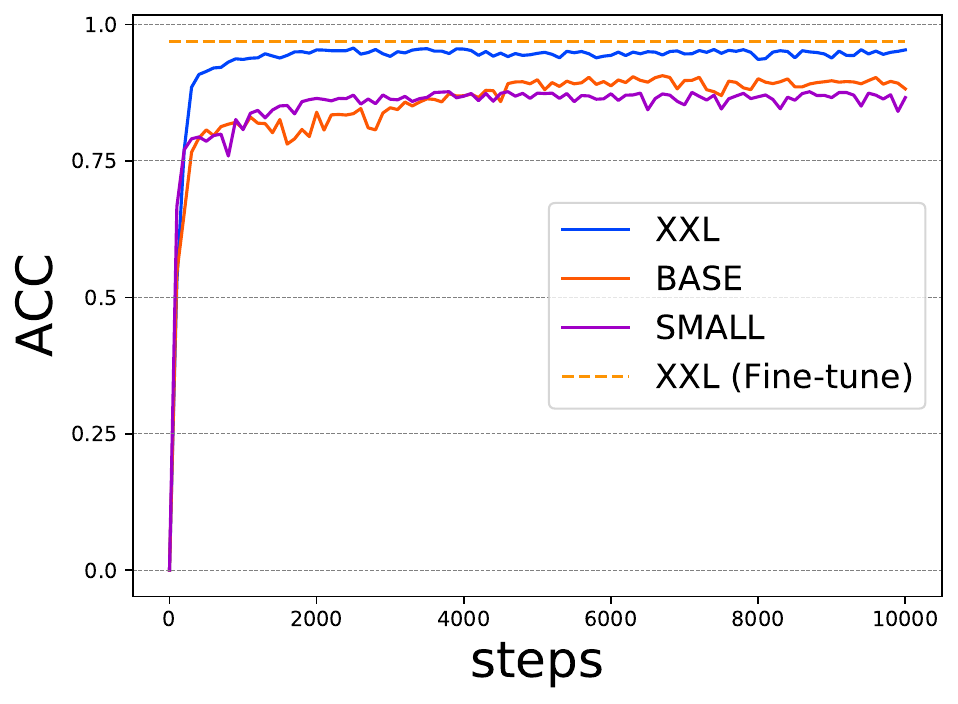}
    }
    \subfigure[Adapter (SST-2).]{
	    \includegraphics[width=0.31\textwidth]{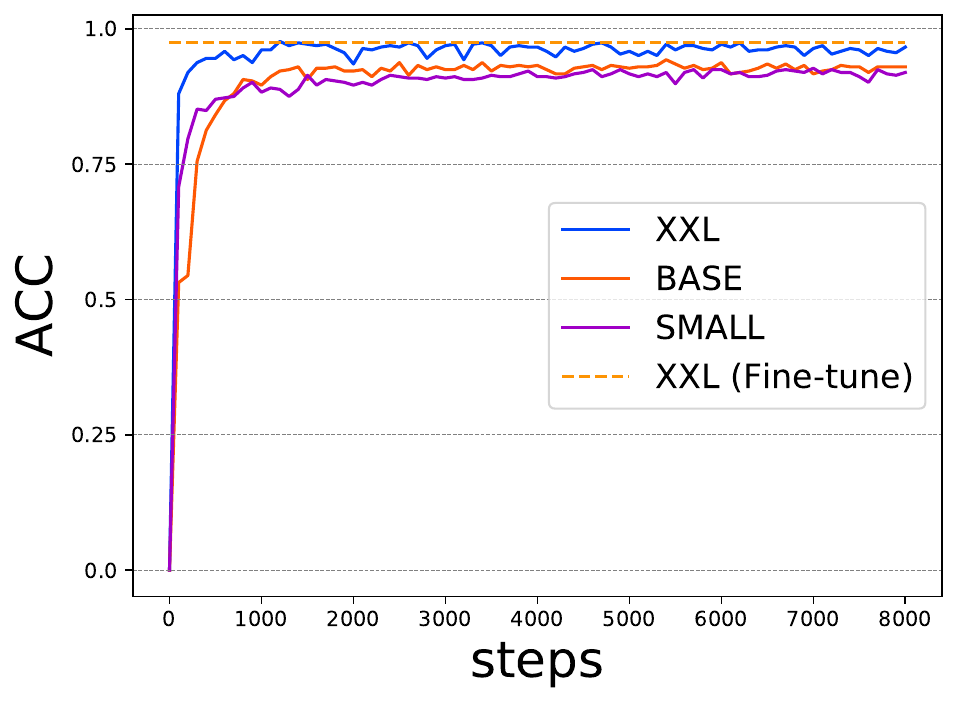}
    }

    \subfigure[LoRA (MNLI).]{
	    \includegraphics[width=0.31\textwidth]{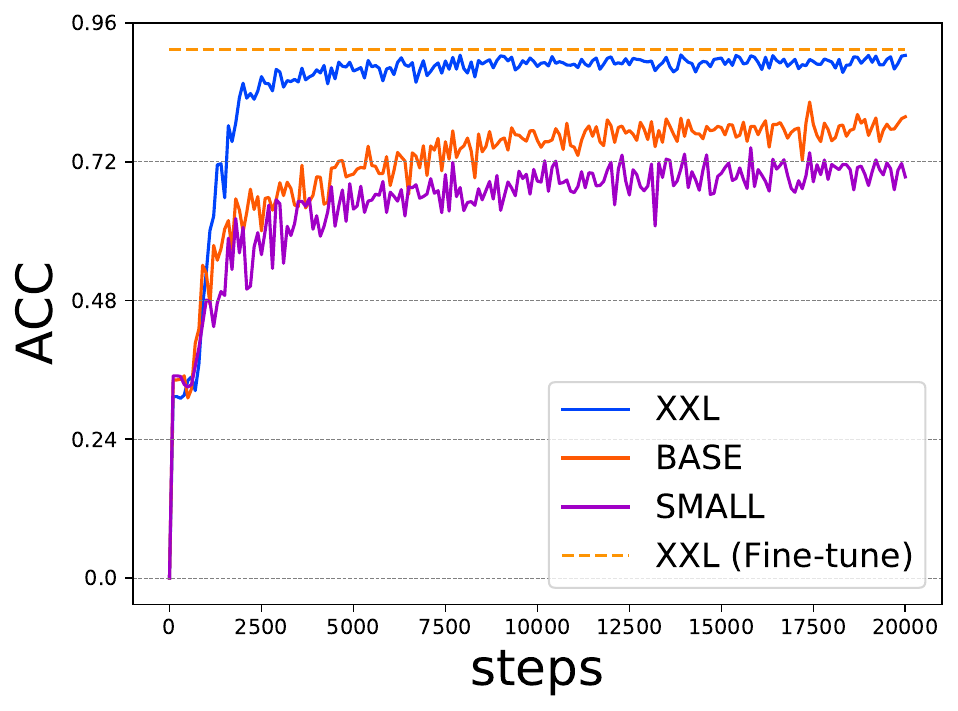}
    }
    \subfigure[LoRA (QNLI).]{
	    \includegraphics[width=0.31\textwidth]{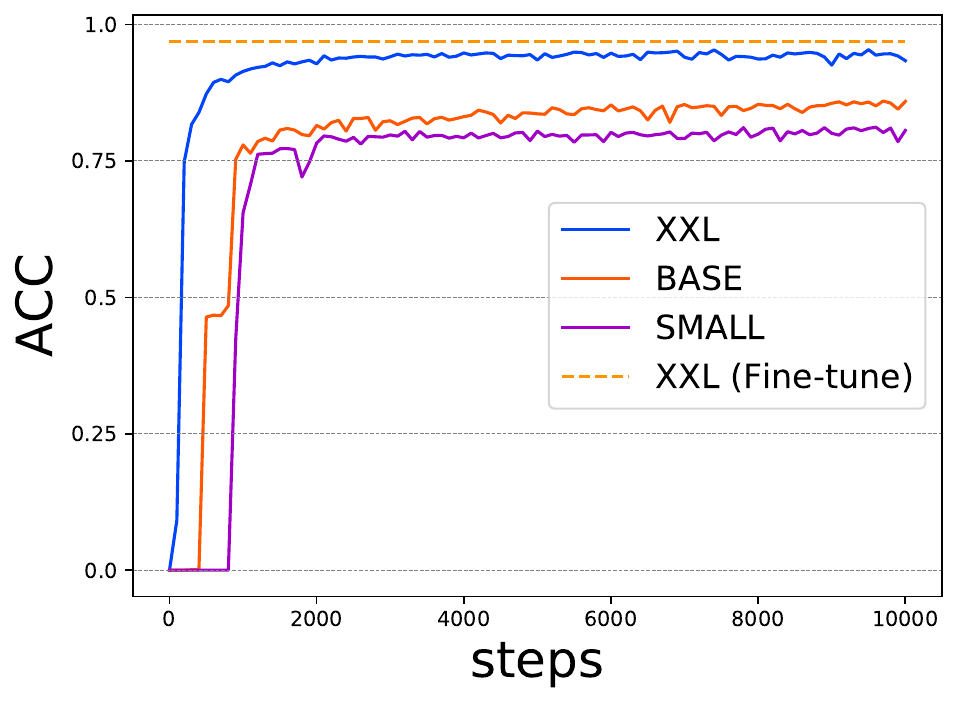}
    }
    \subfigure[LoRA (SST-2).]{
	    \includegraphics[width=0.31\textwidth]{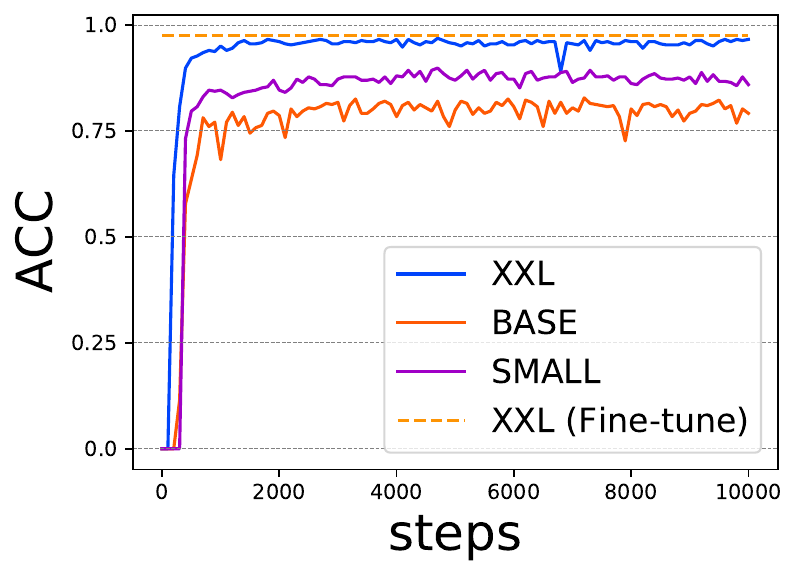}
    }

    \subfigure[Prefix-tuning (MNLI).]{
	    \includegraphics[width=0.31\textwidth]{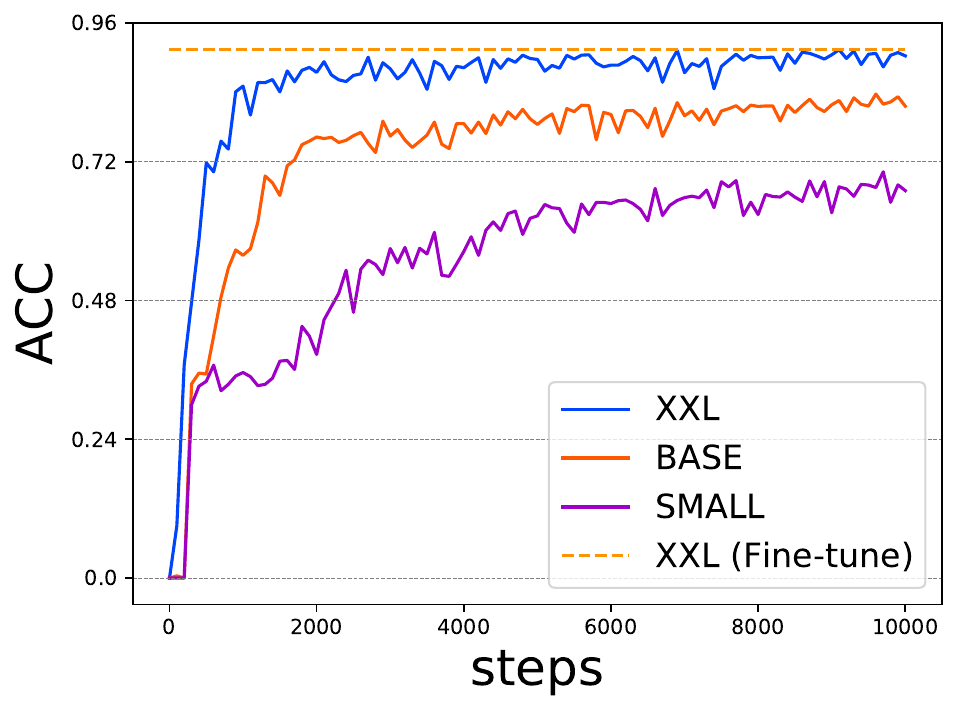}
    }
    \subfigure[Prefix-tuning (QNLI).]{
	    \includegraphics[width=0.31\textwidth]{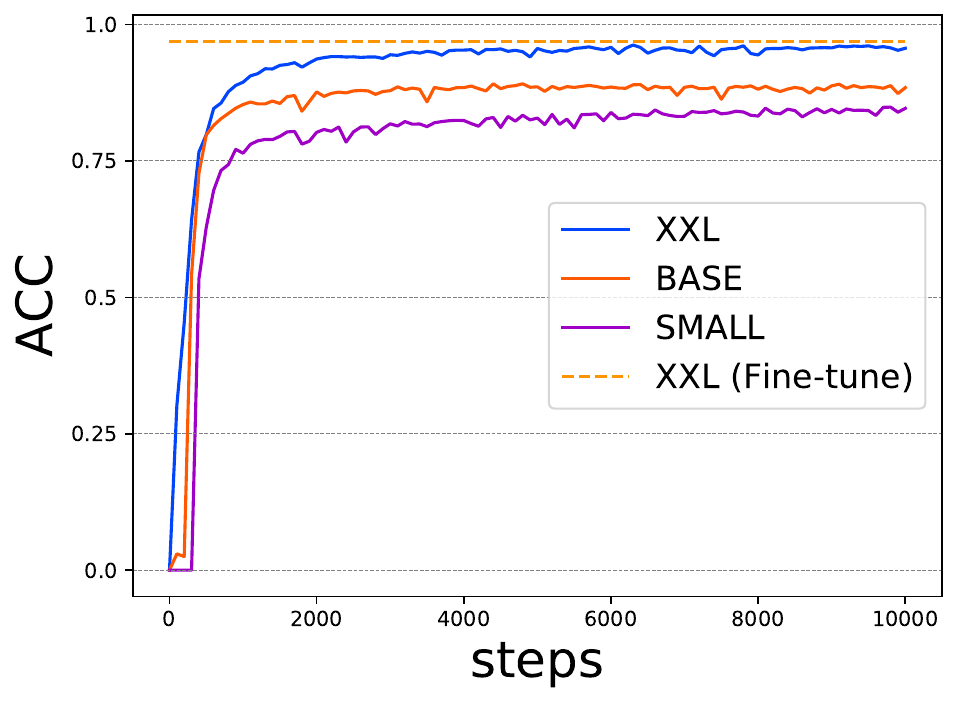}
    }
    \subfigure[Prefix-tuning (SST-2).]{
	    \includegraphics[width=0.31\textwidth]{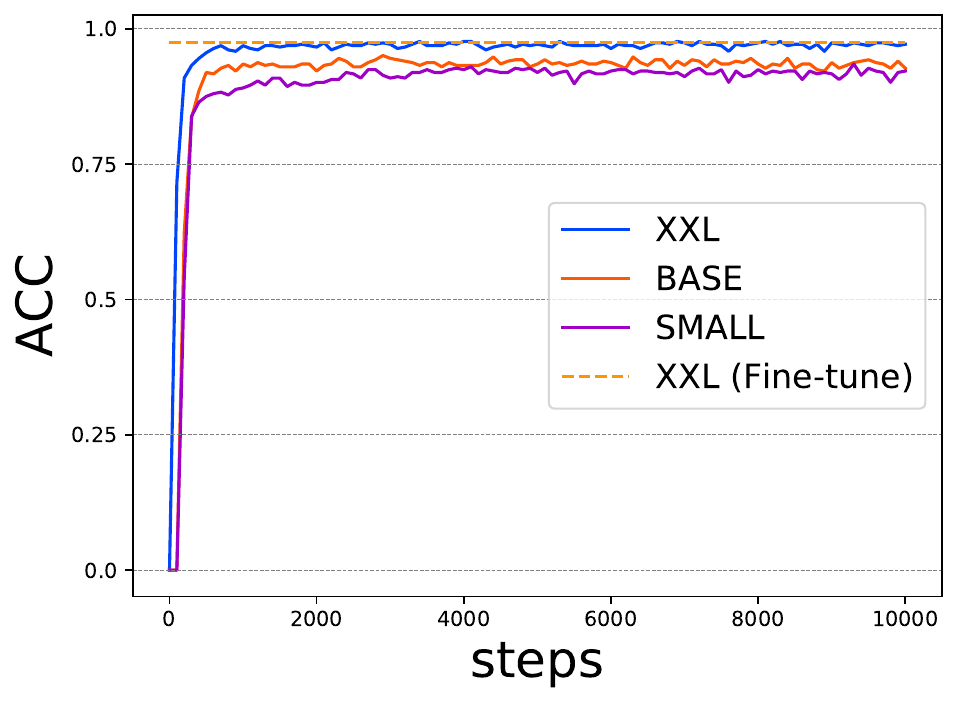}
    }

    \subfigure[Prompt Tuning (MNLI).]{
	    \includegraphics[width=0.31\textwidth]{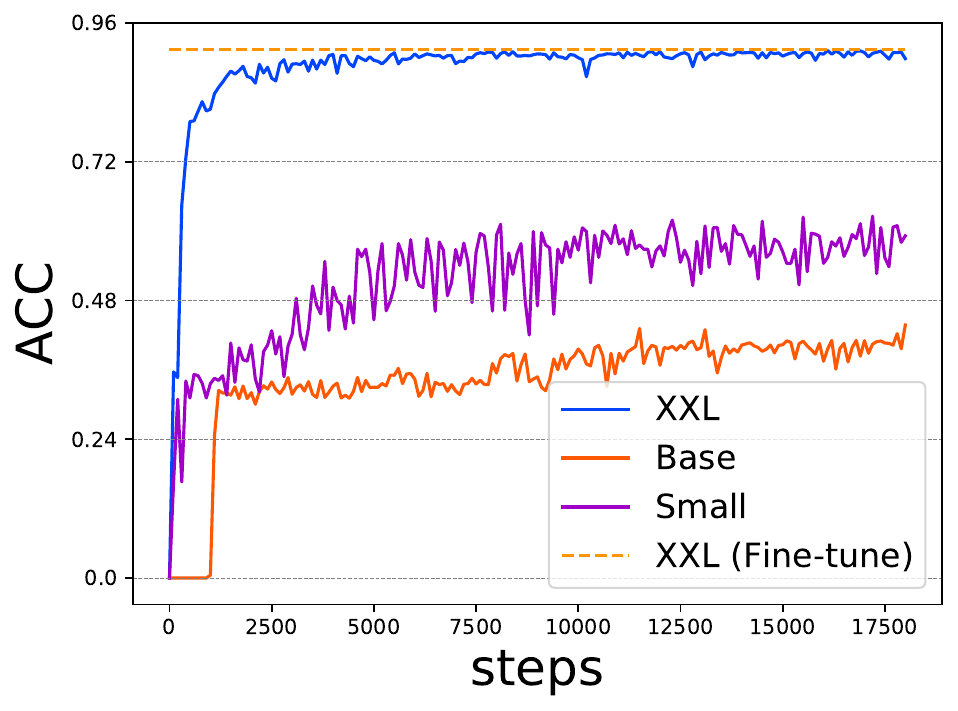}
    }
    \subfigure[Prompt Tuning (QNLI).]{
	    \includegraphics[width=0.31\textwidth]{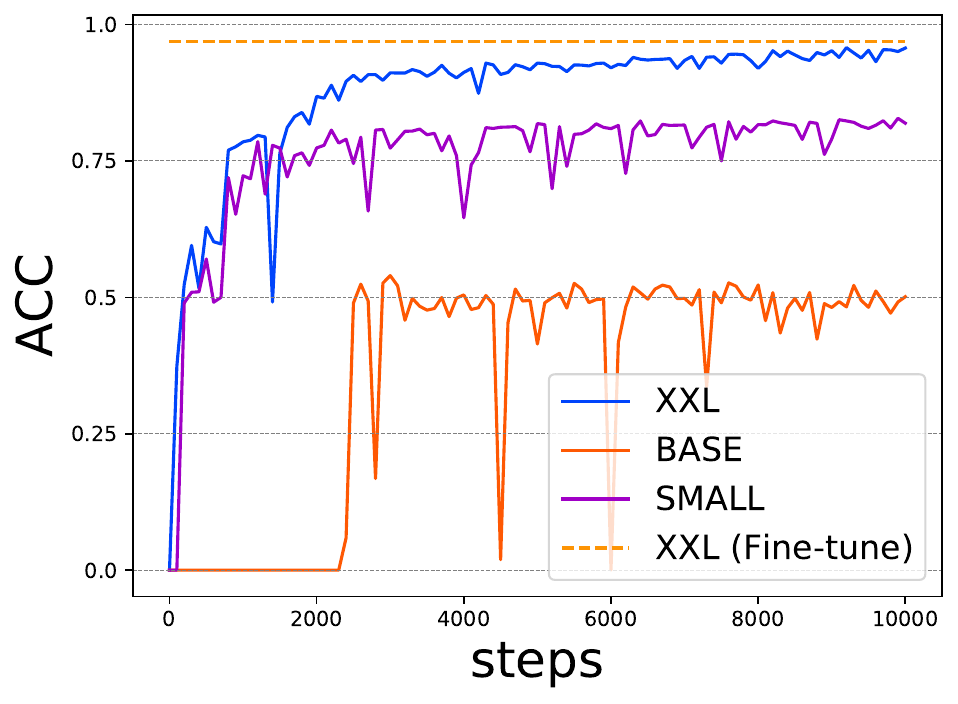}
    }
    \subfigure[Prompt Tuning (SST-2).]{
	    \includegraphics[width=0.31\textwidth]{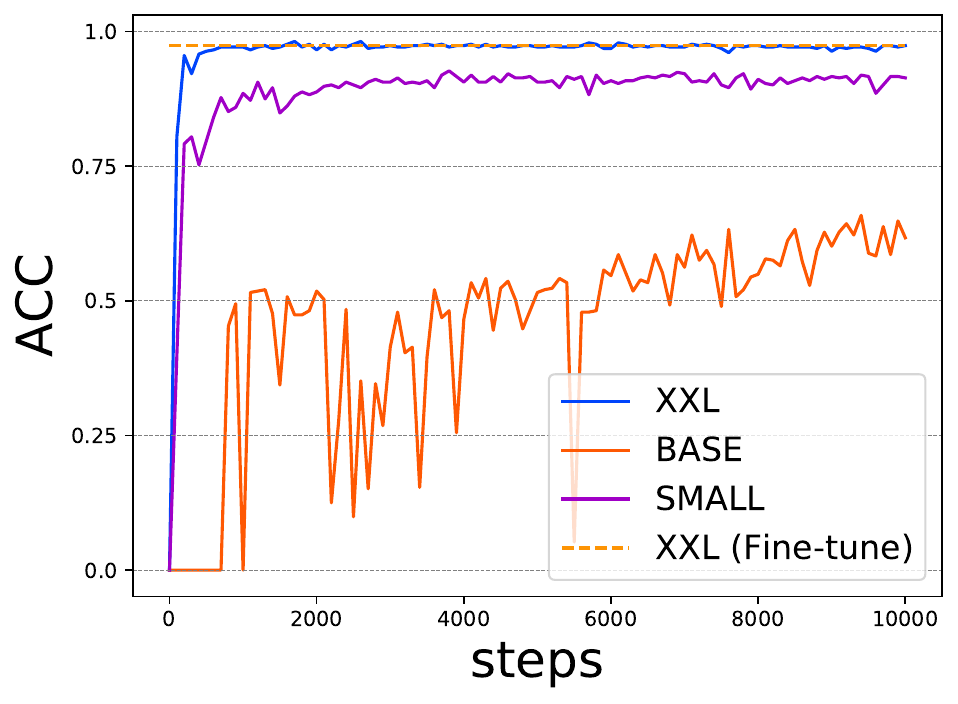}
    }

    \subfigure[Last Layer Tuning (MNLI).]{
	    \includegraphics[width=0.31\textwidth]{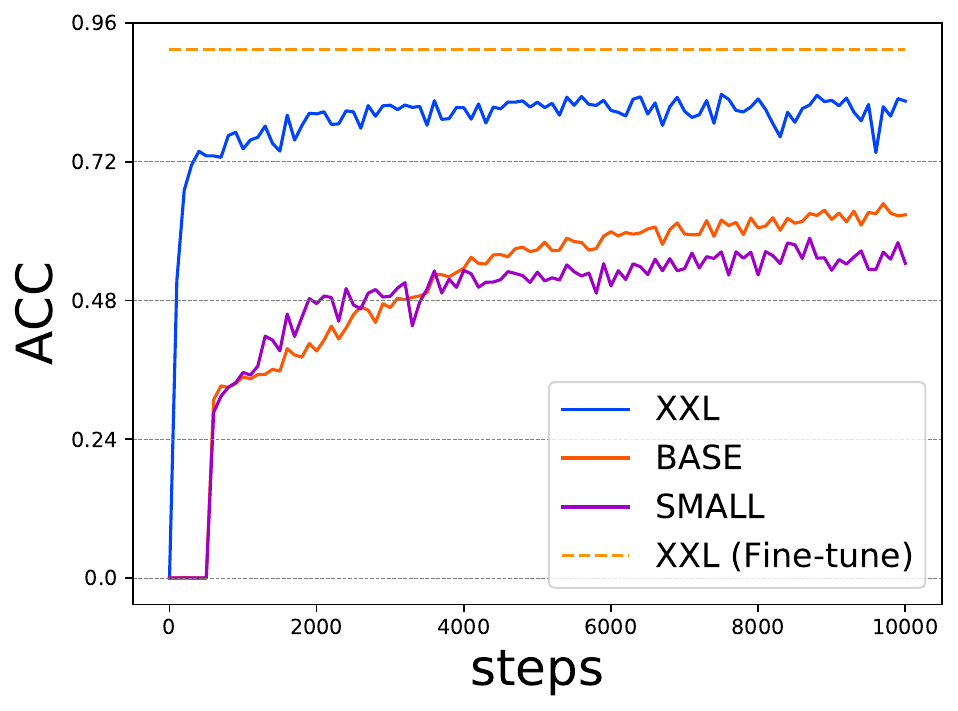}
    }
    \subfigure[Last Layer Tuning (QNLI).]{
	    \includegraphics[width=0.31\textwidth]{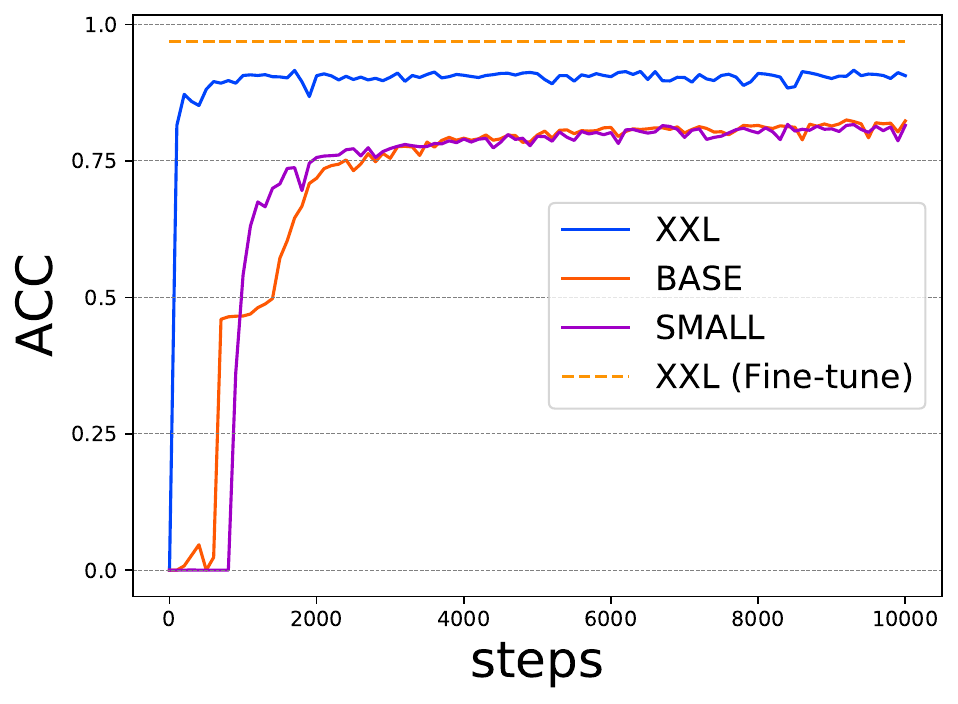}
    }
    \subfigure[Last Layer Tuning (SST-2).]{
	    \includegraphics[width=0.31\textwidth]{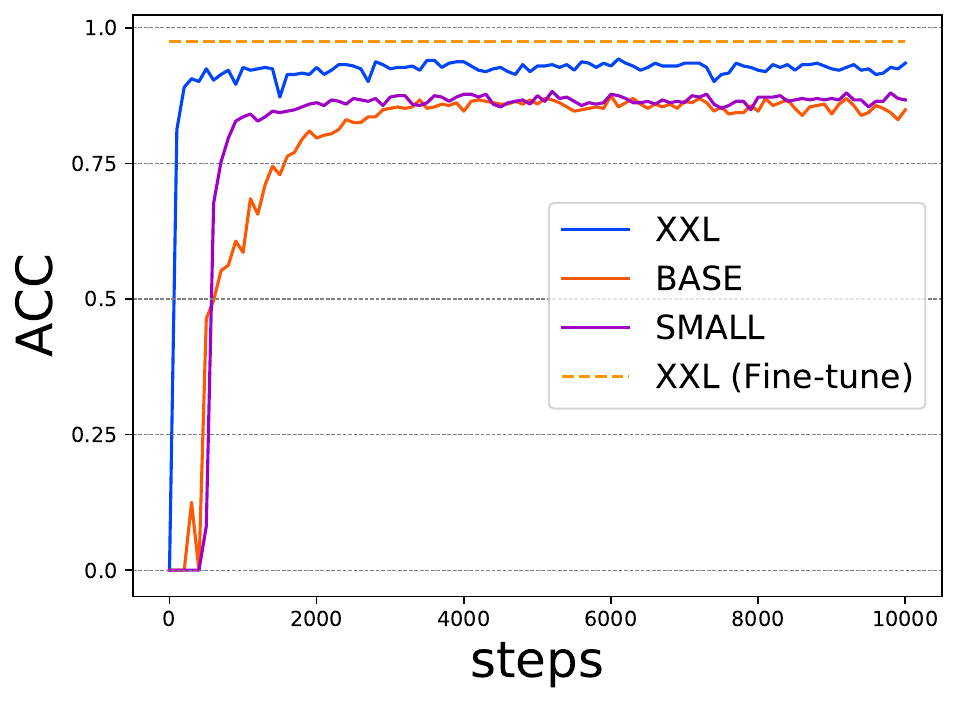}
    }

    % \subfigure[Selective Module Tuning (MNLI).]{
	   % \includegraphics[width=0.31\textwidth]{Figures/power_of_scale/random_eval_acc_MNLI.pdf}
    % }
    % \subfigure[Selective Module Tuning (QNLI).]{
	   % \includegraphics[width=0.31\textwidth]{Figures/power_of_scale/random_eval_acc_QNLI.pdf}
    % }
    % \subfigure[Selective Module Tuning (SST-2).]{
	   % \includegraphics[width=0.31\textwidth]{Figures/power_of_scale/random_eval_acc_SST-2.pdf}
    % }
    
    % \caption{We perform all delta tuning methods conditioned on different scales of T5: $\text{T5}_{\texttt{SMALL}}$(\textcolor{Purple}{\textemdash}), $\text{T5}_{\texttt{BASE}}$(\textcolor{Red}{\textemdash}), and $\text{T5}_{\texttt{XXL}}$(\textcolor{Blue}{\textemdash}). From the above figures, we can observe that with the scale of T5 increasing, all delta tuning methods could converge faster and achieve better performance on both MNLI and QNLI datasets.} 
    % \label{fig:the_power_of_the_scale}
\end{figure*}

\begin{figure*}[thbp]
    \subfigure[Selective Module Tuning (MNLI).]{
	    \includegraphics[width=0.31\textwidth]{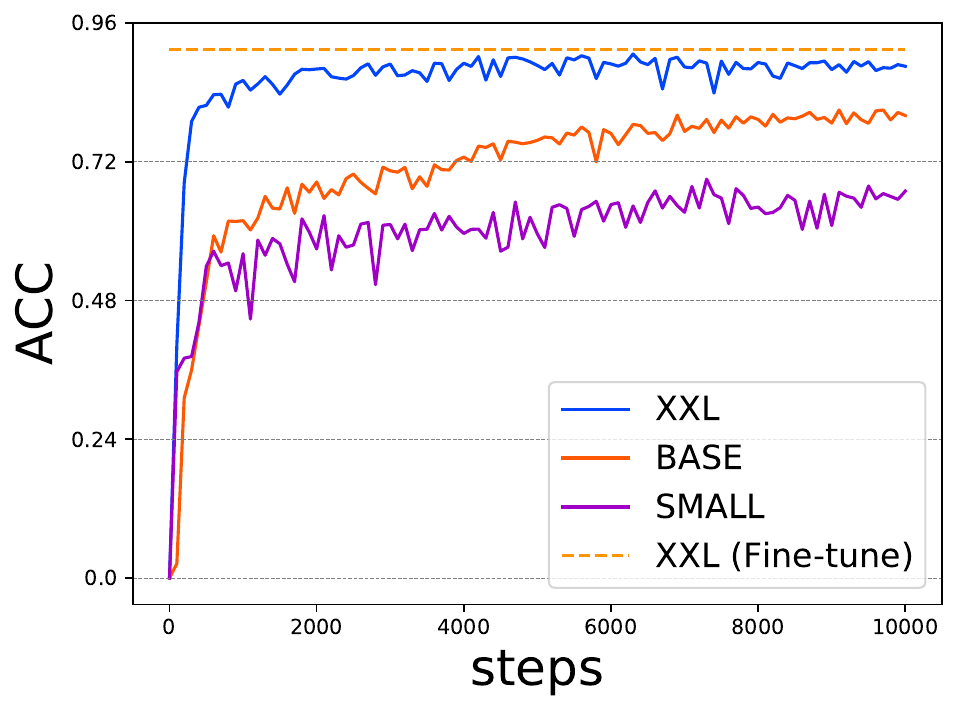}
    }
    \subfigure[Selective Module Tuning (QNLI).]{
	    \includegraphics[width=0.31\textwidth]{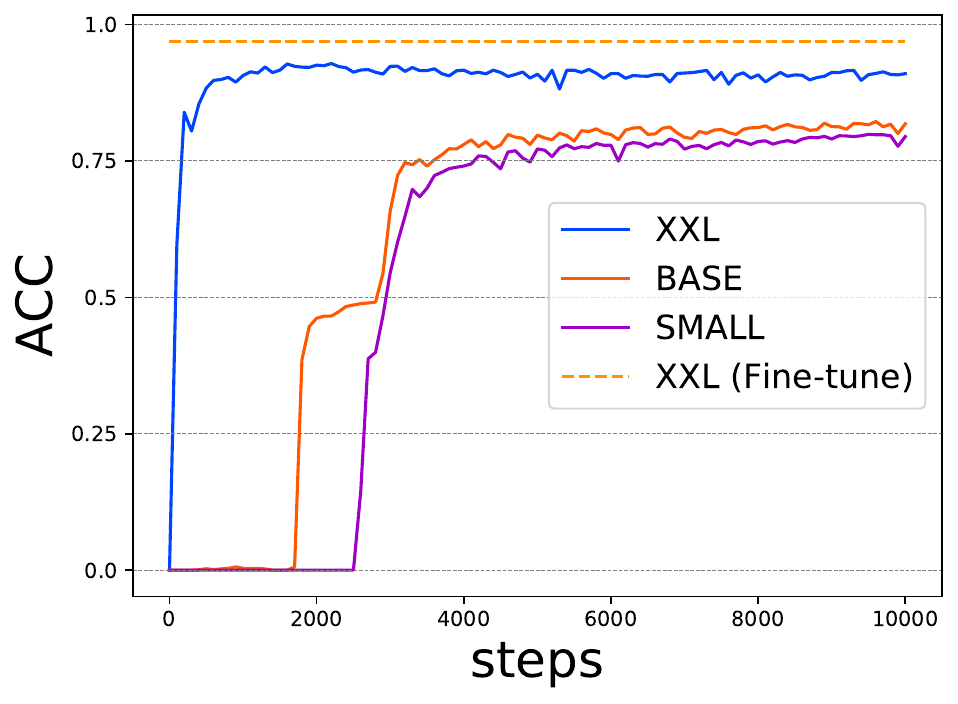}
    }
    \subfigure[Selective Module Tuning (SST-2).]{
	    \includegraphics[width=0.31\textwidth]{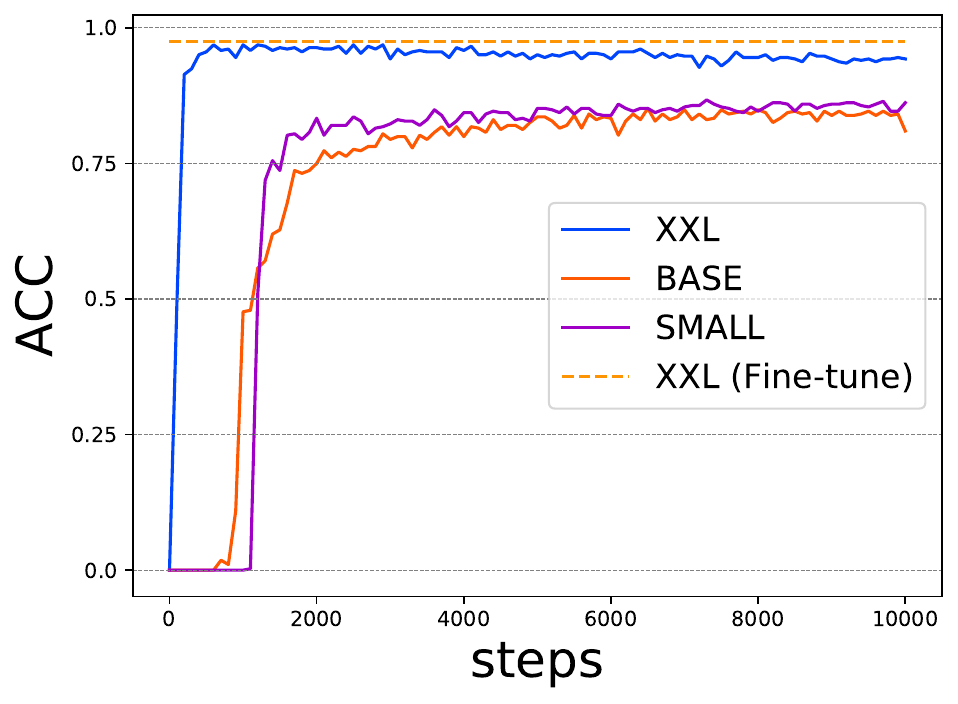}
    }
    
    \caption{We perform all delta tuning methods conditioned on different scales of T5: $\text{T5}_{\texttt{SMALL}}$(\textcolor{Purple}{\textemdash}), $\text{T5}_{\texttt{BASE}}$(\textcolor{Red}{\textemdash}), and $\text{T5}_{\texttt{XXL}}$(\textcolor{Blue}{\textemdash}). From the above figures, we can observe that with the scale of T5 increasing, all delta tuning methods could converge faster and achieve better performance on MNLI, QNLI, and SST-2.}
    \label{fig:the_power_of_the_scale}
\end{figure*}

\subsection{The Power of Scale for Delta Tuning}
\label{sec:scale}
% In the previous sections, we mentioned that delta tuning could be regarded as leveraging extra parameters to stimulate PLMs. As the extra parameters to stimulate PLMs, we are curious about the delta tuning methods' effectiveness in various scales of PLMs. 

% In the previous works, \citet{su2021transferability} and \citet{lester2021power} firstly surveyed prompt tuning on various scales of PLMs and respectively found that with the increasing size of PLMs, (1) the prompt tuning takes fewer training steps and (2) achieves the better performance on downstream tasks.

%and \citet{su2021transferability}

Recently, \citet{lester2021power} found that with the scale of the backbone PLM growing, prompt tuning becomes more and more competitive in performance, and would even achieve comparable performance than fine-tuning for a PLM with over $10$ billion parameters. Besides, \citet{su2021transferability} indicated that the convergence speed of prompt tuning benefits from the scaling law. In this section, we explore whether other delta tuning methods also exhibit such power of scale. Specifically, we experiment on the task of MNLI~\citep{williams-etal-2018-broad}, QNLI, and SST-2, and choose three PLMs ($\text{T5}_{\texttt{SMALL}}$, $\text{T5}_{\texttt{BASE}}$, $\text{T5}_{\texttt{XXL}}$) of increasing sizes, and evaluate the performance of six representative delta tuning methods (adapter, LoRA, prefix-tuning, prompt tuning, last layer tuning, and selective module tuning). Besides, we give the the percentages of the tuned parameters for various methods in every scale of the PLM as shown in Table ~\ref{tab:tuned_parameter_ratio}. We describe more training tails of this section in \refsec{Scale Details}.

The results are visualized in Figure~\ref{fig:the_power_of_the_scale}. From Figure~\ref{fig:the_power_of_the_scale} (a-i), we could observe that, with the scale of the PLM backbone growing, both the performance and the convergence of all delta tuning methods are significantly improved; (2) in addition, Figure~\ref{fig:the_power_of_the_scale} (j-l) indicates that compared with other delta tuning methods, prompt tuning tends to perform extremely bad for small-scale PLMs ($\text{T5}_{\texttt{SMALL}}$ and $\text{T5}_{\texttt{BASE}}$). However, as found in~\refsec{performance}, other delta tuning methods tend to perform comparable with fine-tuning even for a small-scale PLM ($\text{T5}_{\texttt{BASE}}$); (3) based on existing results, in Figure~\ref{fig:the_power_of_the_scale} (m-o) and (p-r), we further design two delta tuning methods: last layer tuning and selective module tuning. For last layer tuning, we optimize the last layer in T5 encoder; for selective module tuning, we randomly choose some modules (e.g., the feed-forward layer, query / key / value matrix in the attention layer, or a layer norm) in the T5 model to be tunable. Both methods show promising results especially when the scale of the PLM is extremely large, with selective module tuning slightly better than last layer tuning. These results suggest that confining the optimization within a specific layer may not be a good strategy (e.g., the case of prompt tuning and last layer tuning). On the other hand, randomly choosing modules across different layers could achieve excellent performance when the scale of PLMs grows extremely large. % These results indicate that, to better stimulate the knowledge contained in PLMs, it is necessary to optimize diverse modules across different layers; 

In general, the above results imply that, the power of scale may be a common phenomenon for delta tuning. We hypothesize the existence of such a phenomenon is because, larger PLMs generally have smaller intrinsic dimensionalities~\citep{aghajanyan2020intrinsic}, therefore, merely tuning minimal parameters could obtain a strong enough representation ability to achieve non-trivial performance in downstream tasks; besides, the over-parameterization and large-scale pre-training may make PLMs more unlikely to get stuck in a local optimum during downstream optimization, and thus the convergence is accelerated.

\begin{figure*}[!th]
    \centering
    \subfigure[Prompt Tuning]{
	    \includegraphics[width=0.482\textwidth]{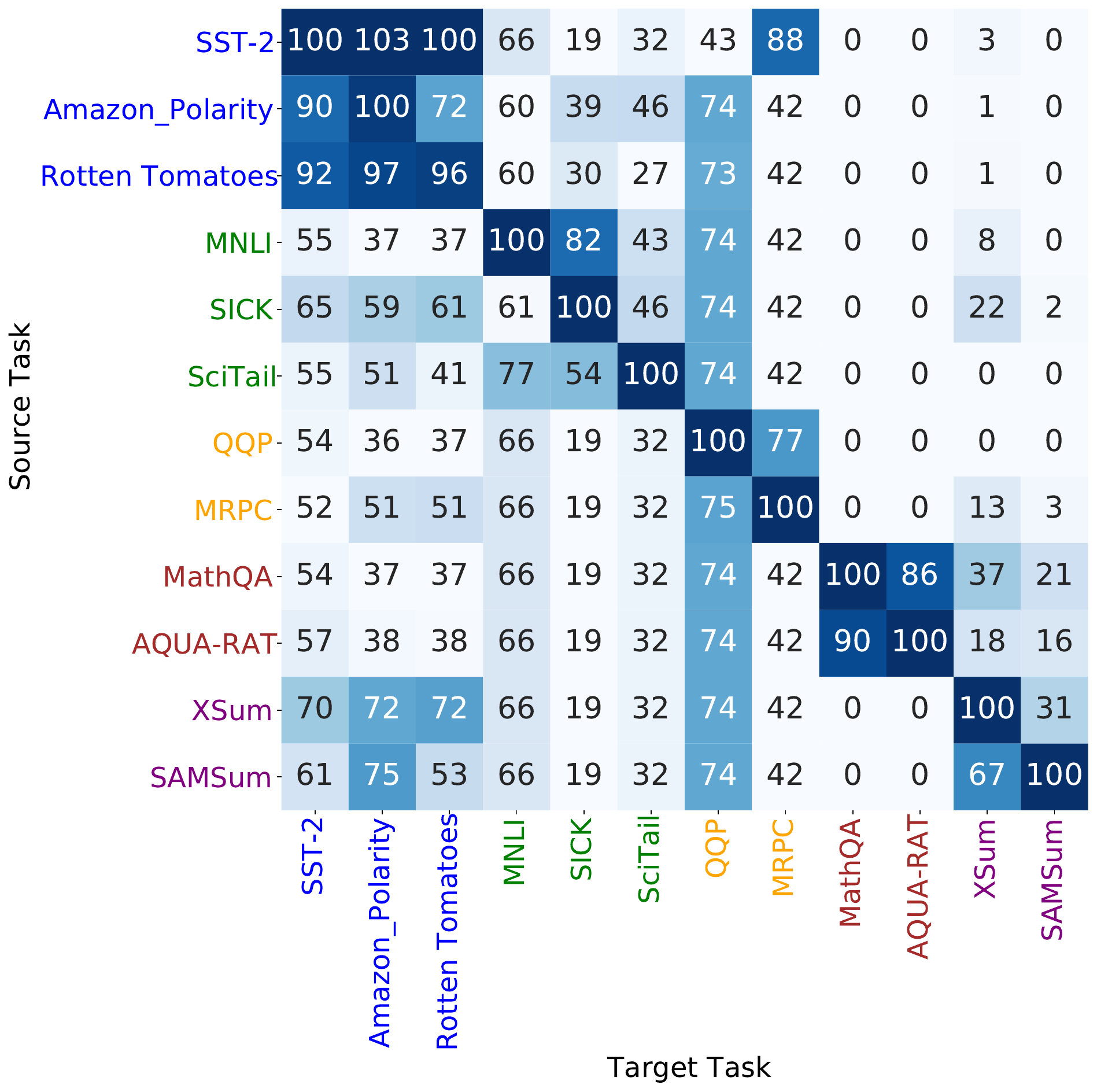}
    }
    \subfigure[Prefix-tuning]{
	    \includegraphics[width=0.482\textwidth]{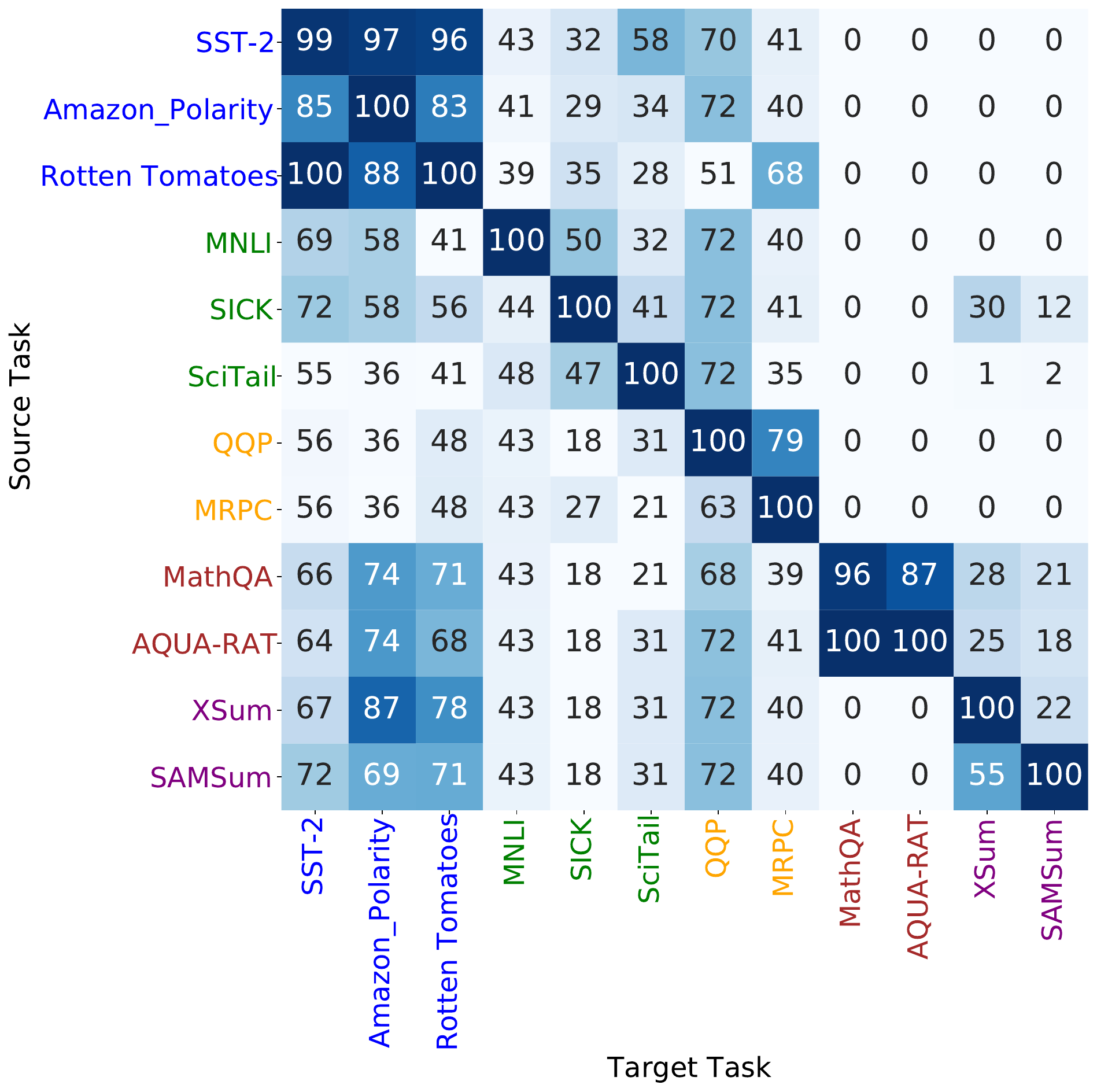}
    }
    \subfigure[Adapter]{
	    \includegraphics[width=0.482\textwidth]{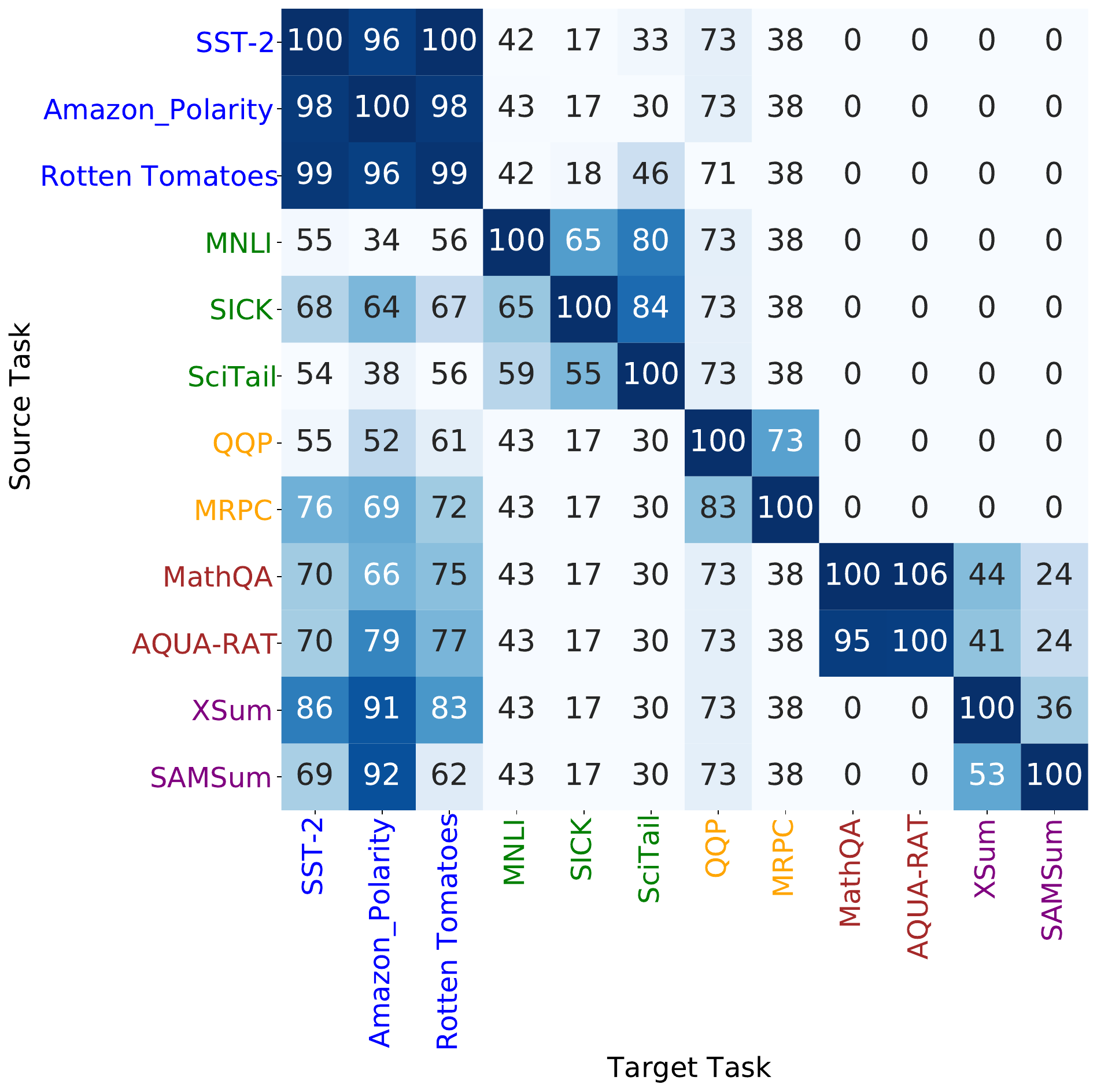}
    }
    \subfigure[LoRA]{
	    \includegraphics[width=0.482\textwidth]{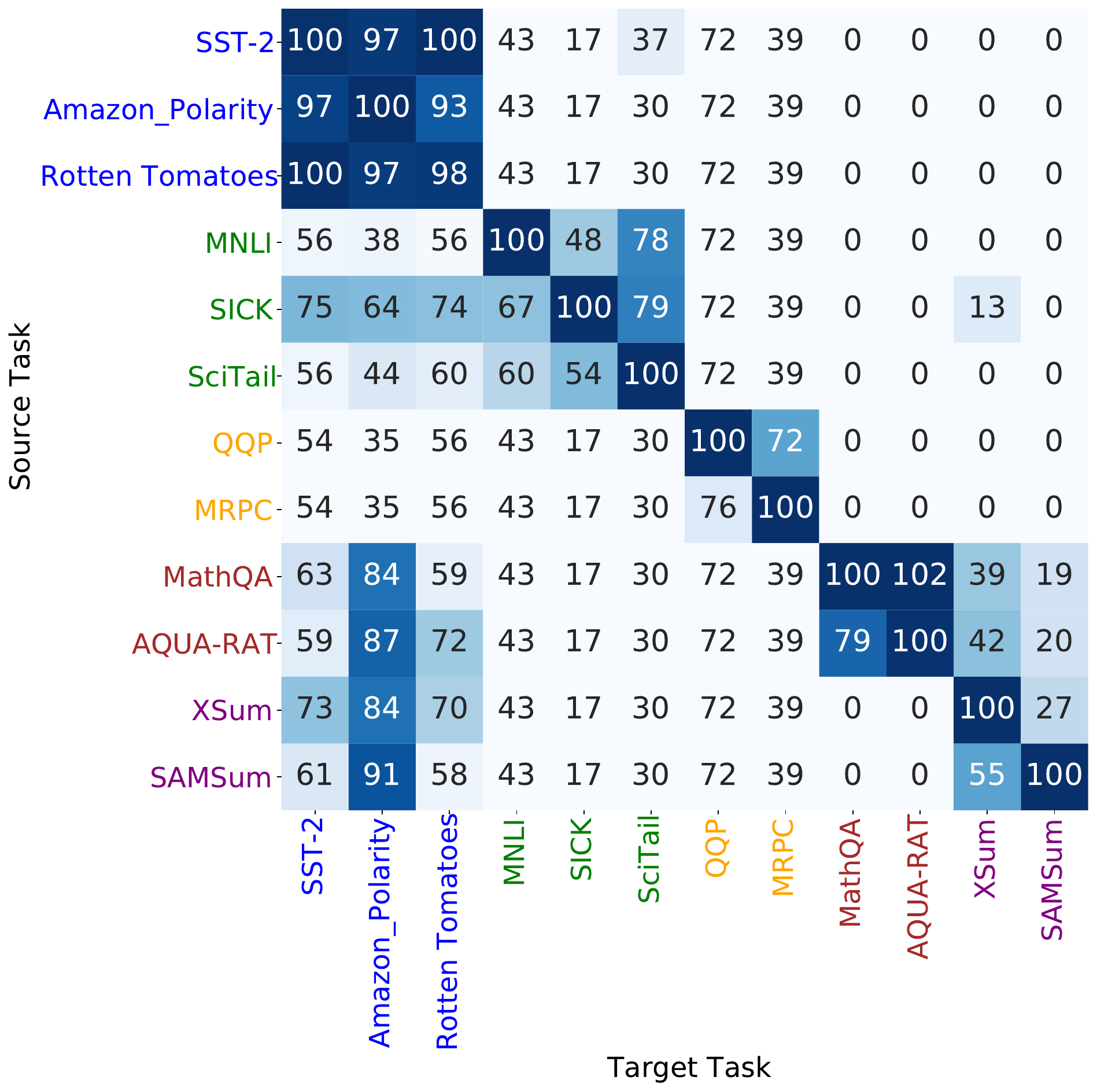}
    }
    \caption{Zero-shot transferring performance of four delta tuning methods using $\text{T5}_{\texttt{BASE}}$. We report relative performance (zero-shot transferring performance / original performance) (\%) on the target tasks (columns) when delta parameters are transferred from the source tasks (rows). Colors of the task names indicate the task types. \textcolor{Blue}{Blue}: sentiment analysis, \textcolor{OliveGreen}{Green}: natural language inference, \textcolor{YellowOrange}{Orange}: paraphrase identification, \textcolor{Brown}{Brown}: question answering, and \textcolor{Purple}{Purple}: summarization.}
    \label{fig:appendix_zero-shot-task-transfer_1}
\end{figure*}

\subsection{Task-level Transferability Evaluation}
\label{sec:transferability}
Recently, \citet{su2021transferability} and \citet{vu2021spot} demonstrate the cross-task transferability of prompt tuning. To verify whether cross-task transferability also exists in various delta tuning methods, we investigate four delta tuning methods (prompt tuning, prefix-tuning, adapter, and LoRA) and $12$ tasks of $5$ different types (sentiment analysis, natural language inference, paraphrase identification, question answering, summarization) by transferring the trained delta parameters to the unseen target tasks. More training and dataset details are left in \refsec{Transferability Details}.

% \citet{vu2021spot} and \citet{su2021transferability} explore the cross-task transferability of prompt tuning and find that the trained prompt on a source task can be reused on the similar target tasks in the zero-shot setting. 

In experiments, we report their relative performance (zero-shot transferring performance / original performance). The results are shown in Figure \ref{fig:appendix_zero-shot-task-transfer_1}, from which we can observe that: (1) for the tasks belonging to the same category, transferring tuned parameters among them generally performs well; (2) for the tasks of different types, transferring delta parameters among them generally achieves poor performance; (3) interestingly, we find that transferring tuned parameters from the text generation tasks such as \textcolor{Brown}{question answering} and  \textcolor{Purple}{summarization} can achieve non-trivial performance on \textcolor{blue}{sentiment analysis}, indicating that text generation tasks might be a more complex task that includes the knowledge required to solve the sentiment analysis tasks. These exciting results verify some common subspace among various tasks introduced in \refsec{optimization}, and demonstrate that it is promising to utilize trained delta parameters for similar tasks through knowledge transfer.

\section{Applications}
\label{sec:applications}
Delta tuning has been successfully applied to a variety of application scenarios. In this section, we briefly introduce several real-world applications, emphasizing on different advantages of delta tuning.

\begin{figure}[thbp]
    \centering
    \subfigure{
        \includegraphics[width=0.45\textwidth]{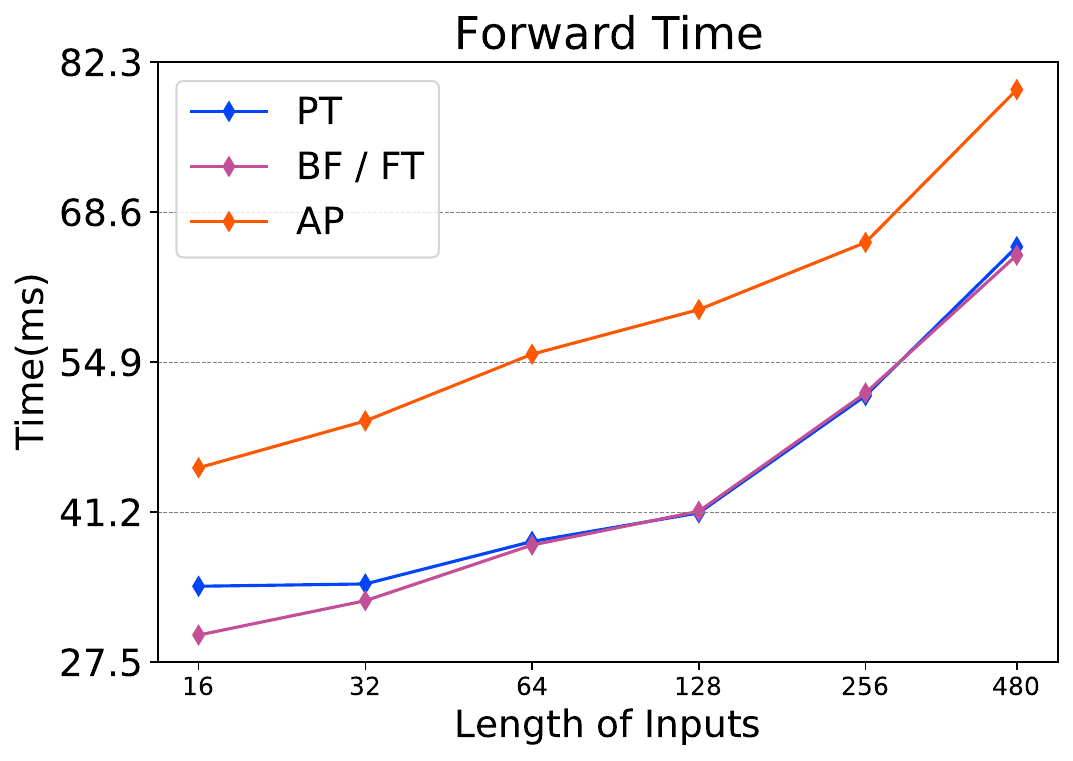}
    }
    \subfigure{
        \includegraphics[width=0.45\textwidth]{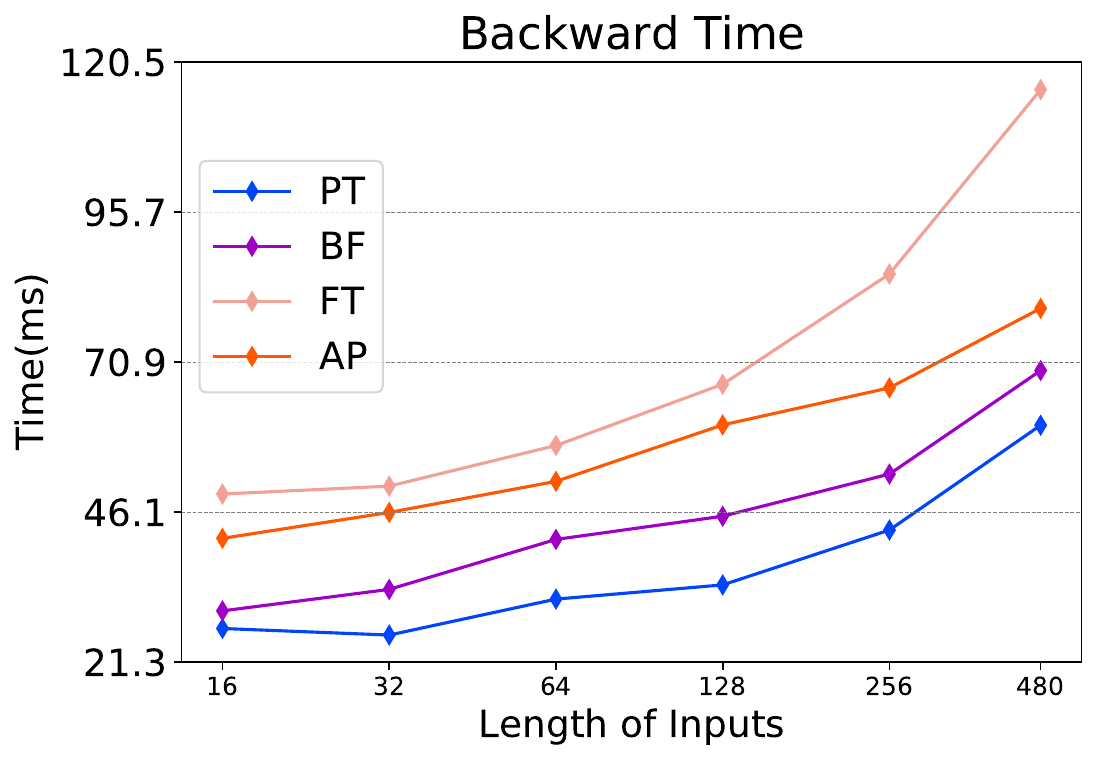}
    }
    \caption{Time consumption for fine-tuning (FT) and different delta tuning methods, including BitFit (BF), adapter (AP) and prompt tuning (PT). We report the results with different input length.}
    \label{fig:time}
\end{figure}

\paragraph{Fast Training and Shareable Checkpoints.}
Transformer-based models, although inherently parallelizable, are very slow to train due to their huge sizes, especially under the current era when ever-larger PLMs constantly emerge. Although delta tuning may converge slower than the traditional fine-tuning, the computations of the tunable parameters during backward propagation are significantly reduced, which conduces to speeding up training, as visualized in Figure~\ref{fig:time}. For instance, \citet{ruckle-etal-2021-adapterdrop} show that using adapters for downstream tuning could reduce training time to $40$\% while maintaining comparable performance than fine-tuning; \citet{mahabadi2021compacter} also indicate that a series of delta tuning methods significantly reduce both the training time for each epoch and the peak GPU memory, which is of paramount importance for practical applications. Another observation is that the structures of delta tuning methods could have considerable impact on the time of a single forward or backward process. Since \textbf{AP} injects additional neural modules to each layer of the Transformer model, the path of data flow has indeed become longer and further lead to inference latency. And such latency could be relatively reduced as the model scales.

Due to the lightweight nature, the tuned delta parameters could also save the storage space, making it easier to share the trained delta checkpoints among practitioners. With the help of delta tuning, researchers could easily scale up experiments to extremely large models containing even billions of parameters. Recently, researchers have been spending huge efforts to create a community of shareable delta tuning checkpoints, such as (1) AdapterHub\footnote{\url{https://adapterhub.ml}}~\citep{pfeiffer2020adapterhub}, an implementation of different adapter variants and a host for adapter checkpoints, and (2) OpenDelta\footnote{\url{https://github.com/thunlp/OpenDelta}}, an emerging plug-and-play library that is compatible with almost all PLMs based on PyTorch\footnote{\url{https://pytorch.org}}.% The much smaller checkpoints produced by delta tuning could be easily shared, reducing the extra effort that later researchers have to pay for model scaling.

% Transformer-based models, although inherently parallelizable, are very slow to train due to their large scale. ~\cite{ruckle-etal-2021-adapterdrop} reports that using adapters can reduce training time to 40\% while maintaining performance. Therefore, delta tuning  methods can be used to speed up training and model selection for training-intensive tasks. We encourage more researchers to scale up their experiments to larger models with the help of delta tuning. The much smaller checkpoints produced by delta tuning are easily shared, reducing the extra effort of later researchers for model scaling.

% add a figure for memory reduction

\paragraph{Multi-task Learning.}
Building a general-purpose AI system has always been the goal of researchers. Recently, extremely large PLMs, such as GPT-3~\citep{brown2020language}, have demonstrated the spectacular ability in fitting different data distributions simultaneously and promoting the downstream performance of various tasks. Multi-task learning has thus received a growing amount of attention under the era of large-scale pre-training. As a parameter-efficient substitution of full-model fine-tuning, delta tuning exhibits excellent ability for multi-task learning and in the meantime, maintains a relatively low additional storage. Successful applications include (1) multi-lingual learning: \citet{pfeiffer2020mad} propose to learn a series of invertible adapters between embeddings in the source and target languages to mitigate lexical differences across languages. The invertible adapter could well support knowledge transfer among multiple subtasks and maintain a low parameter budget, and (2) question answering: \citet{friedman2021single} prove that using a set of adapters that specialize in different QA formats performs favorably to a single language model that is fine-tuned on a mixture of QA formats. In addition, the simple average of specialized adapters also exhibits strong zero-shot transferring ability. Recently, \citet{liu2021pre,sun2021paradigm,karimi-mahabadi-etal-2021-parameter} also demonstrate that delta tuning could not only unify the tasks belonging to the same ontology, but also the tasks with substantially different formats so that different tasks could benefit from each other through knowledge transfer. Expanding from this idea, delta tuning can well support adaptations of large PLMs for multi-lingual and multi-domain scenarios. 

% Under this paradigm, multitask modeling, with the assistance of delta-tuning, is much more tractable to study.

\paragraph{Catastrophic Forgetting Mitigation.}
The language abilities acquired during pre-training are stored in parameters. As a consequence, updating all parameters in PLMs without regularization may lead to catastrophic forgetting when PLMs are sequentially trained across a suite of tasks~\citep{jin2021lifelong,qin2021knowledge,anony2021lifelong}. Since delta tuning only tunes minimal parameters, it could be a potential solution for mitigating the problem of catastrophic forgetting. For instance, MultiEURLEX~\citep{chalkidis2021multieurlex} introduce delta tuning into multilingual transfer learning, and demonstrate that using delta tuning methods rather than full-parameter fine-tuning boosts the performance of zero-shot transfer learning between the source language and the target language; \citet{jin2021lifelong} propose to introduce adapters into PLMs and maintain the original parameters fixed, so that PLMs could be trained in a lifelong manner for emerging data.

\paragraph{Language Model as Services and In-batch Parallel Computing.}
From the practical perspective, extremely large PLMs are generally released as services~\citep{brown2020language, nakano2021webgpt, sun2022black}, that is, users use the model by interacting with the released APIs rather than editing the source code. 
Considering the unaffordable communication costs between users and the service provider, delta tuning is apparently a more competitive choice over the traditional fine-tuning due to its lightweight nature. On one hand, the service provider could support training downstream tasks required by multiple users while consuming much fewer computations and storage space. In addition, considering that several delta tuning algorithms, such as prompt tuning~\citep{lester2021power} and prefix-tuning~\citep{li2021prefix} are inherently parallelizable, such a service could become more practical since delta tuning could well support \textit{in-batch parallel computing} by allowing instances from multiple users to be trained / evaluated in the same batch. Recent works~\citep{he2022unified} also show that most of the delta tuning methods, if not parallelizable inherently, could be modified to support parallel computing, e.g., parallel adapter~\citep{he2022unified}. On the other hand, when the gradients of the central PLM are not available to users, delta tuning still exhibits extraordinary talents in optimizing PLMs through derivative-free algorithms by only accessing the model inference APIs. Recently, \citet{sun2022black,diao2022black} pioneered to propose black-box tuning and show that their method could not only outperform manual prompts and GPT-3’s in-context learning, but also surpass the gradient-based counterparts.

% \section{Future Directions}
% \label{sec:future}

% \begin{itemize}
%     \item Unified and flexible implementations
%     \item Optimal delta parameters
%     \item Theory of model adaptation
% \end{itemize}

\section{Conclusion}
\label{sec:conclusion}

This paper focuses on parameter-efficient methods, i.e., delta tuning, for pre-trained language models. We first describe the problem and provide a categorization to systematically survey the development of delta tuning. Captivated by the empirical evidence, we propose two frameworks to theoretically discuss delta tuning from the optimization and the optimal control perspectives. Our discussion not only sheds light on the theoretical references of a novel design for delta tuning methods, but also implies that we could grasp the essential mechanisms of PLMs through deep analysis. Empirically, we conduct extensive experiments across 100+ NLP tasks to fairly evaluate and explore the combinatorial property, influence of scale, and transferability for delta tuning. Furthermore, we discuss the value of the applications of this paradigm. In summary, delta tuning exhibits significant potential to stimulate extremely large PLMs, and we hope that the paradigm could be further theoretically studied and empirically practiced. 

%We first describe the problem and survey the developement of delta tuning, which is categorized to addition-based, specification-based and reparameterization-based methods. %We then give a theoretical analysis from the optimization and optimal control perspectives.

\addcontentsline{toc}{section}{Broader Impacts}
\section*{Broader Impacts}
\label{sec:impacts}
%TBD

Delta tuning focuses on the efficient adaptation of pre-trained language models, which has both positive applications and potential harms for society. On the bright side, PLMs have exhibited unprecedented capability of natural language understanding (represented by BERT~\citep{devlin2018bert}) and generation (represented by GPT-3~\citep{brown2020language}), which empowers numerous real-world applications such as search engines, question-answering systems, intelligent writing systems, information extraction systems, and code completion, etc.
Recent research also shows that such large-scale PLMs could mimic the behavior to use search engines to answer difficult questions~\citep{nakano2021webgpt}. 
However, the risk is often hidden in the great successes,
on the other hand, PLMs may present biases in terms of gender, race, religion, etc, and even directly produce machine-written language with attacks, profanities, and insults~\citep{weidinger2021ethical}. 
This is because PLMs are pre-trained with large-scale realistic corpora, and these pre-trained data are likely to contain inherent bias. Language is a carrier of human views, so the prejudice and discrimination that exist in human society can easily be mapped onto language, and how to alleviate such challenges of fairness is a question that is well worth further study. Efforts could be made in two ways, first, directly by normalizing the training corpus to remove as many potential biases as possible, and second, by modifying model representations or outputs to reduce the risks. Outside of research, more comprehensive and improved treaties and norms for the use and modification of language models should be established by the community. 

So far, there is no clear evidence that delta tuning mitigates or exacerbates the potential hazards of PLMs. It is likely that the delta tuning methods will still inherit the potential risks of the base language model.  But the delta tuning methods seem to have considerable potential for correcting model bias. Under this circumstance, delta tuning is not only applied to efficiently adapt PLMs to downstream tasks but also could be utilized to specifically process risky information inside the models with a small number of parameters changed. In fact, it has been shown that it is possible to modify the factual errors made by the model in a computationally efficient way~\citep{mitchell2021fast}, signaling that the fairness issue can be potentially addressed through delta tuning. 
At the same time, we have to worry that such a strategy can also be used to further contaminate the language models to produce undesirable predictions. Here, we strongly encourage the community to conduct further research to comprehensively explore the various effects that delta tuning may have on PLMs.

When it comes to environmental issues, given that pre-training, fine-tuning, and storage of PLMs is a resource-intensive process, delta tuning attempts to minimize this impact from the outset. After probing the memory (Figure~\ref{fig:gpu}) and time consumption (Figure~\ref{fig:time}) of delta tuning in our work, we find that such methods could substantially reduce the computational cost. However, in the convergence analysis (Figure~\ref{fig:convergence}, \ref{fig:convergence_2}, \ref{fig:convergence_3}) we conclude that delta tuning methods tend to need more time to converge, although this phenomenon becomes insignificant as the model scales. In order to reduce unnecessary carbon emissions, we will open source all tools, code and checkpoints used in the experiment.

% \addcontentsline{toc}{section}{Reproducibility}
% \section*{Reproducibility}

\addcontentsline{toc}{section}{Acknowledgments}
\section*{Acknowledgments}
The authors would like to thank Junxian He for the constructive comments of the theoretical and empirical parts of the paper; Pengfei Liu for the valuable suggestions of the organization and presentation of the paper; Tianxiang Sun for the thoughtful suggestions of the criterion of delta tuning and experimental configurations; Chenglong Bao for the valuable suggestions about the optimization perspective of delta tuning; Xu Han, Huadong Wang, and Yufei Huang for their overall comments for the paper; Liwei Wang and Cong Fang for the discussion about the theoretical issues of delta tuning; Ruiqi Shao for helping us design the conceptual figure of delta tuning. 
Thanks to all the pioneering researchers who developed the structures, objectives, and delta tuning methods for pre-trained models. Ning Ding is supported by Baidu Scholarship.

\addcontentsline{toc}{section}{Contributions}
\section*{Contributions}
\label{sec:contributions}
The contributions of all authors are listed as follows: 
 Ning Ding, Yujia Qin and Zhiyuan Liu initiated and organized the research. The term ``delta tuning'' was coined by Shengding Hu and recognized by other authors for its vividness. Ning Ding drafted abstract, \refsec{introduction} and \refsec{delta tuning}, Yulin Chen and Ning Ding drafted ~\refsec{preliminaries}. Shengding Hu, Xiaozhi Wang, and Yujia Qin added contents to ~\refsec{addition} and ~\refsec{reparameterization}.
Weilin Zhao, Ning Ding, Yulin Chen, and Shengding Hu manually annotated the randomly selected 1,000 papers and created Table~\ref{tab:stat}, as well as Table~\ref{tab:different_adapters}. 
Fuchao Wei, Zonghan Yang, Ning Ding, Yujia Qin, Shengding Hu and Jianfei Chen discussed the scope and content of ~\refsec{theory}. Fuchao Wei developed the optimization framework and drafted ~\refsec{optimization}, Zonghan Yang and Yang Liu proposed the optimal control framework and drafted ~\refsec{optimal control}. Ning Ding verified the formula derivation. 
Yujia Qin led the empirical study part. And Yujia Qin, Guang Yang, Yusheng Su, Weize Chen, Jing Yi, Chi-Min Chan, and Ning Ding drafted ~\refsec{experiments}. Yujia Qin, Guang Yang, Weize Chen, Jing Yi, and Shengding Hu conducted the experiments for overall performance and combination (\refsecs{performance}{combination}). Yusheng Su and Chi-Min Chan conducted and wrote experiments for transferability and power of scale (\refsecs{scale}{transferability}). Shengding Hu and Yujia Qin drafted~\refsec{applications}.
Zhiyuan Liu, Hai-Tao Zheng, Yang Liu, Jie Tang, Juanzi Li, Maosong Sun advised the project, suggested the theoretical and empirical study and participated in the discussion. Ning Ding and Yujia Qin participated in all the sections and proofread the whole paper.

\newpage
\bibliographystyle{my}
\bibliography{custom}

\begin{thebibliography}{145}
\providecommand{\natexlab}[1]{#1}
\providecommand{\url}[1]{\texttt{#1}}
\expandafter\ifx\csname urlstyle\endcsname\relax
  \providecommand{\doi}[1]{doi: #1}\else
  \providecommand{\doi}{doi: \begingroup \urlstyle{rm}\Url}\fi

\bibitem[Aghajanyan et~al.(2021)Aghajanyan, Gupta, and
  Zettlemoyer]{aghajanyan2020intrinsic}
Armen Aghajanyan, Sonal Gupta, and Luke Zettlemoyer.
\newblock Intrinsic dimensionality explains the effectiveness of language model
  fine-tuning.
\newblock In \emph{Proceedings of the 59th Annual Meeting of the Association
  for Computational Linguistics and the 11th International Joint Conference on
  Natural Language Processing (Volume 1: Long Papers)}, pp.\  7319--7328,
  Online, 2021. Association for Computational Linguistics.
\newblock \doi{10.18653/v1/2021.acl-long.568}.
\newblock URL \url{https://aclanthology.org/2021.acl-long.568}.

\bibitem[Almeida et~al.(2011)Almeida, Hidalgo, and Yamakami]{sms_spam}
Tiago~A. Almeida, Jos\'{e} Mar\'{\i}a~G. Hidalgo, and Akebo Yamakami.
\newblock Contributions to the study of sms spam filtering: New collection and
  results.
\newblock In \emph{Proceedings of the 11th ACM Symposium on Document
  Engineering}, DocEng '11, pp.\  259–262, New York, NY, USA, 2011.
  Association for Computing Machinery.
\newblock ISBN 9781450308632.
\newblock \doi{10.1145/2034691.2034742}.
\newblock URL \url{https://doi.org/10.1145/2034691.2034742}.

\bibitem[Amini et~al.(2019)Amini, Gabriel, Lin, Koncel-Kedziorski, Choi, and
  Hajishirzi]{amini-etal-2019-mathqa}
Aida Amini, Saadia Gabriel, Shanchuan Lin, Rik Koncel-Kedziorski, Yejin Choi,
  and Hannaneh Hajishirzi.
\newblock {M}ath{QA}: Towards interpretable math word problem solving with
  operation-based formalisms.
\newblock In \emph{Proceedings of the 2019 Conference of the North {A}merican
  Chapter of the Association for Computational Linguistics: Human Language
  Technologies, Volume 1 (Long and Short Papers)}, pp.\  2357--2367,
  Minneapolis, Minnesota, 2019. Association for Computational Linguistics.
\newblock \doi{10.18653/v1/N19-1245}.
\newblock URL \url{https://aclanthology.org/N19-1245}.

\bibitem[Ang et~al.(2005)Ang, Chong, and Li]{controller-book-2}
Kiam~Heong Ang, Gregory Chong, and Yun Li.
\newblock Pid control system analysis, design, and technology.
\newblock \emph{IEEE transactions on control systems technology}, 13\penalty0
  (4):\penalty0 559--576, 2005.

\bibitem[Ba et~al.(2016)Ba, Kiros, and Hinton]{ba2016layer}
Jimmy~Lei Ba, Jamie~Ryan Kiros, and Geoffrey~E. Hinton.
\newblock Layer normalization, 2016.
\newblock URL \url{https://arxiv.org/abs/1607.06450}.

\bibitem[Bar-Haim et~al.(2006)Bar-Haim, Dagan, Dolan, Ferro, Giampiccolo,
  Magnini, and Szpektor]{bar2006second}
Roy Bar-Haim, Ido Dagan, Bill Dolan, Lisa Ferro, Danilo Giampiccolo, Bernardo
  Magnini, and Idan Szpektor.
\newblock The second pascal recognising textual entailment challenge.
\newblock In \emph{Proceedings of the second PASCAL challenges workshop on
  recognising textual entailment}, volume~6, pp.\  6--4. Venice, 2006.
\newblock URL
  \url{http://citeseerx.ist.psu.edu/viewdoc/download?doi=10.1.1.60.8552&rep=rep1&type=pdf}.

\bibitem[Barbieri et~al.(2020)Barbieri, Camacho-Collados, Espinosa~Anke, and
  Neves]{barbieri-etal-2020-tweeteval}
Francesco Barbieri, Jose Camacho-Collados, Luis Espinosa~Anke, and Leonardo
  Neves.
\newblock {T}weet{E}val: Unified benchmark and comparative evaluation for tweet
  classification.
\newblock In \emph{Findings of the Association for Computational Linguistics:
  EMNLP 2020}, pp.\  1644--1650, Online, 2020. Association for Computational
  Linguistics.
\newblock \doi{10.18653/v1/2020.findings-emnlp.148}.
\newblock URL \url{https://aclanthology.org/2020.findings-emnlp.148}.

\bibitem[Bengio et~al.(2000)Bengio, Ducharme, and Vincent]{bengio2000neural}
Yoshua Bengio, R{\'e}jean Ducharme, and Pascal Vincent.
\newblock A neural probabilistic language model.
\newblock \emph{Advances in Neural Information Processing Systems}, 13, 2000.
\newblock URL
  \url{https://proceedings.neurips.cc/paper/2000/hash/728f206c2a01bf572b5940d7d9a8fa4c-Abstract.html}.

\bibitem[Berant et~al.(2013)Berant, Chou, Frostig, and
  Liang]{berant-etal-2013-semantic}
Jonathan Berant, Andrew Chou, Roy Frostig, and Percy Liang.
\newblock Semantic parsing on {F}reebase from question-answer pairs.
\newblock In \emph{Proceedings of the 2013 Conference on Empirical Methods in
  Natural Language Processing}, pp.\  1533--1544, Seattle, Washington, USA,
  2013. Association for Computational Linguistics.
\newblock URL \url{https://aclanthology.org/D13-1160}.

\bibitem[Bommasani et~al.(2021)Bommasani, Hudson, Adeli, Altman, Arora, von
  Arx, Bernstein, Bohg, Bosselut, Brunskill,
  et~al.]{bommasani2021opportunities}
Rishi Bommasani, Drew~A Hudson, Ehsan Adeli, Russ Altman, Simran Arora, Sydney
  von Arx, Michael~S Bernstein, Jeannette Bohg, Antoine Bosselut, Emma
  Brunskill, et~al.
\newblock On the opportunities and risks of foundation models.
\newblock \emph{arXiv preprint arXiv:2108.07258}, 2021.
\newblock URL \url{https://arxiv.org/abs/2108.07258}.

\bibitem[Boratko et~al.(2020)Boratko, Li, O{'}Gorman, Das, Le, and
  McCallum]{boratko-etal-2020-protoqa}
Michael Boratko, Xiang Li, Tim O{'}Gorman, Rajarshi Das, Dan Le, and Andrew
  McCallum.
\newblock {P}roto{QA}: A question answering dataset for prototypical
  common-sense reasoning.
\newblock In \emph{Proceedings of the 2020 Conference on Empirical Methods in
  Natural Language Processing (EMNLP)}, pp.\  1122--1136, Online, 2020.
  Association for Computational Linguistics.
\newblock \doi{10.18653/v1/2020.emnlp-main.85}.
\newblock URL \url{https://aclanthology.org/2020.emnlp-main.85}.

\bibitem[Botha et~al.(2018)Botha, Faruqui, Alex, Baldridge, and
  Das]{botha-etal-2018-learning}
Jan~A. Botha, Manaal Faruqui, John Alex, Jason Baldridge, and Dipanjan Das.
\newblock Learning to split and rephrase from {W}ikipedia edit history.
\newblock In \emph{Proceedings of the 2018 Conference on Empirical Methods in
  Natural Language Processing}, pp.\  732--737, Brussels, Belgium, 2018.
  Association for Computational Linguistics.
\newblock \doi{10.18653/v1/D18-1080}.
\newblock URL \url{https://aclanthology.org/D18-1080}.

\bibitem[Boyd \& Barratt(1991)Boyd and Barratt]{controller-book-1}
Stephen~P Boyd and Craig~H Barratt.
\newblock \emph{Linear controller design: limits of performance}, volume~7.
\newblock Citeseer, 1991.

\bibitem[Brown et~al.(2020)Brown, Mann, Ryder, Subbiah, Kaplan, Dhariwal,
  Neelakantan, Shyam, Sastry, Askell, Agarwal, Herbert{-}Voss, Krueger,
  Henighan, Child, Ramesh, Ziegler, Wu, Winter, Hesse, Chen, Sigler, Litwin,
  Gray, Chess, Clark, Berner, McCandlish, Radford, Sutskever, and
  Amodei]{brown2020language}
Tom~B. Brown, Benjamin Mann, Nick Ryder, Melanie Subbiah, Jared Kaplan,
  Prafulla Dhariwal, Arvind Neelakantan, Pranav Shyam, Girish Sastry, Amanda
  Askell, Sandhini Agarwal, Ariel Herbert{-}Voss, Gretchen Krueger, Tom
  Henighan, Rewon Child, Aditya Ramesh, Daniel~M. Ziegler, Jeffrey Wu, Clemens
  Winter, Christopher Hesse, Mark Chen, Eric Sigler, Mateusz Litwin, Scott
  Gray, Benjamin Chess, Jack Clark, Christopher Berner, Sam McCandlish, Alec
  Radford, Ilya Sutskever, and Dario Amodei.
\newblock Language models are few-shot learners.
\newblock In Hugo Larochelle, Marc'Aurelio Ranzato, Raia Hadsell,
  Maria{-}Florina Balcan, and Hsuan{-}Tien Lin (eds.), \emph{Advances in Neural
  Information Processing Systems 33: Annual Conference on Neural Information
  Processing Systems 2020, NeurIPS 2020, December 6-12, 2020, virtual}, 2020.
\newblock URL
  \url{https://proceedings.neurips.cc/paper/2020/hash/1457c0d6bfcb4967418bfb8ac142f64a-Abstract.html}.

\bibitem[Cer et~al.(2017)Cer, Diab, Agirre, Lopez-Gazpio, and
  Specia]{cer-etal-2017-semeval}
Daniel Cer, Mona Diab, Eneko Agirre, I{\~n}igo Lopez-Gazpio, and Lucia Specia.
\newblock {S}em{E}val-2017 task 1: Semantic textual similarity multilingual and
  crosslingual focused evaluation.
\newblock In \emph{Proceedings of the 11th International Workshop on Semantic
  Evaluation ({S}em{E}val-2017)}, pp.\  1--14, Vancouver, Canada, August 2017.
  Association for Computational Linguistics.
\newblock \doi{10.18653/v1/S17-2001}.
\newblock URL \url{https://aclanthology.org/S17-2001}.

\bibitem[Chalkidis et~al.(2021)Chalkidis, Fergadiotis, and
  Androutsopoulos]{chalkidis2021multieurlex}
Ilias Chalkidis, Manos Fergadiotis, and Ion Androutsopoulos.
\newblock Multieurlex--a multi-lingual and multi-label legal document
  classification dataset for zero-shot cross-lingual transfer.
\newblock \emph{ArXiv preprint}, abs/2109.00904, 2021.
\newblock URL \url{https://arxiv.org/abs/2109.00904}.

\bibitem[Chatterjee et~al.(2019)Chatterjee, Narahari, Joshi, and
  Agrawal]{chatterjee-etal-2019-semeval}
Ankush Chatterjee, Kedhar~Nath Narahari, Meghana Joshi, and Puneet Agrawal.
\newblock {S}em{E}val-2019 task 3: {E}mo{C}ontext contextual emotion detection
  in text.
\newblock In \emph{Proceedings of the 13th International Workshop on Semantic
  Evaluation}, pp.\  39--48, Minneapolis, Minnesota, USA, 2019. Association for
  Computational Linguistics.
\newblock \doi{10.18653/v1/S19-2005}.
\newblock URL \url{https://aclanthology.org/S19-2005}.

\bibitem[Chen et~al.(2021)Chen, Li, and Zhang]{close-loop-control-robustness}
Zhuotong Chen, Qianxiao Li, and Zheng Zhang.
\newblock Towards robust neural networks via close-loop control.
\newblock In \emph{International Conference on Learning Representations}, 2021.
\newblock URL \url{https://openreview.net/forum?id=2AL06y9cDE-}.

\bibitem[Dagan et~al.(2005)Dagan, Glickman, and Magnini]{dagan2005pascal}
Ido Dagan, Oren Glickman, and Bernardo Magnini.
\newblock The pascal recognising textual entailment challenge.
\newblock In \emph{Machine Learning Challenges Workshop}, pp.\  177--190.
  Springer, 2005.
\newblock URL \url{https://link.springer.com/chapter/10.1007/11736790_9}.

\bibitem[Davidson et~al.(2017)Davidson, Warmsley, Macy, and
  Weber]{hateoffensive}
Thomas Davidson, Dana Warmsley, Michael Macy, and Ingmar Weber.
\newblock Automated hate speech detection and the problem of offensive
  language.
\newblock \emph{ArXiv preprint}, abs/1703.04009, 2017.
\newblock URL \url{https://arxiv.org/abs/1703.04009}.

\bibitem[Devlin et~al.(2019)Devlin, Chang, Lee, and Toutanova]{devlin2018bert}
Jacob Devlin, Ming-Wei Chang, Kenton Lee, and Kristina Toutanova.
\newblock {BERT}: Pre-training of deep bidirectional transformers for language
  understanding.
\newblock In \emph{Proceedings of the 2019 Conference of the North {A}merican
  Chapter of the Association for Computational Linguistics: Human Language
  Technologies, Volume 1 (Long and Short Papers)}, pp.\  4171--4186,
  Minneapolis, Minnesota, 2019. Association for Computational Linguistics.
\newblock \doi{10.18653/v1/N19-1423}.
\newblock URL \url{https://aclanthology.org/N19-1423}.

\bibitem[Diao et~al.(2022)Diao, Li, Lin, Huang, and Zhang]{diao2022black}
Shizhe Diao, Xuechun Li, Yong Lin, Zhichao Huang, and Tong Zhang.
\newblock Black-box prompt learning for pre-trained language models.
\newblock \emph{arXiv preprint arXiv:2201.08531}, 2022.
\newblock URL \url{https://arxiv.org/abs/2201.08531}.

\bibitem[Diggelmann et~al.(2020)Diggelmann, Boyd-Graber, Bulian, Ciaramita, and
  Leippold]{Diggelmann2020CLIMATEFEVERAD}
T.~Diggelmann, Jordan~L. Boyd-Graber, Jannis Bulian, Massimiliano Ciaramita,
  and Markus Leippold.
\newblock Climate-fever: A dataset for verification of real-world climate
  claims.
\newblock \emph{ArXiv preprint}, abs/2012.00614, 2020.
\newblock URL \url{https://arxiv.org/abs/2012.00614}.

\bibitem[Ding et~al.(2021)Ding, Hu, Zhao, Chen, Liu, Zheng, and
  Sun]{ding2021openprompt}
Ning Ding, Shengding Hu, Weilin Zhao, Yulin Chen, Zhiyuan Liu, Hai-Tao Zheng,
  and Maosong Sun.
\newblock Openprompt: An open-source framework for prompt-learning.
\newblock \emph{ArXiv preprint}, abs/2111.01998, 2021.
\newblock URL \url{https://arxiv.org/abs/2111.01998}.

\bibitem[Dolan \& Brockett(2005)Dolan and
  Brockett]{dolan-brockett-2005-automatically}
William~B. Dolan and Chris Brockett.
\newblock Automatically constructing a corpus of sentential paraphrases.
\newblock In \emph{Proceedings of IWP Workshop}, 2005.
\newblock URL \url{https://aclanthology.org/I05-5002}.

\bibitem[Dunn et~al.(2017)Dunn, Sagun, Higgins, G{\"u}ney, Cirik, and
  Cho]{Dunn2017SearchQAAN}
Matthew Dunn, Levent Sagun, Mike Higgins, V.~U. G{\"u}ney, Volkan Cirik, and
  Kyunghyun Cho.
\newblock Searchqa: A new q\&a dataset augmented with context from a search
  engine.
\newblock \emph{ArXiv preprint}, abs/1704.05179, 2017.
\newblock URL \url{https://arxiv.org/abs/1704.05179}.

\bibitem[Elfwing et~al.(2018)Elfwing, Uchibe, and Doya]{elfwing2018sigmoid}
Stefan Elfwing, Eiji Uchibe, and Kenji Doya.
\newblock Sigmoid-weighted linear units for neural network function
  approximation in reinforcement learning.
\newblock \emph{Neural Networks}, 107:\penalty0 3--11, 2018.
\newblock URL
  \url{https://www.sciencedirect.com/science/article/pii/S0893608017302976}.

\bibitem[Fabbri et~al.(2019)Fabbri, Li, She, Li, and
  Radev]{fabbri-etal-2019-multi}
Alexander Fabbri, Irene Li, Tianwei She, Suyi Li, and Dragomir Radev.
\newblock Multi-news: A large-scale multi-document summarization dataset and
  abstractive hierarchical model.
\newblock In \emph{Proceedings of the 57th Annual Meeting of the Association
  for Computational Linguistics}, pp.\  1074--1084, Florence, Italy, 2019.
  Association for Computational Linguistics.
\newblock \doi{10.18653/v1/P19-1102}.
\newblock URL \url{https://aclanthology.org/P19-1102}.

\bibitem[Fan et~al.(2019)Fan, Jernite, Perez, Grangier, Weston, and
  Auli]{fan-etal-2019-eli5}
Angela Fan, Yacine Jernite, Ethan Perez, David Grangier, Jason Weston, and
  Michael Auli.
\newblock {ELI}5: Long form question answering.
\newblock In \emph{Proceedings of the 57th Annual Meeting of the Association
  for Computational Linguistics}, pp.\  3558--3567, Florence, Italy, 2019.
  Association for Computational Linguistics.
\newblock \doi{10.18653/v1/P19-1346}.
\newblock URL \url{https://aclanthology.org/P19-1346}.

\bibitem[Friedman et~al.(2021)Friedman, Dodge, and Chen]{friedman2021single}
Dan Friedman, Ben Dodge, and Danqi Chen.
\newblock Single-dataset experts for multi-dataset question answering.
\newblock \emph{ArXiv preprint}, abs/2109.13880, 2021.
\newblock URL \url{https://arxiv.org/abs/2109.13880}.

\bibitem[Gao et~al.(2021)Gao, Fisch, and Chen]{Gao2021MakingPL}
Tianyu Gao, Adam Fisch, and Danqi Chen.
\newblock Making pre-trained language models better few-shot learners.
\newblock In \emph{Proceedings of the 59th Annual Meeting of the Association
  for Computational Linguistics and the 11th International Joint Conference on
  Natural Language Processing (Volume 1: Long Papers)}, pp.\  3816--3830,
  Online, 2021. Association for Computational Linguistics.
\newblock \doi{10.18653/v1/2021.acl-long.295}.
\newblock URL \url{https://aclanthology.org/2021.acl-long.295}.

\bibitem[Giampiccolo et~al.(2007)Giampiccolo, Magnini, Dagan, and
  Dolan]{giampiccolo2007third}
Danilo Giampiccolo, Bernardo Magnini, Ido Dagan, and Bill Dolan.
\newblock The third {PASCAL} recognizing textual entailment challenge.
\newblock In \emph{Proceedings of the {ACL}-{PASCAL} Workshop on Textual
  Entailment and Paraphrasing}, pp.\  1--9, Prague, 2007. Association for
  Computational Linguistics.
\newblock URL \url{https://aclanthology.org/W07-1401}.

\bibitem[Gliwa et~al.(2019)Gliwa, Mochol, Biesek, and
  Wawer]{gliwa-etal-2019-samsum}
Bogdan Gliwa, Iwona Mochol, Maciej Biesek, and Aleksander Wawer.
\newblock {SAMS}um corpus: A human-annotated dialogue dataset for abstractive
  summarization.
\newblock In \emph{Proceedings of the 2nd Workshop on New Frontiers in
  Summarization}, pp.\  70--79, Hong Kong, China, 2019. Association for
  Computational Linguistics.
\newblock \doi{10.18653/v1/D19-5409}.
\newblock URL \url{https://aclanthology.org/D19-5409}.

\bibitem[Gordon et~al.(2012)Gordon, Kozareva, and
  Roemmele]{gordon-etal-2012-semeval}
Andrew Gordon, Zornitsa Kozareva, and Melissa Roemmele.
\newblock {S}em{E}val-2012 task 7: Choice of plausible alternatives: An
  evaluation of commonsense causal reasoning.
\newblock In \emph{*{SEM} 2012: The First Joint Conference on Lexical and
  Computational Semantics {--} Volume 1: Proceedings of the main conference and
  the shared task, and Volume 2: Proceedings of the Sixth International
  Workshop on Semantic Evaluation ({S}em{E}val 2012)}, pp.\  394--398,
  Montr{\'e}al, Canada, 2012. Association for Computational Linguistics.
\newblock URL \url{https://aclanthology.org/S12-1052}.

\bibitem[Grefenstette et~al.(2014)Grefenstette, Blunsom,
  et~al.]{grefenstette2014convolutional}
Edward Grefenstette, Phil Blunsom, et~al.
\newblock A convolutional neural network for modelling sentences.
\newblock In \emph{ACL}, 2014.
\newblock URL \url{https://arxiv.org/abs/1404.2188}.

\bibitem[Gu et~al.(2021)Gu, Han, Liu, and Huang]{gu2021ppt}
Yuxian Gu, Xu~Han, Zhiyuan Liu, and Minlie Huang.
\newblock Ppt: Pre-trained prompt tuning for few-shot learning.
\newblock \emph{ArXiv preprint}, abs/2109.04332, 2021.
\newblock URL \url{https://arxiv.org/abs/2109.04332}.

\bibitem[Guo et~al.(2021)Guo, Rush, and Kim]{guo2021parameter}
Demi Guo, Alexander Rush, and Yoon Kim.
\newblock Parameter-efficient transfer learning with diff pruning.
\newblock In \emph{Proceedings of the 59th Annual Meeting of the Association
  for Computational Linguistics and the 11th International Joint Conference on
  Natural Language Processing (Volume 1: Long Papers)}, pp.\  4884--4896,
  Online, 2021. Association for Computational Linguistics.
\newblock \doi{10.18653/v1/2021.acl-long.378}.
\newblock URL \url{https://aclanthology.org/2021.acl-long.378}.

\bibitem[Gurulingappa et~al.(2012)Gurulingappa, Rajput, Roberts, Fluck,
  Hofmann-Apitius, and Toldo]{GURULINGAPPA2012885}
Harsha Gurulingappa, Abdul~Mateen Rajput, Angus Roberts, Juliane Fluck, Martin
  Hofmann-Apitius, and Luca Toldo.
\newblock Development of a benchmark corpus to support the automatic extraction
  of drug-related adverse effects from medical case reports.
\newblock \emph{Journal of Biomedical Informatics}, 45\penalty0 (5):\penalty0
  885--892, 2012.
\newblock ISSN 1532-0464.
\newblock \doi{https://doi.org/10.1016/j.jbi.2012.04.008}.
\newblock URL
  \url{https://www.sciencedirect.com/science/article/pii/S1532046412000615}.
\newblock Text Mining and Natural Language Processing in Pharmacogenomics.

\bibitem[Han et~al.(2021{\natexlab{a}})Han, Pang, and Wu]{han2021robust}
Wenjuan Han, Bo~Pang, and Ying~Nian Wu.
\newblock Robust transfer learning with pretrained language models through
  adapters.
\newblock In \emph{Proceedings of the 59th Annual Meeting of the Association
  for Computational Linguistics and the 11th International Joint Conference on
  Natural Language Processing (Volume 2: Short Papers)}, pp.\  854--861,
  Online, 2021{\natexlab{a}}. Association for Computational Linguistics.
\newblock \doi{10.18653/v1/2021.acl-short.108}.
\newblock URL \url{https://aclanthology.org/2021.acl-short.108}.

\bibitem[Han et~al.(2021{\natexlab{b}})Han, Zhang, Ding, Gu, Liu, Huo, Qiu,
  Zhang, Han, Huang, Jin, Lan, Liu, Liu, Lu, Qiu, Song, Tang, Wen, Yuan, Zhao,
  and Zhu]{HAN2021}
Xu~Han, Zhengyan Zhang, Ning Ding, Yuxian Gu, Xiao Liu, Yuqi Huo, Jiezhong Qiu,
  Liang Zhang, Wentao Han, Minlie Huang, Qin Jin, Yanyan Lan, Yang Liu, Zhiyuan
  Liu, Zhiwu Lu, Xipeng Qiu, Ruihua Song, Jie Tang, Ji-Rong Wen, Jinhui Yuan,
  Wayne~Xin Zhao, and Jun Zhu.
\newblock Pre-trained models: Past, present and future.
\newblock \emph{AI Open}, 2021{\natexlab{b}}.
\newblock ISSN 2666-6510.
\newblock \doi{https://doi.org/10.1016/j.aiopen.2021.08.002}.
\newblock URL
  \url{https://www.sciencedirect.com/science/article/pii/S2666651021000231}.

\bibitem[He et~al.(2022)He, Zhou, Ma, Berg-Kirkpatrick, and
  Neubig]{he2022unified}
Junxian He, Chunting Zhou, Xuezhe Ma, Taylor Berg-Kirkpatrick, and Graham
  Neubig.
\newblock Towards a unified view of parameter-efficient transfer learning.
\newblock In \emph{International Conference on Learning Representations}, 2022.
\newblock URL \url{https://openreview.net/forum?id=0RDcd5Axok}.

\bibitem[He et~al.(2016)He, Zhang, Ren, and Sun]{he2016deep}
Kaiming He, Xiangyu Zhang, Shaoqing Ren, and Jian Sun.
\newblock Deep residual learning for image recognition.
\newblock In \emph{2016 {IEEE} Conference on Computer Vision and Pattern
  Recognition, {CVPR} 2016, Las Vegas, NV, USA, June 27-30, 2016}, pp.\
  770--778. {IEEE} Computer Society, 2016.
\newblock \doi{10.1109/CVPR.2016.90}.
\newblock URL \url{https://doi.org/10.1109/CVPR.2016.90}.

\bibitem[He et~al.(2021)He, Liu, Ye, Tan, Ding, Cheng, Low, Bing, and
  Si]{he2021effectiveness}
Ruidan He, Linlin Liu, Hai Ye, Qingyu Tan, Bosheng Ding, Liying Cheng, Jiawei
  Low, Lidong Bing, and Luo Si.
\newblock On the effectiveness of adapter-based tuning for pretrained language
  model adaptation.
\newblock In \emph{Proceedings of the 59th Annual Meeting of the Association
  for Computational Linguistics and the 11th International Joint Conference on
  Natural Language Processing (Volume 1: Long Papers)}, pp.\  2208--2222,
  Online, 2021. Association for Computational Linguistics.
\newblock \doi{10.18653/v1/2021.acl-long.172}.
\newblock URL \url{https://aclanthology.org/2021.acl-long.172}.

\bibitem[Hochreiter \& Schmidhuber(1997)Hochreiter and
  Schmidhuber]{hochreiter1997long}
Sepp Hochreiter and J{\"u}rgen Schmidhuber.
\newblock Long short-term memory.
\newblock \emph{Neural computation}, 9\penalty0 (8):\penalty0 1735--1780, 1997.
\newblock URL \url{https://ieeexplore.ieee.org/abstract/document/6795963/}.

\bibitem[Hoffart et~al.(2011)Hoffart, Yosef, Bordino, F{\"u}rstenau, Pinkal,
  Spaniol, Taneva, Thater, and Weikum]{hoffart-etal-2011-robust}
Johannes Hoffart, Mohamed~Amir Yosef, Ilaria Bordino, Hagen F{\"u}rstenau,
  Manfred Pinkal, Marc Spaniol, Bilyana Taneva, Stefan Thater, and Gerhard
  Weikum.
\newblock Robust disambiguation of named entities in text.
\newblock In \emph{Proceedings of the 2011 Conference on Empirical Methods in
  Natural Language Processing}, pp.\  782--792, Edinburgh, Scotland, UK., 2011.
  Association for Computational Linguistics.
\newblock URL \url{https://aclanthology.org/D11-1072}.

\bibitem[Houlsby et~al.(2019)Houlsby, Giurgiu, Jastrzebski, Morrone,
  de~Laroussilhe, Gesmundo, Attariyan, and Gelly]{houlsby2019parameter}
Neil Houlsby, Andrei Giurgiu, Stanislaw Jastrzebski, Bruna Morrone, Quentin
  de~Laroussilhe, Andrea Gesmundo, Mona Attariyan, and Sylvain Gelly.
\newblock Parameter-efficient transfer learning for {NLP}.
\newblock In Kamalika Chaudhuri and Ruslan Salakhutdinov (eds.),
  \emph{Proceedings of the 36th International Conference on Machine Learning,
  {ICML} 2019, 9-15 June 2019, Long Beach, California, {USA}}, volume~97 of
  \emph{Proceedings of Machine Learning Research}, pp.\  2790--2799. {PMLR},
  2019.
\newblock URL \url{http://proceedings.mlr.press/v97/houlsby19a.html}.

\bibitem[Hu et~al.(2021{\natexlab{a}})Hu, Shen, Wallis, Allen-Zhu, Li, Wang,
  Wang, and Chen]{hu2021lora}
Edward~J Hu, Yelong Shen, Phillip Wallis, Zeyuan Allen-Zhu, Yuanzhi Li, Shean
  Wang, Lu~Wang, and Weizhu Chen.
\newblock Lora: Low-rank adaptation of large language models.
\newblock \emph{ArXiv preprint}, abs/2106.09685, 2021{\natexlab{a}}.
\newblock URL \url{https://arxiv.org/abs/2106.09685}.

\bibitem[Hu et~al.(2021{\natexlab{b}})Hu, NNN, Wang, Liu, Li, and
  Sun]{Hu2021KnowledgeablePI}
Shengding Hu, NNN, Huadong Wang, Zhiyuan Liu, Juan-Zi Li, and Maosong Sun.
\newblock Knowledgeable prompt-tuning: Incorporating knowledge into prompt
  verbalizer for text classification.
\newblock \emph{ArXiv}, abs/2108.02035, 2021{\natexlab{b}}.
\newblock URL \url{https://arxiv.org/abs/2108.02035}.

\bibitem[Huang et~al.(2019)Huang, Le~Bras, Bhagavatula, and
  Choi]{huang-etal-2019-cosmos}
Lifu Huang, Ronan Le~Bras, Chandra Bhagavatula, and Yejin Choi.
\newblock Cosmos {QA}: Machine reading comprehension with contextual
  commonsense reasoning.
\newblock In \emph{Proceedings of the 2019 Conference on Empirical Methods in
  Natural Language Processing and the 9th International Joint Conference on
  Natural Language Processing (EMNLP-IJCNLP)}, pp.\  2391--2401, Hong Kong,
  China, 2019. Association for Computational Linguistics.
\newblock \doi{10.18653/v1/D19-1243}.
\newblock URL \url{https://aclanthology.org/D19-1243}.

\bibitem[Jiang et~al.(2019)Jiang, Wu, and Jiang]{jiang-etal-2019-freebaseqa}
Kelvin Jiang, Dekun Wu, and Hui Jiang.
\newblock {F}reebase{QA}: A new factoid {QA} data set matching trivia-style
  question-answer pairs with {F}reebase.
\newblock In \emph{Proceedings of the 2019 Conference of the North {A}merican
  Chapter of the Association for Computational Linguistics: Human Language
  Technologies, Volume 1 (Long and Short Papers)}, pp.\  318--323, Minneapolis,
  Minnesota, 2019. Association for Computational Linguistics.
\newblock \doi{10.18653/v1/N19-1028}.
\newblock URL \url{https://aclanthology.org/N19-1028}.

\bibitem[Jin et~al.(2021)Jin, Zhang, Zhu, Xiao, Li, Wei, Arnold, and
  Ren]{jin2021lifelong}
Xisen Jin, Dejiao Zhang, Henghui Zhu, Wei Xiao, Shang-Wen Li, Xiaokai Wei,
  Andrew Arnold, and Xiang Ren.
\newblock Lifelong pretraining: Continually adapting language models to
  emerging corpora.
\newblock \emph{arXiv preprint arXiv:2110.08534}, 2021.
\newblock URL \url{https://arxiv.org/abs/2110.08534}.

\bibitem[Karimi~Mahabadi et~al.(2021)Karimi~Mahabadi, Ruder, Dehghani, and
  Henderson]{karimi-mahabadi-etal-2021-parameter}
Rabeeh Karimi~Mahabadi, Sebastian Ruder, Mostafa Dehghani, and James Henderson.
\newblock Parameter-efficient multi-task fine-tuning for transformers via
  shared hypernetworks.
\newblock In \emph{Proceedings of the 59th Annual Meeting of the Association
  for Computational Linguistics and the 11th International Joint Conference on
  Natural Language Processing (Volume 1: Long Papers)}, pp.\  565--576, Online,
  2021. Association for Computational Linguistics.
\newblock \doi{10.18653/v1/2021.acl-long.47}.
\newblock URL \url{https://aclanthology.org/2021.acl-long.47}.

\bibitem[Khot et~al.(2018{\natexlab{a}})Khot, Sabharwal, and
  Clark]{Khot2018SciTaiLAT}
Tushar Khot, Ashish Sabharwal, and Peter Clark.
\newblock Scitail: {A} textual entailment dataset from science question
  answering.
\newblock In Sheila~A. McIlraith and Kilian~Q. Weinberger (eds.),
  \emph{Proceedings of the Thirty-Second {AAAI} Conference on Artificial
  Intelligence, (AAAI-18), the 30th innovative Applications of Artificial
  Intelligence (IAAI-18), and the 8th {AAAI} Symposium on Educational Advances
  in Artificial Intelligence (EAAI-18), New Orleans, Louisiana, USA, February
  2-7, 2018}, pp.\  5189--5197. {AAAI} Press, 2018{\natexlab{a}}.
\newblock URL
  \url{https://www.aaai.org/ocs/index.php/AAAI/AAAI18/paper/view/17368}.

\bibitem[Khot et~al.(2018{\natexlab{b}})Khot, Sabharwal, and Clark]{scitail}
Tushar Khot, Ashish Sabharwal, and Peter Clark.
\newblock {SciTail}: A textual entailment dataset from science question
  answering.
\newblock In \emph{AAAI}, 2018{\natexlab{b}}.
\newblock URL
  \url{https://www.aaai.org/ocs/index.php/AAAI/AAAI18/paper/view/17368/0}.

\bibitem[Khot et~al.(2020)Khot, Clark, Guerquin, Jansen, and
  Sabharwal]{khot2020qasc}
Tushar Khot, Peter Clark, Michal Guerquin, Peter Jansen, and Ashish Sabharwal.
\newblock {QASC:} {A} dataset for question answering via sentence composition.
\newblock In \emph{The Thirty-Fourth {AAAI} Conference on Artificial
  Intelligence, {AAAI} 2020, The Thirty-Second Innovative Applications of
  Artificial Intelligence Conference, {IAAI} 2020, The Tenth {AAAI} Symposium
  on Educational Advances in Artificial Intelligence, {EAAI} 2020, New York,
  NY, USA, February 7-12, 2020}, pp.\  8082--8090. {AAAI} Press, 2020.
\newblock URL \url{https://aaai.org/ojs/index.php/AAAI/article/view/6319}.

\bibitem[Kim(2014)]{kim-2014-convolutional}
Yoon Kim.
\newblock Convolutional neural networks for sentence classification.
\newblock In \emph{Proceedings of the 2014 Conference on Empirical Methods in
  Natural Language Processing ({EMNLP})}, pp.\  1746--1751, Doha, Qatar,
  October 2014. Association for Computational Linguistics.
\newblock \doi{10.3115/v1/D14-1181}.
\newblock URL \url{https://aclanthology.org/D14-1181}.

\bibitem[Kingma \& Ba(2015)Kingma and Ba]{kingma2014adam}
Diederik~P. Kingma and Jimmy Ba.
\newblock Adam: {A} method for stochastic optimization.
\newblock In Yoshua Bengio and Yann LeCun (eds.), \emph{3rd International
  Conference on Learning Representations, {ICLR} 2015, San Diego, CA, USA, May
  7-9, 2015, Conference Track Proceedings}, 2015.
\newblock URL \url{http://arxiv.org/abs/1412.6980}.

\bibitem[Kopp(1962)]{pmp}
Richard~E Kopp.
\newblock {Pontryagin maximum principle}.
\newblock In \emph{Mathematics in Science and Engineering}, volume~5, pp.\
  255--279. Elsevier, 1962.
\newblock \doi{10.1016/S0076-5392(08)62095-0}.

\bibitem[Lai et~al.(2017)Lai, Xie, Liu, Yang, and Hovy]{lai-etal-2017-race}
Guokun Lai, Qizhe Xie, Hanxiao Liu, Yiming Yang, and Eduard Hovy.
\newblock {RACE}: Large-scale {R}e{A}ding comprehension dataset from
  examinations.
\newblock In \emph{Proceedings of the 2017 Conference on Empirical Methods in
  Natural Language Processing}, pp.\  785--794, Copenhagen, Denmark, 2017.
  Association for Computational Linguistics.
\newblock \doi{10.18653/v1/D17-1082}.
\newblock URL \url{https://aclanthology.org/D17-1082}.

\bibitem[Lebret et~al.(2016)Lebret, Grangier, and
  Auli]{lebret-etal-2016-neural}
R{\'e}mi Lebret, David Grangier, and Michael Auli.
\newblock Neural text generation from structured data with application to the
  biography domain.
\newblock In \emph{Proceedings of the 2016 Conference on Empirical Methods in
  Natural Language Processing}, pp.\  1203--1213, Austin, Texas, 2016.
  Association for Computational Linguistics.
\newblock \doi{10.18653/v1/D16-1128}.
\newblock URL \url{https://aclanthology.org/D16-1128}.

\bibitem[LeCun et~al.(2015)LeCun, Bengio, and Hinton]{lecun2015deep}
Yann LeCun, Yoshua Bengio, and Geoffrey Hinton.
\newblock Deep learning.
\newblock \emph{nature}, 521\penalty0 (7553):\penalty0 436--444, 2015.

\bibitem[Lee et~al.(2019)Lee, Tang, and Lin]{lee2019would}
Jaejun Lee, Raphael Tang, and Jimmy Lin.
\newblock What would elsa do? freezing layers during transformer fine-tuning.
\newblock \emph{ArXiv preprint}, abs/1911.03090, 2019.
\newblock URL \url{https://arxiv.org/abs/1911.03090}.

\bibitem[Leshno et~al.(1993)Leshno, Lin, Pinkus, and
  Schocken]{leshno1993multilayer}
Moshe Leshno, Vladimir~Ya Lin, Allan Pinkus, and Shimon Schocken.
\newblock Multilayer feedforward networks with a nonpolynomial activation
  function can approximate any function.
\newblock \emph{Neural networks}, 6\penalty0 (6):\penalty0 861--867, 1993.

\bibitem[Lester et~al.(2021)Lester, Al-Rfou, and Constant]{lester2021power}
Brian Lester, Rami Al-Rfou, and Noah Constant.
\newblock The power of scale for parameter-efficient prompt tuning.
\newblock \emph{ArXiv preprint}, abs/2104.08691, 2021.
\newblock URL \url{https://arxiv.org/abs/2104.08691}.

\bibitem[Levesque et~al.(2012)Levesque, Davis, and
  Morgenstern]{levesque2012winograd}
Hector~J. Levesque, Ernest Davis, and Leora Morgenstern.
\newblock The winograd schema challenge.
\newblock In \emph{Proceedings of the Thirteenth International Conference on
  Principles of Knowledge Representation and Reasoning}, KR'12, pp.\
  552–561. AAAI Press, 2012.
\newblock ISBN 9781577355601.
\newblock URL
  \url{https://www.aaai.org/ocs/index.php/KR/KR12/paper/download/4492/4924}.

\bibitem[Lewis et~al.(2020)Lewis, Liu, Goyal, Ghazvininejad, Mohamed, Levy,
  Stoyanov, and Zettlemoyer]{mike2019bart}
Mike Lewis, Yinhan Liu, Naman Goyal, Marjan Ghazvininejad, Abdelrahman Mohamed,
  Omer Levy, Veselin Stoyanov, and Luke Zettlemoyer.
\newblock {BART}: Denoising sequence-to-sequence pre-training for natural
  language generation, translation, and comprehension.
\newblock In \emph{Proceedings of the 58th Annual Meeting of the Association
  for Computational Linguistics}, pp.\  7871--7880, Online, 2020. Association
  for Computational Linguistics.
\newblock \doi{10.18653/v1/2020.acl-main.703}.
\newblock URL \url{https://aclanthology.org/2020.acl-main.703}.

\bibitem[Lhoest et~al.(2021)Lhoest, Villanova~del Moral, Jernite, Thakur, von
  Platen, Patil, Chaumond, Drame, Plu, Tunstall, Davison, {\v{S}}a{\v{s}}ko,
  Chhablani, Malik, Brandeis, Le~Scao, Sanh, Xu, Patry, McMillan-Major, Schmid,
  Gugger, Delangue, Matussi{\`e}re, Debut, Bekman, Cistac, Goehringer, Mustar,
  Lagunas, Rush, and Wolf]{lhoest-etal-2021-datasets}
Quentin Lhoest, Albert Villanova~del Moral, Yacine Jernite, Abhishek Thakur,
  Patrick von Platen, Suraj Patil, Julien Chaumond, Mariama Drame, Julien Plu,
  Lewis Tunstall, Joe Davison, Mario {\v{S}}a{\v{s}}ko, Gunjan Chhablani,
  Bhavitvya Malik, Simon Brandeis, Teven Le~Scao, Victor Sanh, Canwen Xu,
  Nicolas Patry, Angelina McMillan-Major, Philipp Schmid, Sylvain Gugger,
  Cl{\'e}ment Delangue, Th{\'e}o Matussi{\`e}re, Lysandre Debut, Stas Bekman,
  Pierric Cistac, Thibault Goehringer, Victor Mustar, Fran{\c{c}}ois Lagunas,
  Alexander Rush, and Thomas Wolf.
\newblock Datasets: A community library for natural language processing.
\newblock In \emph{Proceedings of the 2021 Conference on Empirical Methods in
  Natural Language Processing: System Demonstrations}, pp.\  175--184, Online
  and Punta Cana, Dominican Republic, 2021. Association for Computational
  Linguistics.
\newblock \doi{10.18653/v1/2021.emnlp-demo.21}.
\newblock URL \url{https://aclanthology.org/2021.emnlp-demo.21}.

\bibitem[Li et~al.(2018)Li, Farkhoor, Liu, and Yosinski]{li2018measuring}
Chunyuan Li, Heerad Farkhoor, Rosanne Liu, and Jason Yosinski.
\newblock Measuring the intrinsic dimension of objective landscapes.
\newblock \emph{arXiv preprint arXiv:1804.08838}, 2018.

\bibitem[Li et~al.(2017)Li, Chen, Tai, and E]{pmp-dl}
Qianxiao Li, Long Chen, Cheng Tai, and Weinan E.
\newblock {Maximum Principle Based Algorithms for Deep Learning}.
\newblock \emph{J. Mach. Learn. Res.}, 18:\penalty0 165:1--165:29, 2017.
\newblock URL \url{http://jmlr.org/papers/v18/17-653.html}.

\bibitem[Li \& Liang(2021)Li and Liang]{li2021prefix}
Xiang~Lisa Li and Percy Liang.
\newblock Prefix-tuning: Optimizing continuous prompts for generation.
\newblock In \emph{Proceedings of the 59th Annual Meeting of the Association
  for Computational Linguistics and the 11th International Joint Conference on
  Natural Language Processing (Volume 1: Long Papers)}, pp.\  4582--4597,
  Online, 2021. Association for Computational Linguistics.
\newblock \doi{10.18653/v1/2021.acl-long.353}.
\newblock URL \url{https://aclanthology.org/2021.acl-long.353}.

\bibitem[Lin et~al.(2020)Lin, Lee, Khanna, and Ren]{lin-etal-2020-birds}
Bill~Yuchen Lin, Seyeon Lee, Rahul Khanna, and Xiang Ren.
\newblock {B}irds have four legs?! {N}umer{S}ense: {P}robing {N}umerical
  {C}ommonsense {K}nowledge of {P}re-{T}rained {L}anguage {M}odels.
\newblock In \emph{Proceedings of the 2020 Conference on Empirical Methods in
  Natural Language Processing (EMNLP)}, pp.\  6862--6868, Online, 2020.
  Association for Computational Linguistics.
\newblock \doi{10.18653/v1/2020.emnlp-main.557}.
\newblock URL \url{https://aclanthology.org/2020.emnlp-main.557}.

\bibitem[Lin(2004)]{lin-2004-rouge}
Chin-Yew Lin.
\newblock {ROUGE}: A package for automatic evaluation of summaries.
\newblock In \emph{Text Summarization Branches Out}, pp.\  74--81, Barcelona,
  Spain, July 2004. Association for Computational Linguistics.
\newblock URL \url{https://aclanthology.org/W04-1013}.

\bibitem[Ling et~al.(2017)Ling, Yogatama, Dyer, and
  Blunsom]{ling-etal-2017-program}
Wang Ling, Dani Yogatama, Chris Dyer, and Phil Blunsom.
\newblock Program induction by rationale generation: Learning to solve and
  explain algebraic word problems.
\newblock In \emph{Proceedings of the 55th Annual Meeting of the Association
  for Computational Linguistics (Volume 1: Long Papers)}, pp.\  158--167,
  Vancouver, Canada, July 2017. Association for Computational Linguistics.
\newblock \doi{10.18653/v1/P17-1015}.
\newblock URL \url{https://aclanthology.org/P17-1015}.

\bibitem[Liu et~al.(2021{\natexlab{a}})Liu, Yuan, Fu, Jiang, Hayashi, and
  Neubig]{liu2021pre}
Pengfei Liu, Weizhe Yuan, Jinlan Fu, Zhengbao Jiang, Hiroaki Hayashi, and
  Graham Neubig.
\newblock Pre-train, prompt, and predict: A systematic survey of prompting
  methods in natural language processing.
\newblock \emph{ArXiv preprint}, abs/2107.13586, 2021{\natexlab{a}}.
\newblock URL \url{https://arxiv.org/abs/2107.13586}.

\bibitem[Liu et~al.(2021{\natexlab{b}})Liu, Ji, Fu, Du, Yang, and
  Tang]{liu2021p}
Xiao Liu, Kaixuan Ji, Yicheng Fu, Zhengxiao Du, Zhilin Yang, and Jie Tang.
\newblock P-tuning v2: Prompt tuning can be comparable to fine-tuning
  universally across scales and tasks.
\newblock \emph{arXiv preprint arXiv:2110.07602}, 2021{\natexlab{b}}.
\newblock URL \url{https://arxiv.org/pdf/2110.07602.pdf}.

\bibitem[Liu et~al.(2019)Liu, Ott, Goyal, Du, Joshi, Chen, Levy, Lewis,
  Zettlemoyer, and Stoyanov]{liu2019roberta}
Yinhan Liu, Myle Ott, Naman Goyal, Jingfei Du, Mandar Joshi, Danqi Chen, Omer
  Levy, Mike Lewis, Luke Zettlemoyer, and Veselin Stoyanov.
\newblock Roberta: A robustly optimized bert pretraining approach.
\newblock \emph{ArXiv preprint}, abs/1907.11692, 2019.
\newblock URL \url{https://arxiv.org/abs/1907.11692}.

\bibitem[Liu et~al.(2022)Liu, An, and Qiu]{liu2022mathcal}
Yitao Liu, Chenxin An, and Xipeng Qiu.
\newblock Y-tuning: An efficient tuning paradigm for large-scale pre-trained
  models via label representation learning.
\newblock \emph{arXiv preprint arXiv:2202.09817}, 2022.
\newblock URL \url{https://arxiv.org/pdf/2202.09817.pdf}.

\bibitem[Mahabadi et~al.(2021{\natexlab{a}})Mahabadi, Henderson, and
  Ruder]{mahabadi2021compacter}
Rabeeh~Karimi Mahabadi, James Henderson, and Sebastian Ruder.
\newblock Compacter: Efficient low-rank hypercomplex adapter layers.
\newblock \emph{ArXiv preprint}, abs/2106.04647, 2021{\natexlab{a}}.
\newblock URL \url{https://arxiv.org/abs/2106.04647}.

\bibitem[Mahabadi et~al.(2021{\natexlab{b}})Mahabadi, Ruder, Dehghani, and
  Henderson]{Mahabadi2021ParameterefficientMF}
Rabeeh~Karimi Mahabadi, Sebastian Ruder, Mostafa Dehghani, and James Henderson.
\newblock Parameter-efficient multi-task fine-tuning for transformers via
  shared hypernetworks.
\newblock In \emph{ACL/IJCNLP}, 2021{\natexlab{b}}.

\bibitem[Marelli et~al.(2014)Marelli, Menini, Baroni, Bentivogli, Bernardi, and
  Zamparelli]{marelli-etal-2014-sick}
Marco Marelli, Stefano Menini, Marco Baroni, Luisa Bentivogli, Raffaella
  Bernardi, and Roberto Zamparelli.
\newblock A {SICK} cure for the evaluation of compositional distributional
  semantic models.
\newblock In \emph{Proceedings of the Ninth International Conference on
  Language Resources and Evaluation ({LREC}'14)}, pp.\  216--223, Reykjavik,
  Iceland, May 2014. European Language Resources Association (ELRA).
\newblock URL
  \url{http://www.lrec-conf.org/proceedings/lrec2014/pdf/363_Paper.pdf}.

\bibitem[Mathew et~al.(2020)Mathew, Saha, Yimam, Biemann, Goyal, and
  Mukherjee]{mathew2020hatexplain}
Binny Mathew, Punyajoy Saha, Seid~Muhie Yimam, Chris Biemann, Pawan Goyal, and
  Animesh Mukherjee.
\newblock Hatexplain: A benchmark dataset for explainable hate speech
  detection.
\newblock \emph{ArXiv preprint}, abs/2012.10289, 2020.
\newblock URL \url{https://arxiv.org/abs/2012.10289}.

\bibitem[McCreery et~al.(2020)McCreery, Katariya, Kannan, Chablani, and
  Amatriain]{medical-qqp}
Clara~H. McCreery, Namit Katariya, Anitha Kannan, Manish Chablani, and Xavier
  Amatriain.
\newblock Effective transfer learning for identifying similar questions:
  Matching user questions to {COVID-19} faqs.
\newblock In Rajesh Gupta, Yan Liu, Jiliang Tang, and B.~Aditya Prakash (eds.),
  \emph{{KDD} '20: The 26th {ACM} {SIGKDD} Conference on Knowledge Discovery
  and Data Mining, Virtual Event, CA, USA, August 23-27, 2020}, pp.\
  3458--3465. {ACM}, 2020.
\newblock URL \url{https://dl.acm.org/doi/10.1145/3394486.3412861}.

\bibitem[Mihaylov et~al.(2018)Mihaylov, Clark, Khot, and
  Sabharwal]{mihaylov-etal-2018-suit}
Todor Mihaylov, Peter Clark, Tushar Khot, and Ashish Sabharwal.
\newblock Can a suit of armor conduct electricity? a new dataset for open book
  question answering.
\newblock In \emph{Proceedings of the 2018 Conference on Empirical Methods in
  Natural Language Processing}, pp.\  2381--2391, Brussels, Belgium, 2018.
  Association for Computational Linguistics.
\newblock \doi{10.18653/v1/D18-1260}.
\newblock URL \url{https://aclanthology.org/D18-1260}.

\bibitem[Mitchell et~al.(2021)Mitchell, Lin, Bosselut, Finn, and
  Manning]{mitchell2021fast}
Eric Mitchell, Charles Lin, Antoine Bosselut, Chelsea Finn, and Christopher~D
  Manning.
\newblock Fast model editing at scale.
\newblock \emph{arXiv preprint arXiv:2110.11309}, 2021.

\bibitem[Mollas et~al.(2020)Mollas, Chrysopoulou, Karlos, and
  Tsoumakas]{Mollas2020ETHOSAO}
Ioannis Mollas, Zoe Chrysopoulou, Stamatis Karlos, and Grigorios Tsoumakas.
\newblock Ethos: an online hate speech detection dataset.
\newblock \emph{ArXiv preprint}, abs/2006.08328, 2020.
\newblock URL \url{https://arxiv.org/abs/2006.08328}.

\bibitem[Nakano et~al.(2021)Nakano, Hilton, Balaji, Wu, Ouyang, Kim, Hesse,
  Jain, Kosaraju, Saunders, et~al.]{nakano2021webgpt}
Reiichiro Nakano, Jacob Hilton, Suchir Balaji, Jeff Wu, Long Ouyang, Christina
  Kim, Christopher Hesse, Shantanu Jain, Vineet Kosaraju, William Saunders,
  et~al.
\newblock Webgpt: Browser-assisted question-answering with human feedback.
\newblock \emph{arXiv preprint arXiv:2112.09332}, 2021.

\bibitem[Narayan et~al.(2018)Narayan, Cohen, and
  Lapata]{narayan-etal-2018-dont}
Shashi Narayan, Shay~B. Cohen, and Mirella Lapata.
\newblock Don{'}t give me the details, just the summary! topic-aware
  convolutional neural networks for extreme summarization.
\newblock In \emph{Proceedings of the 2018 Conference on Empirical Methods in
  Natural Language Processing}, pp.\  1797--1807, Brussels, Belgium, 2018.
  Association for Computational Linguistics.
\newblock \doi{10.18653/v1/D18-1206}.
\newblock URL \url{https://aclanthology.org/D18-1206}.

\bibitem[Nie et~al.(2020)Nie, Williams, Dinan, Bansal, Weston, and
  Kiela]{nie-etal-2020-adversarial}
Yixin Nie, Adina Williams, Emily Dinan, Mohit Bansal, Jason Weston, and Douwe
  Kiela.
\newblock Adversarial {NLI}: A new benchmark for natural language
  understanding.
\newblock In \emph{Proceedings of the 58th Annual Meeting of the Association
  for Computational Linguistics}, pp.\  4885--4901, Online, 2020. Association
  for Computational Linguistics.
\newblock \doi{10.18653/v1/2020.acl-main.441}.
\newblock URL \url{https://aclanthology.org/2020.acl-main.441}.

\bibitem[Othman \& Jemni(2012)Othman and Jemni]{Othman2012EnglishASLGP}
Achraf Othman and Mohamed Jemni.
\newblock English-asl gloss parallel corpus 2012: Aslg-pc12.
\newblock 2012.
\newblock URL
  \url{https://www.achrafothman.net/aslsmt/English-ASL-Gloss-Parallel-Corpus-2012-ASLG-PC12.pdf}.

\bibitem[Pang \& Lee(2005)Pang and Lee]{pang-lee-2005-seeing}
Bo~Pang and Lillian Lee.
\newblock Seeing stars: Exploiting class relationships for sentiment
  categorization with respect to rating scales.
\newblock In \emph{Proceedings of the 43rd Annual Meeting of the Association
  for Computational Linguistics ({ACL}{'}05)}, pp.\  115--124, Ann Arbor,
  Michigan, 2005. Association for Computational Linguistics.
\newblock \doi{10.3115/1219840.1219855}.
\newblock URL \url{https://aclanthology.org/P05-1015}.

\bibitem[Perez et~al.(2018)Perez, Strub, De~Vries, Dumoulin, and
  Courville]{perez2018film}
Ethan Perez, Florian Strub, Harm De~Vries, Vincent Dumoulin, and Aaron
  Courville.
\newblock Film: Visual reasoning with a general conditioning layer.
\newblock In \emph{Proceedings of the AAAI Conference on Artificial
  Intelligence}, volume~32, 2018.

\bibitem[Petroni et~al.(2019)Petroni, Rockt{\"a}schel, Riedel, Lewis, Bakhtin,
  Wu, and Miller]{petroni-etal-2019-language}
Fabio Petroni, Tim Rockt{\"a}schel, Sebastian Riedel, Patrick Lewis, Anton
  Bakhtin, Yuxiang Wu, and Alexander Miller.
\newblock Language models as knowledge bases?
\newblock In \emph{Proceedings of the 2019 Conference on Empirical Methods in
  Natural Language Processing and the 9th International Joint Conference on
  Natural Language Processing (EMNLP-IJCNLP)}, pp.\  2463--2473, Hong Kong,
  China, 2019. Association for Computational Linguistics.
\newblock \doi{10.18653/v1/D19-1250}.
\newblock URL \url{https://aclanthology.org/D19-1250}.

\bibitem[Petroni et~al.(2020)Petroni, Lewis, Piktus, Rockt{\"a}schel, Wu,
  Miller, and Riedel]{petroni2020how}
Fabio Petroni, Patrick Lewis, Aleksandra Piktus, Tim Rockt{\"a}schel, Yuxiang
  Wu, Alexander~H. Miller, and Sebastian Riedel.
\newblock How context affects language models' factual predictions.
\newblock In \emph{Automated Knowledge Base Construction}, 2020.
\newblock URL \url{https://openreview.net/forum?id=025X0zPfn}.

\bibitem[Pfeiffer et~al.(2020{\natexlab{a}})Pfeiffer, R{\"u}ckl{\'e}, Poth,
  Kamath, Vuli{\'c}, Ruder, Cho, and Gurevych]{pfeiffer2020adapterhub}
Jonas Pfeiffer, Andreas R{\"u}ckl{\'e}, Clifton Poth, Aishwarya Kamath, Ivan
  Vuli{\'c}, Sebastian Ruder, Kyunghyun Cho, and Iryna Gurevych.
\newblock {A}dapter{H}ub: A framework for adapting transformers.
\newblock In \emph{Proceedings of the 2020 Conference on Empirical Methods in
  Natural Language Processing: System Demonstrations}, pp.\  46--54, Online,
  2020{\natexlab{a}}. Association for Computational Linguistics.
\newblock \doi{10.18653/v1/2020.emnlp-demos.7}.
\newblock URL \url{https://aclanthology.org/2020.emnlp-demos.7}.

\bibitem[Pfeiffer et~al.(2020{\natexlab{b}})Pfeiffer, Vuli{\'c}, Gurevych, and
  Ruder]{pfeiffer2020mad}
Jonas Pfeiffer, Ivan Vuli{\'c}, Iryna Gurevych, and Sebastian Ruder.
\newblock {MAD-X}: {A}n {A}dapter-{B}ased {F}ramework for {M}ulti-{T}ask
  {C}ross-{L}ingual {T}ransfer.
\newblock In \emph{Proceedings of the 2020 Conference on Empirical Methods in
  Natural Language Processing (EMNLP)}, pp.\  7654--7673, Online,
  2020{\natexlab{b}}. Association for Computational Linguistics.
\newblock \doi{10.18653/v1/2020.emnlp-main.617}.
\newblock URL \url{https://aclanthology.org/2020.emnlp-main.617}.

\bibitem[Pfeiffer et~al.(2021)Pfeiffer, Kamath, R{\"u}ckl{\'e}, Cho, and
  Gurevych]{pfeiffer2020adapterfusion}
Jonas Pfeiffer, Aishwarya Kamath, Andreas R{\"u}ckl{\'e}, Kyunghyun Cho, and
  Iryna Gurevych.
\newblock {A}dapter{F}usion: Non-destructive task composition for transfer
  learning.
\newblock In \emph{Proceedings of the 16th Conference of the European Chapter
  of the Association for Computational Linguistics: Main Volume}, pp.\
  487--503, Online, 2021. Association for Computational Linguistics.
\newblock URL \url{https://aclanthology.org/2021.eacl-main.39}.

\bibitem[Pilehvar \& Camacho-Collados(2019)Pilehvar and
  Camacho-Collados]{pilehvar-camacho-collados-2019-wic}
Mohammad~Taher Pilehvar and Jose Camacho-Collados.
\newblock {W}i{C}: the word-in-context dataset for evaluating context-sensitive
  meaning representations.
\newblock In \emph{Proceedings of the 2019 Conference of the North {A}merican
  Chapter of the Association for Computational Linguistics: Human Language
  Technologies, Volume 1 (Long and Short Papers)}, pp.\  1267--1273,
  Minneapolis, Minnesota, 2019. Association for Computational Linguistics.
\newblock \doi{10.18653/v1/N19-1128}.
\newblock URL \url{https://aclanthology.org/N19-1128}.

\bibitem[Pouran Ben~Veyseh et~al.(2020)Pouran Ben~Veyseh, Dernoncourt, Tran,
  and Nguyen]{pouran-ben-veyseh-etal-2020-acronym}
Amir Pouran Ben~Veyseh, Franck Dernoncourt, Quan~Hung Tran, and Thien~Huu
  Nguyen.
\newblock What does this acronym mean? introducing a new dataset for acronym
  identification and disambiguation.
\newblock In \emph{Proceedings of the 28th International Conference on
  Computational Linguistics}, pp.\  3285--3301, Barcelona, Spain (Online),
  2020. International Committee on Computational Linguistics.
\newblock \doi{10.18653/v1/2020.coling-main.292}.
\newblock URL \url{https://aclanthology.org/2020.coling-main.292}.

\bibitem[Qin et~al.(2021{\natexlab{a}})Qin, Lin, Yi, Zhang, Han, Zhang, Su,
  Liu, Li, Sun, et~al.]{qin2021knowledge}
Yujia Qin, Yankai Lin, Jing Yi, Jiajie Zhang, Xu~Han, Zhengyan Zhang, Yusheng
  Su, Zhiyuan Liu, Peng Li, Maosong Sun, et~al.
\newblock Knowledge inheritance for pre-trained language models.
\newblock \emph{arXiv preprint arXiv:2105.13880}, 2021{\natexlab{a}}.

\bibitem[Qin et~al.(2021{\natexlab{b}})Qin, Wang, Su, Lin, Ding, Liu, Li, Hou,
  Li, Sun, et~al.]{qin2021exploring}
Yujia Qin, Xiaozhi Wang, Yusheng Su, Yankai Lin, Ning Ding, Zhiyuan Liu, Juanzi
  Li, Lei Hou, Peng Li, Maosong Sun, et~al.
\newblock Exploring low-dimensional intrinsic task subspace via prompt tuning.
\newblock \emph{ArXiv preprint}, abs/2110.07867, 2021{\natexlab{b}}.
\newblock URL \url{https://arxiv.org/abs/2110.07867}.

\bibitem[Qin et~al.(2021{\natexlab{c}})Qin, Zhang, Lin, Liu, Li, Sun, and
  Zhou]{anony2021lifelong}
Yujia Qin, Jiajie Zhang, Yankai Lin, Zhiyuan Liu, Peng Li, Maosong Sun, and Jie
  Zhou.
\newblock Elle: Efficient lifelong pre-training for emerging data.
\newblock \emph{OpenReview preprint}, 2021{\natexlab{c}}.
\newblock URL \url{https://openreview.net/forum?id=UF7a5kIdzk}.

\bibitem[Radford et~al.(2018)Radford, Narasimhan, Salimans, and
  Sutskever]{radford2018improving}
Alec Radford, Karthik Narasimhan, Tim Salimans, and Ilya Sutskever.
\newblock Improving language understanding by generative pre-training.
\newblock 2018.

\bibitem[Radford et~al.(2019)Radford, Wu, Child, Luan, Amodei, Sutskever,
  et~al.]{radford2019language}
Alec Radford, Jeffrey Wu, Rewon Child, David Luan, Dario Amodei, Ilya
  Sutskever, et~al.
\newblock Language models are unsupervised multitask learners.
\newblock \emph{OpenAI blog}, 1\penalty0 (8):\penalty0 9, 2019.

\bibitem[Raffel et~al.(2019)Raffel, Shazeer, Roberts, Lee, Narang, Matena,
  Zhou, Li, and Liu]{raffel2019exploring}
Colin Raffel, Noam Shazeer, Adam Roberts, Katherine Lee, Sharan Narang, Michael
  Matena, Yanqi Zhou, Wei Li, and Peter~J Liu.
\newblock Exploring the limits of transfer learning with a unified text-to-text
  transformer.
\newblock \emph{ArXiv preprint}, abs/1910.10683, 2019.
\newblock URL \url{https://arxiv.org/abs/1910.10683}.

\bibitem[Rajani et~al.(2019)Rajani, McCann, Xiong, and
  Socher]{rajani-etal-2019-explain}
Nazneen~Fatema Rajani, Bryan McCann, Caiming Xiong, and Richard Socher.
\newblock Explain yourself! leveraging language models for commonsense
  reasoning.
\newblock In \emph{Proceedings of the 57th Annual Meeting of the Association
  for Computational Linguistics}, pp.\  4932--4942, Florence, Italy, 2019.
  Association for Computational Linguistics.
\newblock \doi{10.18653/v1/P19-1487}.
\newblock URL \url{https://aclanthology.org/P19-1487}.

\bibitem[Rajpurkar et~al.(2016)Rajpurkar, Zhang, Lopyrev, and
  Liang]{rajpurkar-etal-2016-squad}
Pranav Rajpurkar, Jian Zhang, Konstantin Lopyrev, and Percy Liang.
\newblock {SQ}u{AD}: 100,000+ questions for machine comprehension of text.
\newblock In \emph{Proceedings of the 2016 Conference on Empirical Methods in
  Natural Language Processing}, pp.\  2383--2392, Austin, Texas, 2016.
  Association for Computational Linguistics.
\newblock \doi{10.18653/v1/D16-1264}.
\newblock URL \url{https://aclanthology.org/D16-1264}.

\bibitem[Rebuffi et~al.(2017)Rebuffi, Bilen, and Vedaldi]{rebuffi2017learning}
Sylvestre-Alvise Rebuffi, Hakan Bilen, and Andrea Vedaldi.
\newblock Learning multiple visual domains with residual adapters.
\newblock \emph{Advances in neural information processing systems}, 30, 2017.

\bibitem[R{\"u}ckl{\'e} et~al.(2021)R{\"u}ckl{\'e}, Geigle, Glockner, Beck,
  Pfeiffer, Reimers, and Gurevych]{ruckle-etal-2021-adapterdrop}
Andreas R{\"u}ckl{\'e}, Gregor Geigle, Max Glockner, Tilman Beck, Jonas
  Pfeiffer, Nils Reimers, and Iryna Gurevych.
\newblock {AdapterDrop}: {O}n the efficiency of adapters in transformers.
\newblock In \emph{Proceedings of EMNLP}, pp.\  7930--7946, 2021.
\newblock URL \url{https://aclanthology.org/2021.emnlp-main.626}.

\bibitem[Saha et~al.(2018)Saha, Aralikatte, Khapra, and
  Sankaranarayanan]{saha-etal-2018-duorc}
Amrita Saha, Rahul Aralikatte, Mitesh~M. Khapra, and Karthik Sankaranarayanan.
\newblock {D}uo{RC}: Towards complex language understanding with paraphrased
  reading comprehension.
\newblock In \emph{Proceedings of the 56th Annual Meeting of the Association
  for Computational Linguistics (Volume 1: Long Papers)}, pp.\  1683--1693,
  Melbourne, Australia, 2018. Association for Computational Linguistics.
\newblock \doi{10.18653/v1/P18-1156}.
\newblock URL \url{https://aclanthology.org/P18-1156}.

\bibitem[Sakaguchi et~al.(2020)Sakaguchi, Bras, Bhagavatula, and
  Choi]{Sakaguchi_Le_Bras_Bhagavatula_Choi_2020}
Keisuke Sakaguchi, Ronan~Le Bras, Chandra Bhagavatula, and Yejin Choi.
\newblock Winogrande: An adversarial winograd schema challenge at scale.
\newblock In \emph{The Thirty-Fourth {AAAI} Conference on Artificial
  Intelligence, {AAAI} 2020, The Thirty-Second Innovative Applications of
  Artificial Intelligence Conference, {IAAI} 2020, The Tenth {AAAI} Symposium
  on Educational Advances in Artificial Intelligence, {EAAI} 2020, New York,
  NY, USA, February 7-12, 2020}, pp.\  8732--8740. {AAAI} Press, 2020.
\newblock URL \url{https://aaai.org/ojs/index.php/AAAI/article/view/6399}.

\bibitem[Scao \& Rush(2021)Scao and Rush]{Scao2021HowMD}
Teven~Le Scao and Alexander~M. Rush.
\newblock How many data points is a prompt worth?
\newblock In \emph{NAACL}, 2021.

\bibitem[Schick \& Sch{\"u}tze(2021)Schick and
  Sch{\"u}tze]{schick-schutze-2021-exploiting}
Timo Schick and Hinrich Sch{\"u}tze.
\newblock Exploiting cloze-questions for few-shot text classification and
  natural language inference.
\newblock In \emph{Proceedings of the 16th Conference of the European Chapter
  of the Association for Computational Linguistics: Main Volume}, pp.\
  255--269, Online, 2021. Association for Computational Linguistics.
\newblock URL \url{https://aclanthology.org/2021.eacl-main.20}.

\bibitem[Sharma et~al.(2019)Sharma, Graesser, Nangia, and
  Evci]{sharma2019natural}
Lakshay Sharma, Laura Graesser, Nikita Nangia, and Utku Evci.
\newblock Natural language understanding with the quora question pairs dataset.
\newblock \emph{arXiv e-prints}, 2019.
\newblock URL \url{https://arxiv.org/abs/1907.01041}.

\bibitem[Shazeer \& Stern(2018)Shazeer and Stern]{pmlr-v80-shazeer18a}
Noam Shazeer and Mitchell Stern.
\newblock Adafactor: Adaptive learning rates with sublinear memory cost.
\newblock In Jennifer~G. Dy and Andreas Krause (eds.), \emph{Proceedings of the
  35th International Conference on Machine Learning, {ICML} 2018,
  Stockholmsm{\"{a}}ssan, Stockholm, Sweden, July 10-15, 2018}, volume~80 of
  \emph{Proceedings of Machine Learning Research}, pp.\  4603--4611. {PMLR},
  2018.
\newblock URL \url{http://proceedings.mlr.press/v80/shazeer18a.html}.

\bibitem[Sileo et~al.(2019)Sileo, Van De~Cruys, Pradel, and
  Muller]{sileo-etal-2019-mining}
Damien Sileo, Tim Van De~Cruys, Camille Pradel, and Philippe Muller.
\newblock Mining discourse markers for unsupervised sentence representation
  learning.
\newblock In \emph{Proceedings of the 2019 Conference of the North {A}merican
  Chapter of the Association for Computational Linguistics: Human Language
  Technologies, Volume 1 (Long and Short Papers)}, pp.\  3477--3486,
  Minneapolis, Minnesota, 2019. Association for Computational Linguistics.
\newblock \doi{10.18653/v1/N19-1351}.
\newblock URL \url{https://aclanthology.org/N19-1351}.

\bibitem[Socher et~al.(2013)Socher, Perelygin, Wu, Chuang, Manning, Ng, and
  Potts]{socher-etal-2013-recursive}
Richard Socher, Alex Perelygin, Jean Wu, Jason Chuang, Christopher~D. Manning,
  Andrew Ng, and Christopher Potts.
\newblock Recursive deep models for semantic compositionality over a sentiment
  treebank.
\newblock In \emph{Proceedings of the 2013 Conference on Empirical Methods in
  Natural Language Processing}, pp.\  1631--1642, Seattle, Washington, USA,
  2013. Association for Computational Linguistics.
\newblock URL \url{https://aclanthology.org/D13-1170}.

\bibitem[Stickland \& Murray(2019)Stickland and Murray]{stickland2019bert}
Asa~Cooper Stickland and Iain Murray.
\newblock {BERT} and pals: Projected attention layers for efficient adaptation
  in multi-task learning.
\newblock In Kamalika Chaudhuri and Ruslan Salakhutdinov (eds.),
  \emph{Proceedings of the 36th International Conference on Machine Learning,
  {ICML} 2019, 9-15 June 2019, Long Beach, California, {USA}}, volume~97 of
  \emph{Proceedings of Machine Learning Research}, pp.\  5986--5995. {PMLR},
  2019.
\newblock URL \url{http://proceedings.mlr.press/v97/stickland19a.html}.

\bibitem[Su et~al.(2021)Su, Wang, Qin, Chan, Lin, Liu, Li, Li, Hou, Sun, and
  Zhou]{su2021transferability}
Yusheng Su, Xiaozhi Wang, Yujia Qin, Chi-Min Chan, Yankai Lin, Zhiyuan Liu,
  Peng Li, Juanzi Li, Lei Hou, Maosong Sun, and Jie Zhou.
\newblock On transferability of prompt tuning for natural language
  understanding.
\newblock \emph{ArXiv preprint}, abs/2111.06719, 2021.
\newblock URL \url{https://arxiv.org/abs/2111.06719}.

\bibitem[Sun et~al.(2021)Sun, Liu, Qiu, and Huang]{sun2021paradigm}
Tianxiang Sun, Xiangyang Liu, Xipeng Qiu, and Xuanjing Huang.
\newblock Paradigm shift in natural language processing.
\newblock \emph{ArXiv preprint}, abs/2109.12575, 2021.
\newblock URL \url{https://arxiv.org/abs/2109.12575}.

\bibitem[Sun et~al.(2022)Sun, Shao, Qian, Huang, and Qiu]{sun2022black}
Tianxiang Sun, Yunfan Shao, Hong Qian, Xuanjing Huang, and Xipeng Qiu.
\newblock Black-box tuning for language-model-as-a-service.
\newblock \emph{arXiv preprint arXiv:2201.03514}, 2022.
\newblock URL \url{https://arxiv.org/abs/2201.03514}.

\bibitem[Tafjord et~al.(2019{\natexlab{a}})Tafjord, Clark, Gardner, Yih, and
  Sabharwal]{Tafjord_Clark_Gardner_Yih_Sabharwal_2019}
Oyvind Tafjord, Peter Clark, Matt Gardner, Wen-tau Yih, and Ashish Sabharwal.
\newblock Quarel: A dataset and models for answering questions about
  qualitative relationships.
\newblock \emph{Proceedings of the AAAI Conference on Artificial Intelligence},
  33\penalty0 (01):\penalty0 7063--7071, Jul. 2019{\natexlab{a}}.
\newblock \doi{10.1609/aaai.v33i01.33017063}.
\newblock URL \url{https://ojs.aaai.org/index.php/AAAI/article/view/4687}.

\bibitem[Tafjord et~al.(2019{\natexlab{b}})Tafjord, Gardner, Lin, and
  Clark]{tafjord-etal-2019-quartz}
Oyvind Tafjord, Matt Gardner, Kevin Lin, and Peter Clark.
\newblock {Q}ua{RT}z: An open-domain dataset of qualitative relationship
  questions.
\newblock In \emph{Proceedings of the 2019 Conference on Empirical Methods in
  Natural Language Processing and the 9th International Joint Conference on
  Natural Language Processing (EMNLP-IJCNLP)}, pp.\  5941--5946, Hong Kong,
  China, 2019{\natexlab{b}}. Association for Computational Linguistics.
\newblock \doi{10.18653/v1/D19-1608}.
\newblock URL \url{https://aclanthology.org/D19-1608}.

\bibitem[Tan et~al.(2021)Tan, Zhang, Wang, and Liu]{tan2021msp}
Zhixing Tan, Xiangwen Zhang, Shuo Wang, and Yang Liu.
\newblock {MSP}: Multi-stage prompting for making pre-trained language models
  better translators.
\newblock \emph{arXiv preprint arXiv:2110.06609}, 2021.

\bibitem[Vaswani et~al.(2017)Vaswani, Shazeer, Parmar, Uszkoreit, Jones, Gomez,
  Kaiser, and Polosukhin]{vaswani2017attention}
Ashish Vaswani, Noam Shazeer, Niki Parmar, Jakob Uszkoreit, Llion Jones,
  Aidan~N. Gomez, Lukasz Kaiser, and Illia Polosukhin.
\newblock Attention is all you need.
\newblock In Isabelle Guyon, Ulrike von Luxburg, Samy Bengio, Hanna~M. Wallach,
  Rob Fergus, S.~V.~N. Vishwanathan, and Roman Garnett (eds.), \emph{Advances
  in Neural Information Processing Systems 30: Annual Conference on Neural
  Information Processing Systems 2017, December 4-9, 2017, Long Beach, CA,
  {USA}}, pp.\  5998--6008, 2017.
\newblock URL
  \url{https://proceedings.neurips.cc/paper/2017/hash/3f5ee243547dee91fbd053c1c4a845aa-Abstract.html}.

\bibitem[Vu et~al.(2021)Vu, Lester, Constant, Al-Rfou, and Cer]{vu2021spot}
Tu~Vu, Brian Lester, Noah Constant, Rami Al-Rfou, and Daniel Cer.
\newblock Spot: Better frozen model adaptation through soft prompt transfer.
\newblock \emph{ArXiv preprint}, abs/2110.07904, 2021.
\newblock URL \url{https://arxiv.org/abs/2110.07904}.

\bibitem[Wang et~al.(2019)Wang, Singh, Michael, Hill, Levy, and
  Bowman]{wang2018glue}
Alex Wang, Amanpreet Singh, Julian Michael, Felix Hill, Omer Levy, and
  Samuel~R. Bowman.
\newblock {GLUE:} {A} multi-task benchmark and analysis platform for natural
  language understanding.
\newblock In \emph{7th International Conference on Learning Representations,
  {ICLR} 2019, New Orleans, LA, USA, May 6-9, 2019}. OpenReview.net, 2019.
\newblock URL \url{https://openreview.net/forum?id=rJ4km2R5t7}.

\bibitem[Wang(2017)]{wang-2017-liar}
William~Yang Wang.
\newblock {``}liar, liar pants on fire{''}: A new benchmark dataset for fake
  news detection.
\newblock In \emph{Proceedings of the 55th Annual Meeting of the Association
  for Computational Linguistics (Volume 2: Short Papers)}, pp.\  422--426,
  Vancouver, Canada, 2017. Association for Computational Linguistics.
\newblock \doi{10.18653/v1/P17-2067}.
\newblock URL \url{https://aclanthology.org/P17-2067}.

\bibitem[Warstadt et~al.(2019)Warstadt, Singh, and
  Bowman]{warstadt-etal-2019-neural}
Alex Warstadt, Amanpreet Singh, and Samuel~R. Bowman.
\newblock Neural network acceptability judgments.
\newblock \emph{Transactions of the Association for Computational Linguistics},
  7:\penalty0 625--641, 2019.
\newblock \doi{10.1162/tacl_a_00290}.
\newblock URL \url{https://aclanthology.org/Q19-1040}.

\bibitem[Warstadt et~al.(2020)Warstadt, Parrish, Liu, Mohananey, Peng, Wang,
  and Bowman]{warstadt2019blimp}
Alex Warstadt, Alicia Parrish, Haokun Liu, Anhad Mohananey, Wei Peng, Sheng-Fu
  Wang, and Samuel~R. Bowman.
\newblock {BL}i{MP}: The benchmark of linguistic minimal pairs for {E}nglish.
\newblock \emph{Transactions of the Association for Computational Linguistics},
  8:\penalty0 377--392, 2020.
\newblock \doi{10.1162/tacl_a_00321}.
\newblock URL \url{https://aclanthology.org/2020.tacl-1.25}.

\bibitem[Weidinger et~al.(2021)Weidinger, Mellor, Rauh, Griffin, Uesato, Huang,
  Cheng, Glaese, Balle, Kasirzadeh, et~al.]{weidinger2021ethical}
Laura Weidinger, John Mellor, Maribeth Rauh, Conor Griffin, Jonathan Uesato,
  Po-Sen Huang, Myra Cheng, Mia Glaese, Borja Balle, Atoosa Kasirzadeh, et~al.
\newblock Ethical and social risks of harm from language models.
\newblock \emph{arXiv preprint arXiv:2112.04359}, 2021.

\bibitem[Williams et~al.(2018)Williams, Nangia, and
  Bowman]{williams-etal-2018-broad}
Adina Williams, Nikita Nangia, and Samuel Bowman.
\newblock A broad-coverage challenge corpus for sentence understanding through
  inference.
\newblock In \emph{Proceedings of the 2018 Conference of the North {A}merican
  Chapter of the Association for Computational Linguistics: Human Language
  Technologies, Volume 1 (Long Papers)}, pp.\  1112--1122, New Orleans,
  Louisiana, 2018. Association for Computational Linguistics.
\newblock \doi{10.18653/v1/N18-1101}.
\newblock URL \url{https://aclanthology.org/N18-1101}.

\bibitem[Wright \& Ma(2021)Wright and Ma]{wright2021high}
John Wright and Yi~Ma.
\newblock High-dimensional data analysis with low-dimensional models:
  Principles, computation, and applications.
\newblock 2021.

\bibitem[Yang et~al.(2015)Yang, Yih, and Meek]{yang-etal-2015-wikiqa}
Yi~Yang, Wen-tau Yih, and Christopher Meek.
\newblock {W}iki{QA}: A challenge dataset for open-domain question answering.
\newblock In \emph{Proceedings of the 2015 Conference on Empirical Methods in
  Natural Language Processing}, pp.\  2013--2018, Lisbon, Portugal, 2015.
  Association for Computational Linguistics.
\newblock \doi{10.18653/v1/D15-1237}.
\newblock URL \url{https://aclanthology.org/D15-1237}.

\bibitem[Yang \& Liu(2022)Yang and Liu]{yang2022on}
Zonghan Yang and Yang Liu.
\newblock {On Robust Prefix-Tuning for Text Classification}.
\newblock In \emph{International Conference on Learning Representations}, 2022.
\newblock URL \url{https://openreview.net/forum?id=eBCmOocUejf}.

\bibitem[Ye et~al.(2021)Ye, Lin, and Ren]{ye2021crossfit}
Qinyuan Ye, Bill~Yuchen Lin, and Xiang Ren.
\newblock {C}ross{F}it: A few-shot learning challenge for cross-task
  generalization in {NLP}.
\newblock In \emph{Proceedings of the 2021 Conference on Empirical Methods in
  Natural Language Processing}, pp.\  7163--7189, Online and Punta Cana,
  Dominican Republic, 2021. Association for Computational Linguistics.
\newblock URL \url{https://aclanthology.org/2021.emnlp-main.572}.

\bibitem[Ye et~al.(2020)Ye, Wu, Zhou, Yang, Tan, Xu, Song, Bao, and
  Ma]{ye2020light}
Shaokai Ye, Kailu Wu, Mu~Zhou, Yunfei Yang, Sia~Huat Tan, Kaidi Xu, Jiebo Song,
  Chenglong Bao, and Kaisheng Ma.
\newblock Light-weight calibrator: {A} separable component for unsupervised
  domain adaptation.
\newblock In \emph{2020 {IEEE/CVF} Conference on Computer Vision and Pattern
  Recognition, {CVPR} 2020, Seattle, WA, USA, June 13-19, 2020}, pp.\
  13733--13742. {IEEE}, 2020.
\newblock \doi{10.1109/CVPR42600.2020.01375}.
\newblock URL \url{https://doi.org/10.1109/CVPR42600.2020.01375}.

\bibitem[Yu et~al.(2018)Yu, Zhang, Yang, Yasunaga, Wang, Li, Ma, Li, Yao,
  Roman, Zhang, and Radev]{yu-etal-2018-spider}
Tao Yu, Rui Zhang, Kai Yang, Michihiro Yasunaga, Dongxu Wang, Zifan Li, James
  Ma, Irene Li, Qingning Yao, Shanelle Roman, Zilin Zhang, and Dragomir Radev.
\newblock {S}pider: A large-scale human-labeled dataset for complex and
  cross-domain semantic parsing and text-to-{SQL} task.
\newblock In \emph{Proceedings of the 2018 Conference on Empirical Methods in
  Natural Language Processing}, pp.\  3911--3921, Brussels, Belgium, 2018.
  Association for Computational Linguistics.
\newblock \doi{10.18653/v1/D18-1425}.
\newblock URL \url{https://aclanthology.org/D18-1425}.

\bibitem[Zaken et~al.(2021)Zaken, Ravfogel, and Goldberg]{zaken2021bitfit}
Elad~Ben Zaken, Shauli Ravfogel, and Yoav Goldberg.
\newblock Bitfit: Simple parameter-efficient fine-tuning for transformer-based
  masked language-models.
\newblock \emph{ArXiv preprint}, abs/2106.10199, 2021.
\newblock URL \url{https://arxiv.org/abs/2106.10199}.

\bibitem[Zellers et~al.(2018)Zellers, Bisk, Schwartz, and
  Choi]{zellers-etal-2018-swag}
Rowan Zellers, Yonatan Bisk, Roy Schwartz, and Yejin Choi.
\newblock {SWAG}: A large-scale adversarial dataset for grounded commonsense
  inference.
\newblock In \emph{Proceedings of the 2018 Conference on Empirical Methods in
  Natural Language Processing}, pp.\  93--104, Brussels, Belgium, 2018.
  Association for Computational Linguistics.
\newblock \doi{10.18653/v1/D18-1009}.
\newblock URL \url{https://aclanthology.org/D18-1009}.

\bibitem[Zellers et~al.(2019)Zellers, Holtzman, Bisk, Farhadi, and
  Choi]{zellers-etal-2019-hellaswag}
Rowan Zellers, Ari Holtzman, Yonatan Bisk, Ali Farhadi, and Yejin Choi.
\newblock {H}ella{S}wag: Can a machine really finish your sentence?
\newblock In \emph{Proceedings of the 57th Annual Meeting of the Association
  for Computational Linguistics}, pp.\  4791--4800, Florence, Italy, 2019.
  Association for Computational Linguistics.
\newblock \doi{10.18653/v1/P19-1472}.
\newblock URL \url{https://aclanthology.org/P19-1472}.

\bibitem[Zhang et~al.(2019{\natexlab{a}})Zhang, Zhang, Lu, Zhu, and
  Dong]{Control_Theory_NEURIPS2019}
Dinghuai Zhang, Tianyuan Zhang, Yiping Lu, Zhanxing Zhu, and Bin Dong.
\newblock You only propagate once: Accelerating adversarial training via
  maximal principle.
\newblock In H.~Wallach, H.~Larochelle, A.~Beygelzimer, F.~d\textquotesingle
  Alch\'{e}-Buc, E.~Fox, and R.~Garnett (eds.), \emph{Advances in Neural
  Information Processing Systems}, volume~32. Curran Associates, Inc.,
  2019{\natexlab{a}}.
\newblock URL
  \url{https://proceedings.neurips.cc/paper/2019/file/812b4ba287f5ee0bc9d43bbf5bbe87fb-Paper.pdf}.

\bibitem[Zhang et~al.(2020)Zhang, Ro, and Sproat]{zhang-etal-2020-semi}
Hao Zhang, Jae Ro, and Richard Sproat.
\newblock Semi-supervised {URL} segmentation with recurrent neural networks
  pre-trained on knowledge graph entities.
\newblock In \emph{Proceedings of the 28th International Conference on
  Computational Linguistics}, pp.\  4667--4675, Barcelona, Spain (Online),
  2020. International Committee on Computational Linguistics.
\newblock \doi{10.18653/v1/2020.coling-main.411}.
\newblock URL \url{https://aclanthology.org/2020.coling-main.411}.

\bibitem[Zhang et~al.(2018)Zhang, Liu, Liu, Gao, Duh, and
  Durme]{Zhang2018ReCoRDBT}
Sheng Zhang, X.~Liu, J.~Liu, Jianfeng Gao, Kevin Duh, and Benjamin~Van Durme.
\newblock Record: Bridging the gap between human and machine commonsense
  reading comprehension.
\newblock \emph{ArXiv preprint}, abs/1810.12885, 2018.
\newblock URL \url{https://arxiv.org/abs/1810.12885}.

\bibitem[Zhang et~al.(2019{\natexlab{b}})Zhang, Baldridge, and
  He]{zhang-etal-2019-paws}
Yuan Zhang, Jason Baldridge, and Luheng He.
\newblock {PAWS}: Paraphrase adversaries from word scrambling.
\newblock In \emph{Proceedings of the 2019 Conference of the North {A}merican
  Chapter of the Association for Computational Linguistics: Human Language
  Technologies, Volume 1 (Long and Short Papers)}, pp.\  1298--1308,
  Minneapolis, Minnesota, 2019{\natexlab{b}}. Association for Computational
  Linguistics.
\newblock \doi{10.18653/v1/N19-1131}.
\newblock URL \url{https://aclanthology.org/N19-1131}.

\bibitem[Zhao et~al.(2020)Zhao, Lin, Mi, Jaggi, and
  Sch{\"u}tze]{zhao2020masking}
Mengjie Zhao, Tao Lin, Fei Mi, Martin Jaggi, and Hinrich Sch{\"u}tze.
\newblock Masking as an efficient alternative to finetuning for pretrained
  language models.
\newblock In \emph{Proceedings of the 2020 Conference on Empirical Methods in
  Natural Language Processing (EMNLP)}, pp.\  2226--2241, Online, 2020.
  Association for Computational Linguistics.
\newblock \doi{10.18653/v1/2020.emnlp-main.174}.
\newblock URL \url{https://aclanthology.org/2020.emnlp-main.174}.

\end{thebibliography}

\newpage
\appendix
\section{Implementation Details}
\label{sec:details}
\subsection{Performance and Convergence}
\label{sec:performance details}
Among the NLP datasets downloaded from Hugginface datasets, for those datasets without publicly released test set, we evenly divide the original development sets into two halves as the new development set and test set; for those datasets without publicly released development set and test set, we divide the original training set with a ratio of $8:1:1$ into the new training set, development set and test set.

For \textbf{PF}, \textbf{LR}, \textbf{AP} and \textbf{FT}, we use AdamW~\citep{kingma2014adam} as the optimizer, set the maximum training steps to $20,000$ with early stop, and save the checkpoint for evaluation on development set every $100$ steps. After that, we evaluate the best checkpoint using the development set on the test set. We experiment on the combinations of different batch sizes ($\{16, 32\}$) and learning rates ($\{1\times10^{-3}, 1\times10^{-4}, 5\times10^{-4}\}$), and report the best performance. Since we found empirically that \textbf{PT} converges much slower than the other tuning methods, we set the maximum training step of \textbf{PT} to $100,000$ steps without early stop, and evaluate the performance on development set for every $1,000$ steps. Following \citet{lester2021power}, we choose Adafactor~\citep{pmlr-v80-shazeer18a} as the optimizer. All the experiments are conducted under the same environment.

\subsection{Combinations of Delta Tuning Methods}
\label{sec:combination details}
For prompt tuning, we prepend $10$ tunable virtual tokens into the input text; for adapter, we set the reduction factor to $16$; for BitFit, all the bias components in PLMs are optimized.

\paragraph{Simultaneous Combination.} For all delta tuning methods on $\text{RoBERTa}_{\texttt{LARGE}}$, we choose AdamW~\citep{kingma2014adam} as the optimizer, set the maximum training steps to $6,000$, and save the checkpoint for evaluation on development set every $200$ steps. After that, we select the best checkpoint based on the development set, and evaluate it on the test set. For the full-data setting, we set the training batch size to $16$ and experiment on the combinations of different learning rates ($\{1\times 10^{-2}, 1\times10^{-3}, 1\times10^{-4}, 1\times10^{-5}\}$); for the few-shot setting, we set the training batch size to $4$ and experiment on the combination of $4$ different learning rates, which are listed in Table~\ref{tab:16_lr}.

For STS-B, which is a regression task, we convert it into a binary classification problem. Specifically, assume that the original output value is bounded by $[v_l, v_u]$ and the new labels are $\{y_l, y_u\}$, the original value is reformulated as:
\[
y = v_l \cdot p(y_l|x_{in}) + v_u \cdot p(y_u|x_{in}).
\]
During optimization, we minimize the KL-divergence between prediction distribution $p(y_u|x_{\text{in}})$ and the 
ground truth $(y-v_l)/(v_u-v_l)$.

\paragraph{Sequential Combination.} We choose AdamW~\citep{kingma2014adam} as the optimizer, set the batch size to $64$ and the learning rate to $1\times 10^{-2}$ for prompt tuning, $1\times 10^{-4}$ for BitFit and $1\times 10^{-5}$ for adapter.

  \begin{table}[thbp]
  \centering
  \caption{Learning rate setting of $\text{RoBERTa}_{\texttt{LARGE}}$ on 16-shot GLUE datasets.}
    \begin{tabu}{l|r r r r r r r r}
    \toprule
    \textbf{Prompt tuning} & \multicolumn{1}{c|}{\color{BrickRed}{\XSolidBrush}} & \multicolumn{1}{c|}{\color{BrickRed}{\XSolidBrush}} & \multicolumn{1}{c|}{\color{BrickRed}{\XSolidBrush}} & \multicolumn{1}{c|}{\color{BrickRed}{\XSolidBrush}} & \multicolumn{1}{c|}{\color{ForestGreen}{\Checkmark}} & \multicolumn{1}{c|}{\color{ForestGreen}{\Checkmark}} & \multicolumn{1}{c|}{\color{ForestGreen}{\Checkmark}} & \multicolumn{1}{c}{\color{ForestGreen}{\Checkmark}} \\
    
    \textbf{BitFit} & \multicolumn{1}{c|}{\color{BrickRed}{\XSolidBrush}} & \multicolumn{1}{c|}{\color{BrickRed}{\XSolidBrush}} &
    \multicolumn{1}{c|}{\color{ForestGreen}{\Checkmark}} & \multicolumn{1}{c|}{\color{ForestGreen}{\Checkmark}} &
    \multicolumn{1}{c|}{\color{BrickRed}{\XSolidBrush}} & \multicolumn{1}{c|}{\color{BrickRed}{\XSolidBrush}} &  \multicolumn{1}{c|}{\color{ForestGreen}{\Checkmark}} & \multicolumn{1}{c}{\color{ForestGreen}{\Checkmark}} \\
    
    \textbf{Adapter} & \multicolumn{1}{c|}{\color{BrickRed}{\XSolidBrush}} & \multicolumn{1}{c|}{\color{ForestGreen}{\Checkmark}} & \multicolumn{1}{c|}{\color{BrickRed}{\XSolidBrush}} & \multicolumn{1}{c|}{\color{ForestGreen}{\Checkmark}} & \multicolumn{1}{c|}{\color{BrickRed}{\XSolidBrush}} & \multicolumn{1}{c|}{\color{ForestGreen}{\Checkmark}} & \multicolumn{1}{c|}{\color{BrickRed}{\XSolidBrush}} & \multicolumn{1}{c}{\color{ForestGreen}{\Checkmark}} \\ \midrule 
    \multicolumn{9}{l}{Learning Rates} \\
    \midrule
    1e-2 &- &\color{BrickRed}{\XSolidBrush} & \color{BrickRed}{\XSolidBrush}& \color{BrickRed}{\XSolidBrush}& \color{ForestGreen}{\Checkmark}& \color{BrickRed}{\XSolidBrush}& \color{BrickRed}{\XSolidBrush}&\color{BrickRed}{\XSolidBrush} \\
    3e-3 & -& \color{BrickRed}{\XSolidBrush}& \color{ForestGreen}{\Checkmark}&\color{BrickRed}{\XSolidBrush} &\color{ForestGreen}{\Checkmark} &\color{BrickRed}{\XSolidBrush} &\color{BrickRed}{\XSolidBrush} &\color{BrickRed}{\XSolidBrush} \\
    1e-3 & -&\color{ForestGreen}{\Checkmark} & \color{ForestGreen}{\Checkmark}&\color{ForestGreen}{\Checkmark} & \color{ForestGreen}{\Checkmark}& \color{ForestGreen}{\Checkmark}& \color{ForestGreen}{\Checkmark}& \color{ForestGreen}{\Checkmark}\\
    3e-4 & -&\color{ForestGreen}{\Checkmark} &\color{ForestGreen}{\Checkmark} & \color{ForestGreen}{\Checkmark}& \color{ForestGreen}{\Checkmark}&\color{ForestGreen}{\Checkmark} &\color{ForestGreen}{\Checkmark} &\color{ForestGreen}{\Checkmark} \\
    1e-4 & -&\color{ForestGreen}{\Checkmark} &\color{ForestGreen}{\Checkmark} & \color{ForestGreen}{\Checkmark}&\color{BrickRed}{\XSolidBrush} & \color{ForestGreen}{\Checkmark}&\color{ForestGreen}{\Checkmark} & \color{ForestGreen}{\Checkmark}\\
    3e-5 & -&\color{ForestGreen}{\Checkmark} &\color{BrickRed}{\XSolidBrush} &\color{ForestGreen}{\Checkmark}  &\color{BrickRed}{\XSolidBrush} &\color{ForestGreen}{\Checkmark} &\color{ForestGreen}{\Checkmark} & \color{ForestGreen}{\Checkmark}\\
    \midrule
    \end{tabu}
    \label{tab:16_lr}
  \end{table}

\subsection{The Power of Scale for Delta Tuning}
\label{sec:Scale Details}

  \begin{table}[thbp]
  \small
      \centering
      \caption{The percentages of the tuned parameters (parameters participating optimizing in a PLM / all parameters in a PLM) during the training.}
      \begin{tabu}{r r r r}
           \toprule
           \textbf{} & \textbf{SMALL} & \textbf{BASE} & \textbf{XXL} \\
           \midrule
            \textbf{Adapter} &  1.70\%&  1.20\%& 0.28\%\\
            \textbf{LoRA} &  0.73\% &  0.64\% & 0.26\% \\
            \textbf{Prefix-tuning} &  0.50\% &  0.47\% & 0.11\%\\
            \textbf{Prompt Tuning} &  0.06\%&  0.03\%& 0.01\%\\
            \textbf{Last Layer Tuning} &  6.30\% &  4.20\% & 2.10\% \\
            \textbf{Selective Module Tuning} &  2.10\% &  4.20\% & 2.40\% \\
           \bottomrule
      \end{tabu}
      \label{tab:tuned_parameter_ratio}
  \end{table}

Apart from the delta tuning methods (prompt tuning, adapter, LoRA and prefix-tuning) introduced in the previous sections, we additionally design two delta tuning methods, i.e., last layer tuning and selective module tuning, to investigate the power of scale for delta tuning. For last layer tuning, we only select the last layer of the encoder in T5 to optimize. For selective module tuning, we manually choose some modules (e.g.,
the feed-forward layer, query / key / value matrix in the attention layer, or a layer norm) in T5 to optimize. We set the training batch size to $64$ for all delta tuning methods. For the different scales of T5, we use the same learning rates during training: $5\times 10^{-3}$ (prompt tuning), $5\times 10^{-4}$ (adapter), $5\times 10^{-5}$ (LoRA) $5\times 10^{-3}$ (prefix-tuning),  $5\times 10^{-5}$ (last layer tuning), and  $5\times 10^{-5}$ (selective module tuning). The percentage of the tunable parameters for each method / model is listed in Table~\ref{tab:tuned_parameter_ratio}.

\subsection{Task-level Transferability Evaluation}
\label{sec:Transferability Details}

In the cross-task transferability experiments, we utilize $12$ tasks of $5$ different types as follows:

\paragraph{Sentiment Analysis.} Given a sentence, a PLM identifies the sentiment polarity in this sentence. We choose \texttt{SST-2} \citep{socher-etal-2013-recursive}, \texttt{Amazon/Polarity}, and \texttt{Rotten Tomatoes} \citep{pang-lee-2005-seeing} to analyze.

\paragraph{Natural Language Inference.} Given a premise and hypothesis pair, a PLM determines whether the hypothesis is entailed, contradict, or undetermined by the premise. We choose \texttt{MNLI} \citep{williams-etal-2018-broad}, \texttt{SICK} \citep{marelli-etal-2014-sick}, and \texttt{SciTail} \citep{scitail} to analyze.

\paragraph{Paraphrase Identification.} Given a pair of sentences, a PLM judges whether they are semantically identical. We choose \texttt{QQP} \citep{sharma2019natural} and \texttt{MRPC} \citep{dolan-brockett-2005-automatically} to analyze.

\paragraph{Question Answering.} Given a question, a PLM answers the question based on context. We choose \texttt{MathQA} \citep{amini-etal-2019-mathqa} and \texttt{AQUA-RAT} \citep{ling-etal-2017-program} to analyze. 

\paragraph{Summarization.} Given an article, a PLM summarizes it. We choose \texttt{Multi-News} \citep{fabbri-etal-2019-multi}, and \texttt{SAMSum} \citep{gliwa-etal-2019-samsum} to analyze.

\paragraph{Evaluation Metrics.}
%\label{ssec:appendix_evaluation_metric}
For sentiment analysis, natural language inference, and paraphrase identification tasks, we choose accuracy (Acc.) as their evaluation metric in the experiments. For question answering and summarization, we utilize F1 and ROUGE-L \citep{lin-2004-rouge}, respectively. Finally, we report their relative performance (transferring zero-shot performance / original performance) (\%). 

\section{Tasks Evaluated in Experiments}
\label{sec:tasks}

\LTcapwidth=\textwidth
\small
    \begin{longtable}{llll}
    \caption{The tasks evaluated in our experiments in Table~\ref{tab:overall}. We refer to \citet{ye2021crossfit} for task ontology.} 
    \label{tab:ontology_1}%
    \endfirsthead
    \endhead
    \toprule
    \textbf{Ontology} & \textbf{Task Name} & \textbf{Reference}\\
    \midrule
    \multirow{2}[2]{*}{cls/sentiment analysis} 
          & glue-sst2 & \citealt{socher-etal-2013-recursive} \\
          %& imdb  & \citealt{maas-etal-2011-learning}   \\
          & rotten\_tomatoes & \citealt{pang-lee-2005-seeing} \\
    \midrule
    \multirow{10}[2]{*}{cls/emotion} 
          & emo & \citealt{chatterjee-etal-2019-semeval}    \\
          %& tweet\_eval-emoji & \citealt{barbieri-etal-2020-tweeteval}    \\
          & tweet\_eval-hate &	\citealt{barbieri-etal-2020-tweeteval} \\
          & tweet\_eval-irony &	\citealt{barbieri-etal-2020-tweeteval} \\
          & tweet\_eval-offensive &	\citealt{barbieri-etal-2020-tweeteval} \\
          & tweet\_eval-sentiment  &	\citealt{barbieri-etal-2020-tweeteval} \\
          & tweet\_eval-stance\_abortion &	\citealt{barbieri-etal-2020-tweeteval} \\
          & tweet\_eval-stance\_atheism &	\citealt{barbieri-etal-2020-tweeteval} \\
          & tweet\_eval-stance\_climate &	\citealt{barbieri-etal-2020-tweeteval} \\
          & tweet\_eval-stance\_feminist &	\citealt{barbieri-etal-2020-tweeteval} \\
          & tweet\_eval-stance\_hillary &	\citealt{barbieri-etal-2020-tweeteval} \\
    \midrule
    \multirow{6}[2]{*}{cls/hate speech detection} 
          &   ethos-disability   &	\citealt{Mollas2020ETHOSAO} \\
          &   ethos-gender   &	\citealt{Mollas2020ETHOSAO} \\
          &   ethos-national\_origin &	\citealt{Mollas2020ETHOSAO}   \\
          &   ethos-religion   &	\citealt{Mollas2020ETHOSAO} \\
          %&   ethos-sexual\_orientation   &	\citealt{Mollas2020ETHOSAO} \\
          &   hate\_speech18  &	\citealt{hateoffensive} \\
          &   hatexplain &	\citealt{mathew2020hatexplain}  \\
    \midrule
    \multirow{6}[2]{*}{cls/NLI} 
          &   anli & \citealt{nie-etal-2020-adversarial}  \\
          &  glue-mnli & \citealt{williams-etal-2018-broad}   \\
          &  glue-qnli &	\citealt{rajpurkar-etal-2016-squad}   \\
          &  glue-rte  &	\begin{tabular}[c]{@{}l@{}}\citealt{dagan2005pascal, bar2006second}\\\citealt{giampiccolo2007third}\end{tabular}  \\
          %&  glue-wnli &	\citealt{faruqui-das-2018-identifying}   \\
          &  scitail  & \citealt{Khot2018SciTaiLAT}  \\
          &  superglue-rte & \begin{tabular}[c]{@{}l@{}}\citealt{dagan2005pascal, bar2006second}\\\end{tabular}   \\
    \midrule
    \multirow{2}[2]{*}{cls/fact checking} 
          &  climate\_fever  &	\citealt{Diggelmann2020CLIMATEFEVERAD}  \\
          %&   kilt\_fever &	\citealt{thorne-etal-2018-fever}  \\
          &   liar  &	\citealt{wang-2017-liar} \\
    \midrule
    \multirow{3}[2]{*}{cls/paraphrase} 
          &   glue-qqp & \href{https://quoradata.quora.com/First-Quora-Dataset-Release-Question-Pairs}{(link)}  \\
          &  medical\_questions\_pairs & \citealt{medical-qqp}  \\
          &  paws & \citealt{zhang-etal-2019-paws}   \\
    \midrule
    \multirow{1}[2]{*}{cls/topic} 
          &   ag\_news  &	\href{http://groups.di.unipi.it/~gulli/AG_corpus_of_news_articles.html}{Gulli (link)} \\
          %&   dbpedia\_14 &	\citealt{Lehmann2015DBpediaA}  \\
    \midrule
    \multirow{7}[2]{*}{cls/other} 
          &  ade\_corpus\_v2-classification &	\citealt{GURULINGAPPA2012885}   \\
          &   discovery  &	\citealt{sileo-etal-2019-mining} \\
          &   glue-cola  &	\citealt{warstadt-etal-2019-neural} \\
          %&   google\_wellformed\_query &	\citealt{faruqui-das-2018-identifying}  \\
          &   sms\_spam &	\citealt{sms_spam}   \\
          &   superglue-wic  &	\citealt{pilehvar-camacho-collados-2019-wic} \\
          &   superglue-wsc &	\citealt{levesque2012winograd}  \\
          &   wiki\_qa &	\citealt{yang-etal-2015-wikiqa}  \\
    \midrule
    \multirow{8}[2]{*}{qa/closed-book qa} 
          & freebase\_qa  &	\citealt{jiang-etal-2019-freebaseqa}    \\
          %&  jeopardy &	\href{https://www.reddit.com/r/datasets/comments/1uyd0t/200000_jeopardy_questions_in_a_json_file/}{(link)}   \\
          %&  kilt\_hotpotqa  &	\citealt{yang-etal-2018-hotpotqa}  \\
          %&  kilt\_nq  &	\citealt{kwiatkowski-etal-2019-natural}  \\
          %&  kilt\_trex  &	\citealt{elsahar-etal-2018-rex}  \\
          %&  kilt\_zsre &	\citealt{levy-etal-2017-zero}   \\
          &  lama-conceptnet &	\citealt{petroni-etal-2019-language,petroni2020how}   \\
          &  lama-google\_re &	\citealt{petroni-etal-2019-language,petroni2020how}   \\
          &  lama-squad &	\citealt{petroni-etal-2019-language,petroni2020how}    \\
          &  lama-trex &	\citealt{petroni-etal-2019-language,petroni2020how}   \\
          &  numer\_sense &	\citealt{lin-etal-2020-birds}   \\
          &  search\_qa  &	\citealt{Dunn2017SearchQAAN}  \\
          %&  squad-no\_context  &	\citealt{rajpurkar-etal-2016-squad}  \\
          &  web\_questions  &	\citealt{berant-etal-2013-semantic}  \\
    %\midrule
    %\multirow{2}[2]{*}{qa/binary} 
          %&  boolq &	\citealt{clark-etal-2019-boolq}   \\
          %&  mc\_taco  &	\citealt{zhou-etal-2019-going}  \\
    \midrule
    \multirow{13}[2]{*}{qa/multiple-choice qa} 
          %&  ai2\_arc &	\citealt{Clark2018ThinkYH}   \\
          %&  aqua\_rat &	\citealt{ling-etal-2017-program}  \\
          %&  codah  &	\citealt{chen-etal-2019-codah}  \\
          %&  commonsense\_qa &	\citealt{talmor-etal-2019-commonsenseqa}   \\
          &  cosmos\_qa  &	\citealt{huang-etal-2019-cosmos}  \\
          &  dream &	\citealt{saha-etal-2018-duorc}   \\
          &  hellaswag &	\citealt{zellers-etal-2019-hellaswag}   \\
          %&  math\_qa &	\citealt{amini-etal-2019-mathqa}   \\
          &  openbookqa &	\citealt{mihaylov-etal-2018-suit}   \\
          &  qasc  &	\citealt{khot2020qasc}  \\
          %&  quail  &	\citealt{Rogers_Kovaleva_Downey_Rumshisky_2020}  \\
          &  quarel &	\citealt{Tafjord_Clark_Gardner_Yih_Sabharwal_2019}   \\
          &  quartz-no\_knowledge &	\citealt{tafjord-etal-2019-quartz}   \\
          &  quartz-with\_knowledge &	\citealt{tafjord-etal-2019-quartz}   \\
          &  race-high &	\citealt{lai-etal-2017-race}   \\
          &  race-middle &	\citealt{lai-etal-2017-race}   \\
          %&  social\_i\_qa  &	\citealt{sap-etal-2019-social}  \\
          &  superglue-copa  &	\citealt{gordon-etal-2012-semeval}  \\
          %&  superglue-multirc &	\citealt{khashabi-etal-2018-looking}   \\
          &  swag  &	\citealt{zellers-etal-2018-swag}  \\
          &  wino\_grande &	\citealt{Sakaguchi_Le_Bras_Bhagavatula_Choi_2020}   \\
    \midrule
    \multirow{3}[2]{*}{qa/long-form qa} 
          & eli5-askh   & \citealt{fan-etal-2019-eli5}   \\
          &  eli5-asks & \citealt{fan-etal-2019-eli5}   \\
          &  eli5-eli5  & \citealt{fan-etal-2019-eli5}  \\
    \midrule
    \multirow{1}[2]{*}{qa/MRC} 
          %&  biomrc  &	\citealt{pappas-etal-2020-biomrc}  \\
          %&  quoref &	\citealt{dasigi-etal-2019-quoref}   \\
          %&  ropes &	\citealt{lin-etal-2019-reasoning}   \\
          &  superglue-record  & \citealt{Zhang2018ReCoRDBT}  
          \\
    %\midrule
    \multirow{3}[2]{*}{cg/summarization} 
          %& gigaword  &	\citealt{napoles-etal-2012-annotated}     \\
          &  multi\_news & \citealt{fabbri-etal-2019-multi}   \\
          &  samsum &	\citealt{gliwa-etal-2019-samsum}   \\
          &  xsum &	\citealt{narayan-etal-2018-dont}   \\
    % \midrule
    %\multirow{2}[2]{*}{cg/dialogue} 
          %&   empathetic\_dialogues & \citealt{rashkin-etal-2019-towards}  \\
          %&  kilt\_wow &	\citealt{dinan2018wizard}   \\
    \midrule
    \multirow{3}[2]{*}{cg/other} 
          &  spider &	\citealt{yu-etal-2018-spider}   \\
          &  wiki\_bio &	\citealt{lebret-etal-2016-neural}   \\
          &  wiki\_split & \citealt{botha-etal-2018-learning}    \\
          %&  wikisql &	\citealt{zhongSeq2SQL2017}   \\
    \midrule
    %\parbox[c]{2cm}{other/linguistic phenomenon} 
    \multirow{3}[2]{*}{other/linguistic phenomenon}
          %&  \parbox[c]{4cm}{
          & blimp-anaphor\_gender\_agreement  & \citealt{warstadt2019blimp}  \\
          & blimp-ellipsis\_n\_bar\_1 & \citealt{warstadt2019blimp}   \\
          & blimp-sentential\_negation\_npi\_scope & \citealt{warstadt2019blimp} \\%}  \\
    % \multirow{3}[2]{*}{other/linguistic phenomenon} &  blimp-anaphor\_gender\_agreement    \\
    %       &  blimp-ellipsis\_n\_bar\_1    \\
    %       &  blimp-sentential\_negation\_npi\_scope    \\
    \midrule
    \parbox[c]{2cm}{other/generate explanation} 
          %&  \parbox[c]{4cm}{cos\_e}  \\
          & cos\_e & \citealt{rajani-etal-2019-explain}  \\
    \midrule
    \multirow{2}[2]{*}{other/slot\_filling} 
          &  ade\_corpus\_v2-dosage & \citealt{GURULINGAPPA2012885}   \\
          &  ade\_corpus\_v2-effect & \citealt{GURULINGAPPA2012885}   \\
    \midrule
    \multirow{1}[2]{*}{other/entity linking} 
          &  kilt\_ay2 &	\citealt{hoffart-etal-2011-robust}   \\
    \midrule
    \multirow{4}[2]{*}{other/other} 
          &   acronym\_identification  & \citealt{pouran-ben-veyseh-etal-2020-acronym}  \\
          %&  art &	\citealt{bhagavatula2020abductive}    \\
          &  aslg\_pc12  &	\citealt{Othman2012EnglishASLGP}  \\
          %&  break-QDMR &	\citealt{wolfson-etal-2020-break}   \\
          %&  break-QDMR-high-level  & \citealt{wolfson-etal-2020-break}  \\
          %&  common\_gen  &	\citealt{lin-etal-2020-commongen}  \\
          &  crawl\_domain &	\citealt{zhang-etal-2020-semi}   \\
          %&  crows\_pairs  &	\citealt{nangia-etal-2020-crows}  \\
          %&  definite\_pronoun\_resolution &	\citealt{rahman-ng-2012-resolving}   \\
          %&  e2e\_nlg\_cleaned &	\citealt{dusek.etal2020:csl, dusek-etal-2019-semantic}   \\
          %&  limit  &	\citealt{manotas-etal-2020-limit}  \\
          %&  piqa  &	\citealt{Bisk2020}   \\
          &  proto\_qa &	\citealt{boratko-etal-2020-protoqa}   \\
          %&  qa\_srl &	\citealt{he-etal-2015-question}   \\
    \bottomrule
    \end{longtable}%

\end{document}